\documentclass{article}
\usepackage{silence}
\usepackage[letterpaper, left=1in, right=1in, top=1in, bottom=1in]{geometry}
\usepackage{parskip}
\usepackage[T1]{fontenc}
\usepackage{times}
\usepackage{inconsolata}
\AddToHook{ttfamily}{\fontshape{\updefault}}
\usepackage[final,nopatch=footnote]{microtype}
\usepackage{amsmath}
\usepackage{amssymb}
\usepackage{mathtools}
\usepackage{amsthm}
\usepackage{bm}
\allowdisplaybreaks

\usepackage{graphicx}
\usepackage{booktabs}
\usepackage{xcolor}
\usepackage{pifont}
\AddToHook{env/table/begin}{\setlength{\abovecaptionskip}{0pt}\setlength{\belowcaptionskip}{6pt}}

\usepackage[algo2e,ruled]{algorithm2e}
\makeatletter
\let\For\@undefined
\let\If\@undefined
\let\Else\@undefined
\let\Return\@undefined
\let\While\@undefined
\let\Repeat\@undefined
\let\Until\@undefined
\makeatother
\usepackage[noend]{algpseudocode}
\algrenewcommand{\algorithmicindent}{1.25em}
\algrenewcommand{\alglinenumber}[1]{\scriptsize\color{black}#1}
\algrenewcommand{\algorithmiccomment}[1]{\hfill{\footnotesize\color{black}$\triangleright$~\textit{#1}}}
\SetAlFnt{\small}
\SetAlCapFnt{\small}
\SetAlCapNameFnt{\small\bfseries}
\makeatletter
\AddToHook{env/algorithmic/begin}{\def\@currentcounter{ALG@line}}
\makeatother

\usepackage[round]{natbib}
\usepackage{hyperref}
\definecolor{NavyBlue}{cmyk}{0.94,0.54,0,0}
\hypersetup{
  colorlinks=true,
  allcolors=NavyBlue,
  breaklinks=true,
  hypertexnames=false}
\usepackage[nameinlink,capitalize]{cleveref}
\usepackage{autonum}
\crefname{equation}{Eq.}{Eqs.}
\Crefname{equation}{Eq.}{Eqs.}
\crefformat{equation}{#2Eq.\,(#1)#3}
\Crefformat{equation}{#2Eq.\,(#1)#3}
\crefname{assumption}{Assumption}{Assumptions}
\Crefname{assumption}{Assumption}{Assumptions}
\crefname{algocf}{Algorithm}{Algorithms}
\Crefname{algocf}{Algorithm}{Algorithms}
\crefalias{ALG@line}{line}
\AddToHook{cmd/appendix/after}{%
  \crefalias{section}{appendix}%
  \crefalias{subsection}{appendix}%
  \crefalias{subsubsection}{appendix}}

\theoremstyle{plain}
\newtheorem{theorem}{Theorem}[section]
\newtheorem{lemma}{Lemma}[section]

\newtheorem{definition}{Definition}[section]
\newtheorem{assumption}{Assumption}[section]
\theoremstyle{definition}

\renewenvironment{proof}[1][\proofname]{%
  \par\addvspace{6pt}\noindent\textit{#1.}\enspace\ignorespaces}{%
  \ifhmode\unskip\nobreak\hfill\fi\ensuremath{\square}\par\addvspace{6pt}}
\let\originalparagraph\paragraph
\renewcommand{\paragraph}[1]{\originalparagraph{#1.}}
\theoremstyle{plain}
\newtheorem{proposition}{Proposition}[section]
\theoremstyle{definition}
\newtheorem{example}{Example}[section]
\AtEndEnvironment{example}{\unskip\nobreak\hfill$\triangleleft$}

\definecolor{actup}{RGB}{70,120,255}
\definecolor{actdown}{RGB}{60,200,90}
\definecolor{actleft}{RGB}{180,90,255}
\definecolor{actright}{RGB}{255,70,70}
\definecolor{actstay}{RGB}{255,170,40}

\newcommand{\uptri}{\textcolor{actup}{$\blacktriangle$}}
\newcommand{\downtri}{\textcolor{actdown}{$\blacktriangledown$}}
\newcommand{\lefttri}{\textcolor{actleft}{$\blacktriangleleft$}}
\newcommand{\righttri}{\textcolor{actright}{$\blacktriangleright$}}
\newcommand{\staycirc}{\textcolor{actstay}{$\bullet$}}

\newcommand{\OCUPO}{\textnormal{\texttt{OCUPO}}}
\newcommand{\OCUPOON}{\textnormal{\texttt{OCUPO-On}}}
\newcommand{\OCUPOOFF}{\textnormal{\texttt{OCUPO-Off}}}
\newcommand{\OCUPOLIN}{\textnormal{\texttt{OCUPO-Lin}}}
\newcommand{\TME}{\textnormal{\texttt{TME}}}
\newcommand{\BTE}{\textnormal{\texttt{BTE}}}
\newcommand{\STE}{\textnormal{\texttt{STE}}}
\newcommand{\PRE}{\textnormal{\texttt{PRE}}}
\newcommand{\OPE}{\textnormal{\texttt{OPE}}}
\newcommand{\BFR}{\textnormal{\texttt{BFR}}}
\newcommand{\NPG}{\textnormal{\texttt{NPG}}}
\newcommand{\PNPG}{\textnormal{\texttt{PNPG}}}

\hypersetup{
  pdftitle={Provably Efficient Concave Utility Reinforcement Learning with Linear Function Approximation},
  pdfauthor={Max Qiushi Lin, Ahmed Magd, Sharan Vaswani}}

\title{Provably Efficient Concave Utility Reinforcement Learning \\ with Linear Function Approximation}
\author{%
  \begin{tabular}{c}
    Max Qiushi Lin\textsuperscript{1}\hspace{2.5em}
    Ahmed Magd\textsuperscript{1}\hspace{2.5em}
    Sharan Vaswani\textsuperscript{1}
  \end{tabular}
}
\date{}

\begin{document}

\maketitle
\footnotetext[1]{Simon Fraser University. \texttt{\{maxqslin, a7med1magd2, vaswani.sharan\}@gmail.com}.}
\setcounter{footnote}{1}

\begin{abstract}
Concave utility reinforcement learning (CURL) provides a unified framework for a diverse range of problems, including standard reinforcement learning (RL), imitation learning, and maximum entropy exploration. Unlike standard RL, CURL generally requires estimating high-dimensional state-action occupancy measures, making it a challenging problem. Consequently, existing methods for CURL either sidestep the exploration problem, are restricted to the tabular setting, or lack computationally efficient policy optimization. To that end, we first study CURL in finite-horizon episodic bilinear MDPs, where the transition model admits a low-dimensional representation in known feature spaces. In particular, we use the bilinear structure to efficiently estimate occupancy measures via low-dimensional feature expectations, and propose a model-based optimistic concave utility policy optimization (\texttt{OCUPO}) framework. We first instantiate \texttt{OCUPO} in the on-policy setting, in which the algorithm uses independent trajectories in each iteration to estimate the concave utility policy gradient. We prove that the resulting algorithm achieves an $\widetilde{\mathcal{O}}(\epsilon^{-4})$ sample complexity. Moreover, in order to improve the sample efficiency, we instantiate \texttt{OCUPO} in the off-policy setting. We propose a novel backward feature regression subroutine that reuses trajectories collected by different policies to estimate the current policy's occupancy measure. Under suitable technical assumptions on the policy class (satisfied, e.g., by log-linear policies with sufficiently expressive features), we prove that the resulting algorithm achieves an $\widetilde{\mathcal{O}}(\epsilon^{-2})$ sample complexity, up to a controllable projection error. Furthermore, we extend our framework to handle linear MDPs, and show that in the off-policy setting, the resulting algorithm retains the same sample complexity. In these settings, the dependence on $\epsilon$ matches that for standard RL with linear function approximation. Moreover, our algorithms do not rely on the integral oracle commonly required by prior work.
\end{abstract}

\section{Introduction}

While standard reinforcement learning (RL) succeeds at maximizing the cumulative reward, complex real-world applications often require different objectives, such as maximizing the entropy of the state visitation distribution~\citep{hazan2018provably}, mimicking demonstrations from experts~\citep{abbeel2004apprenticeship,argall2009survey}, or keeping the state visitation within safety limits~\citep{altman1999constrained,qin2021density}. To address these diverse objectives, a body of literature~\citep{zhang2020variational,kumar2022policy,barakat2023reinforcement,ying2023scalable,marin2024metacurl,moreno2024efficient,moreno2025online,chen2024robust,barakat2024towards} has considered the problem of concave utility reinforcement learning (CURL). In particular, concave utility functions of the state-action occupancy measure encapsulate the standard RL problem and a wide range of the above objectives (see~\cref{app:utility_examples} and~\citet[Table 1]{zahavy2021reward}). Consequently, developing principled and efficient algorithms for solving the CURL problem is important.

However, solving CURL is challenging since it generally requires estimating the high-dimensional state-action occupancy measure. Consequently, most existing theoretical work~\citep{moreno2024efficient,marin2024metacurl,zhang2020variational,geist2021concave,kumar2022policy} on CURL focuses on tabular settings where the state and action spaces are relatively small and discrete. Furthermore, most prior works~\citep{zhang2020variational,geist2021concave,kumar2022policy,barakat2023reinforcement,barakat2024towards} focus on the planning setting and assume that it is possible to accurately estimate the value function for all the state-action pairs, thereby bypassing the need for strategic exploration. A notable exception is the work by~\cite{moreno2025online} that proposes a theoretically principled approach to explicitly address the exploration problem in CURL. However, their algorithms and corresponding guarantees are strictly restricted to the tabular setting, relying on explicit representations of finite state and action spaces, and cannot be used for environments with large state-action spaces. Beyond the tabular setting, \citet{huang2023density} estimate occupancies and explore in low-rank MDPs with known density features, with applications to convex RL, but their approach does not provide a computationally efficient policy search.

To address these issues, we consider the setting of linear function approximation~\citep{jin2023provably,cai2020provably,yang2020reinforcement}. By assuming that the environment's structure can be succinctly expressed in a low-dimensional feature space, linear function approximation enables generalization across unseen states and overcomes the curse of dimensionality. Specifically, we seek to answer the following fundamental question:
\begin{center}
\textit{Can we design provably efficient algorithms for CURL with linear function approximation?}
\end{center}
In this paper, we answer this question affirmatively in the context of finite-horizon episodic bilinear MDPs~\citep{yang2020reinforcement}, where the transition model admits a low-dimensional representation in known feature spaces. In this setting, we make the following contributions.

\paragraph{Contribution 1} In \cref{section:generic_framework}, we use the policy gradient theorem~\citep{kumar2022policy} and develop optimistic concave utility policy optimization ($\OCUPO$), a generic model-based framework for solving CURL problems. At each iteration, $\OCUPO$ estimates the transition model and the occupancy measure of the current policy, which together determine a policy-dependent pseudo-reward. The framework then uses these pseudo-rewards and optimistic policy evaluation to construct the state-action value functions that guide strategic exploration. Finally, $\OCUPO$ uses the resulting state-action value estimates to update the policy using natural policy gradient~\citep[NPG;][]{kakade2001natural}.

\paragraph{Contribution 2} In \cref{section:ocupo-on}, we instantiate $\OCUPO$ for finite-horizon episodic bilinear MDPs in the on-policy setting. The resulting algorithm is referred to as $\OCUPOON$ and relies on sampling multiple independent trajectories in each iteration. $\OCUPOON$ uses the gathered data to perform vector-valued ridge regression and efficiently estimate the transition model. Estimating the high-dimensional occupancy measure is a challenging but crucial problem for computing the pseudo-reward. By exploiting the bilinear structure of the MDP, $\OCUPOON$ approximates the occupancy measure using the estimated transition model and low-dimensional feature expectations. These feature expectations can be efficiently estimated by directly averaging the features over multiple independent trajectories collected by the current policy. For evaluating a given policy under the pseudo-reward derived from the estimated occupancy measures, $\OCUPOON$ adopts a novel approach to efficiently construct optimistic state-action value estimates. To derive theoretical guarantees for $\OCUPOON$, we use concavity to reduce CURL to standard RL with the policy-dependent pseudo-rewards, and control the errors arising from pseudo-reward estimation and policy evaluation. We show that $\OCUPOON$ can be implemented efficiently and achieves an $\widetilde{\mathcal{O}}(\epsilon^{-4})$ sample complexity for the CURL problem. To the best of our knowledge, this is the first sample complexity guarantee for CURL in bilinear MDPs achieved by a computationally efficient algorithm without any coverage assumption. The resulting dependence on $\epsilon$ matches the on-policy guarantee for standard RL with linear function approximation~\citep{liu2023optimistic}.

\paragraph{Contribution 3} To improve the sample complexity, in \cref{section:ocupo-off}, we instantiate $\OCUPO$ in the off-policy setting and refer to the resulting algorithm as $\OCUPOOFF$. In this setting, the algorithm collects only one trajectory per iteration and stores it in a replay buffer. Unlike the on-policy setting, since the trajectories were collected by different past policies, averaging the features over the transitions in the replay buffer does not lead to a valid estimate of the occupancy measure of the current policy. To overcome this challenge, we propose backward feature regression, a novel subroutine that exploits the low-dimensional structure of the MDP and can reuse the data collected by past policies to estimate the current occupancy measure. Analyzing this algorithm introduces several technical challenges, such as controlling the error accumulated through the backward regression and establishing uniform concentration over the policy class. We address these challenges for parametric policy classes with bounded covering number, and detail the approach for log-linear policies. Consequently, we prove that $\OCUPOOFF$ achieves an improved sample complexity of $\widetilde{\mathcal{O}}(\epsilon^{-2})$, up to a controllable projection error determined by the policy class. To the best of our knowledge, this is the first such rate for CURL in bilinear MDPs achieved by a computationally efficient algorithm without any coverage assumption. This dependence on $\epsilon$ matches the guarantees for policy optimization in standard RL with linear function approximation~\citep{sherman2023rate,cassel2024warm,lin2026optimistic} and for CURL in tabular MDPs~\citep{moreno2025online}. Moreover, in contrast to prior approaches for bilinear MDPs~\citep{yang2020reinforcement} or linear mixture MDPs~\citep{ayoub2020model,cai2020provably}, our algorithms do not require access to an impractical integral oracle.

\paragraph{Contribution 4} In \cref{section:ocupo-lin}, we instantiate $\OCUPO$ in linear MDPs~\citep{jin2023provably}, a more general setting, and refer to the resulting algorithm as $\OCUPOLIN$. In linear MDPs, the transition probability is linear in the known features of the current state-action pair, but each next state has its own unknown coefficient vector. Hence, a direct estimate of the transition model consists of one vector per state and requires memory proportional to the number of states. To address this issue, we propose sparse transition estimation, which estimates these vectors by a ridge regression over the transitions in the replay buffer. The resulting estimate is zero for every state that does not appear as a next state in the replay buffer. Hence, we can store it with memory proportional to the number of collected transitions, and evaluate it at any state without enumerating the state space. Using this estimate together with backward feature regression, we estimate the occupancy measure of the current policy from the replay buffer, and reuse the policy evaluation and optimization in $\OCUPOOFF$. Since the number of nonzero coefficient vectors in this estimate grows with the size of the replay buffer, analyzing $\OCUPOLIN$ introduces further technical challenges, such as controlling the pseudo-reward error and establishing uniform concentration over the value functions without any dependence on the number of states or the size of the replay buffer. We address these challenges by comparing the estimated pseudo-rewards and value functions with their counterparts under the true transition model and adapting the level of optimism to the collected data. Consequently, we prove that the resulting algorithm, $\OCUPOLIN$, retains the $\widetilde{\mathcal{O}}(\epsilon^{-2})$ sample complexity, up to the same projection error. To the best of our knowledge, this is the first sample complexity guarantee for CURL in linear MDPs.

Furthermore, in~\cref{sec:experiments}, we conduct extensive experiments with multiple CURL objectives across various environments to empirically verify the effectiveness of our proposed algorithms. 

\section{Preliminaries} \label{sec:preliminaries}

In this section, we introduce the finite-horizon MDP setting, formally define concave utility reinforcement learning, and describe the bilinear transition model considered in this paper. We compare our assumptions and guarantees with those of prior work in~\cref{app:related_work}.

\subsection{Finite-Horizon MDPs}

We consider a finite-horizon MDP with a finite state space $\mathcal{S}$, a finite action set $\mathcal{A}$, horizon $H$, transition model $P^\star\colon \mathcal{S} \times \mathcal{A} \to \Delta(\mathcal{S})$, and known initial state distribution $\rho \in \Delta(\mathcal{S})$. The agent starts at an initial state $s_1 \sim \rho$. At step $h$, it observes the current state $s_h \in \mathcal{S}$, takes an action $a_h \in \mathcal{A}$, and transitions to $s_{h+1} \sim P^\star(\cdot \mid s_h, a_h)$. The agent follows a policy $\pi \in \Delta(\mathcal{A})^{\mathcal{S} \times [H]}$, where $\pi_h(\cdot \mid s) \in \Delta(\mathcal{A})$ specifies the action distribution at state $s$ and step $h$. We assume that $P^\star$ is unknown to the agent. The state occupancy measure $d^\pi$ and the state-action occupancy measure $\mu^\pi$ are defined by $d_h^\pi(s)\coloneq\Pr[s_h=s\mid\pi,P^\star]$ and $\mu_h^\pi(s,a)\coloneq\Pr[s_h=s,a_h=a\mid\pi,P^\star]$. For a reward function $R\colon \mathcal{S}\times\mathcal{A}\times[H]\to\mathbb{R}$, we define the value function as $V_h^{\pi,R}(s)\coloneq\mathbb{E}_{\pi,P^\star}[\sum_{t=h}^H R_t(s_t,a_t)\mid s_h=s]$ and the state-action value function as $Q_h^{\pi,R}(s,a)\coloneq\mathbb{E}_{\pi,P^\star}[\sum_{t=h}^H R_t(s_t,a_t)\mid s_h=s,a_h=a]$. The expectations are taken over the randomness in the policy and the transition dynamics. These functions represent the expected cumulative reward from step $h$ onward, starting from state $s$ or state-action pair $(s,a)$ and subsequently following $\pi$.

\subsection{Concave Utility Reinforcement Learning} \label{sec:curl}

We study concave utility reinforcement learning (CURL), where the objective is to find a policy that maximizes a concave utility functional of its state-action occupancy measure.
\begin{definition}[CURL]
The CURL problem in a finite-horizon MDP is $\max_{\pi\in\Delta(\mathcal{A})^{\mathcal{S}\times[H]}}F(\mu^\pi)$, where $F(\mu^\pi)\coloneq\sum_{h=1}^H f_h(\mu_h^\pi)$ and $f_h\colon [0,1]^{\mathcal{S}\times\mathcal{A}}\to\mathbb{R}$ is concave for every $h\in[H]$.
\end{definition}
CURL subsumes several problems of interest, including the following examples.
\begin{example}[Standard RL] \label{eg:standard_rl}
$F(\mu^\pi)=\sum_{h=1}^H\langle\mu_h^\pi,r_h\rangle$, where $\{r_h\}_{h=1}^H$ are bounded reward functions. Each $f_h$ is $G$-Lipschitz and $L$-smooth with $G=\max_{h}\|r_h\|_\infty$ and $L=0$.
\end{example}

\begin{example}[Imitation Learning] \label{eg:imitation_learning}
$F(\mu^\pi)=-\sum_{h=1}^H\|\mu_h^\pi-\mu_h^{\exp}\|_2^2$, where $\mu^{\exp}$ is the occupancy measure of an expert. Each $f_h$ is $G$-Lipschitz and $L$-smooth with $G=2$ and $L=2$.
\end{example}

\begin{example}[Maximum Entropy] \label{eg:pure_exploration}
$F(\mu^\pi)=-\sum_{h=1}^H\sum_{s,a}(\mu_h^\pi(s,a)+\tau)\log(\mu_h^\pi(s,a)+\tau)$ with $\tau>0$. Each $f_h$ is $G$-Lipschitz and $L$-smooth with $G=\max\{|1+\log\tau|,1+\log(1+\tau)\}$ and $L=1/\tau$.
\end{example}
Standard RL is recovered when the objective is linear in the state-action occupancy measure (\hyperref[eg:standard_rl]{Examples~\ref*{eg:standard_rl}} and~\labelcref{eg:multi_objective_rl}). Penalties on deviations from an expert's occupancy measure capture imitation learning (\hyperref[eg:imitation_learning]{Examples~\ref*{eg:imitation_learning}} and~\labelcref{eg:kl_imitation}). Entropy maximization captures exploration objectives (\hyperref[eg:pure_exploration]{Examples~\ref*{eg:pure_exploration}}, \labelcref{eg:tsallis_entropy} and~\labelcref{eg:group_entropy}). Penalties on violations of state occupancy constraints give a penalized form of safety-constrained reinforcement learning (\cref{eg:state_density_constraints}). See~\cref{app:utility_examples} for more examples and details.

We then impose the following assumptions on each $f_h$. Lipschitzness and smoothness control how changes in the occupancy measure affect the utility values and their gradients.
\begin{assumption}[Utility Lipschitzness and Smoothness] \label{asm:lipschitzness_and_smoothness}
The utility functions are differentiable. There exist constants $G > 0$ and $L\geq0$ such that, for every $h\in[H]$ and all $\boldsymbol{u},\boldsymbol{u}^\prime\in[0,1]^{\mathcal{S}\times\mathcal{A}}$, $|f_h(\boldsymbol{u})-f_h(\boldsymbol{u}^\prime)|\leq G\|\boldsymbol{u}-\boldsymbol{u}^\prime\|_1$ and $\|\nabla f_h(\boldsymbol{u})-\nabla f_h(\boldsymbol{u}^\prime)\|_\infty\leq L\|\boldsymbol{u}-\boldsymbol{u}^\prime\|_1$. That is, $f_h$ is $G$-Lipschitz with respect to the $\ell_1$ norm and $L$-smooth with respect to the $\ell_1/\ell_\infty$ norm geometry.
\end{assumption}
Moreover, coordinate separability allows one to estimate each gradient coordinate from a single occupancy coordinate, without enumerating the state-action space.
\begin{assumption}[Coordinate Separability] \label{asm:coordinate_separability}
For every $h\in[H]$, $(s,a)\in\mathcal{S}\times\mathcal{A}$, and $\boldsymbol{u}\in[0,1]^{\mathcal{S}\times\mathcal{A}}$, the gradient coordinate $[\nabla f_h(\boldsymbol{u})](s,a)$ depends only on $\boldsymbol{u}(s,a)$.\footnote{\cref{asm:coordinate_separability} can be relaxed to~\cref{asm:gradient_locality}, where each gradient coordinate depends on a bounded number of occupancy coordinates. All results in this paper then hold with $L$ multiplied by this number.}
\end{assumption}
Note that~\cref{eg:standard_rl,eg:imitation_learning,eg:pure_exploration} (and more examples in~\cref{app:utility_examples}) satisfy both assumptions, with a fixed smoothing parameter $\tau>0$ for the entropy and KL objectives.

\subsection{Bilinear MDPs}

We consider bilinear MDPs~\citep{yang2020reinforcement}, where the transition model admits a low-dimensional representation in two known feature spaces.

\begin{definition}[Bilinear MDP; {\citealp[Assumptions $1$ and $2^\prime$]{yang2020reinforcement}}] \label{def:bilienar_mdp}
An MDP is a bilinear MDP if there exist feature maps $\boldsymbol{\phi}\colon \mathcal{S}\times\mathcal{A}\to\mathbb{R}^{d_\phi}$ and $\boldsymbol{\psi}\colon \mathcal{S}\to\mathbb{R}^{d_\psi}$, and a transition core matrix $\mathbf{M}^\star\in\mathbb{R}^{d_\phi\times d_\psi}$, such that $P^\star(s^\prime\mid s,a)=\boldsymbol{\phi}(s,a)^\top\mathbf{M}^\star\boldsymbol{\psi}(s^\prime)$. Moreover, there exist positive constants $C_{\mathbf{M}}$, $C_\phi$, $C_\psi$, and $C_{\mathbf{K}}$ such that the following feature regularity conditions hold: (i) $\|\mathbf{M}^\star\|_F\leq C_{\mathbf{M}}\sqrt{d_\phi}$, (ii) for every $(s,a)\in\mathcal{S}\times\mathcal{A}$, $\|\boldsymbol{\phi}(s,a)\|_2\leq C_\phi\sqrt{d_\phi}$, and (iii) for every function $V\colon \mathcal{S}\to\mathbb{R}$, $\|\sum_{s\in\mathcal{S}}\boldsymbol{\psi}(s)V(s)\|_2\leq C_\psi\|V\|_\infty$, and the matrix $\mathbf{K}_\psi\coloneq\sum_{s\in\mathcal{S}}\boldsymbol{\psi}(s)\boldsymbol{\psi}(s)^\top$ is invertible and satisfies $\max_{s\in\mathcal{S}}\lVert \boldsymbol{\psi}(s)^\top\mathbf{K}_\psi^{-1} \rVert_2\leq C_{\mathbf{K}}$.
\end{definition}
We assume that $\boldsymbol{\phi}$, $\boldsymbol{\psi}$, and $\mathbf{K}_\psi$ are available to the agent, whereas $\mathbf{M}^\star$ is unknown. Throughout the paper, for any $m\in\mathbb{Z}^+$ and any $\boldsymbol{\zeta}\colon \mathcal{S}\to\mathbb{R}^m$, we define $P^\star\boldsymbol{\zeta}(s,a)\coloneq\sum_{s'\in\mathcal{S}}P^\star(s'\mid s,a) \boldsymbol{\zeta}(s')$.

\subsection{Learning Objective}
We study CURL in the online setting where the agent sequentially interacts with the environment for $K$ iterations and aims to minimize the \textit{cumulative regret} $\operatorname{Reg}_K\coloneq\sum_{k=1}^K\bigl(F(\mu^{\pi^\star})-F(\mu^{\pi_k})\bigr)$, where $\pi^\star\in\operatorname*{arg\,max}_{\pi\in\Delta(\mathcal{A})^{\mathcal{S}\times[H]}}F(\mu^\pi)$ denotes an optimal policy. Let $\overline{\pi}_K$ denote the mixture policy that selects a policy uniformly at random from $\{\pi_1,\ldots,\pi_K\}$. By the concavity of $F$, the \textit{suboptimality gap} (defined below) of $\overline{\pi}_K$ satisfies
\begin{align}
\operatorname{Sub\text{-}Opt}(\overline{\pi}_K)\coloneq F(\mu^{\pi^\star})-F(\mu^{\overline{\pi}_K})\leq\mathbb{E}_{k\sim\operatorname{Unif}([K])}\bigl[F(\mu^{\pi^\star})-F(\mu^{\pi_k})\bigr]=\frac{\operatorname{Reg}_K}{K}.
\end{align}
Our goal is to learn an $\epsilon$-optimal mixture policy, which can be achieved by establishing a sublinear regret bound. We refer to the number of trajectories required to output such a policy as the sample complexity.

\section{A Generic Framework for Solving CURL} \label{section:generic_framework}

Unlike standard RL, CURL objectives are, in general, nonlinear in the state-action occupancy measure, and hence cannot be written as an expected cumulative sum of any fixed rewards. Moreover, the optimal policy in CURL may be stochastic (\cref{prop:stochastic_optimal_policy}). Consequently, standard value-based methods cannot be applied. Instead, we adopt a policy optimization approach. Following~\citet{kumar2022policy}, the policy gradient satisfies, for every $h\in[H]$ and $(s,a)\in\mathcal{S}\times\mathcal{A}$,
\begin{align}
\frac{\partial F(\mu^\pi)}{\partial\pi_h(a\mid s)}=d_h^\pi(s)Q_h^{\pi,R^\pi}(s,a), \; \text{where} \; R_h^\pi\coloneq\nabla f_h(\mu_h^\pi).
\end{align}
This is analogous to the policy gradient theorem in standard RL~\citep{sutton1999policy}, except that the $Q$-function is evaluated with respect to the policy-dependent pseudo-reward $R^\pi$. Since the transition model and therefore the occupancy measure are unknown, we construct an estimate of the occupancy measure $\widehat{\mu}_k$ at iteration $k$ and form $\widehat{R}_k$ as an estimate of the pseudo-reward:
\begin{align}
\widehat{R}_{k,h}(s,a)&=[\nabla f_h(\widehat{\mu}_{k,h})](s,a). \label{eq:pseudo_reward_estimate}
\end{align}

\begin{algorithm2e}[H]
\caption{Optimistic Concave Utility Policy Optimization (\OCUPO)}
\label{alg:curl_framework}
\begin{algorithmic}[1]
\State \textbf{Input:} number of iterations $K$, data collection batch size $N$ (only for on-policy).
\State \textbf{Initialization:} $\pi_1\gets\pi_{\mathrm{Unif}}$, $\mathcal{D}_0\gets\emptyset$.
\For{$k=1,\ldots,K$}
    \State $\mathcal{D}_k\gets\begin{cases}
    \text{On-Policy:} & \{N\text{ fresh traj.}\stackrel{\mathrm{i.i.d.}}{\sim}(\pi_k,P^\star)\},\\
    \text{Off-Policy:} & \mathcal{D}_{k-1}\cup\{1\text{ traj.}\sim(\pi_k,P^\star)\}.
    \end{cases}$ \label{line:generic_data} \Comment{Data Collection}
    \State $\widehat{P}_k\gets\TME{}(\mathcal{D}_k)$. \label{line:generic_tme} \Comment{Transition Model Estimation}
    \State $\widehat{R}_k\gets\PRE{}(\pi_k,\widehat{P}_k,\mathcal{D}_k)$. \label{line:generic_pre} \Comment{Pseudo-Reward Estimation}
    \State $\widehat{Q}_k\gets\OPE{}(\pi_k,\widehat{R}_k,\mathcal{D}_k)$. \label{line:generic_ope} \Comment{Optimistic Policy Evaluation}
    \State $\pi_{k+1}\gets\NPG{}(\pi_k,\widehat{Q}_k)$. \label{line:generic_po} \Comment{Policy Optimization}
\EndFor
\State \Return\quad $\{\pi_1,\ldots,\pi_K\}$.
\end{algorithmic}
\end{algorithm2e}

We present the complete framework in~\cref{alg:curl_framework}. Starting with the uniform policy $\pi_1$, at each iteration $k$, the agent interacts with the environment using $\pi_k$ (\cref{line:generic_data}). In the on-policy setting, it collects $N$ fresh trajectories. In the off-policy setting, it collects one trajectory and adds it to the replay buffer. Using the collected data, transition model estimation (\TME{}) fits a model $\widehat{P}_k$ (\cref{line:generic_tme}). Pseudo-reward estimation (\PRE{}) then uses the learned model and the collected data to estimate the occupancy measure $\mu^{\pi_k}$ and construct the utility gradient using~\cref{eq:pseudo_reward_estimate} (\cref{line:generic_pre}). However, evaluating the policy with these estimates alone could undervalue regions of the state-action space whose transitions have rarely been observed and thereby hinder exploration. Therefore, optimistic policy evaluation (\OPE{}) adds a confidence bonus to account for the uncertainty in the estimated value of the next state when constructing $\widehat{Q}_k$ (\cref{line:generic_ope}), which encourages the policy to explore these regions. Finally, natural policy gradient (\NPG{}) uses these estimates to compute the next policy $\pi_{k+1}$ (\cref{line:generic_po}). \cref{section:ocupo-on,section:ocupo-off} instantiate this framework in the on-policy and off-policy settings, respectively.

\section{On-Policy Optimization for CURL} \label{section:ocupo-on}

In this section, we instantiate $\OCUPO$ in the on-policy setting (referred to as $\OCUPOON$). We present its algorithm design in~\cref{subsec:on_policy_algorithm} and analyze its sample complexity in~\cref{subsec:on_policy_sample_complexity,subsec:on_policy_proof_outline}.

\subsection{Algorithm Design} \label{subsec:on_policy_algorithm}

At iteration $k$, $\OCUPOON$ collects $N$ fresh trajectories using $\pi_k$. We define each trajectory as $\tau_i=(s_{i,1},a_{i,1},\ldots,\allowbreak s_{i,H},a_{i,H},s_{i,H+1})$, where $s_{i,h}$ and $a_{i,h}$ are the state and action at step $h$, and denote by $\boldsymbol{\phi}_{i,h}=\boldsymbol{\phi}(s_{i,h},a_{i,h})$ the observed feature. Based on the model-based framework $\OCUPO$ (\cref{alg:curl_framework}), we subsequently instantiate \TME{}, \PRE{}, \OPE{}, and \NPG{}.

\paragraph{Transition Model Estimation}
To learn the transition model, we note that the bilinear structure reduces this task to estimating the unknown core $\mathbf{M}^\star\in\mathbb{R}^{d_\phi\times d_\psi}$. Following~\citet{yang2020reinforcement}, in~\cref{lem:bilinear_regression_response}, we first show that, for a next state $s'\sim P^\star(\cdot\mid s,a)$, the row vector $\boldsymbol{\psi}(s')^\top\mathbf{K}_\psi^{-1}\in\mathbb{R}^{1\times d_\psi}$ is an unbiased estimator of $\boldsymbol{\phi}(s,a)^\top\mathbf{M}^\star\in\mathbb{R}^{1\times d_\psi}$. This allows us to estimate $\mathbf{M}^\star$ by a vector-valued ridge regression using all the trajectories $\tau_i\in\mathcal{D}_k$:
\begin{align}
\mathbf{M}_k&\coloneq\operatorname*{arg\,min}_{\mathbf{M}\in\mathbb{R}^{d_\phi\times d_\psi}}\biggl\{\sum_{\tau_i\in\mathcal{D}_k}\sum_{h=1}^H\|\boldsymbol{\psi}(s_{i,h+1})^\top\mathbf{K}_\psi^{-1}-\boldsymbol{\phi}_{i,h}^\top\mathbf{M}\|_2^2+\|\mathbf{M}\|_F^2\biggr\}\nonumber\\
&=\boldsymbol{\Lambda}_k^{-1}\sum_{\tau_i\in\mathcal{D}_k}\sum_{h=1}^H\boldsymbol{\phi}_{i,h}\boldsymbol{\psi}(s_{i,h+1})^\top\mathbf{K}_\psi^{-1}, \label{eq:bilinear_tme_estimator}
\end{align}
where $\boldsymbol{\Lambda}_k=I+\sum_{\tau_i\in\mathcal{D}_k}\sum_{h=1}^H\boldsymbol{\phi}_{i,h}\boldsymbol{\phi}_{i,h}^\top$ is the Gram matrix. This enables us to express the estimated transition model as $\widehat{P}_k(s'\mid s,a)=\boldsymbol{\phi}(s,a)^\top\mathbf{M}_k\boldsymbol{\psi}(s')$. See~\cref{app:on_tme_details} for the pseudocode.

\paragraph{Pseudo-Reward Estimation}
The bilinear structure of the MDP results in a low-dimensional representation of the occupancy measures. In particular, in~\cref{lem:occupancy_measure}, we prove that, for every policy $\pi$, $h\geq2$, and $(s,a)\in\mathcal{S}\times\mathcal{A}$,
\begin{align}
\mu_h^\pi(s,a)=\pi_h(a\mid s)(\boldsymbol{\omega}_{h-1}^\pi)^\top\mathbf{M}^\star\boldsymbol{\psi}(s), \label{eq:bilinear_occupancy_decomposition}
\end{align}
where $\boldsymbol{\omega}_{h-1}^\pi\coloneq\mathbb{E}_{(s,a)\sim\mu_{h-1}^\pi}[\boldsymbol{\phi}(s,a)]\in\mathbb{R}^{d_\phi}$ is the feature expectation under policy $\pi$. Given the estimated transition core $\mathbf{M}_k$, it therefore suffices to estimate $\boldsymbol{\omega}^{\pi_k}$. Using the $N$ fresh trajectories, for every iteration $k$ and step $h$, we can compute $\widehat{\boldsymbol{\omega}}_{k,h}=N^{-1}\sum_{i=1}^N\boldsymbol{\phi}_{i,h}$ as an unbiased estimator of $\boldsymbol{\omega}_h^{\pi_k}$. Combining $\widehat{\boldsymbol{\omega}}_{k,h-1}$ with $\mathbf{M}_k$, we obtain the estimated occupancy measure:
\begin{align}
\widehat{\mu}_{k,h}(s,a)&=\begin{cases}\rho(s)\pi_{k,1}(a\mid s),&h=1,\\
\operatorname{Clip}(\pi_{k,h}(a\mid s)\widehat{\boldsymbol{\omega}}_{k,h-1}^\top\mathbf{M}_k\boldsymbol{\psi}(s),[0,1]),&h\in\{2,\ldots,H\}.\end{cases} \label{eq:bilinear_occupancy_reconstruction}
\end{align}
The pseudo-reward $\widehat{R}_k$ can then be constructed using~\cref{eq:pseudo_reward_estimate}. Since $\mathbf{M}_k$ is not guaranteed to be a valid transition core matrix, the estimated occupancy measures are clipped to ensure that they lie in the correct domain $[0,1]$. By~\cref{asm:coordinate_separability}, each pseudo-reward coordinate $\widehat{R}_{k,h}(s,a)$ depends only on $\widehat{\mu}_{k,h}(s,a)$ and can therefore be computed efficiently. See~\cref{app:on_pre_details} for the pseudocode.

\paragraph{Optimistic Policy Evaluation}
Using the estimated pseudo-reward, we next evaluate the state-action values required for updating the policy. The Bellman relation $Q_h^{\pi,R}=R_h+P^\star V_{h+1}^{\pi,R}$ separates the immediate pseudo-reward from the expected next-state value. In \cref{lem:bilinear_next_state_values}, we show that in bilinear MDPs, the latter is linear in the features. That is, for any function $V\colon \mathcal{S}\to\mathbb{R}$, $(P^\star V)(s,a)=\boldsymbol{\phi}(s,a)^\top\boldsymbol{v}^\star(V)$, where $\boldsymbol{v}^\star(V)\coloneq\mathbf{M}^\star\sum_{s'}\boldsymbol{\psi}(s')V(s')\in\mathbb{R}^{d_\phi}$. By this relation, estimating $\widehat{Q}_{k,h}$ requires an estimate of $P^\star\widehat{V}_{k,h+1}$, which we can obtain from $\widehat{P}_k$ and hence from $\mathbf{M}_k$. However, computing $\widehat{P}_k\widehat{V}_{k,h+1}(s,a)=\boldsymbol{\phi}(s,a)^\top\mathbf{M}_k\sum_{s' \in \mathcal{S}}\boldsymbol{\psi}(s')\widehat{V}_{k,h+1}(s')$ still requires a sum across the entire state space. Prior methods~\citep{yang2020reinforcement,ayoub2020model,zhou2021nearly,li2024prior} evaluate such sums assuming access to an integral oracle. Without such an oracle, explicit summation may require $\mathcal{O}(|\mathcal{S}|)$ computation. To avoid such an assumption and additional computation, we follow a different procedure. In particular, in \cref{lem:bilinear_value_response}, we show that for a state-action pair $(s,a)$ and any fixed function $V\colon \mathcal{S}\to\mathbb{R}$, $V(s')$, where $s'\sim P^\star(\cdot\mid s,a)$, is an unbiased estimator of $P^\star V(s,a)=\boldsymbol{\phi}(s,a)^\top\boldsymbol{v}^\star(V)$. This motivates estimating $\boldsymbol{v}^\star(\widehat{V}_{k,h+1})$ directly by the following ridge regression:
\begin{align}
\widehat{\boldsymbol{v}}_{k,h}\coloneq\operatorname*{arg\,min}_{\boldsymbol{v}\in\mathbb{R}^{d_\phi}}\biggl\{\sum_{\tau_i\in\mathcal{D}_k}\bigl(\boldsymbol{\phi}_{i,h}^\top\boldsymbol{v}-\widehat{V}_{k,h+1}(s_{i,h+1})\bigr)^2+\|\boldsymbol{v}\|_2^2\biggr\}. \label{eq:ope_ridge_objective}
\end{align}

\begin{algorithm2e}[H]
\caption{Optimistic Policy Evaluation (\OPE{})} \label{alg:ope}
\begin{algorithmic}[1]
\State \textbf{Input:} policy $\pi_k$, pseudo-reward $\widehat{R}_k$, dataset $\mathcal{D}_k$, radii $\{\alpha_h\}_{h=1}^{H-1}$.
\State $\widehat{Q}_{k,H}(s,a)\gets\operatorname{Clip}(\widehat{R}_{k,H}(s,a),[-G,G])$, for every queried $(s,a)$.
\State $\widehat{V}_{k,H}(s)\gets\sum_a\pi_{k,H}(a\mid s)\widehat{Q}_{k,H}(s,a)$, for every queried $s$.
\For{$h=H-1,\ldots,1$}
    \State $\boldsymbol{\Gamma}_{k,h}\gets I+\sum_{\tau_i\in\mathcal{D}_k}\boldsymbol{\phi}_{i,h}\boldsymbol{\phi}_{i,h}^\top$. \label{line:ope_gram_matrix}
    \State $\widehat{\boldsymbol{v}}_{k,h}\gets\boldsymbol{\Gamma}_{k,h}^{-1}\sum_{\tau_i\in\mathcal{D}_k}\boldsymbol{\phi}_{i,h}\widehat{V}_{k,h+1}(s_{i,h+1})$. \label{line:vtr_estimator}
    \State $\widetilde{Q}_{k,h}(s,a)\gets\widehat{R}_{k,h}(s,a)+\boldsymbol{\phi}(s,a)^\top\widehat{\boldsymbol{v}}_{k,h}+\alpha_h\|\boldsymbol{\phi}(s,a)\|_{\boldsymbol{\Gamma}_{k,h}^{-1}}$, for every queried $(s,a)$. \label{line:optimistic_Q}
    \State $\widehat{Q}_{k,h}(s,a)\gets\operatorname{Clip}(\widetilde{Q}_{k,h}(s,a),[-(H-h+1)G,(H-h+1)G])$, for every queried $(s,a)$. \label{line:clipped_Q}
    \State $\widehat{V}_{k,h}(s)\gets\sum_a\pi_{k,h}(a\mid s)\widehat{Q}_{k,h}(s,a)$, for every queried $s$. \label{line:clipped_V}
\EndFor
\State \Return\quad $\widehat{Q}_k$.
\end{algorithmic}
\end{algorithm2e}
In \cref{alg:ope}, \cref{line:ope_gram_matrix,line:vtr_estimator} compute the explicit solution to~\cref{eq:ope_ridge_objective} in terms of the Gram matrix $\boldsymbol{\Gamma}_{k,h}=I+\sum_{\tau_i\in\mathcal{D}_k}\boldsymbol{\phi}_{i,h}\boldsymbol{\phi}_{i,h}^\top$ for each step $h$. 
To encourage exploration in under-explored feature directions, \cref{line:optimistic_Q} adds the bonus $\alpha_h\|\boldsymbol{\phi}(s,a)\|_{\boldsymbol{\Gamma}_{k,h}^{-1}}$, where $\alpha_h$ is a confidence radius to be set later. With the added bonus, the resulting state-action values $\widetilde{Q}_{k,h}$ may fall outside the valid range of the true state-action values. In \cref{line:clipped_Q}, we therefore clip them to this range, $[-(H-h+1)G,(H-h+1)G]$. In \cref{line:clipped_V}, we construct $\widehat{V}_{k,h}$ from the policy $\pi_{k,h}$ and the clipped state-action values $\widehat{Q}_{k,h}$. We then use $\widehat{V}_{k,h}$ to estimate $\widehat{Q}_{k,h-1}$ and $\widehat{V}_{k,h-1}$. Repeating this procedure, \cref{alg:ope} performs a backward induction from step $H$ to $1$, starting from $\widehat{V}_{k,H+1}=0$ with $\alpha_H=0$.

\paragraph{Natural Policy Gradient}
For \cref{line:generic_po} of \cref{alg:curl_framework}, we adopt the canonical \NPG{} update for softmax policies~\citep{kakade2001natural,agarwal2021theory}, which takes the form
\begin{align}
\pi_{k+1,h}(a\mid s)=\frac{\pi_{k,h}(a\mid s)\exp(\eta\widehat{Q}_{k,h}(s,a))}{\sum_{a'}\pi_{k,h}(a'\mid s)\exp(\eta\widehat{Q}_{k,h}(s,a'))}, \label{eq:npg_update}
\end{align}
for every $h\in[H]$ and $(s,a)\in\mathcal{S}\times\mathcal{A}$, where $\eta>0$ is the step size.

\subsection{Sample Complexity} \label{subsec:on_policy_sample_complexity}

We now state the sample complexity of $\OCUPOON$ in the following theorem.
\begin{theorem}[Sample Complexity of $\OCUPOON$] \label{thm:on_policy}
Suppose that the MDP is a bilinear MDP (\cref{def:bilienar_mdp}) and that~\cref{asm:lipschitzness_and_smoothness,asm:coordinate_separability} hold. Running $\OCUPOON$ for $K$ iterations with $N=\widetilde{\mathcal{O}}(d_\phi(d_\phi^2+d_\psi)K)$ fresh trajectories per iteration, step size $\eta=\sqrt{2\log|\mathcal{A}|/(KH^2G^2)}$, and appropriate choices of the other parameters yields that with probability at least $1-\delta$,
\begin{align}
\operatorname{Sub\text{-}Opt}(\overline{\pi}_K) &\leq\widetilde{\mathcal{O}}\left(\frac{H^2(L+G)}{\sqrt K}\right).
\end{align}
Hence, for any $\epsilon>0$, $\OCUPOON$ with $K=\widetilde{\mathcal{O}}(H^4(L+G)^2/\epsilon^2)$ yields an $\epsilon$-optimal mixture policy, and therefore requires $KN=\widetilde{\mathcal{O}}(H^8d_\phi(d_\phi^2+d_\psi)(L+G)^4/\epsilon^4)$ trajectories.
\end{theorem}
\cref{thm:on_policy} shows that $\OCUPOON$ solves CURL in bilinear MDPs with an $\widetilde{\mathcal{O}}(\epsilon^{-4})$ sample complexity. To the best of our knowledge, $\OCUPOON$ is the first computationally efficient algorithm with a sample complexity guarantee for CURL in bilinear MDPs that requires no coverage assumption. Its dependence on $\epsilon$ matches the on-policy guarantee of~\citet{liu2023optimistic} for standard RL with linear function approximation. See~\cref{tab:standard_rl_comparison} for a detailed comparison in standard RL. Moreover, unlike prior methods~\citep{yang2020reinforcement,ayoub2020model,zhou2021nearly,li2024prior}, $\OCUPOON$ does not rely on an integral oracle, thereby avoiding a potential $\mathcal{O}(|\mathcal{S}|)$ computational cost.

\subsection{\texorpdfstring{Proof Sketch of \cref{thm:on_policy}}{Proof Sketch of Theorem \ref*{thm:on_policy}}} \label{subsec:on_policy_proof_outline}

We first decompose the cumulative regret into the following three terms. 

\paragraph{Regret Decomposition}
In \cref{lem:curl_regret_decomposition}, we first use the concavity of each $f_h$ to bound the suboptimality gap of $\pi_k$, $F(\mu^{\pi^\star})-F(\mu^{\pi_k})$, by the gap in value functions between $\pi^\star$ and $\pi_k$, $\mathbb{E}_\rho[V_1^{\pi^\star,R_k}-V_1^{\pi_k,R_k}]$, under the true pseudo-reward $R_{k,h}=\nabla f_h(\mu_h^{\pi_k})$. Then, we apply the extended performance difference lemma (\cref{lem:extended_pdl}) and obtain the following decomposition of the cumulative regret:
\begin{align}
\operatorname{Reg}_K&\leq\underbrace{\sum_{k,h}\mathbb{E}_{d_h^{\pi^\star}}\bigl\langle \widehat{Q}_{k,h},\pi_h^\star-\pi_{k,h} \bigr\rangle}_{\text{(I)}}+\underbrace{\sum_{k,h}\mathbb{E}_{\mu_h^{\pi^\star}}[\xi_{k,h}]}_{\text{(II)}}+\underbrace{\sum_{k,h}\mathbb{E}_{\mu_h^{\pi_k}} [ -\xi_{k,h} ]}_{\text{(III)}}, \label{eq:regret_decomposition}
\end{align}
where $\xi_{k,h}=R_{k,h}+P^\star\widehat{V}_{k,h+1}-\widehat{Q}_{k,h}$ is the Bellman error. Term (I) corresponds to the regret of the policies $\pi_1,\ldots,\pi_K$ with respect to the value estimates $\widehat{Q}_1,\ldots,\widehat{Q}_K$. Term (II) is the expected Bellman error under $\pi^\star$. Term (III) is the expected negative Bellman error under $\pi_k$. We then bound Term (I) directly and bound Terms (II) and (III) under a good event.

\paragraph{Bounding Term (I)}
Since \NPG{} is an instance of online mirror descent (OMD) at each state and step, we can apply the OMD regret lemma (\cref{thm:npg_regret}) with $u=\pi^\star$. Since $|\widehat{Q}_{k,h}|\leq HG$ and we set $\eta=\sqrt{2\log|\mathcal{A}|/(KH^2G^2)}$, we obtain
\begin{align}
\mathrm{Term~(I)}&\leq\frac{H\log|\mathcal{A}|}{\eta}+\frac{\eta KH^3G^2}{2}=\mathcal{O}\Bigl(H^2G\sqrt{K\log|\mathcal{A}|}\Bigr). \label{eq:on_oracle_outline}
\end{align}

\paragraph{The Good Event}
To bound Terms (II) and (III), we define the following good event, under which the three estimates used by $\OCUPO$ are accurate.

\begin{definition}[Good Event] \label{def:good_event}
For radii $\beta_k,\Delta_{k,h},\alpha_{k,h}\geq0$, let $\mathcal{E}\coloneq\mathcal{E}_{\mathrm{P}}\cap\mathcal{E}_{\mathrm{R}}\cap\mathcal{E}_{\mathrm{V}}$, where
\begin{align}
\mathcal{E}_{\mathrm{P}}&\coloneq\bigl\{\forall k\in[K]\colon\|\boldsymbol{\Lambda}_k^{1/2}(\mathbf{M}_k-\mathbf{M}^\star)\|_F\leq\sqrt{\beta_k}\bigr\},\\
\mathcal{E}_{\mathrm{R}}&\coloneq\bigl\{\forall k\in[K],\ h\in[H]\colon\|\widehat{R}_{k,h}-R_{k,h}\|_\infty\leq\Delta_{k,h}\bigr\},\\
\mathcal{E}_{\mathrm{V}}&\coloneq\bigl\{\forall k\in[K],\ h<H\colon\|\widehat{\boldsymbol{v}}_{k,h}-\boldsymbol{v}^\star(\widehat{V}_{k,h+1})\|_{\boldsymbol{\Gamma}_{k,h}}\leq\alpha_{k,h}\bigr\}.
\end{align}
\end{definition}
The event $\mathcal{E}_{\mathrm{P}}$ states that the true transition core matrix $\mathbf{M}^\star$ lies in a confidence ball around the estimate $\mathbf{M}_k$ in~\cref{eq:bilinear_tme_estimator}. The event $\mathcal{E}_{\mathrm{R}}$ states that the estimated pseudo-reward is close to the true pseudo-reward. The event $\mathcal{E}_{\mathrm{V}}$ states that the ridge regression in~\cref{eq:ope_ridge_objective} accurately estimates $P^\star\widehat{V}_{k,h+1}$.
In particular, for $\OCUPOON$, we denote the radii for $\mathcal{E}_{\mathrm{P}}$, $\mathcal{E}_{\mathrm{R}}$, and $\mathcal{E}_{\mathrm{V}}$ by $\beta^{\mathrm{On}}$, $\Delta_{k,h}^{\mathrm{On}}$, and $\alpha_h^{\mathrm{On}}$, respectively. We next describe each event and set their corresponding radii.

\paragraph{Transition Model Error ($\mathcal{E}_{\mathrm{P}}$)}
In~\cref{lem:good_estimation_event_P}, we can directly apply matrix concentration (\cref{lem:lin_matrix_self_normalized}) to show that $\mathcal{E}_{\mathrm{P}}$ holds with probability at least $1-\delta/4$ with $\beta^{\mathrm{On}}=\widetilde{\mathcal{O}}(d_\phi)$.

\paragraph{Pseudo-Reward Error ($\mathcal{E}_{\mathrm{R}}$)}
Bounding both Terms (II) and (III) requires controlling the pseudo-reward estimation error. Using~\cref{eq:pseudo_reward_estimate} and the smoothness of $f_h$, we first have $\|\widehat{R}_{k,h}-R_{k,h}\|_\infty\leq L\|\widehat{\mu}_{k,h}-\mu_h^{\pi_k}\|_1$. Therefore, it suffices to bound $\|\widehat{\mu}_{k,h}-\mu_h^{\pi_k}\|_1$. At $h=1$, the occupancy is known. Hence, this error is zero, and we set $\Delta_{k,1}^{\mathrm{On}}\coloneq0$. For $h\in\{2,\ldots,H\}$, we use the occupancy decomposition in~\cref{eq:bilinear_occupancy_decomposition} to show that, under the event $\mathcal{E}_{\mathrm{P}}$, with probability at least $1-\delta/4$,
\begin{align}
\|\widehat{R}_{k,h}-R_{k,h}\|_\infty&\leq LC_\psi\Bigl(\sqrt{\beta^{\mathrm{On}}}\,\|\widehat{\boldsymbol{\omega}}_{k,h-1}\|_{\boldsymbol{\Lambda}_k^{-1}}+C_{\mathbf{M}}\sqrt{d_\phi}\,\|\widehat{\boldsymbol{\omega}}_{k,h-1}-\boldsymbol{\omega}_{h-1}^{\pi_k}\|_2\Bigr) \label{eq:on_reward_omega_decomposition}\\
&\leq LC_\psi\left(\sqrt{\beta^{\mathrm{On}}}\,\widehat{e}_{k,h-1}+C_{\mathbf{M}}C_\phi d_\phi\sqrt{\frac{8\log(8KH/\delta)}{N}}\right) \eqcolon \Delta_{k,h}^{\mathrm{On}}, \label{eq:on_reward_order_main}
\end{align}
where $\widehat{e}_{k,h}\coloneq N^{-1}\sum_{i=1}^N\|\boldsymbol{\phi}_{i,h}\|_{\boldsymbol{\Lambda}_k^{-1}}$. To obtain~\cref{eq:on_reward_omega_decomposition}, we split $\|\widehat{\mu}_{k,h}-\mu_h^{\pi_k}\|_1$ into the transition model error and the error in estimating $\boldsymbol{\omega}_{h-1}^{\pi_k}$. To derive~\cref{eq:on_reward_order_main} from~\cref{eq:on_reward_omega_decomposition}, we bound the first term by the triangle inequality $\|\widehat{\boldsymbol{\omega}}_{k,h-1}\|_{\boldsymbol{\Lambda}_k^{-1}}\leq\widehat{e}_{k,h-1}$, and for the second term, we use vector concentration (\cref{lem:vector_azuma_hoeffding}) to show that $\|\widehat{\boldsymbol{\omega}}_{k,h-1}-\boldsymbol{\omega}_{h-1}^{\pi_k}\|_2\leq\widetilde{\mathcal{O}}(\sqrt{d_\phi/N})$. Combining these bounds, we show in~\cref{lem:on_policy_reward_error} that, under the event $\mathcal{E}_{\mathrm{P}}$, $\mathcal{E}_{\mathrm{R}}$ holds with probability at least $1-\delta/4$.

\paragraph{Value Estimation Error ($\mathcal{E}_{\mathrm{V}}$)}
Establishing $\mathcal{E}_{\mathrm{V}}$ requires covering the value function class. To see why, recall that \OPE{} estimates the expected next-state value $P^\star\widehat{V}_{k,h+1}$ through the ridge regression in~\cref{eq:ope_ridge_objective}. The regression target $\widehat{V}_{k,h+1}$ is constructed by the backward induction of \OPE{} after $\mathcal{D}_k$ has been collected. Consequently, $\widehat{V}_{k,h+1}$ is correlated with the states $s_{i,h+1}$ in the dataset, and in general $\mathbb{E}_{s_{i,h+1}\sim P^\star(\cdot\mid s_{i,h},a_{i,h})}[\widehat{V}_{k,h+1}(s_{i,h+1})\mid s_{i,h},a_{i,h}]\neq P^\star\widehat{V}_{k,h+1}(s_{i,h},a_{i,h})$. Hence, concentration for a fixed target (\cref{lem:self_normalized_concentration}) does not apply. Similar to the analysis of LSVI-UCB~\citep{jin2023provably}, we therefore cover the class of all value functions that \OPE{} can produce and establish concentration uniformly over this value function class. In~\cref{lem:on_value_covering_formal}, we show that the logarithmic covering number of this class is bounded by $\widetilde{\mathcal{O}}(d_\phi^2+d_\psi)$, which determines the order of $\alpha_h^{\mathrm{On}}$. With $\alpha_h^{\mathrm{On}}=\widetilde{\mathcal{O}}((H-h)G\sqrt{d_\phi^2+d_\psi})$, \cref{lem:good_estimation_event_V} shows that $\mathcal{E}_{\mathrm{V}}$ holds with probability at least $1-\delta/4$. In contrast, \citet{liu2023optimistic} avoid this covering argument in standard RL by splitting the fresh trajectories into $H$ disjoint subsets, one for each step, whereas $\OCUPOON$ reuses the same $N$ trajectories at every step.

\paragraph{Bounding Terms (II) and (III) under the Good Event}
We present the following lemma to bound the Bellman error under the good event.

\begin{lemma}[Bellman Error Bounds] \label{lem:on_bellman_error}
Suppose that \OPE{} uses the radii $\alpha_{k,h}$ of $\mathcal{E}_{\mathrm{V}}$ in~\cref{def:good_event}, and let $\alpha_{k,H}\coloneq0$. Then, under the good event $\mathcal{E}$, for every $k\in[K]$, $h\in[H]$, and $(s,a)\in\mathcal{S}\times\mathcal{A}$,
\begin{align}
\xi_{k,h}(s,a)&=(R_{k,h}+P^\star\widehat{V}_{k,h+1}-\widehat{Q}_{k,h})(s,a)\leq\Delta_{k,h}, \label{eq:on_bellman_optimism}\\
-\xi_{k,h}(s,a)&=(\widehat{Q}_{k,h}-R_{k,h}-P^\star\widehat{V}_{k,h+1})(s,a)\leq\Delta_{k,h}+2\alpha_{k,h}\|\boldsymbol{\phi}(s,a)\|_{\boldsymbol{\Gamma}_{k,h}^{-1}}. \label{eq:on_bellman_cost}
\end{align}
\end{lemma}
\cref{eq:on_bellman_optimism} shows that $\widehat{Q}_{k,h}$ is optimistic up to $\Delta_{k,h}$, and~\cref{eq:on_bellman_cost} bounds the cost of this optimism by $\Delta_{k,h}$ plus twice the bonus $\alpha_{k,h}\|\boldsymbol{\phi}(s,a)\|_{\boldsymbol{\Gamma}_{k,h}^{-1}}$. For $\OCUPOON$, we apply~\cref{lem:on_bellman_error} with $\Delta_{k,h}=\Delta_{k,h}^{\mathrm{On}}$ and $\alpha_{k,h}=\alpha_h^{\mathrm{On}}$. Summing over $k\in[K]$ and $h\in[H]$, we obtain
\begin{align}
\mathrm{Term~(II)}+\mathrm{Term~(III)}&\leq2\sum_{k=1}^K\sum_{h=1}^H\Delta_{k,h}^{\mathrm{On}}+2\sum_{k=1}^K\sum_{h=1}^{H-1}\alpha_h^{\mathrm{On}}\mathbb{E}_{\mu_h^{\pi_k}}\|\boldsymbol{\phi}\|_{\boldsymbol{\Gamma}_{k,h}^{-1}}. \label{eq:on_uncertainty_outline}
\end{align}
Summing $\Delta_{k,h}^{\mathrm{On}}$ in~\cref{eq:on_reward_order_main} over $k\in[K]$ and $h\in[H]$ and then applying the Cauchy--Schwarz inequality to $\widehat{e}_{k,h-1}$ yields $\sum_{k,h}\Delta_{k,h}^{\mathrm{On}}\leq\widetilde{\mathcal{O}}(KLHd_\phi/\sqrt N)$. Using the elliptical potential lemma (\cref{lem:elliptical_potential}), we show that, with probability at least $1-\delta/4$, $\mathbb{E}_{\mu_h^{\pi_k}}\|\boldsymbol{\phi}\|_{\boldsymbol{\Gamma}_{k,h}^{-1}}\leq\widetilde{\mathcal{O}}(\sqrt{d_\phi/N}+d_\phi^{3/2}/N)$ for all $k\in[K]$ and $h<H$.

\paragraph{Putting Everything Together}
Setting $N=\widetilde{\mathcal{O}}(d_\phi(d_\phi^2+d_\psi)K)$, we combine~\cref{eq:on_oracle_outline,eq:on_uncertainty_outline} and take a union bound to show that, with probability at least $1-\delta$, $\operatorname{Reg}_K\leq\widetilde{\mathcal{O}}(H^2(L+G)\sqrt K)$, and hence $\operatorname{Sub\text{-}Opt}(\overline{\pi}_K)\leq\operatorname{Reg}_K/K\leq\widetilde{\mathcal{O}}(H^2(L+G)/\sqrt K)$. Therefore, running $K=\widetilde{\mathcal{O}}(H^4(L+G)^2/\epsilon^2)$ iterations of $\OCUPOON$, each using $N$ fresh trajectories, yields the $\widetilde{\mathcal{O}}(H^8d_\phi(d_\phi^2+d_\psi)(L+G)^4/\epsilon^4)$ sample complexity. See~\cref{app:regret_analysis_on} for the complete proof.

\section{Off-Policy Optimization for CURL} \label{section:ocupo-off}

To improve the sample complexity, we now consider the off-policy setting, where we collect one trajectory under $\pi_k$ in each iteration $k$ and reuse the data gathered in the previous iterations. However, reusing trajectories raises a new challenge. The occupancy measure of the current policy must now be estimated from data collected by different policies. To address this challenge, we develop an algorithm, $\OCUPOOFF$, for off-policy optimization. We first describe its algorithm design in~\cref{subsec:off_policy_algorithm} and then analyze its sample complexity in~\cref{subsec:off_policy_sample_complexity,subsec:off_policy_proof_outline}.

\subsection{Algorithm Design} \label{subsec:off_policy_algorithm}

Recall from~\cref{eq:pseudo_reward_estimate,eq:bilinear_occupancy_reconstruction} that constructing $\widehat{R}_k$ requires the estimated occupancy measure $\widehat{\mu}_k$, which in turn requires $\mathbf{M}_k$ and $\widehat{\boldsymbol{\omega}}_{k,h}$. All trajectories in $\mathcal{D}_k$ are generated by the same transition model. Hence, we can still estimate $\mathbf{M}_k$ via~\cref{eq:bilinear_tme_estimator} and apply \OPE{} (\cref{alg:ope}) using $\mathcal{D}_k$. On the other hand, since the trajectories in $\mathcal{D}_k$ were collected under different policies, averaging their features does not lead to an unbiased estimator of $\boldsymbol{\omega}_h^{\pi_k}$. Instead, we present a new algorithm, backward feature regression (\BFR{}), which uses the previous trajectories to estimate $\boldsymbol{\omega}_h^{\pi_k}$.

\paragraph{Backward Feature Regression}
In order to estimate $\boldsymbol{\omega}_h^{\pi_k}$, where $\boldsymbol{\omega}_h^{\pi}\coloneq\mathbb{E}_{(s,a)\sim\mu_h^{\pi}}[\boldsymbol{\phi}(s,a)]$, we first define $\boldsymbol{g}_{h,t}^{\pi}(s)\coloneq\mathbb{E}_{\pi,P^\star}[\boldsymbol{\phi}(s_h,a_h)\mid s_t=s]$ for any step $h$ and any $t\leq h$. This is the expected feature at step $h$ when starting from state $s$ at step $t$ and following $\pi$. In~\cref{lem:backward_feature_recursion}, we show that $\boldsymbol{\omega}_h^{\pi}=\mathbb{E}_{s_1\sim\rho}[\boldsymbol{g}_{h,1}^{\pi}(s_1)]$. We also prove that $\boldsymbol{g}_{h,h}^{\pi}(s)=\sum_a\pi_h(a\mid s)\boldsymbol{\phi}(s,a)$, and for all $1\leq t<h$,
\begin{align}
\boldsymbol{g}_{h,t}^{\pi}(s)=\sum\nolimits_a\pi_t(a\mid s)P^\star\boldsymbol{g}_{h,t+1}^{\pi}(s,a). \label{eq:backward_feature_recursion}
\end{align}
Given the known starting point $\boldsymbol{g}_{h,h}^{\pi}$ (computed directly from $\pi_h$ and $\boldsymbol{\phi}$), this enables us to obtain $\boldsymbol{g}_{h,1}^{\pi}$ via backward induction from $t=h-1$ to $t=1$. Let $\widehat{\boldsymbol{g}}_{h,t}^{\pi_k}$ denote an estimate of $\boldsymbol{g}_{h,t}^{\pi_k}$ at a step $t<h$, with $\widehat{\boldsymbol{g}}_{h,h}^{\pi_k}=\boldsymbol{g}_{h,h}^{\pi_k}$. At each step of the backward induction, we can obtain $\widehat{\boldsymbol{g}}_{h,t}^{\pi_k}$ using~\cref{eq:backward_feature_recursion}, which requires us to estimate $P^\star\widehat{\boldsymbol{g}}_{h,t+1}^{\pi_k}$. To this end, in~\cref{lem:bilinear_bfr_linearity}, we show that $P^\star\widehat{\boldsymbol{g}}_{h,t+1}^{\pi_k}$ is linear in the features. That is, $P^\star\widehat{\boldsymbol{g}}_{h,t+1}^{\pi_k}(s,a)=\mathbf{B}^\star(\widehat{\boldsymbol{g}}_{h,t+1}^{\pi_k})^\top\boldsymbol{\phi}(s,a)$, where $\mathbf{B}^\star(\widehat{\boldsymbol{g}}_{h,t+1}^{\pi_k})\coloneq\mathbf{M}^\star\sum_{s'}\boldsymbol{\psi}(s')\widehat{\boldsymbol{g}}_{h,t+1}^{\pi_k}(s')^\top\in\mathbb{R}^{d_\phi\times d_\phi}$. Moreover, in~\cref{lem:bilinear_bfr_response}, we show that, for any fixed function $\boldsymbol{g}\colon \mathcal{S}\to\mathbb{R}^{d_\phi}$, $\boldsymbol{g}(s')$, where $s'\sim P^\star(\cdot\mid s,a)$, is an unbiased estimator of $P^\star\boldsymbol{g}(s,a)$. This motivates us to estimate $\mathbf{B}^\star(\widehat{\boldsymbol{g}}_{h,t+1}^{\pi_k})$ through the following vector-valued ridge regression for every step $h$ and every $t<h$ in the backward induction:
\begin{align}
\widehat{\mathbf{B}}_{h,t}^{\pi_k}\coloneq\operatorname*{arg\,min}_{\mathbf{B}\in\mathbb{R}^{d_\phi\times d_\phi}}\biggl\{\sum_{\tau_i\in\mathcal{D}_k}\|\widehat{\boldsymbol{g}}_{h,t+1}^{\pi_k}(s_{i,t+1})-\mathbf{B}^\top\boldsymbol{\phi}_{i,t}\|_2^2+\|\mathbf{B}\|_F^2\biggr\}. \label{eq:bfr_ridge_objective}
\end{align}
After obtaining $\widehat{\mathbf{B}}_{h,t}^{\pi_k}$, an estimate of $\mathbf{B}^\star(\widehat{\boldsymbol{g}}_{h,t+1}^{\pi_k})$, we can estimate $P^\star\widehat{\boldsymbol{g}}_{h,t+1}^{\pi_k}$ by $(\widehat{\mathbf{B}}_{h,t}^{\pi_k})^\top\boldsymbol{\phi}$, which can then be used to compute $\widehat{\boldsymbol{g}}_{h,t}^{\pi_k}$ via~\cref{eq:backward_feature_recursion}. This completes the backward induction at step $t$.

\begin{algorithm2e}[H]
\caption{Backward Feature Regression (\BFR{})} \label{alg:rfe}
\begin{algorithmic}[1]
\State \textbf{Input:} policy $\pi_k$, model core $\mathbf{M}_k$, replay dataset $\mathcal{D}_k=\{\tau_i\}_{i=1}^k$, projection radius $R_\phi$.
\State $\boldsymbol{\Gamma}_{k,t}\gets I+\sum_{i=1}^k\boldsymbol{\phi}_{i,t}\boldsymbol{\phi}_{i,t}^\top$ for $t\in[H-1]$.
\State $\overline{\boldsymbol{\phi}}_{\pi_k,h}(s)\gets\sum_a\pi_{k,h}(a\mid s)\boldsymbol{\phi}(s,a)$ for $h\in[H-1]$, for every queried $s$. \label{line:rfe_policy_feature}
\For{$h=1,\ldots,H-1$}
    \State $\widehat{\boldsymbol{g}}_{h,h}^{\pi_k}\gets\overline{\boldsymbol{\phi}}_{\pi_k,h}$. \label{line:rfe_base_target}
    \For{$t=h-1,\ldots,1$}
        \State $\widehat{\mathbf{B}}_{h,t}^{\pi_k}\gets\boldsymbol{\Gamma}_{k,t}^{-1}\sum_{i=1}^k\boldsymbol{\phi}_{i,t}\widehat{\boldsymbol{g}}_{h,t+1}^{\pi_k}(s_{i,t+1})^\top$. \label{line:rfe_regression}
        \State $\widehat{\boldsymbol{q}}_{h,t}^{\pi_k}(s,a)\gets\operatorname{Proj}_{\mathbb{B}_2(R_\phi)}((\widehat{\mathbf{B}}_{h,t}^{\pi_k})^\top\boldsymbol{\phi}(s,a))$, for every queried $(s,a)$. \label{line:rfe_projection}
        \State $\widehat{\boldsymbol{g}}_{h,t}^{\pi_k}(s)\gets\sum_a\pi_{k,t}(a\mid s)\widehat{\boldsymbol{q}}_{h,t}^{\pi_k}(s,a)$, for every queried $s$. \label{line:rfe_policy_average}
    \EndFor
    \State $\widehat{\boldsymbol{\omega}}_{k,h}\gets k^{-1}\sum_{i=1}^k\widehat{\boldsymbol{g}}_{h,1}^{\pi_k}(s_{i,1})$. \label{line:rfe_omega}
\EndFor
\State Construct $\widehat{\mu}_k,\widehat{R}_k$ using~\cref{eq:bilinear_occupancy_reconstruction,eq:pseudo_reward_estimate}, for every queried $(h,s,a)$. \label{line:rfe_initial_occupancy} \label{line:rfe_mu} \label{line:rfe_reward}
\State \Return\quad $\widehat{R}_k$.
\end{algorithmic}
\end{algorithm2e}
We describe the complete procedure in~\cref{alg:rfe}. In~\cref{line:rfe_policy_feature,line:rfe_base_target}, we initialize $\widehat{\boldsymbol{g}}_{h,h}^{\pi_k}=\boldsymbol{g}_{h,h}^{\pi_k}$ for each step $h$. In~\cref{line:rfe_regression}, we solve~\cref{eq:bfr_ridge_objective} in closed form to obtain $\widehat{\mathbf{B}}_{h,t}^{\pi_k}$. However, the norm of $(\widehat{\mathbf{B}}_{h,t}^{\pi_k})^\top\boldsymbol{\phi}(s,a)$ may fall outside the valid range $[0,C_\phi\sqrt{d_\phi}]$ (since $\|\boldsymbol{\phi}(s,a)\|_2\leq C_\phi\sqrt{d_\phi}$). Analogous to clipping in \OPE{} (\cref{line:clipped_Q} of \cref{alg:ope}), we therefore project $(\widehat{\mathbf{B}}_{h,t}^{\pi_k})^\top\boldsymbol{\phi}(s,a)$ onto the closed Euclidean ball $\mathbb{B}_2(R_\phi)$ with $R_\phi=C_\phi\sqrt{d_\phi}$ to obtain $\widehat{\boldsymbol{q}}_{h,t}^{\pi_k}$ (\cref{line:rfe_projection}). In~\cref{line:rfe_policy_average}, we use~\cref{eq:backward_feature_recursion} with $\widehat{\boldsymbol{q}}_{h,t}^{\pi_k}$ to obtain $\widehat{\boldsymbol{g}}_{h,t}^{\pi_k}$. We repeat the above steps for $t=h-1,\ldots,1$. After reaching $t=1$, since every initial state $s_{i,1}$ in $\mathcal{D}_k$ is drawn from $\rho$, we estimate $\boldsymbol{\omega}_h^{\pi_k}$ by averaging $\widehat{\boldsymbol{g}}_{h,1}^{\pi_k}$ over the initial states, i.e., $\widehat{\boldsymbol{\omega}}_{k,h}=k^{-1}\sum_{i=1}^k\widehat{\boldsymbol{g}}_{h,1}^{\pi_k}(s_{i,1})$ (\cref{line:rfe_omega}). Finally, in~\cref{line:rfe_reward}, using $\widehat{\boldsymbol{\omega}}_{k,h}$ and $\mathbf{M}_k$, we construct $\widehat{\mu}_k$ as in~\cref{eq:bilinear_occupancy_reconstruction} and compute $\widehat{R}_k$ via~\cref{eq:pseudo_reward_estimate}.
Note that, for each step $h\in[H-1]$, \cref{alg:rfe} performs $h-1$ backward regressions in~\cref{line:rfe_regression}. Hence, the computational cost of~\cref{alg:rfe} at iteration $k$ is $\mathcal{O}(H^2(k|\mathcal{A}|d_\phi^2+d_\phi^3))$, counting each evaluation of the features and the policy as one operation, independent of $|\mathcal{S}|$.

\paragraph{Projected Natural Policy Gradient}
Since $\OCUPOOFF$ reuses the same trajectories across iterations, the policies it selects depend on the replay data, and its estimates must remain accurate for every policy it may produce. However, the \NPG{} update in~\cref{eq:npg_update} accumulates all past value estimates and can therefore produce increasingly complex policies. To this end, $\OCUPOOFF$ uses projected \NPG{}~\citep[\PNPG{};][]{lin2026optimistic}, which keeps every policy in a given policy class $\Pi$. For a specific projection operator $\operatorname{Proj}_\Pi$, \PNPG{} performs the ordinary \NPG{} update and then projects the result onto $\Pi$ at each iteration,
\begin{align}
\pi_{k+1/2,h}(a\mid s)=\frac{\pi_{k,h}(a\mid s)\exp(\eta\widehat{Q}_{k,h}(s,a))}{\sum_{a'}\pi_{k,h}(a'\mid s)\exp(\eta\widehat{Q}_{k,h}(s,a'))},\qquad\pi_{k+1}=\operatorname{Proj}_\Pi(\pi_{k+1/2}). \label{eq:pnpg_update}
\end{align}
In~\cref{app:pnpg_update}, we instantiate $\Pi$ with log-linear policies and design the corresponding $\operatorname{Proj}_\Pi$. This projection fits the logits on a small coreset of state-action pairs, which depends only on the policy features and is constructed once before learning, without interacting with the environment. Constructing it may require a pass over $\mathcal{S}\times\mathcal{A}$ or an oracle that maximizes a quadratic form of the policy features (\cref{app:pnpg_update}). This one-time cost does not grow with $K$.

\subsection{Sample Complexity} \label{subsec:off_policy_sample_complexity}

Because of the projection, we cannot apply the regret bound of OMD (\cref{thm:npg_regret}) to \PNPG{} directly. In order to bound the NPG regret, we instead assume that \PNPG{} obeys the following assumption. 
\begin{assumption}[Regret up to Projection Error] \label{asm:pnpg}
There exists a constant $\varepsilon_{\mathrm{proj}}\geq0$ such that, for every comparator policy $u$ and every sequence of value estimates generated by \cref{alg:curl_framework}, \PNPG{} satisfies $\sum_{k,h}\mathbb{E}_{d_h^u}\langle \widehat{Q}_{k,h},u_h-\pi_{k,h}\rangle\leq\mathcal{O}(H^2G\sqrt{K\log|\mathcal{A}|}+K\varepsilon_{\mathrm{proj}})$.
\end{assumption}
This assumption states that \PNPG{} admits the OMD regret bound in~\cref{eq:on_oracle_outline} up to an additive constant $\varepsilon_{\mathrm{proj}}$ per iteration, where $\varepsilon_{\mathrm{proj}}$ accounts for the projection error.
We also impose the following assumption on the given policy class $\Pi$.
\begin{assumption}[Covering Number] \label{asm:policy_class}
For every $\varepsilon\in(0,1]$, $\log\mathcal{N}(\varepsilon,\Pi,\mathrm{dist}_\Pi)\leq\mathcal{O}(d_\Pi\log(1+K/\varepsilon))$, where $d_\Pi\ll\min\{K,|\mathcal{S}|\}$ and $\mathrm{dist}_\Pi(\pi,\pi')\coloneq\max_{h\in[H]}\sup_{s\in\mathcal{S}}\|\pi_h(\cdot\mid s)-\pi'_h(\cdot\mid s)\|_1$.
\end{assumption}
In~\cref{app:pnpg_guarantees}, we verify~\cref{asm:pnpg,asm:policy_class} for log-linear policies and control the projection error $\varepsilon_{\mathrm{proj}}$ (\cref{lem:off_projection_error,prop:pnpg_oracle}). Given the above assumptions, the sample complexity of $\OCUPOOFF$ is stated in the following theorem.
\begin{theorem}[Sample Complexity of $\OCUPOOFF$] \label{thm:off_policy}
Suppose that the MDP is a bilinear MDP (\cref{def:bilienar_mdp}) and that~\cref{asm:lipschitzness_and_smoothness,asm:coordinate_separability,asm:pnpg,asm:policy_class} hold. Running $\OCUPOOFF$ for $K$ iterations with appropriate choices of parameters yields that with probability at least $1-\delta$,
\begin{align}
\operatorname{Sub\text{-}Opt}(\overline{\pi}_K)&\leq\widetilde{\mathcal{O}}\left(\frac{H^2\Bigl[G\sqrt{d_\phi(d_\phi^2+d_\psi+d_\Pi)}+Ld_\phi^{3/2}\sqrt{d_\phi^2+d_\Pi}\Bigr]}{\sqrt K}+\varepsilon_{\mathrm{proj}}\right).
\end{align}
Hence, for any $\epsilon>0$, $\OCUPOOFF$ with $K=\widetilde{\mathcal{O}}(H^4[G^2d_\phi(d_\phi^2+d_\psi+d_\Pi)+L^2d_\phi^3(d_\phi^2+d_\Pi)]/\epsilon^2)$ yields an $\mathcal{O}(\epsilon+\varepsilon_{\mathrm{proj}})$-optimal mixture policy, and therefore requires $K$ trajectories, one per iteration.
\end{theorem}
\cref{thm:off_policy} shows that $\OCUPOOFF$ improves the sample complexity of $\OCUPOON$ from $\widetilde{\mathcal{O}}(\epsilon^{-4})$ to $\widetilde{\mathcal{O}}(\epsilon^{-2})$, up to the projection error $\varepsilon_{\mathrm{proj}}$. To the best of our knowledge, $\OCUPOOFF$ is the first computationally efficient algorithm that achieves this sample complexity for CURL in bilinear MDPs without any coverage assumption. This dependence on $\epsilon$ matches the guarantees of policy optimization methods for standard RL in linear MDPs~\citep{sherman2023rate,cassel2024warm,lin2026optimistic} and of~\citet{moreno2025online} for CURL in tabular MDPs. See~\cref{tab:standard_rl_comparison} for a detailed comparison in standard RL.

\subsection{\texorpdfstring{Proof Sketch of \cref{thm:off_policy}}{Proof Sketch of Theorem \ref*{thm:off_policy}}} \label{subsec:off_policy_proof_outline}

We follow the proof structure of~\cref{thm:on_policy} and use the same regret decomposition in~\cref{eq:regret_decomposition}.

\paragraph{Bounding Term (I)}
By~\cref{asm:pnpg} with $u=\pi^\star$, which \PNPG{} satisfies with $\eta=\sqrt{2\log|\mathcal{A}|/(KH^2G^2)}$ (\cref{prop:pnpg_oracle}), we obtain
\begin{align}
\mathrm{Term~(I)}\leq\mathcal{O}\Bigl(H^2G\sqrt{K\log|\mathcal{A}|}+K\varepsilon_{\mathrm{proj}}\Bigr). \label{eq:off_term1_outline}
\end{align}

\paragraph{The Good Event}
To bound Terms (II) and (III), we use the good event $\mathcal{E}$ in~\cref{def:good_event} with the radii $\beta_k^{\mathrm{Off}}$, $\Delta_{k,h}^{\mathrm{Off}}$, and $\alpha_h^{\mathrm{Off}}$ for $\mathcal{E}_{\mathrm{P}}$, $\mathcal{E}_{\mathrm{R}}$, and $\mathcal{E}_{\mathrm{V}}$, respectively.

\paragraph{Transition Model Error ($\mathcal{E}_{\mathrm{P}}$)}
As in $\OCUPOON$, in~\cref{lem:off_model_event_main}, we use matrix concentration (\cref{lem:lin_matrix_self_normalized}) to show that $\mathcal{E}_{\mathrm{P}}$ holds with probability at least $1-\delta/4$ with $\beta_k^{\mathrm{Off}}=\widetilde{\mathcal{O}}(d_\phi)$.

\paragraph{Pseudo-Reward Error ($\mathcal{E}_{\mathrm{R}}$)}
As in $\OCUPOON$, the pseudo-reward error is zero at $h=1$, and we set $\Delta_{k,1}^{\mathrm{Off}}\coloneq0$. For $h\in\{2,\ldots,H\}$, we use the occupancy decomposition in~\cref{eq:bilinear_occupancy_decomposition} and obtain that under the event $\mathcal{E}_{\mathrm{P}}$,
\begin{align}
\|\widehat{R}_{k,h}-R_{k,h}\|_\infty&\leq LC_\psi\biggl[\sqrt{\beta_k^{\mathrm{Off}}}\,\|\boldsymbol{\omega}_{h-1}^{\pi_k}\|_{\boldsymbol{\Lambda}_k^{-1}}+\Bigl(\sqrt{\beta_k^{\mathrm{Off}}}+C_{\mathbf{M}}\sqrt{d_\phi}\Bigr)\|\widehat{\boldsymbol{\omega}}_{k,h-1}-\boldsymbol{\omega}_{h-1}^{\pi_k}\|_2\biggr]. \label{eq:off_reward_omega_decomposition}
\end{align}
The first term in~\cref{eq:off_reward_omega_decomposition} is related to the transition model error. To bound it, we use Jensen's inequality to show that $\|\boldsymbol{\omega}_{h-1}^{\pi_k}\|_{\boldsymbol{\Lambda}_k^{-1}}\leq\widehat{\ell}_{k,h-1}$, where $\widehat{\ell}_{k,t}\coloneq\mathbb{E}_{(s,a)\sim\mu_t^{\pi_k}}\|\boldsymbol{\phi}(s,a)\|_{\boldsymbol{\Gamma}_{k-1,t}^{-1}}$. The second term depends on the error in estimating $\boldsymbol{\omega}_{h-1}^{\pi_k}$. Unlike in $\OCUPOON$, we cannot control this error through vector concentration, since the trajectories in $\mathcal{D}_k$ were collected by past policies rather than by $\pi_k$. Instead, for every step $j<H$, we use $\widehat{\boldsymbol{\omega}}_{k,j}=k^{-1}\sum_{i=1}^k\widehat{\boldsymbol{g}}_{j,1}^{\pi_k}(s_{i,1})$ (\cref{line:rfe_omega} of \cref{alg:rfe}) and the triangle inequality to decompose
\begin{align}
\bigl\lVert \widehat{\boldsymbol{\omega}}_{k,j}-\boldsymbol{\omega}_j^{\pi_k} \bigr\rVert_2&\leq\biggl\lVert \frac1k\sum_{i=1}^k\widehat{\boldsymbol{g}}_{j,1}^{\pi_k}(s_{i,1})-\mathbb{E}_\rho[\widehat{\boldsymbol{g}}_{j,1}^{\pi_k}(s_1)] \biggr\rVert_2+\bigl\lVert \mathbb{E}_\rho[\widehat{\boldsymbol{g}}_{j,1}^{\pi_k}(s_1)]-\boldsymbol{\omega}_j^{\pi_k} \bigr\rVert_2. \label{eq:off_feature_error_outline}
\end{align}
The first term in~\cref{eq:off_feature_error_outline} is the error of averaging $\widehat{\boldsymbol{g}}_{j,1}^{\pi_k}$ over the $k$ initial states in $\mathcal{D}_k$. Using vector concentration (\cref{lem:lin_matrix_self_normalized}) over these initial states, we bound it by $\gamma_\rho/\sqrt k$, where $\gamma_\rho$ is the constant of this concentration bound. The second term is the error accumulated by the backward regressions. For each regression at step $t<j$ (\cref{line:rfe_regression} of \cref{alg:rfe}), we use self-normalized concentration (\cref{lem:lin_uniform_matrix_regression}) to show that $\|\boldsymbol{\Gamma}_{k,t}^{1/2}(\widehat{\mathbf{B}}_{j,t}^{\pi_k}-\mathbf{B}^\star(\widehat{\boldsymbol{g}}_{j,t+1}^{\pi_k}))\|_F\leq\kappa$, where $\kappa$ is the constant of this concentration bound. Propagating these errors through~\cref{eq:backward_feature_recursion} from $t=j-1$ to $t=1$, we bound the second term by $\kappa\sum_{t=1}^{j-1}\widehat{\ell}_{k,t}$. Substituting these bounds with $j=h-1$ into~\cref{eq:off_reward_omega_decomposition}, we show in~\cref{lem:off_reward_error_main} that, under the event $\mathcal{E}_{\mathrm{P}}$, with probability at least $1-\delta/4$,
\begin{align}
\|\widehat{R}_{k,h}-R_{k,h}\|_\infty\leq\Delta_{k,h}^{\mathrm{Off}}&\coloneq LC_\psi\Biggl[\sqrt{\beta_k^{\mathrm{Off}}}\,\widehat{\ell}_{k,h-1}+\Bigl(\sqrt{\beta_k^{\mathrm{Off}}}+C_{\mathbf{M}}\sqrt{d_\phi}\Bigr)\biggl(\frac{\gamma_\rho}{\sqrt k}+\kappa\sum_{t=1}^{h-2}\widehat{\ell}_{k,t}\biggr)\Biggr]. \label{eq:off_reward_outline}
\end{align}

\paragraph{Value Estimation Error ($\mathcal{E}_{\mathrm{V}}$)}
As in $\OCUPOON$, establishing $\mathcal{E}_{\mathrm{V}}$ requires covering the value functions that \OPE{} can produce. Unlike in $\OCUPOON$, the regression in~\cref{eq:ope_ridge_objective} now uses the replay data $\mathcal{D}_k$, which also contains the trajectories used to compute $\pi_k$. Hence, $\widehat{V}_{k,h+1}(s)=\sum_a\pi_{k,h+1}(a\mid s)\widehat{Q}_{k,h+1}(s,a)$ depends on the data through $\pi_k$ as well, and covering these value functions also requires covering the policy class $\Pi$. Given~\cref{asm:policy_class}, we show in~\cref{lem:off_covering_numbers_formal} that the logarithmic covering number of this class is $\widetilde{\mathcal{O}}(d_\phi^2+d_\psi+d_\Pi)$. With $\alpha_h^{\mathrm{Off}}=\widetilde{\mathcal{O}}((H-h)G\sqrt{d_\phi^2+d_\psi+d_\Pi})$, we show in~\cref{lem:off_value_event_main} that $\mathcal{E}_{\mathrm{V}}$ holds with probability at least $1-\delta/4$.

\paragraph{Bounding Terms (II) and (III) under the Good Event}
On the good event $\mathcal{E}$ with the radii $\beta_k^{\mathrm{Off}}$, $\Delta_{k,h}^{\mathrm{Off}}$, and $\alpha_h^{\mathrm{Off}}$, \cref{lem:on_bellman_error} shows that~\cref{eq:on_bellman_optimism,eq:on_bellman_cost} hold with $\Delta_{k,h}=\Delta_{k,h}^{\mathrm{Off}}$ and $\alpha_{k,h}=\alpha_h^{\mathrm{Off}}$. Summing over $k \in [K]$ and $h \in [H]$, we obtain
\begin{align}
\MoveEqLeft
\mathrm{Term~(II)}+\mathrm{Term~(III)}\leq2\sum_{k=1}^K\sum_{h=1}^H\Delta_{k,h}^{\mathrm{Off}}+2\sum_{k=1}^K\sum_{h=1}^{H-1}\alpha_h^{\mathrm{Off}}\widehat{\ell}_{k,h}\nonumber\\
&\leq\widetilde{\mathcal{O}}\biggl(\Bigl(LH\kappa\bigl(\sqrt{\beta^{\mathrm{Off}}}+C_{\mathbf{M}}\sqrt{d_\phi}\bigr)+\alpha^{\mathrm{Off}}\Bigr)\sum_{k=1}^K\sum_{h=1}^{H-1}\widehat{\ell}_{k,h}+LH\gamma_\rho\bigl(\sqrt{\beta^{\mathrm{Off}}}+C_{\mathbf{M}}\sqrt{d_\phi}\bigr)\sqrt K\biggr)\nonumber\\
&\leq\widetilde{\mathcal{O}}\biggl(H^2\Bigl[G\sqrt{d_\phi(d_\phi^2+d_\psi+d_\Pi)}+Ld_\phi^{3/2}\sqrt{d_\phi^2+d_\Pi}\Bigr]\sqrt K\biggr). \label{eq:off_uncertainty_outline}
\end{align}
The second inequality substitutes $\Delta_{k,h}^{\mathrm{Off}}$ from~\cref{eq:off_reward_outline}, where $\beta^{\mathrm{Off}}\coloneq\max_{k\in[K]}\beta_k^{\mathrm{Off}}$ and $\alpha^{\mathrm{Off}}\coloneq\max_{h<H}\alpha_h^{\mathrm{Off}}$. For the last inequality, we use the elliptical potential lemma (\cref{lem:elliptical_potential}) to show that, with probability at least $1-\delta/4$, $\sum_{k=1}^K\sum_{h=1}^{H-1}\widehat{\ell}_{k,h}\leq\widetilde{\mathcal{O}}(H\sqrt{d_\phi K})$. Finally, we substitute the orders of the radii and concentration constants established above, namely $\sqrt{\beta^{\mathrm{Off}}}=\widetilde{\mathcal{O}}(\sqrt{d_\phi})$, $\gamma_\rho,\kappa=\widetilde{\mathcal{O}}(\sqrt{d_\phi(d_\phi^2+d_\Pi)})$, and $\alpha^{\mathrm{Off}}=\widetilde{\mathcal{O}}(HG\sqrt{d_\phi^2+d_\psi+d_\Pi})$.

\paragraph{Putting Everything Together}
We combine~\cref{eq:off_term1_outline,eq:off_uncertainty_outline} and take a union bound to show that, with probability at least $1-\delta$, $\operatorname{Sub\text{-}Opt}(\overline{\pi}_K)\leq\operatorname{Reg}_K/K$ satisfies the bound in~\cref{thm:off_policy}. Therefore, running $K=\widetilde{\mathcal{O}}(H^4[G^2d_\phi(d_\phi^2+d_\psi+d_\Pi)+L^2d_\phi^3(d_\phi^2+d_\Pi)]/\epsilon^2)$ iterations of $\OCUPOOFF$, each collecting one trajectory, yields an $\mathcal{O}(\epsilon+\varepsilon_{\mathrm{proj}})$-optimal mixture policy. See~\cref{app:ucgu_off} for the complete proof.

\section{Extension to Linear MDPs} \label{section:ocupo-lin}

In this section, we consider linear MDPs, a more general setting than bilinear MDPs, and instantiate $\OCUPO$ in this setting (referred to as $\OCUPOLIN$). We first introduce the setting in~\cref{subsec:linear_mdp}, describe the algorithm design in~\cref{subsec:lin_algorithm}, and then analyze its sample complexity in~\cref{subsec:lin_sample_complexity,subsec:lin_proof_outline}.

\subsection{Linear MDPs} \label{subsec:linear_mdp}

\begin{definition}[Linear MDP; {\citealp{jin2023provably}}] \label{def:linear_mdp}
An MDP is a linear MDP if there exist a feature map $\boldsymbol{\phi}\colon \mathcal{S}\times\mathcal{A}\to\mathbb{R}^{d_\phi}$ and a matrix $\mathbf{T}^\star\in\mathbb{R}^{d_\phi\times|\mathcal{S}|}$ such that $P^\star(s^\prime\mid s,a)=\boldsymbol{\phi}(s,a)^\top\mathbf{T}^\star(s^\prime)$, where, for a matrix $\mathbf{T}$ with columns indexed by states, $\mathbf{T}(s)$ denotes its column indexed by $s$. Moreover, there exist positive constants $C_\phi$ and $C_{\mathbf{T}}$ such that the following feature regularity conditions hold: (i) for every $(s,a)\in\mathcal{S}\times\mathcal{A}$, $\|\boldsymbol{\phi}(s,a)\|_2\leq C_\phi\sqrt{d_\phi}$, and (ii) for every function $g\colon \mathcal{S}\to\mathbb{R}$, $\|\sum_{s\in\mathcal{S}}\mathbf{T}^\star(s)g(s)\|_2\leq C_{\mathbf{T}}\sqrt{d_\phi}\|g\|_\infty$.
\end{definition}
We assume that $\boldsymbol{\phi}$ is available to the agent, whereas $\mathbf{T}^\star$ is unknown. Then, $P^\star(\cdot\mid s,a)=(\mathbf{T}^\star)^\top\boldsymbol{\phi}(s,a)$. A bilinear MDP (\cref{def:bilienar_mdp}) is a linear MDP with $\mathbf{T}^\star=\mathbf{M}^\star\boldsymbol{\Psi}$, where $\boldsymbol{\Psi}\in\mathbb{R}^{d_\psi\times|\mathcal{S}|}$ is the matrix that has $\boldsymbol{\psi}(s)$ as each column $s$. In bilinear MDPs, $\boldsymbol{\Psi}$ is known, and estimating $\mathbf{T}^\star$ can be reduced to estimating $\mathbf{M}^\star\in\mathbb{R}^{d_\phi\times d_\psi}$. However, $\mathbf{T}^\star$ is unknown in linear MDPs, making it a more challenging problem.

\subsection{Algorithm Design} \label{subsec:lin_algorithm}

We instantiate $\OCUPO$ in linear MDPs by modifying $\OCUPOOFF$. $\OCUPOOFF$ requires the bilinear structure in only two places: estimating the transition model (\cref{eq:bilinear_tme_estimator}) and constructing $\widehat{\mu}_k$ from $\mathbf{M}_k$ (\cref{line:rfe_reward} of \cref{alg:rfe}). In particular, \OPE{} does not use the estimated transition model directly. Hence, once $\widehat{R}_k$ is constructed, we can reuse \OPE{} (\cref{alg:ope}) with a different confidence radius. Therefore, we only need to instantiate the transition model estimation, redesign~\cref{line:rfe_reward} of~\cref{alg:rfe}, and adjust the confidence radius.

\paragraph{Sparse Transition Estimation}
In order to estimate $\mathbf{T}^\star$, we denote by $\boldsymbol{e}_s\in\mathbb{R}^{|\mathcal{S}|}$ the one-hot vector of a state $s$. In~\cref{lem:linear_regression_response}, we show that, for a next state $s'\sim P^\star(\cdot\mid s,a)$, $\boldsymbol{e}_{s'}$ is an unbiased estimator of $P^\star(\cdot\mid s,a)=(\mathbf{T}^\star)^\top\boldsymbol{\phi}(s,a)$. This motivates us to develop sparse transition estimation (\STE{}), which solves the following ridge regression over all the transitions in $\mathcal{D}_k$:
\begin{align}
\widehat{\mathbf{T}}_k\coloneq\operatorname*{arg\,min}_{\mathbf{T}\in\mathbb{R}^{d_\phi\times|\mathcal{S}|}}\biggl\{\sum_{i=1}^k\sum_{h=1}^H\bigl\lVert \mathbf{T}^\top\boldsymbol{\phi}_{i,h}-\boldsymbol{e}_{s_{i,h+1}} \bigr\rVert_2^2+\lVert \mathbf{T} \rVert_F^2\biggr\}=\boldsymbol{\Lambda}_k^{-1}\sum_{i=1}^k\sum_{h=1}^H\boldsymbol{\phi}_{i,h}\boldsymbol{e}_{s_{i,h+1}}^\top, \label{eq:ste_estimator}
\end{align}
where $\boldsymbol{\Lambda}_k\coloneq I+\sum_{i=1}^k\sum_{h=1}^H\boldsymbol{\phi}_{i,h}\boldsymbol{\phi}_{i,h}^\top$ is the Gram matrix. Although $\widehat{\mathbf{T}}_k$ is a matrix with $|\mathcal{S}|$ columns, solving for $\widehat{\mathbf{T}}_k$ does not require $\mathcal{O}(|\mathcal{S}|)$ computation, since $\widehat{\mathbf{T}}_k$ has at most $kH$ nonzero columns. To see this, we denote by $\mathcal{S}_k^+\coloneq\{s_{i,h+1}\mid i\in[k],h\in[H]\}$ the set of next states that appear in $\mathcal{D}_k$. Since $\boldsymbol{e}_{s_{i,h+1}}$ is nonzero only at $s_{i,h+1}$, we obtain that
\begin{align}
\widehat{\mathbf{T}}_k(s)=\begin{cases}\boldsymbol{\Lambda}_k^{-1}\sum_{(i,h):s_{i,h+1}=s}\boldsymbol{\phi}_{i,h},&s\in\mathcal{S}_k^+,\\
\boldsymbol{0},&s\notin\mathcal{S}_k^+.\end{cases} \label{eq:ste_columns}
\end{align}
Hence, we only need to store $\widehat{\mathbf{T}}_k(s)$ for $s\in\mathcal{S}_k^+$, and we can evaluate $\widehat{\mathbf{T}}_k(s)$ at any other state by returning zero, without enumerating the state space. Since every transition in $\mathcal{D}_k$ contributes to exactly one column, the computational cost of \STE{} at iteration $k$ is $\mathcal{O}(kHd_\phi^2+d_\phi^3)$, counting each feature evaluation as one operation. See~\cref{alg:tme_linear} in~\cref{app:lin_tme_details} for the pseudocode.

\paragraph{Pseudo-Reward Estimation}
We reuse \BFR{} (\cref{alg:rfe}) to obtain $\widehat{\boldsymbol{\omega}}_{k,h}$ (\cref{line:rfe_omega}). The only line that we need to change is \cref{line:rfe_reward}, which constructs $\widehat{\mu}_k$. Analogous to~\cref{eq:bilinear_occupancy_decomposition}, in~\cref{lem:linear_occupancy_measure}, we show that, for every policy $\pi$, $h\geq2$, and $(s,a)\in\mathcal{S}\times\mathcal{A}$,
\begin{align}
\mu_h^\pi(s,a)=\pi_h(a\mid s)(\boldsymbol{\omega}_{h-1}^\pi)^\top\mathbf{T}^\star(s). \label{eq:lin_occupancy_decomposition}
\end{align}
Replacing $\mathbf{T}^\star(s)$ by $\widehat{\mathbf{T}}_k(s)$, we obtain the estimated occupancy measure
\begin{align}
\widehat{\mu}_{k,h}(s,a)&=\begin{cases}\rho(s)\pi_{k,1}(a\mid s),&h=1,\\
\operatorname{Clip}(\pi_{k,h}(a\mid s)\widehat{\boldsymbol{\omega}}_{k,h-1}^\top\widehat{\mathbf{T}}_k(s),[0,1]),&h\in\{2,\ldots,H\},\end{cases} \label{eq:lin_occupancy_reconstruction}
\end{align}
and construct $\widehat{R}_k$ using~\cref{eq:pseudo_reward_estimate}.

\paragraph{Optimistic Policy Evaluation}
We reuse \OPE{} (\cref{alg:ope}) to compute $\widehat{Q}_k$ from $\widehat{R}_k$ and $\mathcal{D}_k$, with the confidence radius specified in~\cref{subsec:lin_proof_outline}.

\paragraph{Projected Natural Policy Gradient}
We update the policy by \PNPG{} in~\cref{eq:pnpg_update} as in $\OCUPOOFF$, under the same assumptions (\cref{asm:pnpg,asm:policy_class}).

\subsection{Sample Complexity} \label{subsec:lin_sample_complexity}

We now state the sample complexity of $\OCUPOLIN$ in the following theorem.
\begin{theorem}[Sample Complexity of $\OCUPOLIN$] \label{thm:lin_off}
Suppose that the MDP is a linear MDP (\cref{def:linear_mdp}) and that~\cref{asm:lipschitzness_and_smoothness,asm:coordinate_separability,asm:pnpg,asm:policy_class} hold. Running $\OCUPOLIN$ for $K$ iterations with appropriate choices of parameters yields that with probability at least $1-\delta$,
\begin{align}
\operatorname{Sub\text{-}Opt}(\overline{\pi}_K)&\leq\widetilde{\mathcal{O}}\left(\frac{H^2\Bigl[G\sqrt{d_\phi(d_\phi^2+d_\Pi)}+Ld_\phi^3\sqrt{d_\phi^2+d_\Pi}\Bigr]}{\sqrt K}+\varepsilon_{\mathrm{proj}}\right). \label{eq:lin_off_rate}
\end{align}
Hence, for any $\epsilon>0$, $\OCUPOLIN$ with $K=\widetilde{\mathcal{O}}(H^4[G^2d_\phi(d_\phi^2+d_\Pi)+L^2d_\phi^6(d_\phi^2+d_\Pi)]/\epsilon^2)$ yields an $\mathcal{O}(\epsilon+\varepsilon_{\mathrm{proj}})$-optimal mixture policy, and therefore requires $K$ trajectories, one per iteration.
\end{theorem}
\cref{thm:lin_off} shows that $\OCUPOLIN$ retains the $\widetilde{\mathcal{O}}(\epsilon^{-2})$ sample complexity in linear MDPs. To the best of our knowledge, this is the first sample complexity guarantee for CURL in linear MDPs. In standard RL, where $L=0$, \cref{thm:lin_off} recovers the $\epsilon^{-2}$ dependence of~\citet{lin2026optimistic}. See~\cref{tab:standard_rl_comparison} for a detailed comparison in standard RL.

\subsection{\texorpdfstring{Proof Sketch of \cref{thm:lin_off}}{Proof Sketch of Theorem \ref*{thm:lin_off}}} \label{subsec:lin_proof_outline}

We use the same regret decomposition as in~\cref{eq:regret_decomposition}. As in $\OCUPOOFF$, we apply~\cref{asm:pnpg} with $u=\pi^\star$ and the same step size to obtain
\begin{align}
\mathrm{Term~(I)}\leq\mathcal{O}\Bigl(H^2G\sqrt{K\log|\mathcal{A}|}+K\varepsilon_{\mathrm{proj}}\Bigr). \label{eq:lin_term1_outline}
\end{align}
It suffices to bound Terms (II) and (III).

\paragraph{The Good Event}
To this end, we define the good event $\mathcal{E}\coloneq\mathcal{E}_{\mathrm{P}}\cap\mathcal{E}_{\mathrm{R}}\cap\mathcal{E}_{\mathrm{V}}$ with radii $\beta^{\mathrm{Lin}}$, $\Delta_{k,h}^{\mathrm{Lin}}$, and $\alpha_{k,h}^{\mathrm{Lin}}$, where
\begin{align}
\mathcal{E}_{\mathrm{P}}&\coloneq\Bigl\{\forall k\in[K]\colon\bigl\lVert \boldsymbol{\Lambda}_k^{1/2}(\widehat{\mathbf{T}}_k-\mathbf{T}^\star) \bigr\rVert_F\leq\sqrt{\beta^{\mathrm{Lin}}}\Bigr\},\\
\mathcal{E}_{\mathrm{R}}&\coloneq\bigl\{\forall k\in[K],\ h\in[H]\colon\|\widehat{R}_{k,h}-R_{k,h}\|_\infty\leq\Delta_{k,h}^{\mathrm{Lin}}\bigr\},\\
\mathcal{E}_{\mathrm{V}}&\coloneq\Bigl\{\forall k\in[K],\ h<H\colon\bigl\lVert \widehat{\boldsymbol{v}}_{k,h}-\boldsymbol{v}^\star(\widehat{V}_{k,h+1}) \bigr\rVert_{\boldsymbol{\Gamma}_{k,h}}\leq\alpha_{k,h}^{\mathrm{Lin}}\Bigr\}.
\end{align}
This good event is analogous to~\cref{def:good_event}, with the only difference in $\mathcal{E}_{\mathrm{P}}$. Since a linear MDP has no transition core, we can only represent the transition model error by the estimation error of $\widehat{\mathbf{T}}_k$, instead of that of $\mathbf{M}_k$.

\paragraph{Transition Model Error ($\mathcal{E}_{\mathrm{P}}$)}
Using the closed form of $\widehat{\mathbf{T}}_k$ in~\cref{eq:ste_estimator} and the triangle inequality, we obtain
\begin{align}
\bigl\lVert \boldsymbol{\Lambda}_k^{1/2}(\widehat{\mathbf{T}}_k-\mathbf{T}^\star) \bigr\rVert_F\leq\biggl\lVert \boldsymbol{\Lambda}_k^{-1/2}\sum_{i=1}^k\sum_{h=1}^H\boldsymbol{\phi}_{i,h}\boldsymbol{y}_{i,h}^\top \biggr\rVert_F+\bigl\lVert \boldsymbol{\Lambda}_k^{-1/2}\mathbf{T}^\star \bigr\rVert_F, \label{eq:lin_model_error_outline}
\end{align}
where $\boldsymbol{y}_{i,h}\coloneq\boldsymbol{e}_{s_{i,h+1}}-P^\star(\cdot\mid s_{i,h},a_{i,h})$. The first term is the error in the observed transitions. Since $\boldsymbol{y}_{i,h}$ has zero conditional mean given $(s_{i,h},a_{i,h})$ and Euclidean norm at most $2$ regardless of $|\mathcal{S}|$, we control it by self-normalized concentration (\cref{lem:lin_matrix_self_normalized}). The second term is the contribution of regularization. We apply $\boldsymbol{\Lambda}_k\succeq I$ and the feature regularity in~\cref{def:linear_mdp} to obtain that $\|\boldsymbol{\Lambda}_k^{-1/2}\mathbf{T}^\star\|_F\leq\|\mathbf{T}^\star\|_F\leq C_{\mathbf{T}}\sqrt{d_\phi}$. Consequently, in~\cref{lem:lin_off_model_event}, we show that $\mathcal{E}_{\mathrm{P}}$ holds with probability at least $1-\delta/4$ with $\beta^{\mathrm{Lin}}=\widetilde{\mathcal{O}}((1+C_{\mathbf{T}}^2)d_\phi)$.

\paragraph{Pseudo-Reward Error ($\mathcal{E}_{\mathrm{R}}$)}
As in $\OCUPOOFF$, the pseudo-reward error is zero at $h=1$, and we set $\Delta_{k,1}^{\mathrm{Lin}}\coloneq0$. For $h\in\{2,\ldots,H\}$, a direct approach applies the smoothness of $f_h$ (\cref{asm:lipschitzness_and_smoothness}), the occupancy decomposition in~\cref{lem:linear_occupancy_measure}, the nonexpansiveness of clipping, and the triangle inequality to obtain
\begin{align}
\|\widehat{R}_{k,h}-R_{k,h}\|_\infty\leq L\|\widehat{\mu}_{k,h}-\mu_h^{\pi_k}\|_1\leq L\sum_{s\in\mathcal{S}}\Bigl|\widehat{\boldsymbol{\omega}}_{k,h-1}^\top\big(\widehat{\mathbf{T}}_k(s)-\mathbf{T}^\star(s)\big)\Bigr|+L\sum_{s\in\mathcal{S}}\Bigl|\big(\widehat{\boldsymbol{\omega}}_{k,h-1}-\boldsymbol{\omega}_{h-1}^{\pi_k}\big)^\top\mathbf{T}^\star(s)\Bigr|.
\end{align}
The first term sums the error of $\widehat{\mathbf{T}}_k(s)$ over all states and can therefore grow with $|\mathcal{S}|$. Instead, we use~\cref{asm:coordinate_separability} to bound each pseudo-reward coordinate by the error of the corresponding occupancy coordinate, which replaces the sums over $\mathcal{S}$ by suprema. Using~\cref{eq:lin_occupancy_decomposition,eq:lin_occupancy_reconstruction}, we obtain
\begin{align}
\|\widehat{R}_{k,h}-R_{k,h}\|_\infty&\leq L\sup_{s\in\mathcal{S}}\Bigl|\widehat{\boldsymbol{\omega}}_{k,h-1}^\top\big(\widehat{\mathbf{T}}_k(s)-\mathbf{T}^\star(s)\big)\Bigr|+L\sup_{s\in\mathcal{S}}\Bigl|\big(\widehat{\boldsymbol{\omega}}_{k,h-1}-\boldsymbol{\omega}_{h-1}^{\pi_k}\big)^\top\mathbf{T}^\star(s)\Bigr|. \label{eq:lin_reward_split_outline}
\end{align}
For the first term of~\cref{eq:lin_reward_split_outline}, which is related to the transition model error, we use $\mathcal{E}_{\mathrm{P}}$. In particular, the following derivation removes the supremum over $\mathcal{S}$ by bounding the error at every state by the transition model error in $\mathcal{E}_{\mathrm{P}}$:
\begin{align}
\MoveEqLeft L\sup_{s\in\mathcal{S}}\Bigl|\widehat{\boldsymbol{\omega}}_{k,h-1}^\top\big(\widehat{\mathbf{T}}_k(s)-\mathbf{T}^\star(s)\big)\Bigr| \leq L\|\widehat{\boldsymbol{\omega}}_{k,h-1}\|_{\boldsymbol{\Lambda}_k^{-1}}\sup_{s\in\mathcal{S}}\|\widehat{\mathbf{T}}_k(s)-\mathbf{T}^\star(s)\|_{\boldsymbol{\Lambda}_k}\nonumber\\
&\qquad \leq L\|\widehat{\boldsymbol{\omega}}_{k,h-1}\|_{\boldsymbol{\Lambda}_k^{-1}}\bigl\lVert \boldsymbol{\Lambda}_k^{1/2}(\widehat{\mathbf{T}}_k-\mathbf{T}^\star) \bigr\rVert_F\leq L\sqrt{\beta^{\mathrm{Lin}}}\|\widehat{\boldsymbol{\omega}}_{k,h-1}\|_{\boldsymbol{\Lambda}_k^{-1}},
\end{align}
where the first inequality follows from the Cauchy--Schwarz inequality, the second from~\cref{lem:lin_column_error}, and the last from the definition of $\mathcal{E}_{\mathrm{P}}$. For the second term of~\cref{eq:lin_reward_split_outline}, we directly apply the feature regularity in~\cref{def:linear_mdp}. Putting the two terms together yields
\begin{align}
\|\widehat{R}_{k,h}-R_{k,h}\|_\infty&\leq L\Bigl[\sqrt{\beta^{\mathrm{Lin}}}\,\|\boldsymbol{\omega}_{h-1}^{\pi_k}\|_{\boldsymbol{\Lambda}_k^{-1}}+\bigl(\sqrt{\beta^{\mathrm{Lin}}}+C_{\mathbf{T}}\sqrt{d_\phi}\bigr)\|\widehat{\boldsymbol{\omega}}_{k,h-1}-\boldsymbol{\omega}_{h-1}^{\pi_k}\|_2\Bigr]\nonumber\\
&\leq\Delta_{k,h}^{\mathrm{Lin}}\coloneq L\Biggl[\sqrt{\beta^{\mathrm{Lin}}}\,\widehat{\ell}_{k,h-1}+\bigl(\sqrt{\beta^{\mathrm{Lin}}}+C_{\mathbf{T}}\sqrt{d_\phi}\bigr)\biggl(\frac{\gamma_\rho}{\sqrt k}+\kappa\sum_{t=1}^{h-2}\widehat{\ell}_{k,t}\biggr)\Biggr], \label{eq:lin_reward_outline}
\end{align}
where the first inequality has the same form as~\cref{eq:off_reward_omega_decomposition} for $\OCUPOOFF$, and for the second inequality, we control both $\|\boldsymbol{\omega}_{h-1}^{\pi_k}\|_{\boldsymbol{\Lambda}_k^{-1}}$ and $\|\widehat{\boldsymbol{\omega}}_{k,h-1}-\boldsymbol{\omega}_{h-1}^{\pi_k}\|_2$ in the same way as in $\OCUPOOFF$.

\paragraph{Value Estimation Error ($\mathcal{E}_{\mathrm{V}}$)}
As in $\OCUPOOFF$, the bound on $\|\widehat{\boldsymbol{v}}_{k,h}-\boldsymbol{v}^\star(\widehat{V}_{k,h+1})\|_{\boldsymbol{\Gamma}_{k,h}}$ in $\mathcal{E}_{\mathrm{V}}$ is determined by $\|\sum_{i=1}^k\boldsymbol{\phi}_{i,h}[\widehat{V}_{k,h+1}(s_{i,h+1})-P^\star\widehat{V}_{k,h+1}(s_{i,h},a_{i,h})]\|_{\boldsymbol{\Gamma}_{k,h}^{-1}}$.
Since $\widehat{V}_{k,h+1}$ is computed from the same data, establishing $\mathcal{E}_{\mathrm{V}}$ requires bounding this quantity uniformly over the value functions that \OPE{} can produce, which in turn requires covering them. However, each $\widehat{V}_{k,h+1}$ depends on the columns of $\widehat{\mathbf{T}}_k$ through the pseudo-reward $\widehat{R}_{k,h+1}$. The number of nonzero columns grows with $k$, and covering them would introduce a dependence on the number of next states in $\mathcal{D}_k$. To address this issue, we replace $\widehat{\mathbf{T}}_k(s)$ by the true columns $\mathbf{T}^\star(s)$ in the pseudo-reward. Specifically, for every iteration $k$ and $h<H$, let $\mu_{k,h+1}^\circ(s,a)\coloneq\operatorname{Clip}(\pi_{k,h+1}(a\mid s)\widehat{\boldsymbol{\omega}}_{k,h}^\top\mathbf{T}^\star(s),[0,1])$ and $R_{k,h+1}^\circ\coloneq\nabla f_{h+1}(\mu_{k,h+1}^\circ)$. We define $V_{k,h+1}^\circ$ as the value function computed by \OPE{} at step $h+1$ with $\widehat{R}_{k,h+1}$ replaced by $R_{k,h+1}^\circ$. That is, $V_{k,H}^\circ(s)\coloneq\sum_a\pi_{k,H}(a\mid s)\operatorname{Clip}(R_{k,H}^\circ(s,a),[-G,G])$ and, for $h<H-1$,
\begin{align}
V_{k,h+1}^\circ(s)&\coloneq\sum_a\pi_{k,h+1}(a\mid s)\operatorname{Clip}\Bigl(R_{k,h+1}^\circ(s,a)+\boldsymbol{\phi}(s,a)^\top\widehat{\boldsymbol{v}}_{k,h+1}\\
&\qquad+\alpha_{k,h+1}^{\mathrm{Lin}}\|\boldsymbol{\phi}(s,a)\|_{\boldsymbol{\Gamma}_{k,h+1}^{-1}},[-(H-h)G,(H-h)G]\Bigr).
\end{align}
Unlike $\widehat{V}_{k,h+1}$, $V_{k,h+1}^\circ$ depends on the data only through the policy $\pi_k$ and the low-dimensional parameters $\widehat{\boldsymbol{\omega}}_{k,h}$, $\widehat{\boldsymbol{v}}_{k,h+1}$, $\boldsymbol{\Gamma}_{k,h+1}$, and $\alpha_{k,h+1}^{\mathrm{Lin}}$. Let $\mathcal{V}_{h+1}^\circ$ be the class of all such functions as these parameters range over bounded sets (see~\cref{app:lin_covering_arguments} for the formal definition). Since $\mathbf{T}^\star$ is fixed, this class is fixed before data collection. Hence, it can be covered without any dependence on $k$ or $|\mathcal{S}|$. We then apply~\cref{lem:lin_target_perturbation} to obtain that
\begin{align}
\MoveEqLeft\biggl\|\sum_{i=1}^k\boldsymbol{\phi}_{i,h}\bigl[\widehat{V}_{k,h+1}(s_{i,h+1})-P^\star\widehat{V}_{k,h+1}(s_{i,h},a_{i,h})\bigr]\biggr\|_{\boldsymbol{\Gamma}_{k,h}^{-1}}\nonumber\\*
&\leq\biggl\|\sum_{i=1}^k\boldsymbol{\phi}_{i,h}\bigl[V_{k,h+1}^\circ(s_{i,h+1})-P^\star V_{k,h+1}^\circ(s_{i,h},a_{i,h})\bigr]\biggr\|_{\boldsymbol{\Gamma}_{k,h}^{-1}}+2\sqrt{k}\bigl\|\widehat{V}_{k,h+1}-V_{k,h+1}^\circ\bigr\|_\infty.
\end{align}
We bound the first term uniformly over $\mathcal{V}_{h+1}^\circ$ by covering $\mathcal{V}_{h+1}^\circ$ as in $\OCUPOOFF$. For the second term, on the event $\mathcal{E}_{\mathrm{P}}$, we have $\|\widehat{V}_{k,h+1}-V_{k,h+1}^\circ\|_\infty\leq L\sqrt{\beta^{\mathrm{Lin}}}\|\widehat{\boldsymbol{\omega}}_{k,h}\|_{\boldsymbol{\Lambda}_k^{-1}}$. Therefore, we set
\begin{align}
\alpha_{k,h}^{\mathrm{Lin}}=\alpha_0^{\mathrm{Lin}}+2\sqrt{k}\,L\sqrt{\beta^{\mathrm{Lin}}}\bigl\lVert \widehat{\boldsymbol{\omega}}_{k,h} \bigr\rVert_{\boldsymbol{\Lambda}_k^{-1}}, \label{eq:lin_value_radius_outline}
\end{align}
which includes a data-dependent correction computable from $\boldsymbol{\Lambda}_k$ and $\widehat{\boldsymbol{\omega}}_{k,h}$. Given~\cref{asm:policy_class}, we show in~\cref{lem:lin_off_value_cover} that the logarithmic covering number of $\mathcal{V}_{h+1}^\circ$ is $\widetilde{\mathcal{O}}(d_\phi^2+d_\Pi)$. We show in~\cref{lem:lin_off_value_event} that, with $\alpha_0^{\mathrm{Lin}}=\widetilde{\mathcal{O}}(HG\sqrt{d_\phi^2+d_\Pi})$ and under the event $\mathcal{E}_{\mathrm{P}}$, $\mathcal{E}_{\mathrm{V}}$ holds with probability at least $1-\delta/4$.

\paragraph{Bounding Terms (II) and (III) under the Good Event}
On the good event $\mathcal{E}$ with the radii $\beta^{\mathrm{Lin}}$, $\Delta_{k,h}^{\mathrm{Lin}}$, and $\alpha_{k,h}^{\mathrm{Lin}}$, \cref{lem:on_bellman_error} shows that~\cref{eq:on_bellman_optimism,eq:on_bellman_cost} hold with $\Delta_{k,h}=\Delta_{k,h}^{\mathrm{Lin}}$ and $\alpha_{k,h}=\alpha_{k,h}^{\mathrm{Lin}}$. Summing over $k\in[K]$ and $h\in[H]$, we obtain
\begin{align}
\MoveEqLeft\mathrm{Term~(II)}+\mathrm{Term~(III)}\leq2\sum_{k=1}^K\sum_{h=1}^H\Delta_{k,h}^{\mathrm{Lin}}+2\sum_{k=1}^K\sum_{h=1}^{H-1}\alpha_{k,h}^{\mathrm{Lin}}\widehat{\ell}_{k,h}\nonumber\\
&\leq\widetilde{\mathcal{O}}\Biggl(\biggl(LH\kappa\Bigl(\sqrt{\beta^{\mathrm{Lin}}}+C_{\mathbf{T}}\sqrt{d_\phi}\Bigr)+\alpha_0^{\mathrm{Lin}}\biggr)\sum_{k=1}^K\sum_{h=1}^{H-1}\widehat{\ell}_{k,h}+LH\gamma_\rho\Bigl(\sqrt{\beta^{\mathrm{Lin}}}+C_{\mathbf{T}}\sqrt{d_\phi}\Bigr)\sqrt K\nonumber\\
&\qquad+L\sqrt{\beta^{\mathrm{Lin}}}\sum_{k=1}^K\sum_{h=1}^{H-1}\sqrt{k}\bigl\lVert \widehat{\boldsymbol{\omega}}_{k,h} \bigr\rVert_{\boldsymbol{\Lambda}_k^{-1}}\widehat{\ell}_{k,h}\Biggr)\nonumber\\
&\leq\widetilde{\mathcal{O}}\biggl(H^2\Bigl[G\sqrt{d_\phi(d_\phi^2+d_\Pi)}+Ld_\phi^3\sqrt{d_\phi^2+d_\Pi}\Bigr]\sqrt K\biggr). \label{eq:lin_uncertainty_outline}
\end{align}
In the second inequality, we substitute $\Delta_{k,h}^{\mathrm{Lin}}$ in~\cref{eq:lin_reward_outline} and $\alpha_{k,h}^{\mathrm{Lin}}$ in~\cref{eq:lin_value_radius_outline}. Compared with $\OCUPOOFF$, the last term is new and comes from the term $2\sqrt{k}\,L\sqrt{\beta^{\mathrm{Lin}}}\|\widehat{\boldsymbol{\omega}}_{k,h}\|_{\boldsymbol{\Lambda}_k^{-1}}$ in~\cref{eq:lin_value_radius_outline}. To bound it, we apply $\|\widehat{\boldsymbol{\omega}}_{k,h}\|_{\boldsymbol{\Lambda}_k^{-1}}\leq\widehat{\ell}_{k,h}+\gamma_\rho/\sqrt k+\kappa\sum_{t=1}^{h-1}\widehat{\ell}_{k,t}$, which follows as in the proof of~\cref{lem:lin_off_reward_error}, and $\sum_{k=1}^K\widehat{\ell}_{k,h}^2\leq\widetilde{\mathcal{O}}(d_\phi^2)$ for every $h$, which follows from~\cref{lem:lin_off_martingale} using an extension of the elliptical potential lemma (\cref{lem:lin_predictable_leverage}). For the last inequality, we use $\sum_{k=1}^K\sum_{h=1}^{H-1}\widehat{\ell}_{k,h}\leq\widetilde{\mathcal{O}}(H\sqrt{d_\phi K})$, which also follows from~\cref{lem:lin_off_martingale}, analogous to~\cref{lem:off_martingale} for $\OCUPOOFF$. Finally, we substitute the orders of the radii and concentration constants established above, namely $\sqrt{\beta^{\mathrm{Lin}}}=\widetilde{\mathcal{O}}(\sqrt{d_\phi})$, $\gamma_\rho,\kappa=\widetilde{\mathcal{O}}(\sqrt{d_\phi(d_\phi^2+d_\Pi)})$, and $\alpha_0^{\mathrm{Lin}}=\widetilde{\mathcal{O}}(HG\sqrt{d_\phi^2+d_\Pi})$.

\paragraph{Putting Everything Together}
We combine~\cref{eq:lin_term1_outline,eq:lin_uncertainty_outline} and take a union bound to show that, with probability at least $1-\delta$, $\operatorname{Sub\text{-}Opt}(\overline{\pi}_K)\leq\operatorname{Reg}_K/K$ satisfies the bound in~\cref{thm:lin_off}. Therefore, running $K=\widetilde{\mathcal{O}}(H^4[G^2d_\phi(d_\phi^2+d_\Pi)+L^2d_\phi^6(d_\phi^2+d_\Pi)]/\epsilon^2)$ iterations of $\OCUPOLIN$, each collecting one trajectory, yields an $\mathcal{O}(\epsilon+\varepsilon_{\mathrm{proj}})$-optimal mixture policy. See~\cref{app:lin_regret_analysis_off} for the complete proof.
\section{Conclusion}

We propose $\OCUPO$, a model-based policy optimization framework for solving CURL with linear function approximation. In bilinear MDPs, we exploit the low-dimensional structure of the transition model to estimate the occupancy measure through low-dimensional feature expectations, without enumerating the state space. We first design the on-policy algorithm $\OCUPOON$, which collects fresh trajectories in each iteration and achieves an $\widetilde{\mathcal{O}}(\epsilon^{-4})$ sample complexity. By reusing trajectories collected by past policies through backward feature regression, the off-policy algorithm $\OCUPOOFF$ improves the sample complexity to $\widetilde{\mathcal{O}}(\epsilon^{-2})$ up to a projection error, under suitable assumptions on the policy class. We further extend $\OCUPOOFF$ to linear MDPs and show that the resulting algorithm, $\OCUPOLIN$, retains the same sample complexity. In each setting, the dependence on $\epsilon$ matches that of policy optimization methods for standard RL with linear function approximation. To the best of our knowledge, our algorithms are the first computationally efficient policy optimization methods with strategic exploration for CURL beyond the tabular setting. Moreover, since standard RL is a special case of CURL, our methods also apply to standard RL in bilinear MDPs without an integral oracle. In future work, we aim to derive similar guarantees for infinite-horizon problems and for general function approximation, which requires occupancy estimators and exploration bonuses suited to nonlinear representations.

\bibliography{references}

\clearpage
\appendix

\addtocontents{toc}{\protect\setcounter{tocdepth}{3}}
{\small\tableofcontents}
\clearpage

\section{Objectives of CURL} \label{app:utility_examples}

This appendix collects further examples of CURL beyond~\cref{eg:standard_rl,eg:imitation_learning,eg:pure_exploration}. They include apprenticeship learning, exploration, and distribution-matching objectives in~\citet[Table 1]{zahavy2021reward}, together with linearly scalarized multi-objective RL and a penalized form of safety-constrained RL. We use additive smoothing for Shannon entropy and KL divergence. All constants hold on $[0,1]^{\mathcal{S}\times\mathcal{A}}$. We first list examples that satisfy~\cref{asm:lipschitzness_and_smoothness,asm:coordinate_separability}.

\begin{example}[Linearly Scalarized Multi-Objective RL] \label{eg:multi_objective_rl}
\leavevmode\par\noindent
$F(\mu^\pi)=\sum_{j=1}^m w_j\sum_{h=1}^H\langle \mu_h^\pi,r_h^{(j)}\rangle$, where $r^{(1)},\ldots,r^{(m)}$ are bounded reward functions and $\boldsymbol{w}\in\Delta([m])$ specifies fixed preference weights. This linear scalarization is equivalent to standard RL with reward $\sum_{j=1}^m w_jr_h^{(j)}$. \cref{asm:lipschitzness_and_smoothness,asm:coordinate_separability} hold with $G=\max_{h\in[H]}\|\sum_{j=1}^m w_jr_h^{(j)}\|_\infty$ and $L=0$.
\end{example}

\begin{example}[Imitation Learning via KL Divergence] \label{eg:kl_imitation}
\leavevmode\par\noindent
$F(\mu^\pi)=-\sum_{h=1}^H\sum_{s,a}(\mu_h^\pi(s,a)+\tau)\log\!\frac{\mu_h^\pi(s,a)+\tau}{\mu_h^{\exp}(s,a)+\tau}$, where $\mu^{\exp}$ is the expert occupancy measure and $\tau>0$ is a smoothing parameter. \cref{asm:lipschitzness_and_smoothness,asm:coordinate_separability} hold with $G=1+\log((1+\tau)/\tau)$ and $L=1/\tau$.
\end{example}

\begin{example}[Maximum Tsallis Entropy] \label{eg:tsallis_entropy}
\leavevmode\par\noindent
$F(\mu^\pi)=\sum_{h=1}^H\frac{1-\sum_{s,a}(\mu_h^\pi(s,a))^q}{q-1}$, where $q\geq2$ is the entropy order~\citep{tsallis1988possible}. \cref{asm:lipschitzness_and_smoothness,asm:coordinate_separability} hold with $G=q/(q-1)$ and $L=q$. The case $q=2$ gives quadratic entropy.
\end{example}

Moreover, \cref{asm:coordinate_separability} can be relaxed to the following gradient locality assumption.
\begin{assumption}[Gradient Locality] \label{asm:gradient_locality}
For every $h\in[H]$ and $(s,a)\in\mathcal{S}\times\mathcal{A}$, there exists a known finite set $\mathcal{G}_h(s,a)\subseteq\mathcal{S}\times\mathcal{A}$ with $|\mathcal{G}_h(s,a)|\leq D_{\mathcal{G}}\ll|\mathcal{S}|$, such that, for every $\boldsymbol{u}\in[0,1]^{\mathcal{S}\times\mathcal{A}}$, the gradient coordinate $[\nabla f_h(\boldsymbol{u})](s,a)$ depends only on the coordinates of $\boldsymbol{u}$ indexed by $\mathcal{G}_h(s,a)$.
\end{assumption}
Here, we denote $D_{\mathcal{G}}\coloneq\max_{h\in[H],(s,a)\in\mathcal{S}\times\mathcal{A}}|\mathcal{G}_h(s,a)|$ for the largest number of coordinates accessed by a single gradient query. \cref{asm:coordinate_separability} is the special case $\mathcal{G}_h(s,a)=\{(s,a)\}$ with $D_{\mathcal{G}}=1$. Under~\cref{asm:gradient_locality}, replacing the coordinates of $\boldsymbol{u}$ indexed by $\mathcal{G}_h(s,a)$ with those of $\boldsymbol{u}'$ and applying smoothness gives $|[\nabla f_h(\boldsymbol{u})-\nabla f_h(\boldsymbol{u}')](s,a)|\leq LD_{\mathcal{G}}\|\boldsymbol{u}-\boldsymbol{u}'\|_\infty$. Under~\cref{asm:coordinate_separability}, our analysis uses this bound with $D_{\mathcal{G}}=1$. Moreover, each pseudo-reward query evaluates at most $D_{\mathcal{G}}$ occupancy coordinates instead of one. Consequently, all results hold with $L$ replaced by $LD_{\mathcal{G}}$. The following examples satisfy~\cref{asm:lipschitzness_and_smoothness,asm:gradient_locality} with $D_{\mathcal{G}}>1$.

\begin{example}[Maximum Group Entropy] \label{eg:group_entropy}
\leavevmode\par\noindent
$F(\mu^\pi)=-\sum_{h=1}^H\sum_{p\in\mathcal{P}}(\mu_h^\pi(p)+\tau)\log(\mu_h^\pi(p)+\tau)$, where $\tau>0$ is a smoothing parameter, $\mathcal{P}$ partitions $\mathcal{S}\times\mathcal{A}$ into groups whose maximum size $\max_{p\in\mathcal{P}}|p|$ does not depend on $|\mathcal{S}|$, and $\mu_h^\pi(p)\coloneq\sum_{(s,a)\in p}\mu_h^\pi(s,a)$. On the utility domain, a group sum lies in $[0,\max_{p\in\mathcal{P}}|p|]$. \cref{asm:lipschitzness_and_smoothness,asm:gradient_locality} therefore hold with $G=\max\{|1+\log\tau|,1+\log(\max_{p\in\mathcal{P}}|p|+\tau)\}$, $L=1/\tau$, and $D_{\mathcal{G}}=\max_{p\in\mathcal{P}}|p|$.
\end{example}

\begin{example}[RL with State-Density Constraints] \label{eg:state_density_constraints}
\leavevmode\par\noindent
$F(\mu^\pi)=\sum_{h=1}^H\langle \mu_h^\pi,r_h\rangle-\lambda\sum_{h=1}^H\sum_{s\in\mathcal{S}}([d_h^\pi(s)-u_h(s)]_+^2+[\ell_h(s)-d_h^\pi(s)]_+^2)$, where $\lambda>0$, $d_h^\pi(s)=\sum_{a\in\mathcal{A}}\mu_h^\pi(s,a)$, $[x]_+\coloneq\max\{x,0\}$, and the state-density limits $\ell_h,u_h\colon \mathcal{S}\to[0,1]$ satisfy $\ell_h(s)\leq u_h(s)$. The quadratic penalties discourage violations of state-density limits that encode safety or resource requirements~\citep{qin2021density}. \cref{asm:lipschitzness_and_smoothness,asm:gradient_locality} hold with $G=\max_{h\in[H]}\|r_h\|_\infty+2\lambda|\mathcal{A}|$, $L=2\lambda$, and $D_{\mathcal{G}}=|\mathcal{A}|$.
\end{example}

In addition, we show that the optimal policies of CURL may be stochastic.
\begin{proposition}[Stochastic Optimal Policy] \label{prop:stochastic_optimal_policy}
There exists a CURL problem satisfying~\cref{asm:lipschitzness_and_smoothness,asm:coordinate_separability} whose unique optimal policy is stochastic.
\end{proposition}

\begin{proof}
Consider an MDP with a single state $s$, two actions $a_1$ and $a_2$, horizon $H=1$, and the maximum entropy utility of~\cref{eg:pure_exploration}. Writing $p\coloneq\pi_1(a_1\mid s)$, we have $F(\mu^\pi)=\varphi(p)\coloneq-(p+\tau)\log(p+\tau)-(1-p+\tau)\log(1-p+\tau)$. The function $\varphi$ is strictly concave on $[0,1]$ and satisfies $\varphi(p)=\varphi(1-p)$. Hence, for every $p\neq1/2$,
\begin{align}
\varphi(p)=\frac{\varphi(p)+\varphi(1-p)}{2}<\varphi\left(\frac12\right).
\end{align}
The uniform policy is therefore the unique optimal policy, and every other policy, including both deterministic policies, is strictly suboptimal.
\end{proof}

\section{Additional Related Work} \label{app:related_work}

\addtocontents{toc}{\protect\setcounter{tocdepth}{1}}

This appendix discusses prior work related to $\OCUPO$. In~\cref{app:related_comparison}, we compare our assumptions and guarantees with those of prior CURL methods that go beyond exact tabular planning. We then review prior work on CURL in~\cref{app:related_curl}, occupancy measure estimation in~\cref{app:related_occupancy}, and policy optimization with linear function approximation in~\cref{app:related_lfa}.

\subsection{Comparison with Occupancy-Based Methods} \label{app:related_comparison}
We summarize how our assumptions and guarantees differ from those of prior CURL methods that go beyond exact tabular planning. \citet{moreno2025online} address strategic exploration in tabular CURL with unknown transition dynamics and establish near-optimal regret guarantees. Their methods rely on explicit tabular representations, and their computation scales polynomially with $|\mathcal{S}||\mathcal{A}|$. \citet{barakat2023reinforcement} and \citet{barakat2024towards} approximate occupancy measures within a linear or general function class, which removes the dependence on $|\mathcal{S}|$. However, their assumptions are placed directly on this function class: the occupancy measures of all policies encountered must be well approximated by it, and the sampling distribution must be suitable for fitting it. Since a generic function class need not respect the Bellman flow constraints satisfied by every occupancy measure, this approximability is an additional assumption. In contrast, our assumption is placed on the transition model, and the low-dimensional representation in~\cref{eq:bilinear_occupancy_decomposition} follows from the flow constraints under this model, so that it holds for the occupancy measure of every policy. Moreover, they do not address strategic exploration, and their guarantees concern first-order stationarity or global optimality under additional assumptions on the policy class. \citet{huang2023density} estimate occupancies from non-exploratory data and collect exploratory data in low-rank MDPs with known density features, with applications to convex RL, but their approach does not provide a computationally efficient policy search. \citet{huang2024occupancy} reduce occupancy and gradient estimation to regression and handle differentiable objectives, with guarantees that rely on exploration and coverage conditions. In contrast, we assume a bilinear MDP with known features $\boldsymbol{\phi}$ and $\boldsymbol{\psi}$, which is a low-rank MDP with known density features and additionally known features $\boldsymbol{\phi}$. Under this assumption, $\OCUPO$ explores through optimism, avoids the integral oracle, and has a per-iteration cost independent of $|\mathcal{S}|$. It achieves sample complexities of $\widetilde{\mathcal{O}}(\epsilon^{-4})$ in the on-policy setting and $\widetilde{\mathcal{O}}(\epsilon^{-2})$ up to the projection error in the off-policy setting. Our main technical contribution is thus the estimation and confidence analysis that makes optimistic exploration computationally efficient in this setting: occupancy reconstruction from feature expectations, backward feature regression on replay data, and uniform concentration over the induced value and feature classes.

\subsection{Concave Utility Reinforcement Learning} \label{app:related_curl}

\paragraph{Occupancy Optimization and Reductions to Ordinary RL}
The occupancy formulation makes the feasible set convex while allowing nonlinear objectives. \citet{hazan2018provably} study maximum-entropy exploration through a conditional-gradient reduction to an ordinary RL planning oracle. Their approach shows how a sequence of reward-maximizing policies can optimize a nonlinear exploration objective, with planning and occupancy estimation serving as separate ingredients. \citet{zahavy2021reward} give a general Fenchel-dual formulation of convex MDPs as a game between a policy player and a cost player. Their framework accommodates different no-regret learners, including optimistic RL methods, and organizes many applications within a common formulation. Exploration is therefore already addressed in parts of the convex-RL literature. Our focus is on obtaining explicit estimation and exploration guarantees under the transition representation and local utility assumptions considered here. \citet{geist2021concave} connect CURL to mean-field games, relating concavity of the utility to monotonicity of the associated game. This perspective motivates algorithms based on fictitious play and related policy updates. It provides a structural interpretation of occupancy optimization, while our finite-sample analysis additionally controls an unknown transition model, estimated pseudo-rewards, and value targets fitted using previously collected trajectories.

\paragraph{Policy Gradient Methods}
\citet{zhang2020variational} introduce a variational policy gradient method for general utilities and use the hidden convexity of occupancy optimization to establish global convergence. \citet{zhang2021convergence} develop truncated variance-reduced policy gradients, obtaining stationarity guarantees and global guarantees under additional policy overparameterization conditions. \citet{kumar2022policy} derive a direct policy gradient theorem in which the gradient of the utility plays the role of a policy-dependent reward. This identity underlies our \PRE{}--\OPE{} decomposition: each policy update uses a conventional value function, but its reward depends on the current occupancy and must therefore be estimated.

\citet{barakat2023reinforcement} propose a single-loop normalized policy gradient algorithm with recursive variance reduction. They also study large finite state-action spaces by approximating the occupancy with a linear function class. Their large-space result controls stationarity up to approximation error and requires conditions on the occupancy-regression features and sampling distribution. These guarantees differ from both their global guarantees under additional assumptions and the global utility-suboptimality bounds established here. Our representation assumption is imposed on the transition kernel, and we control exploration through confidence bounds while accessing the nonlinear utility through local gradient queries. Moreover, \citet{barakat2024towards} approximate occupancies by maximum likelihood within a realizable function class and analyze policy gradient with these estimates. Their bounds can depend on the dimension of the occupancy class rather than the number of state-action pairs, and their revised analysis includes global convergence under suitable policy assumptions. Their results thus also establish CURL guarantees with function approximation. Their approach estimates a current-policy occupancy model from samples. Our off-policy construction instead estimates expected future feature vectors by backward regression using adaptively collected replay data, and combines these estimates with optimistic policy evaluation.

\paragraph{Model-Based Mirror Descent and Nonstationarity}
\citet{moreno2024efficient} propose MD-CURL for finite-horizon CURL and Greedy MD-CURL for online interaction with partially unknown dynamics. Their regularization yields a tractable policy update. In the structured-noise setting, observed noise can be reused through a known transition map to estimate the kernel across state-action pairs. They also examine settings with less informative observations. These results provide a model-based precedent, while our analysis develops transition and occupancy confidence bounds under bilinear and linear representations. \citet{marin2024metacurl} extend this direction to nonstationary losses and transitions through a meta-algorithm that combines learners over different intervals. Their dynamic-regret guarantees address environmental changes, whereas our analysis concerns a fixed unknown MDP with linear structure.

\paragraph{Constraints and Penalty Methods}
\citet{bai2023achieving} study concave-utility constrained MDPs using a conservative stochastic primal-dual method. Their zero-violation and sample-complexity guarantees use a generative model. Such access differs from collecting trajectories under successive policies, and their constrained guarantees are not implied by our utility-maximization results.

More recently, \citet{wolf2026global} study constrained maximum-entropy exploration through quadratic-penalty regularization. Their model-free policy gradient method estimates occupancies and constructs pseudo-rewards for the penalized objective. Under regularity assumptions on the policy and occupancy parametrization, together with strong duality, they establish global last-iterate objective and constraint-violation guarantees with an $\widetilde{\mathcal{O}}(\epsilon^{-6})$ sample complexity. Their penalty subproblems are closely related to concave occupancy optimization, although general constraint penalties need not satisfy our coordinate separability or gradient locality assumptions. Their guarantees concern explicit constraints and the last policy iterate, whereas ours concern a fixed utility evaluated at the occupancy of a policy mixture. The rates therefore address different guarantees under different assumptions. Our focus is on exploiting transition structure and trajectory reuse for sample-efficient utility maximization.

\paragraph{Robust General Utilities}
\citet{chen2024robust} consider general utilities under transition uncertainty, formulating a robust minimax problem and establishing gradient-convergence guarantees together with stronger guarantees for particular ambiguity sets. Robustness to an adversarial model set differs from using statistical confidence bounds to learn one fixed environment. \citet{liu2026robust} instead study robustness to utility misspecification, optimizing policies against an uncertainty set of utility functionals. Their formulation includes fixed-utility RL as the special case of a singleton uncertainty set. Their main stochastic policy gradient methods establish first-order stationarity guarantees in a discounted setting with parameterized policies. They also provide a global saddle-gap guarantee for a model-based tabular formulation. Our work studies a fixed utility under unknown bilinear or linear dynamics and establishes global utility-suboptimality bounds for both on-policy and off-policy estimation. The distinction lies in the representation assumptions, estimation procedures, and statistical guarantees, rather than the use of occupancy-dependent pseudo-rewards alone.

\paragraph{Multiple Agents and Alternative Occupancy Objectives}
\citet{ying2023scalable} study multi-agent general utilities with localized policies and spatial correlation decay, obtaining stationarity bounds with an approximation term governed by the communication radius. These structural questions complement the single-agent exploration problem considered here. \citet{mutti2023convex} distinguish a utility evaluated at the population occupancy from the expected utility of an empirical occupancy over finitely many trials. These objectives generally differ and can have different optimal policy classes. Our objective is the former: an episode-wise mixture is evaluated through its population occupancy. \citet{mutny2023active} formulate experimental design under Markovian movement constraints as an occupancy-optimization problem and adapt policies using previous measurements. Such objectives further illustrate the breadth of CURL, although applying our theorems also requires checking smoothness and coordinate separability or gradient locality for the chosen criterion.

\subsection{Occupancy Measure Estimation} \label{app:related_occupancy}

\paragraph{Density Features and Occupancy-Based Gradients}
Estimating the occupancy of the current policy is essential for evaluating a nonlinear utility gradient. With trajectory reuse, the estimator must also account for the difference between the current policy and the policies that generated the stored data.~\citet{huang2023density} study low-rank MDPs through their density features. They develop off-policy occupancy estimation and an online procedure for constructing exploratory data distributions, with applications to convex RL. Their setting assumes known density features on the next-state side of the factorization. A bilinear MDP with known $\boldsymbol{\psi}$ is such a low-rank MDP, and our setting additionally assumes known features $\boldsymbol{\phi}$. This additional structure is what makes $\OCUPO$ computationally efficient, with a per-iteration cost independent of $|\mathcal{S}|$. In linear MDPs (\cref{section:ocupo-lin}), only the features $\boldsymbol{\phi}$ are known, which motivates our sparse transition estimator: the learner stores coefficients only at observed successor states, while the concentration argument controls every queried column. Moreover,~\citet{huang2024occupancy} develop occupancy-based policy gradients for online and offline RL, reducing occupancy and gradient estimation to regression. Their analysis addresses exploration and data coverage, and their method extends to differentiable occupancy objectives. Our analysis follows a different route: a local utility gradient supplies a pseudo-reward, backward feature regression estimates the current occupancy using replay data, and value-confidence bounds control the evaluation errors entering policy optimization. In particular, our guarantees account for the dependence between the estimated occupancy, pseudo-reward, and fitted next-step value function.

\paragraph{Fitted Occupancy-Ratio Estimation}
\citet{van2026fitted} introduce fitted occupancy-ratio evaluation (FORE), which estimates the discounted occupancy ratio of a target policy relative to an offline sampling distribution. Their KL-projected recursion uses one-step transition data and contracts in relative entropy under the stated coverage and ratio-class conditions. The analysis controls statistical error and direct approximation error of the ratio class without requiring Bellman completeness. The fitted ratio supports estimation of occupancy functionals, policy values, and weights for fitted policy evaluation.

This approach is relevant to our occupancy-feature estimation because an occupancy expectation can be expressed as a ratio-weighted expectation under a reference distribution. FORE could therefore motivate an alternative estimator of $\omega_h^\pi$. However, its discounted fixed-policy guarantees do not directly establish the confidence bounds required here. Our setting involves finite-horizon occupancies, adaptively collected trajectories, and target policies that change with the data. Although FORE permits repeated fitting using the same offline dataset, extending its guarantees to this setting would require additional uniform concentration and uncertainty analysis. Our backward feature regression instead exploits the transition representation to estimate the required feature expectations directly, with errors controlled in the feature directions used by the sample-complexity analysis.

\subsection{Policy Optimization with Linear Function Approximation} \label{app:related_lfa}

\paragraph{Optimistic Policy Evaluation}
Linear-transition structure supports efficient exploration through least-squares confidence bounds~\citep{yang2020reinforcement,jin2023provably}. In a bilinear model, directly applying a fitted transition core to a value function requires a feature-weighted sum over the state space. Our \OPE{} in~\cref{alg:ope} estimates the expected next-state value directly from observed successor values. Its regression step is related to least-squares value methods, while its confidence argument must also account for policy-dependent nonlinear pseudo-rewards.

\paragraph{Occupancy-Weighted Fitted Evaluation}
For a fixed policy, \citet{van2025fitted} study fitted $Q$-evaluation using offline data. They show that weighting Bellman regressions by a target-policy discounted occupancy ratio aligns the regression norm with the target-policy dynamics and restores contraction of the population projected Bellman operator. Under their coverage and function-class conditions, they derive finite-sample guarantees with estimated weights and approximation error. Combined with FORE, joint realizability of the value-function and occupancy-ratio classes suffices for consistent evaluation without Bellman completeness.

Although the original CURL objective need not admit a standard Bellman recursion, evaluation under a fixed pseudo-reward does. Occupancy-weighted fitted evaluation is therefore relevant to the evaluation component of our framework. Its guarantees concern accurate evaluation of a fixed target policy, while our analysis additionally requires optimistic confidence bounds for policies and pseudo-rewards estimated using replay data. Their approach obtains stability through occupancy weighting. Ours obtains the confidence bounds needed for exploration by exploiting the assumed transition representation.

\paragraph{Policy Optimization and Projection}
Optimistic policy optimization for ordinary reward maximization provides the regret-based viewpoint used in our analysis~\citep{liu2023optimistic,sherman2023rate,cassel2024warm,zhong2023theoretical}. These methods combine value estimates with policy updates while controlling the errors caused by learning the environment. CURL introduces an additional estimation step because the reward entering policy evaluation depends on the current occupancy.

\cref{tab:standard_rl_comparison} compares the sample complexities of these methods with ours in standard RL with rewards in $[0,1]$, where $L=0$ and $G\leq1$ (\cref{eg:standard_rl}). For the purpose of comparison, we define $d$ such that $d_\phi=d_\psi=d_x=\mathcal{O}(d)$, where $d_x$ is the dimension of the policy features. For log-linear policies, the covering dimension of our policy class is then $d_\Pi=Hd_x$ (\cref{prop:pnpg_oracle}). Some of these methods also allow stage-dependent transitions or handle unknown or adversarial rewards (e.g.,~\citealp{cassel2024warm}), whereas we assume a time-homogeneous transition model and a known utility. Hence,~\cref{tab:standard_rl_comparison} compares rates under the respective settings rather than under identical assumptions.

\begin{table}[t]
\centering
\caption{Sample complexity, measured in trajectories, of policy optimization with linear function approximation in standard RL, where $d_\phi=d_\psi=d_x=\mathcal{O}(d)$. We omit logarithmic factors, $|\mathcal{A}|$, and normalization constants. For regret bounds, we report the number of trajectories after which the average regret is at most $\epsilon$. Here, $\varepsilon_{\mathrm{proj}}$ is the projection error of the log-linear policy class. For the sake of comparison, we report the rates of~\citet{lin2026optimistic} with \PNPG{} as the actor and a UCB critic.}
\label{tab:standard_rl_comparison}
\begin{tabular}{@{}lcccc@{}}
\toprule
Method & MDP & Learning & Sample Complexity & Optimality Gap \\
\midrule
\citet{liu2023optimistic} & Linear & On-policy & $\widetilde{\mathcal{O}}(d^2H^7/\epsilon^4)$ & $\epsilon$ \\
\citet{sherman2023rate} & Linear & Off-policy & $\widetilde{\mathcal{O}}(d^4H^7/\epsilon^2)$ & $\epsilon$ \\
\citet{cassel2024warm} & Linear & Off-policy & $\widetilde{\mathcal{O}}((d^3H^4+dH^5)/\epsilon^2)$ & $\epsilon$ \\
\citet{lin2026optimistic} & Linear & On-policy & $\widetilde{\mathcal{O}}(d^3H^8/\epsilon^4)$ & $\epsilon+\varepsilon_{\mathrm{proj}}$ \\
\citet{lin2026optimistic} & Linear & Off-policy & $\widetilde{\mathcal{O}}(d^3H^4/\epsilon^2)$ & $\epsilon+\varepsilon_{\mathrm{proj}}$ \\
\midrule
$\OCUPOON$ (Ours) & Bilinear & On-policy & $\widetilde{\mathcal{O}}(d^3H^8/\epsilon^4)$ & $\epsilon$ \\
$\OCUPOOFF$ (Ours) & Bilinear & Off-policy & $\widetilde{\mathcal{O}}((d^3H^4+d^2H^5)/\epsilon^2)$ & $\epsilon+\varepsilon_{\mathrm{proj}}$ \\
$\OCUPOLIN$ (Ours) & Linear & Off-policy & $\widetilde{\mathcal{O}}((d^3H^4+d^2H^5)/\epsilon^2)$ & $\epsilon+\varepsilon_{\mathrm{proj}}$ \\
\bottomrule
\end{tabular}
\end{table}

In particular, \citet{lin2026optimistic} analyze optimistic actor-critic with parametric policies and quantify the effect of projecting an ideal \NPG{} update onto a log-linear class. \cref{app:policy_oracle} uses their projection and regret results and verifies~\cref{asm:pnpg,asm:policy_class} for log-linear policies. The on-policy analysis uses only the regret bound of \NPG{}, whereas the off-policy analysis in \cref{section:ocupo-off} additionally uses the covering bound on the policy class.

\addtocontents{toc}{\protect\setcounter{tocdepth}{3}}

\section{Policy Optimization Methods} \label{app:policy_oracle}

This appendix presents the \NPG{} update used by $\OCUPOON$ and its projected variant used by $\OCUPOOFF$. For the projected variant, we instantiate the policy class with log-linear policies, bound the constant $\varepsilon_{\mathrm{proj}}$, and verify~\cref{asm:pnpg,asm:policy_class}.

\subsection{Natural Policy Gradient} \label{app:npg} \label{app:npg_design} \label{app:npg_guarantees}

The \NPG{} update in~\cref{eq:npg_update} is an online mirror descent step with the negative entropy regularizer. For every $h\in[H]$ and $s\in\mathcal{S}$, it solves
\begin{align}
\pi_{k+1,h}(\cdot\mid s)&=\operatorname*{arg\,min}_{p\in\Delta(\mathcal{A})}\Bigl\{-\eta\bigl\langle p,\widehat{Q}_{k,h}(s,\cdot) \bigr\rangle+D_{\mathrm{KL}}(p\|\pi_{k,h}(\cdot\mid s))\Bigr\}, \label{eq:npg_mirror_update}
\end{align}
whose closed-form solution is~\cref{eq:npg_update}. Starting from the uniform policy, $\pi_{k+1,h}(\cdot\mid s)$ is the softmax of $\eta\sum_{i=1}^k\widehat{Q}_{i,h}(s,\cdot)$, which can be evaluated at queried states without maintaining a table over $\mathcal{S}$. The following regret bound is the standard online mirror descent bound with the negative entropy~\citep[Sections 5.3--5.4]{hazan2022introduction}, applied at each state and step, averaged under the comparator occupancy, and summed over steps.

\begin{lemma}[Regret of Online Mirror Descent] \label{thm:npg_regret}
Consider a finite set $\mathcal{X}$ and a sequence of functions $q_k\colon\mathcal{X}\to\mathbb{R}$ with $\|q_k\|_\infty\leq Q_{\max}$ for every $k\in[K]$. Starting from the uniform distribution $p_1\in\Delta(\mathcal{X})$, online mirror descent with step size $\eta>0$ performs the update $p_{k+1}=\operatorname*{arg\,max}_{p\in\Delta(\mathcal{X})}\{\eta\langle q_k,p\rangle-\mathrm{KL}(p\,\|\,p_k)\}$. Then, for every $u\in\Delta(\mathcal{X})$,
\begin{align}
\sum_{k=1}^K\langle q_k,u-p_k\rangle&\leq\frac{\log|\mathcal{X}|}{\eta}+\frac{\eta KQ_{\max}^2}{2}.
\end{align}
\end{lemma}
The \NPG{} update in~\cref{eq:npg_update} is this update applied at each step $h$ and state $s$, with $\mathcal{X}=\mathcal{A}$, $p_k=\pi_{k,h}(\cdot\mid s)$, and $q_k=\widehat{Q}_{k,h}(s,\cdot)$. For a comparator policy $u$, applying~\cref{thm:npg_regret} at each $(s,h)$ with $u_h(\cdot\mid s)$, averaging under $d_h^u$, and summing over $h$ gives
\begin{align}
\sum_{k=1}^K\sum_{h=1}^H\mathbb{E}_{s\sim d_h^u}\bigl\langle \widehat{Q}_{k,h}(s,\cdot),u_h(\cdot\mid s)-\pi_{k,h}(\cdot\mid s) \bigr\rangle&\leq\frac{H\log|\mathcal{A}|}{\eta}+\frac{\eta KHQ_{\max}^2}{2}. \label{eq:npg_regret_bound}
\end{align}
\OPE{} clips $\widehat{Q}_{k,h}$ to $[-(H-h+1)G,(H-h+1)G]$ (\cref{line:clipped_Q} of \cref{alg:ope}). Hence~\cref{eq:npg_regret_bound} holds with $Q_{\max}=HG$ for every sequence of value estimates generated by~\cref{alg:curl_framework}, including the data-dependent ones. Choosing $\eta=\sqrt{2\log|\mathcal{A}|/(KH^2G^2)}$ then gives
\begin{align}
\sum_{k=1}^K\sum_{h=1}^H\mathbb{E}_{s\sim d_h^u}\bigl\langle \widehat{Q}_{k,h}(s,\cdot),u_h(\cdot\mid s)-\pi_{k,h}(\cdot\mid s) \bigr\rangle&\leq H^2G\sqrt{2K\log|\mathcal{A}|}.
\end{align}

\subsection{Projected Natural Policy Gradient for Log-Linear Policies} \label{app:pnpg_update} \label{app:pnpg_design} \label{app:pnpg_guarantees}

\subsubsection{Algorithm Design}

\paragraph{Log-Linear Policies}
Let $\boldsymbol{x}\colon \mathcal{S}\times\mathcal{A}\to\mathbb{R}^{d_x}$ be a known policy feature map with $\|\boldsymbol{x}(s,a)\|_2\leq C_x\sqrt{d_x}$, whose image spans $\mathbb{R}^{d_x}$. For parameters $\boldsymbol{\theta}_h\in\mathbb{R}^{d_x}$, the corresponding log-linear policy is
\begin{align}
\pi_{\boldsymbol{\theta},h}(a\mid s)&=\frac{\exp(\boldsymbol{x}(s,a)^\top\boldsymbol{\theta}_h)}{\sum_{a'}\exp(\boldsymbol{x}(s,a')^\top\boldsymbol{\theta}_h)}.
\end{align}
We initialize $\boldsymbol{\theta}_{1,h}=\boldsymbol{0}$ for every step, which gives the uniform policy.

\paragraph{Projection by Logit Regression}
Let $\pi_k=\pi_{\boldsymbol{\theta}_k}$. Substituting it into the ideal update in~\cref{eq:pnpg_update}, the normalizing constant of $\pi_{\boldsymbol{\theta}_k,h}(\cdot\mid s)$ cancels, and $\pi_{k+1/2,h}(\cdot\mid s)$ is the softmax policy with logits
\begin{align}
z_{k,h}(s,a)\coloneq\boldsymbol{x}(s,a)^\top\boldsymbol{\theta}_{k,h}+\eta\widehat{Q}_{k,h}(s,a).
\end{align}
Since $\widehat{Q}_{k,h}$ need not lie in the span of the policy features, $\pi_{k+1/2}$ need not be log-linear. We therefore project it onto log-linear policies by fitting its logits. A direct fit would control the logit error at every state-action pair through the full actor loss
\begin{align}
\ell_{k,h}^{\mathrm{act}}(\boldsymbol{\theta})&\coloneq\sup_{(s,a)\in\mathcal{S}\times\mathcal{A}}\bigl(\boldsymbol{x}(s,a)^\top\boldsymbol{\theta}-z_{k,h}(s,a)\bigr)^2. \label{eq:pnpg_full_projection}
\end{align}
Minimizing~\cref{eq:pnpg_full_projection} requires evaluating the target at every state-action pair. Instead, we fit the logits on a coreset $\mathcal{C}\subseteq\mathcal{S}\times\mathcal{A}$ weighted by $\rho_{\mathcal{C}}\in\Delta(\mathcal{C})$, through the coreset actor loss
\begin{align}
\widetilde{\ell}_{k,h}^{\mathrm{act}}(\boldsymbol{\theta})&\coloneq\sum_{(s,a)\in\mathcal{C}}\rho_{\mathcal{C}}(s,a)\bigl(\boldsymbol{x}(s,a)^\top\boldsymbol{\theta}-z_{k,h}(s,a)\bigr)^2. \label{eq:pnpg_actor_loss}
\end{align}
\cref{alg:pnpg} forms~\cref{eq:pnpg_actor_loss} (\cref{line:pnpg_loss}) and approximately minimizes it (\cref{line:pnpg_solver}). The resulting parameters define $\pi_{k+1}=\pi_{\boldsymbol{\theta}_{k+1}}$.

\begin{algorithm2e}[htbp]
\caption{Projected Natural Policy Gradient (\PNPG{}) for Log-Linear Policies} \label{alg:pnpg}
\begin{algorithmic}[1]
\State \textbf{Input:} parameters $\boldsymbol{\theta}_k$, values $\widehat{Q}_k$, step size $\eta$, coreset $(\mathcal{C},\rho_{\mathcal{C}})$.
\For{$h=1,\ldots,H$}
    \State $\widetilde{\ell}_{k,h}^{\mathrm{act}}(\boldsymbol{\theta})\gets\sum_{(s,a)\in\mathcal{C}}\rho_{\mathcal{C}}(s,a)\bigl(\boldsymbol{x}(s,a)^\top(\boldsymbol{\theta}-\boldsymbol{\theta}_{k,h})-\eta\widehat{Q}_{k,h}(s,a)\bigr)^2$. \label{line:pnpg_loss}
    \State $\boldsymbol{\theta}_{k+1,h}\approx\operatorname*{arg\,min}_{\boldsymbol{\theta}\in\mathbb{R}^{d_x}}\widetilde{\ell}_{k,h}^{\mathrm{act}}(\boldsymbol{\theta})$. \label{line:pnpg_solver}
\EndFor
\State \Return\quad $\pi_{\boldsymbol{\theta}_{k+1}}$.
\end{algorithmic}
\end{algorithm2e}

\paragraph{Choice of the Coreset}
The coreset must represent every policy-feature direction, rather than merely contain frequently visited states. We define
\begin{align}
\boldsymbol{G}\coloneq\sum_{(s,a)\in\mathcal{C}}\rho_{\mathcal{C}}(s,a)\boldsymbol{x}(s,a)\boldsymbol{x}(s,a)^\top,\qquad\lambda_G\coloneq\lambda_{\min}(\boldsymbol{G}),\qquad\chi_G\coloneq\sup_{(s,a)\in\mathcal{S}\times\mathcal{A}}\|\boldsymbol{x}(s,a)\|_{\boldsymbol{G}^{-1}}.
\end{align}
If $\lambda_G>0$, the coreset actor loss is strongly convex, and for every $\boldsymbol{v}\in\mathbb{R}^{d_x}$ and $(s,a)\in\mathcal{S}\times\mathcal{A}$, the Cauchy--Schwarz inequality transfers a parameter error in the $\boldsymbol{G}$-norm to the logit:
\begin{align}
|\boldsymbol{x}(s,a)^\top\boldsymbol{v}|\leq\|\boldsymbol{x}(s,a)\|_{\boldsymbol{G}^{-1}}\|\boldsymbol{v}\|_{\boldsymbol{G}}\leq\chi_G\|\boldsymbol{v}\|_{\boldsymbol{G}}. \label{eq:pnpg_chi_transfer}
\end{align}
The following theorem guarantees a small coreset with $\lambda_G>0$ and a controlled $\chi_G$.

\begin{theorem}[Kiefer--Wolfowitz; {\citealp{kiefer1960equivalence}; \citealp[Theorem 21.1]{lattimore2019bandit}}] \label{thm:kiefer_wolfowitz}
Let $\mathcal{X}\subset\mathbb{R}^{d}$ be a compact set that spans $\mathbb{R}^{d}$. For $\rho\in\Delta(\mathcal{X})$, define $\boldsymbol{G}(\rho)\coloneq\mathbb{E}_{\boldsymbol{x}\sim\rho}[\boldsymbol{x}\boldsymbol{x}^\top]$ and $g(\rho)\coloneq\sup_{\boldsymbol{x}\in\mathcal{X}}\|\boldsymbol{x}\|_{\boldsymbol{G}(\rho)^{-1}}^2$. The following are equivalent for $\rho^\star\in\Delta(\mathcal{X})$: (i) $\rho^\star$ maximizes $\log\det\boldsymbol{G}(\rho)$, (ii) $\rho^\star$ minimizes $g(\rho)$, and (iii) $g(\rho^\star)=d$. Moreover, some minimizer $\rho^\star$ is supported on at most $d(d+1)/2$ points of $\mathcal{X}$.
\end{theorem}

Applying~\cref{thm:kiefer_wolfowitz} to the image of the policy feature map, which is compact in our finite-state setting, gives a coreset with $|\mathcal{C}|\leq d_x(d_x+1)/2$ and $\chi_G=\sqrt{d_x}$. Here $\lambda_G>0$, since $g(\rho^\star)=d_x$ is finite only if $\boldsymbol{G}(\rho^\star)$ is invertible. The coreset depends only on the known policy features and is fixed before data collection. In practice, the coreset can be computed by the Frank--Wolfe algorithm for the G-optimal design problem~\citep{todd2016minimum}, which returns a design $\rho$ with $g(\rho)\leq2d_x$ supported on $\mathcal{O}(d_x\log\log d_x)$ points~\citep[Chapter 21]{lattimore2019bandit}, at the cost of replacing $\chi_G=\sqrt{d_x}$ by $\sqrt{2d_x}$. Each iteration of this algorithm maximizes $\|\boldsymbol{x}(s,a)\|_{\boldsymbol{G}(\rho)^{-1}}$ over the state-action pairs, which requires either a pass over $\mathcal{S}\times\mathcal{A}$ or an oracle for this maximization. This oracle involves only the known policy features and differs from the integral oracle over the transition model. In both cases, the coreset is computed once before learning, and its cost does not grow with $K$. Evaluating a log-linear policy at a state requires $\mathcal{O}(|\mathcal{A}|d_x)$ operations. Since $\boldsymbol{G}$ is fixed, each update in~\cref{line:pnpg_solver} solves a $d_x$-dimensional least-squares problem over the coreset using a precomputed factorization of $\boldsymbol{G}$.

\subsubsection{Theoretical Guarantees}

\paragraph{Properties of the Actor Losses}
We record three facts used below. First, a $\rho_{\mathcal{C}}$-weighted average of squared logit errors cannot exceed their supremum. Hence, for every $\boldsymbol{\theta}\in\mathbb{R}^{d_x}$,
\begin{align}
\widetilde{\ell}_{k,h}^{\mathrm{act}}(\boldsymbol{\theta})\leq\ell_{k,h}^{\mathrm{act}}(\boldsymbol{\theta}). \label{eq:pnpg_loss_order}
\end{align}
Second, $\widetilde{\ell}_{k,h}^{\mathrm{act}}$ is a quadratic function with Hessian $2\boldsymbol{G}\succ0$. Hence, it has a unique minimizer $\boldsymbol{\theta}_{k+1,h}^\star$. The full actor loss $\ell_{k,h}^{\mathrm{act}}$ is convex and continuous, and by~\cref{eq:pnpg_loss_order} it tends to infinity as $\|\boldsymbol{\theta}\|_2\to\infty$. Hence, it also has a minimizer. Third, since the gradient of $\widetilde{\ell}_{k,h}^{\mathrm{act}}$ vanishes at $\boldsymbol{\theta}_{k+1,h}^\star$, its exact second-order expansion gives, for every $\boldsymbol{\theta}\in\mathbb{R}^{d_x}$,
\begin{align}
\widetilde{\ell}_{k,h}^{\mathrm{act}}(\boldsymbol{\theta})-\widetilde{\ell}_{k,h}^{\mathrm{act}}(\boldsymbol{\theta}_{k+1,h}^\star)=\|\boldsymbol{\theta}-\boldsymbol{\theta}_{k+1,h}^\star\|_{\boldsymbol{G}}^2. \label{eq:pnpg_quadratic_identity}
\end{align}

We make the following standard assumptions on the two actor losses.

\begin{assumption}[\PNPG{} Approximation Error] \label{asm:off_pnpg_bias}
For every possible history and every $k\in[K]$, $h\in[H]$, the full actor loss in~\cref{eq:pnpg_full_projection} satisfies $\min_{\boldsymbol{\theta}\in\mathbb{R}^{d_x}}\ell_{k,h}^{\mathrm{act}}(\boldsymbol{\theta})\leq\varepsilon_{\mathrm{bias}}$.
\end{assumption}

\begin{assumption}[\PNPG{} Optimization Error] \label{asm:off_pnpg_opt}
For every possible history and every $k\in[K]$, $h\in[H]$, the update computed in~\cref{line:pnpg_solver} of~\cref{alg:pnpg} satisfies $\widetilde{\ell}_{k,h}^{\mathrm{act}}(\boldsymbol{\theta}_{k+1,h})-\min_{\boldsymbol{\theta}\in\mathbb{R}^{d_x}}\widetilde{\ell}_{k,h}^{\mathrm{act}}(\boldsymbol{\theta})\leq\varepsilon_{\mathrm{opt}}$.
\end{assumption}

The first assumption is a representability condition on the full actor loss. In practice, $\varepsilon_{\mathrm{bias}}$ can be reduced by choosing higher-dimensional policy features. The second assumption concerns only the accuracy of the solver on the coreset actor loss. Minimizing the coreset actor loss by $T_k$ steps of gradient descent from $\boldsymbol{\theta}_{k,h}$ with step size $1/(2\lambda_{\max}(\boldsymbol{G}))$ ensures $\varepsilon_{\mathrm{opt}}\leq\eta^2H^2G^2\exp(-T_k/\kappa_G)$, where $\kappa_G$ is the condition number of $\boldsymbol{G}$.

\paragraph{Regret with Projection Errors}
The following result is~\citet[Theorem 4.1]{lin2026optimistic} applied to the value vectors $\widehat{Q}_{k,h}(s,\cdot)$, with bound $HG$ in place of $H$. It holds for every sequence of these vectors, including the data-dependent estimates generated by our framework.

\begin{theorem}[Projected \NPG{} Regret; {\citealp[Theorem 4.1]{lin2026optimistic}}] \label{thm:off_projected_npg_regret}
Consider specific $h\in[H]$ and $s\in\mathcal{S}$, and let $\{\widehat{Q}_{k,h}(s,\cdot)\}_{k=1}^K$ be any sequence with $\|\widehat{Q}_{k,h}(s,\cdot)\|_\infty\leq HG$. Suppose $\pi_{1,h}(\cdot\mid s)$ is uniform, $\pi_{k+1/2,h}(\cdot\mid s)$ is given by~\cref{eq:pnpg_update} with step size $\eta>0$, and $\pi_{k+1,h}$ is returned by \cref{alg:pnpg}. For $u\in\Delta(\mathcal{A})$, define the projection error
\begin{align}
\varepsilon_{k,h}^{\mathrm{proj}}(u,s)\coloneq D_{\mathrm{KL}}(u\|\pi_{k+1,h}(\cdot\mid s))-D_{\mathrm{KL}}(u\|\pi_{k+1/2,h}(\cdot\mid s)).
\end{align}
Then, for every $u\in\Delta(\mathcal{A})$,
\begin{align}
\sum_{k=1}^K\bigl\langle \widehat{Q}_{k,h}(s,\cdot),u-\pi_{k,h}(\cdot\mid s) \bigr\rangle\leq\frac{\log|\mathcal{A}|}{\eta}+\frac{\eta KH^2G^2}{2}+\frac{1}{\eta}\sum_{k=1}^K\varepsilon_{k,h}^{\mathrm{proj}}(u,s).
\end{align}
\end{theorem}
\noindent We measure the projection error uniformly by
\begin{align}
\overline{\varepsilon}_{k,h}\coloneq\sup_{s\in\mathcal{S}}\sup_{u\in\Delta(\mathcal{A})}\varepsilon_{k,h}^{\mathrm{proj}}(u,s)\qquad\text{and}\qquad\varepsilon_{\mathrm{proj}}\coloneq\frac{H}{\eta}\max_{k\in[K],h\in[H]}\overline{\varepsilon}_{k,h}. \label{eq:pnpg_kl_error}
\end{align}
Taking $u=\pi_{k+1/2,h}(\cdot\mid s)$, for which the second divergence vanishes, shows that $\overline{\varepsilon}_{k,h}\geq0$. Hence, $\varepsilon_{\mathrm{proj}}\geq0$, as required in~\cref{asm:pnpg}.

\paragraph{Projection Error}
The next lemma is a variant of~\citet[Lemma 4.1]{lin2026optimistic} for the loss-based conditions in \cref{asm:off_pnpg_bias,asm:off_pnpg_opt}.

\begin{lemma}[\PNPG{} Projection Error] \label{lem:off_projection_error}
Under \cref{asm:off_pnpg_bias,asm:off_pnpg_opt}, for every $k\in[K]$ and $h\in[H]$,
\begin{align}
\overline{\varepsilon}_{k,h}\leq2(1+\chi_G)\sqrt{\varepsilon_{\mathrm{bias}}}+2\chi_G\sqrt{\varepsilon_{\mathrm{opt}}}.
\end{align}
\end{lemma}

\begin{proof}
Consider specific $k$, $h$, $s$, and $u$. We first reduce the KL difference to a uniform logit error, and then bound that logit error by the two assumptions.

To this end, we define $z(a)\coloneq z_{k,h}(s,a)$ and $y(a)\coloneq\boldsymbol{x}(s,a)^\top\boldsymbol{\theta}_{k+1,h}$ as the logits of $\pi_{k+1/2,h}(\cdot\mid s)$ and $\pi_{k+1,h}(\cdot\mid s)$, and let $Z(\boldsymbol{w})\coloneq\log\sum_{a\in\mathcal{A}}\exp(w(a))$ denote the log-partition function of a logit vector $\boldsymbol{w}$. The terms $u(a)\log u(a)$ cancel in the KL difference, and $\log\pi_{k+1/2,h}(a\mid s)=z(a)-Z(\boldsymbol{z})$ and $\log\pi_{k+1,h}(a\mid s)=y(a)-Z(\boldsymbol{y})$ give
\begin{align}
D_{\mathrm{KL}}(u\|\pi_{k+1,h}(\cdot\mid s))-D_{\mathrm{KL}}(u\|\pi_{k+1/2,h}(\cdot\mid s))&=\sum_au(a)\bigl(z(a)-y(a)\bigr)-\bigl(Z(\boldsymbol{z})-Z(\boldsymbol{y})\bigr).
\end{align}
The first sum is a $u$-average of the logit differences. The log-partition function is $1$-Lipschitz in the sup norm, since its gradient is a probability vector. Therefore
\begin{align}
\bigl|\varepsilon_{k,h}^{\mathrm{proj}}(u,s)\bigr|&\leq2\sup_{(s',a')\in\mathcal{S}\times\mathcal{A}}\bigl|\boldsymbol{x}(s',a')^\top\boldsymbol{\theta}_{k+1,h}-z_{k,h}(s',a')\bigr|. \label{eq:pnpg_proj_logit}
\end{align}

It remains to bound the uniform logit error on the right-hand side of~\cref{eq:pnpg_proj_logit}. Let $\boldsymbol{\theta}^\dagger$ be a minimizer of the full actor loss $\ell_{k,h}^{\mathrm{act}}$, and let $\boldsymbol{\theta}^\star\coloneq\boldsymbol{\theta}_{k+1,h}^\star$ be the unique minimizer of the coreset actor loss $\widetilde{\ell}_{k,h}^{\mathrm{act}}$. By the triangle inequality, for every $(s',a')$,
\begin{align}
\bigl|\boldsymbol{x}(s',a')^\top\boldsymbol{\theta}_{k+1,h}-z_{k,h}(s',a')\bigr|&\leq\underbrace{\bigl|\boldsymbol{x}(s',a')^\top(\boldsymbol{\theta}_{k+1,h}-\boldsymbol{\theta}^\star)\bigr|}_{\text{(I)}}+\underbrace{\bigl|\boldsymbol{x}(s',a')^\top(\boldsymbol{\theta}^\star-\boldsymbol{\theta}^\dagger)\bigr|}_{\text{(II)}}\\
&\quad+\underbrace{\bigl|\boldsymbol{x}(s',a')^\top\boldsymbol{\theta}^\dagger-z_{k,h}(s',a')\bigr|}_{\text{(III)}}.
\end{align}
For Term (I),~\cref{eq:pnpg_chi_transfer,eq:pnpg_quadratic_identity} and \cref{asm:off_pnpg_opt} give
\begin{align}
\text{Term (I)}\leq\chi_G\|\boldsymbol{\theta}_{k+1,h}-\boldsymbol{\theta}^\star\|_{\boldsymbol{G}}=\chi_G\sqrt{\widetilde{\ell}_{k,h}^{\mathrm{act}}(\boldsymbol{\theta}_{k+1,h})-\widetilde{\ell}_{k,h}^{\mathrm{act}}(\boldsymbol{\theta}^\star)}\leq\chi_G\sqrt{\varepsilon_{\mathrm{opt}}}.
\end{align}
For Term (II), the same two facts give $\text{Term (II)}\leq\chi_G\|\boldsymbol{\theta}^\dagger-\boldsymbol{\theta}^\star\|_{\boldsymbol{G}}$, and
\begin{align}
\|\boldsymbol{\theta}^\dagger-\boldsymbol{\theta}^\star\|_{\boldsymbol{G}}^2=\widetilde{\ell}_{k,h}^{\mathrm{act}}(\boldsymbol{\theta}^\dagger)-\widetilde{\ell}_{k,h}^{\mathrm{act}}(\boldsymbol{\theta}^\star)\leq\widetilde{\ell}_{k,h}^{\mathrm{act}}(\boldsymbol{\theta}^\dagger)\leq\ell_{k,h}^{\mathrm{act}}(\boldsymbol{\theta}^\dagger)\leq\varepsilon_{\mathrm{bias}},
\end{align}
where the last two inequalities are~\cref{eq:pnpg_loss_order} and \cref{asm:off_pnpg_bias}. Hence Term (II) is at most $\chi_G\sqrt{\varepsilon_{\mathrm{bias}}}$. For Term (III), the definition of $\ell_{k,h}^{\mathrm{act}}$ as a supremum and \cref{asm:off_pnpg_bias} give $\text{Term (III)}\leq\sqrt{\ell_{k,h}^{\mathrm{act}}(\boldsymbol{\theta}^\dagger)}\leq\sqrt{\varepsilon_{\mathrm{bias}}}$.

Finally, adding the bounds on Terms (I), (II), and (III), substituting them into~\cref{eq:pnpg_proj_logit}, and taking the supremum over $s$ and $u$ bounds $\sup_{s,u}|\varepsilon_{k,h}^{\mathrm{proj}}(u,s)|\geq\overline{\varepsilon}_{k,h}$, which proves the lemma.
\end{proof}
The lemma holds for any coreset with $\boldsymbol{G}\succ0$. With the coreset from~\cref{thm:kiefer_wolfowitz}, $\chi_G=\sqrt{d_x}$, and hence $\overline{\varepsilon}_{k,h}\leq\mathcal{O}(\sqrt{d_x}(\sqrt{\varepsilon_{\mathrm{bias}}}+\sqrt{\varepsilon_{\mathrm{opt}}}))$.

Combined with~\cref{eq:pnpg_kl_error}, \cref{lem:off_projection_error} gives $\varepsilon_{\mathrm{proj}}\leq(2H/\eta)\bigl((1+\chi_G)\sqrt{\varepsilon_{\mathrm{bias}}}+\chi_G\sqrt{\varepsilon_{\mathrm{opt}}}\bigr)$. Suppose that the policy features are sufficiently expressive, in the sense that they approximate every $\widehat{Q}_{k,h}$ uniformly within $\epsilon/(4H(1+\chi_G))$. Substituting $\boldsymbol{\theta}=\boldsymbol{\theta}_{k,h}+\eta\boldsymbol{w}$ in~\cref{eq:pnpg_full_projection} shows that~\cref{asm:off_pnpg_bias} holds with $\varepsilon_{\mathrm{bias}}=\eta^2\epsilon^2/(16H^2(1+\chi_G)^2)$. Hence, $\varepsilon_{\mathrm{proj}}\leq\epsilon$ whenever, in addition, $\varepsilon_{\mathrm{opt}}\leq\eta^2\epsilon^2/(16H^2\chi_G^2)$, which holds after $T_k\geq\kappa_G\log(16H^4G^2\chi_G^2/\epsilon^2)$ steps of gradient descent. The $\widetilde{\mathcal{O}}(\epsilon^{-2})$ rate in~\cref{thm:off_policy,thm:lin_off} therefore applies to a fixed policy class. If reaching a smaller $\epsilon$ requires richer features, the resulting growth of $d_\Pi=Hd_x$ (\cref{prop:pnpg_oracle}) enters the sample complexity.

\paragraph{Bounded Parameters}
The next lemma shows that the parameters remain bounded. This is not an assumption: \OPE{} clips $\widehat{Q}_{k,h}$. Hence, each \NPG{} step shifts the logits by a bounded amount, and the coreset regression transfers this bound to the parameters.

\begin{lemma}[Bounded Parameters and Policy Cover] \label{lem:pnpg_covering}
Under \cref{asm:off_pnpg_opt}, every parameter $\boldsymbol{\theta}_{k,h}$ produced through update $K$ satisfies
\begin{align}
\|\boldsymbol{\theta}_{k,h}\|_2\leq B_{\theta,K}\coloneq K\left(\frac{\eta C_x\sqrt{d_x}HG}{\lambda_G}+\sqrt{\frac{\varepsilon_{\mathrm{opt}}}{\lambda_G}}\right). \label{eq:pnpg_parameter_envelope}
\end{align}
Consequently, the class $\Pi\coloneq\{\pi_{\boldsymbol{\theta}}\mid \|\boldsymbol{\theta}_h\|_2\leq B_{\theta,K}\text{ for every }h\in[H]\}$ is fixed before data collection, contains $\pi_1$ and every \PNPG{} iterate, and satisfies
\begin{align}
\log\mathcal{N}(\varepsilon,\Pi,\mathrm{dist}_\Pi)&\leq Hd_x\log\left(1+\frac{4C_x\sqrt{d_x}B_{\theta,K}}{\varepsilon}\right). \label{eq:pnpg_policy_cover}
\end{align}
\end{lemma}

\begin{proof}
We first bound each parameter update, and then translate a cover of the parameter ball into a cover of the policies.

We first bound the parameters. Consider specific $k$ and $h$. The gradient of~\cref{eq:pnpg_actor_loss} vanishes at its minimizer $\boldsymbol{\theta}_{k+1,h}^\star$. Since $z_{k,h}(s,a)-\boldsymbol{x}(s,a)^\top\boldsymbol{\theta}_{k,h}=\eta\widehat{Q}_{k,h}(s,a)$, this first-order condition reads
\begin{align}
\boldsymbol{\theta}_{k+1,h}^\star-\boldsymbol{\theta}_{k,h}=\eta\boldsymbol{G}^{-1}\sum_{(s,a)\in\mathcal{C}}\rho_{\mathcal{C}}(s,a)\boldsymbol{x}(s,a)\widehat{Q}_{k,h}(s,a).
\end{align}
The clipping in \cref{alg:ope} gives $|\widehat{Q}_{k,h}(s,a)|\leq(H-h+1)G\leq HG$. Together with $\|\boldsymbol{G}^{-1}\|_{\mathrm{op}}=1/\lambda_G$ and $\|\boldsymbol{x}(s,a)\|_2\leq C_x\sqrt{d_x}$, this gives $\|\boldsymbol{\theta}_{k+1,h}^\star-\boldsymbol{\theta}_{k,h}\|_2\leq\eta C_x\sqrt{d_x}HG/\lambda_G$. By~\cref{eq:pnpg_quadratic_identity}, \cref{asm:off_pnpg_opt}, and $\boldsymbol{G}\succeq\lambda_GI$, the computed update satisfies $\|\boldsymbol{\theta}_{k+1,h}-\boldsymbol{\theta}_{k+1,h}^\star\|_2\leq\sqrt{\varepsilon_{\mathrm{opt}}/\lambda_G}$. The triangle inequality then gives $\|\boldsymbol{\theta}_{k+1,h}-\boldsymbol{\theta}_{k,h}\|_2\leq B_{\theta,K}/K$. Since $\boldsymbol{\theta}_{1,h}=\boldsymbol{0}$, summing at most $K$ such increments gives $\|\boldsymbol{\theta}_{k,h}\|_2\leq B_{\theta,K}$. This argument uses only the clipping range of the fitted values, and it therefore holds for every possible history. Hence $\Pi$ is fixed before data collection and contains $\pi_1$ and every \PNPG{} iterate.

We next cover the policies. Consider specific $h$, $s$, and $\boldsymbol{\theta}_h,\boldsymbol{\theta}_h'\in\mathbb{R}^{d_x}$. Along the path $\boldsymbol{\theta}_h(t)\coloneq\boldsymbol{\theta}_h'+t(\boldsymbol{\theta}_h-\boldsymbol{\theta}_h')$ for $t\in[0,1]$, let $\pi_t$ be the softmax policy at $s$ with logits $z_t(a)\coloneq\boldsymbol{x}(s,a)^\top\boldsymbol{\theta}_h(t)$, and define $\delta(a)\coloneq\boldsymbol{x}(s,a)^\top(\boldsymbol{\theta}_h-\boldsymbol{\theta}_h')$. Differentiating the softmax gives $\mathrm{d}\pi_t(a)/\mathrm{d}t=\pi_t(a)(\delta(a)-\sum_b\pi_t(b)\delta(b))$, and hence $\sum_a|\mathrm{d}\pi_t(a)/\mathrm{d}t|\leq2\max_a|\delta(a)|$. Integrating over $t\in[0,1]$ and applying the Cauchy--Schwarz inequality,
\begin{align}
\sup_s\|\pi_{\boldsymbol{\theta},h}(\cdot\mid s)-\pi_{\boldsymbol{\theta}',h}(\cdot\mid s)\|_1\leq2C_x\sqrt{d_x}\,\|\boldsymbol{\theta}_h-\boldsymbol{\theta}_h'\|_2. \label{eq:pnpg_softmax_lipschitz}
\end{align}
Let $r\coloneq\varepsilon/(2C_x\sqrt{d_x})$. By \cref{lem:euclidean_cover}, the ball $\{\boldsymbol{\theta}_h\mid\|\boldsymbol{\theta}_h\|_2\leq B_{\theta,K}\}$ admits an internal $r$-cover $\mathcal{N}_r$ with at most $(1+4C_x\sqrt{d_x}B_{\theta,K}/\varepsilon)^{d_x}$ points. Choosing the nearest point of $\mathcal{N}_r$ separately at each step and applying~\cref{eq:pnpg_softmax_lipschitz}, the product collection $\{\pi_{\boldsymbol{\theta}'}\mid\boldsymbol{\theta}_h'\in\mathcal{N}_r\text{ for every }h\in[H]\}\subseteq\Pi$ is an $\varepsilon$-cover of $\Pi$ in $\mathrm{dist}_\Pi$. It has at most $|\mathcal{N}_r|^H$ elements, which proves~\cref{eq:pnpg_policy_cover}.
\end{proof}

\paragraph{Verification of \cref{asm:pnpg,asm:policy_class}}

\begin{proposition}[\PNPG{} Satisfies~\cref{asm:pnpg,asm:policy_class}] \label{prop:pnpg_oracle}
Suppose \cref{asm:off_pnpg_bias,asm:off_pnpg_opt} hold, $C_x$, $\lambda_G$, and $\varepsilon_{\mathrm{opt}}$ do not depend on $|\mathcal{S}|$, and $Hd_x\ll\min\{K,|\mathcal{S}|\}$. Let \PNPG{} use the step size $\eta=\sqrt{2\log|\mathcal{A}|/(KH^2G^2)}$, and let $\Pi$ be the class in \cref{lem:pnpg_covering}. Then \PNPG{} satisfies~\cref{asm:pnpg} with $\varepsilon_{\mathrm{proj}}$ in~\cref{eq:pnpg_kl_error}, controlled by~\cref{lem:off_projection_error}, and $\Pi$ satisfies~\cref{asm:policy_class} with $d_\Pi=Hd_x$.
\end{proposition}

\begin{proof}
We verify~\cref{asm:pnpg,asm:policy_class} separately.

We first verify~\cref{asm:pnpg}. Consider specific $h$, $s$, and a comparator policy $u$. The clipping in \cref{alg:ope} gives $\|\widehat{Q}_{k,h}(s,\cdot)\|_\infty\leq HG$. Applying \cref{thm:off_projected_npg_regret} with comparator $u_h(\cdot\mid s)$ and bounding each $\varepsilon_{k,h}^{\mathrm{proj}}(u_h(\cdot\mid s),s)$ by $\overline{\varepsilon}_{k,h}$ in~\cref{eq:pnpg_kl_error} gives
\begin{align}
\sum_{k=1}^K\bigl\langle \widehat{Q}_{k,h}(s,\cdot),u_h(\cdot\mid s)-\pi_{k,h}(\cdot\mid s) \bigr\rangle\leq\frac{\log|\mathcal{A}|}{\eta}+\frac{\eta KH^2G^2}{2}+\frac{K}{\eta}\max_{k\in[K]}\overline{\varepsilon}_{k,h}.
\end{align}
The right-hand side does not depend on $s$. Taking the expectation over $s\sim d_h^u$, summing over $h\in[H]$, and using the step size and the definition of $\varepsilon_{\mathrm{proj}}$,
\begin{align}
\sum_{k=1}^K\sum_{h=1}^H\mathbb{E}_{s\sim d_h^u}\bigl\langle \widehat{Q}_{k,h}(s,\cdot),u_h(\cdot\mid s)-\pi_{k,h}(\cdot\mid s) \bigr\rangle&\leq H^2G\sqrt{2K\log|\mathcal{A}|}+K\varepsilon_{\mathrm{proj}}.
\end{align}
This implies~\cref{asm:pnpg}.

We next verify~\cref{asm:policy_class}. By \cref{lem:pnpg_covering}, $\Pi$ contains every \PNPG{} iterate. With the prescribed step size, $\eta HG=\sqrt{2\log|\mathcal{A}|/K}\leq\sqrt{2\log|\mathcal{A}|}$, and~\cref{eq:pnpg_parameter_envelope} gives $4C_x\sqrt{d_x}B_{\theta,K}\leq c_\Pi K$, where
\begin{align}
c_\Pi\coloneq\frac{4C_x^2d_x\sqrt{2\log|\mathcal{A}|}}{\lambda_G}+4C_x\sqrt{\frac{d_x\varepsilon_{\mathrm{opt}}}{\lambda_G}}.
\end{align}
Substituting into~\cref{eq:pnpg_policy_cover} and using $1+c_\Pi K/\varepsilon\leq(1+c_\Pi)(1+K/\varepsilon)$ and $\log(1+K/\varepsilon)\geq\log2$ gives, for every $\varepsilon\in(0,1]$,
\begin{align}
\log\mathcal{N}(\varepsilon,\Pi,\mathrm{dist}_\Pi)&\leq\left(1+\frac{\log(1+c_\Pi)}{\log2}\right)Hd_x\log\left(1+\frac K\varepsilon\right).
\end{align}
The leading constant depends neither on $K$ nor on $|\mathcal{S}|$. Together with $Hd_x\ll\min\{K,|\mathcal{S}|\}$, this verifies~\cref{asm:policy_class} with $d_\Pi=Hd_x$.
\end{proof}

\section{Deferred Details and Proofs for \texorpdfstring{\cref{section:ocupo-on}}{Section \ref*{section:ocupo-on}}} \label{app:ucgu_on}

We provide the deferred regression derivations and the pseudocode of~\cref{alg:tme_bilinear,alg:pre}, followed by the estimation guarantees and the sample-complexity proof. We first collect the design of the three estimation components and the covering argument for the value functions, then define their good events and prove the corresponding guarantees. Each guarantee states the order of its confidence radius. The exact value is given in its proof. All conditional statements for iteration $k$ condition on the history before its fresh batch is collected. Under this conditioning, $\pi_k$ is fixed and the $N$ trajectories are independent. Constants denoted by $c$ below are numerical and can be increased when necessary.

\subsection{Algorithm Design} \label{app:on_algorithm_design}

\subsubsection{Transition Model Estimation} \label{app:model_on} \label{app:on_tme_details}

Using the collected trajectories, we construct $\mathbf{M}_k$, an estimate of $\mathbf{M}^\star$. The following lemma gives an unbiased regression target built from the known successor features.

\begin{lemma}[Unbiased Transition-Core Regression] \label{lem:bilinear_regression_response}
Suppose the transition kernel satisfies \cref{def:bilienar_mdp}. Let $\mathcal{H}$ contain the history through a current state-action pair $(s,a)$, and let $S'$ have conditional law $P^\star(\cdot\mid s,a)$. Then $\boldsymbol{\psi}(S')^\top\mathbf{K}_\psi^{-1}$ is an unbiased estimator of $\boldsymbol{\phi}(s,a)^\top\mathbf{M}^\star$ conditional on $\mathcal{H}$:
\begin{align}
\mathbb{E}[\boldsymbol{\psi}(S')^\top\mathbf{K}_\psi^{-1}\mid\mathcal{H}]&=\boldsymbol{\phi}(s,a)^\top\mathbf{M}^\star.
\end{align}
This identity holds under any behavior policy, including one chosen from the preceding history.
\end{lemma}

\begin{proof}
Conditional on $\mathcal{H}$, the current pair $(s,a)$ is fixed and only its successor $S'$ is random. Hence
\begin{align}
\mathbb{E}[\boldsymbol{\psi}(S')^\top\mathbf{K}_\psi^{-1}\mid\mathcal{H}]&=\sum_{s'}P^\star(s'\mid s,a)\boldsymbol{\psi}(s')^\top\mathbf{K}_\psi^{-1}\\
&=\boldsymbol{\phi}(s,a)^\top\mathbf{M}^\star\biggl(\sum_{s'}\boldsymbol{\psi}(s')\boldsymbol{\psi}(s')^\top\biggr)\mathbf{K}_\psi^{-1}\\
&=\boldsymbol{\phi}(s,a)^\top\mathbf{M}^\star.
\end{align}
The first equality is the definition of the conditional expectation under the law $P^\star(\cdot\mid s,a)$. The second equality substitutes the bilinear transition $P^\star(s'\mid s,a)=\boldsymbol{\phi}(s,a)^\top\mathbf{M}^\star\boldsymbol{\psi}(s')$ from \cref{def:bilienar_mdp}. The last equality uses $\mathbf{K}_\psi=\sum_{s'}\boldsymbol{\psi}(s')\boldsymbol{\psi}(s')^\top$, which is invertible by condition (iii) of \cref{def:bilienar_mdp}. The behavior policy affects the distribution of the current pair but not the conditional law of its successor, which proves the final assertion.
\end{proof}

Following~\citet{yang2020reinforcement}, this conditional-mean identity motivates the ridge estimator $\mathbf{M}_k$ in~\cref{eq:bilinear_tme_estimator}. Its objective is a strictly convex quadratic in $\mathbf{M}$, and setting its gradient to zero gives the closed form in~\cref{eq:bilinear_tme_estimator}, computed in~\cref{line:tme_gram,line:tme_model} of~\cref{alg:tme_bilinear}, where $\boldsymbol{\Lambda}_k^{-1}$ exists because $\boldsymbol{\Lambda}_k\succeq I$. The Gram matrix $\boldsymbol{\Lambda}_k$ pools all steps because the transition model is shared across the horizon. The returned model is represented by its core and evaluated at queried transitions in~\cref{line:tme_query}.

\begin{algorithm2e}[htbp]
\caption{Bilinear Transition Estimation (\BTE{})} \label{alg:tme_bilinear}
\begin{algorithmic}[1]
\State \textbf{Input:} dataset $\mathcal{D}_k$.
\State $\boldsymbol{\Lambda}_k\gets I+\sum_{\tau_i\in\mathcal{D}_k}\sum_{h=1}^H\boldsymbol{\phi}_{i,h}\boldsymbol{\phi}_{i,h}^\top$. \label{line:tme_gram}
\State $\mathbf{M}_k\gets\boldsymbol{\Lambda}_k^{-1}\sum_{\tau_i\in\mathcal{D}_k}\sum_{h=1}^H\boldsymbol{\phi}_{i,h}\boldsymbol{\psi}(s_{i,h+1})^\top\mathbf{K}_\psi^{-1}$. \label{line:tme_model}
\State $\widehat{P}_k(s'\mid s,a)\gets\boldsymbol{\phi}(s,a)^\top\mathbf{M}_k\boldsymbol{\psi}(s')$, for every queried $(s'\mid s,a)$. \label{line:tme_query}
\State \Return\quad $\widehat{P}_k$, represented by $\mathbf{M}_k$.
\end{algorithmic}
\end{algorithm2e}

\subsubsection{Pseudo-Reward Estimation} \label{app:reward_on} \label{app:on_pre_details}

The occupancy decomposition in \cref{lem:occupancy_measure} shows that, given the transition model, it suffices to estimate the low-dimensional feature expectations $\boldsymbol{\omega}_h^{\pi_k}$ for $h\in[H-1]$. This can be done by rolling out the current policy and averaging the features encountered at each step. \cref{alg:pre} combines these averages with $\mathbf{M}_k$ to estimate the occupancy measure and then evaluates the utility gradient. The estimate in \cref{line:omega} is $\widehat{\boldsymbol{\omega}}_{k,h}=N^{-1}\sum_{i=1}^N\boldsymbol{\phi}_{i,h}$. It is an unbiased estimator of $\boldsymbol{\omega}_h^{\pi_k}$, since each step-$h$ pair of the fresh batch has distribution $\mu_h^{\pi_k}$.

The initial occupancy in \cref{line:mu_init} is known exactly. The estimated core need not define a normalized transition kernel. Hence, its induced occupancy coordinates may lie outside the utility domain. At later steps, coordinatewise clipping in \cref{line:mu} keeps the estimated occupancies in $[0,1]$. The utility domain in~\cref{asm:lipschitzness_and_smoothness} ensures that the resulting gradients are well defined.

These functions are evaluated on demand: by coordinate separability, a query of $\widehat{R}_{k,h}(s,a)$ in \cref{line:R_hat} requires only the occupancy coordinate at $(s,a)$. In particular, \cref{alg:pre} does not enumerate the state-action space.

\begin{algorithm2e}[htbp]
\caption{Pseudo-Reward Estimation by Sampling} \label{alg:pre}
\begin{algorithmic}[1]
\State \textbf{Input:} policy $\pi_k$, model core $\mathbf{M}_k$, fresh dataset $\mathcal{D}_k=\{\tau_i\}_{i=1}^N$.
\State $\widehat{\mu}_{k,1}(s,a)\gets\rho(s)\pi_{k,1}(a\mid s)$, for every queried $(s,a)$. \label{line:mu_init}
\State $\widehat{R}_{k,1}(s,a)\gets[\nabla f_1(\widehat{\mu}_{k,1})](s,a)$, for every queried $(s,a)$.
\For{$h=1,\ldots,H-1$}
    \State $\widehat{\boldsymbol{\omega}}_{k,h}\gets N^{-1}\sum_{i=1}^N\boldsymbol{\phi}(s_{i,h},a_{i,h})$. \label{line:omega}
    \State $\widehat{\mu}_{k,h+1}(s,a)\gets\operatorname{Clip}(\pi_{k,h+1}(a\mid s)\widehat{\boldsymbol{\omega}}_{k,h}^\top\mathbf{M}_k\boldsymbol{\psi}(s),[0,1])$, for every queried $(s,a)$. \label{line:mu}
    \State $\widehat{R}_{k,h+1}(s,a)\gets[\nabla f_{h+1}(\widehat{\mu}_{k,h+1})](s,a)$, for every queried $(s,a)$. \label{line:R_hat}
\EndFor
\State \Return\quad $\widehat{R}_k$.
\end{algorithmic}
\end{algorithm2e}

\subsubsection{Optimistic Policy Evaluation} \label{app:ope_on} \label{app:on_ope_details}

With the pseudo-reward estimate in hand, we next evaluate the action values.
Recall that $Q_h^{\pi,R}(s,a)=R_h(s,a)+P^\star V_{h+1}^{\pi,R}(s,a)$, where $P^\star V(s,a)\coloneq\sum_{s'}P^\star(s'\mid s,a)V(s')$. The following lemma shows that the expected next-state value can be represented by a low-dimensional coefficient, even when $V$ itself is not linear in the features.

\begin{lemma}[Linearity of Expected Next-State Values] \label{lem:bilinear_next_state_values}
Suppose that the MDP is a bilinear MDP (\cref{def:bilienar_mdp}). For any bounded function $V\colon \mathcal{S}\to\mathbb{R}$, let $\boldsymbol{v}^\star(V)\coloneq\mathbf{M}^\star\sum_{s'}\boldsymbol{\psi}(s')V(s')\in\mathbb{R}^{d_\phi}$. Then, for every $(s,a)\in\mathcal{S}\times\mathcal{A}$,
\begin{align}
P^\star V(s,a)&=\boldsymbol{\phi}(s,a)^\top\boldsymbol{v}^\star(V).
\end{align}
\end{lemma}

\begin{proof}
We compute
\begin{align}
P^\star V(s,a)&=\sum_{s'}P^\star(s'\mid s,a)V(s')\\
&=\sum_{s'}\boldsymbol{\phi}(s,a)^\top\mathbf{M}^\star\boldsymbol{\psi}(s')V(s')\\
&=\boldsymbol{\phi}(s,a)^\top\biggl(\mathbf{M}^\star\sum_{s'}\boldsymbol{\psi}(s')V(s')\biggr)\\
&=\boldsymbol{\phi}(s,a)^\top\boldsymbol{v}^\star(V).
\end{align}
The first equality is the definition of $P^\star V$. The second substitutes the bilinear transition $P^\star(s'\mid s,a)=\boldsymbol{\phi}(s,a)^\top\mathbf{M}^\star\boldsymbol{\psi}(s')$ from \cref{def:bilienar_mdp}. The third pulls $\boldsymbol{\phi}(s,a)^\top\mathbf{M}^\star$ out of the sum, since it does not depend on the successor state $s'$. The last equality is the definition of $\boldsymbol{v}^\star(V)$. The coefficient $\boldsymbol{v}^\star(V)$ depends on $V$ and the MDP, but not on the queried pair $(s,a)$, which establishes linearity in the current features.
\end{proof}

In particular, writing $\boldsymbol{v}_{k,h}^\star=\boldsymbol{v}^\star(\widehat{V}_{k,h+1})$, \cref{lem:bilinear_next_state_values} gives the linear representation of expected next-state values used by \OPE{}.

To account for transition uncertainty, we follow~\citet{yang2020reinforcement} and first consider an ideal optimistic backup, which motivates \cref{alg:ope}. In particular, for every iteration $k \in [K]$, let $R_{k,h}$ be the true pseudo-reward derived from $\pi_k$ and $\beta^{\mathrm{On}}>0$ be the radius parameter in \cref{lem:on_model_confidence_formal}. We define the matrix confidence ball as
\begin{align}
B_k\coloneq\{M\in\mathbb{R}^{d_\phi\times d_\psi}\mid\|\boldsymbol{\Lambda}_k^{1/2}(M-\mathbf{M}_k)\|_F\leq\sqrt{\beta^{\mathrm{On}}}\}.
\end{align} If the true pseudo-reward $R_{k,h}$ and $\overline{\boldsymbol{\psi}}_{k,h}\coloneq\sum_{s^\prime\in\mathcal{S}}\boldsymbol{\psi}(s^\prime)\widehat{V}_{k,h+1}(s^\prime)$ were available, the corresponding optimistic backup would be
\begin{align}
\widetilde{Q}_{k,h}(\cdot,\cdot)=R_{k,h}(\cdot,\cdot)+\max_{M\in B_k}\boldsymbol{\phi}(\cdot,\cdot)^\top M\overline{\boldsymbol{\psi}}_{k,h}.
\end{align}
The optimization over the matrix confidence ball admits a closed-form solution. For any $M\in B_k$ and a fixed pair $(s,a)$, we have $\boldsymbol{\phi}(s,a)^\top M\overline{\boldsymbol{\psi}}_{k,h}=\boldsymbol{\phi}(s,a)^\top\mathbf{M}_k\overline{\boldsymbol{\psi}}_{k,h}+\boldsymbol{\phi}(s,a)^\top(M-\mathbf{M}_k)\overline{\boldsymbol{\psi}}_{k,h}$, and
\begin{align}
\boldsymbol{\phi}(s,a)^\top(M-\mathbf{M}_k)\overline{\boldsymbol{\psi}}_{k,h}&=\bigl\langle \boldsymbol{\Lambda}_k^{1/2}(M-\mathbf{M}_k),\boldsymbol{\Lambda}_k^{-1/2}\boldsymbol{\phi}(s,a)\overline{\boldsymbol{\psi}}_{k,h}^\top\bigr\rangle_F\\
&\leq\sqrt{\beta^{\mathrm{On}}}\bigl\lVert \boldsymbol{\phi}(s,a) \bigr\rVert_{\boldsymbol{\Lambda}_k^{-1}}\bigl\lVert \overline{\boldsymbol{\psi}}_{k,h} \bigr\rVert_2,
\end{align}
by the Cauchy--Schwarz inequality, $\|\boldsymbol{u}\boldsymbol{w}^\top\|_F=\|\boldsymbol{u}\|_2\|\boldsymbol{w}\|_2$, and the definition of $B_k$. Equality holds when $\boldsymbol{\Lambda}_k^{1/2}(M-\mathbf{M}_k)$ is a nonnegative multiple of $\boldsymbol{\Lambda}_k^{-1/2}\boldsymbol{\phi}(s,a)\overline{\boldsymbol{\psi}}_{k,h}^\top$ with Frobenius norm $\sqrt{\beta^{\mathrm{On}}}$. Therefore
\begin{align}
\widetilde{Q}_{k,h}(\cdot,\cdot)=R_{k,h}(\cdot,\cdot)+\boldsymbol{\phi}(\cdot,\cdot)^\top\mathbf{M}_k\overline{\boldsymbol{\psi}}_{k,h}+\sqrt{\beta^{\mathrm{On}}}\bigl\lVert \boldsymbol{\phi}(\cdot,\cdot) \bigr\rVert_{\boldsymbol{\Lambda}_k^{-1}}\bigl\lVert \overline{\boldsymbol{\psi}}_{k,h} \bigr\rVert_2.
\end{align}
However, this oracle expression is not directly implementable, since $R_{k,h}$ is unknown, and evaluating $\overline{\boldsymbol{\psi}}_{k,h}$ requires summing over the entire state space. Hence,~\cref{alg:curl_framework} replaces $R_{k,h}$ with its estimate $\widehat{R}_{k,h}$. The pseudo-reward error is then carried by the analysis rather than by the backup. To estimate the expected next-state value without a state-space sum, we regress the observed successor values directly on the current state-action features.

Specifically, for each trajectory $\tau_i\in\mathcal{D}_k$ passed to \OPE{}, define $\boldsymbol{\phi}_{i,h}=\boldsymbol{\phi}(s_{i,h},a_{i,h})$. Together, \hyperref[line:ope_gram_matrix]{\cref{line:ope_gram_matrix}} and~\hyperref[line:vtr_estimator]{\cref{line:vtr_estimator}} of~\cref{alg:ope} solve the ridge-regression problem in~\cref{eq:ope_ridge_objective}.

\begin{lemma}[Unbiased Value Regression] \label{lem:bilinear_value_response}
Under \cref{def:bilienar_mdp}, let $V\colon \mathcal{S}\to\mathbb{R}$ be a bounded target fixed before a successor $S'$ is drawn, and let $\boldsymbol{v}^\star(V)$ be as in \cref{lem:bilinear_next_state_values}. Conditional on a history $\mathcal{H}$ that fixes $V$ and the current pair $(s,a)$, with $S'\sim P^\star(\cdot\mid s,a)$, $V(S')$ is an unbiased estimator of the expected next-state value, which is linear in the current features:
\begin{align}
\mathbb{E}[V(S')\mid\mathcal{H}]&=P^\star V(s,a)=\boldsymbol{\phi}(s,a)^\top\boldsymbol{v}^\star(V).
\end{align}
\end{lemma}

\begin{proof}
Since $V$ and $(s,a)$ are fixed under the conditioning and $S'$ has law $P^\star(\cdot\mid s,a)$, the definition of the conditional expectation gives $\mathbb{E}[V(S')\mid\mathcal{H}]=\sum_{s'}P^\star(s'\mid s,a)V(s')$, which is $P^\star V(s,a)$ by definition. \cref{lem:bilinear_next_state_values} then gives $P^\star V(s,a)=\boldsymbol{\phi}(s,a)^\top\boldsymbol{v}^\star(V)$. Thus $V(S')-P^\star V(s,a)$ has conditional mean zero. The sum defining $\boldsymbol{v}^\star(V)$ is a population coefficient. It is not evaluated to observe $V(S')$.
\end{proof}

The objective in~\cref{eq:ope_ridge_objective} is a strictly convex quadratic in $\boldsymbol{v}$ with gradient $2\sum_{\tau_i\in\mathcal{D}_k}\boldsymbol{\phi}_{i,h}(\boldsymbol{\phi}_{i,h}^\top\boldsymbol{v}-\widehat{V}_{k,h+1}(s_{i,h+1}))+2\boldsymbol{v}$. Setting the gradient to zero gives the normal equations $\boldsymbol{\Gamma}_{k,h}\widehat{\boldsymbol{v}}_{k,h}=\sum_{\tau_i\in\mathcal{D}_k}\boldsymbol{\phi}_{i,h}\widehat{V}_{k,h+1}(s_{i,h+1})$, where $\boldsymbol{\Gamma}_{k,h}=I+\sum_{\tau_i\in\mathcal{D}_k}\boldsymbol{\phi}_{i,h}\boldsymbol{\phi}_{i,h}^\top\succeq I$ is invertible. Their unique solution is
\begin{align}
\widehat{\boldsymbol{v}}_{k,h} =\boldsymbol{\Gamma}_{k,h}^{-1}\sum_{\tau_i\in\mathcal{D}_k}\boldsymbol{\phi}_{i,h}\widehat{V}_{k,h+1}(s_{i,h+1}).
\end{align}
This explains both the design matrix and the coefficient update in \hyperref[line:ope_gram_matrix]{\cref{line:ope_gram_matrix}}--\hyperref[line:vtr_estimator]{\cref{line:vtr_estimator}}. The computation uses only the observed successor values in $\mathcal{D}_k$ and a $d_\phi\times d_\phi$ matrix solve. It requires neither $\mathbf{M}^\star$ nor an integral oracle or enumeration of $\mathcal{S}$. The resulting coefficient estimates the linear prediction $P^\star\widehat{V}_{k,h+1}$.

In the algorithm, $\widehat{V}_{k,h+1}$ is constructed from the same dataset. Hence, the population identity does not assert that the selected regression residuals are conditionally centered on that dataset. We handle target reuse by proving concentration uniformly over the value-function class of \cref{app:on_covering_arguments}, and then substituting the fitted target. No additional trajectories are required.

Finally, \cref{line:optimistic_Q} adds the confidence bonus $\alpha_h^{\mathrm{On}}\|\boldsymbol{\phi}(\cdot,\cdot)\|_{\boldsymbol{\Gamma}_{k,h}^{-1}}$ to the predicted expected next-state value. We clip the resulting $Q$-function in \cref{line:clipped_Q} and average it under $\pi_k$ in \cref{line:clipped_V} to obtain the next value target.

We use the conventions $\widehat{V}_{k,H+1}=0$, $\widehat{\boldsymbol{v}}_{k,H}=0$, and $\alpha_H=0$. The step-$H$ Gram matrix is defined by the same sample sum for the analysis, but no value regression is performed at that step.

\subsection{Covering Arguments} \label{app:on_covering_arguments}

This subsection collects the covering bound used in the proof of \cref{lem:on_value_regression_formal}. Throughout, consider a specific iteration $k$ and condition on the history before $\mathcal{D}_k$ is collected, so that $\pi_k$ is fixed.

\paragraph{Value-Function Class}
For a step $h\in[H]$, a bonus radius $\alpha\geq0$, and parameters $\vartheta=(\boldsymbol{c},\boldsymbol{v},\boldsymbol{\Gamma})$ with $\boldsymbol{c}\in\mathbb{R}^{d_\psi}$, $\boldsymbol{v}\in\mathbb{R}^{d_\phi}$, and a positive definite matrix $\boldsymbol{\Gamma}\in\mathbb{R}^{d_\phi\times d_\phi}$, define the occupancy and the induced pseudo-reward by
\begin{align}
\mu_h^{\boldsymbol{c}}(s,a)\coloneq\begin{cases}\rho(s)\pi_{k,1}(a\mid s),&h=1,\\
\operatorname{Clip}(\pi_{k,h}(a\mid s)\boldsymbol{c}^\top\boldsymbol{\psi}(s),[0,1]),&h\in\{2,\ldots,H\},\end{cases}\qquad R_h^{\boldsymbol{c}}\coloneq\nabla f_h(\mu_h^{\boldsymbol{c}}).
\end{align}
At $h=1$, neither depends on $\boldsymbol{c}$. With the clipping interval $\mathcal{I}_h\coloneq[-(H-h+1)G,(H-h+1)G]$, the corresponding value function is
\begin{align}
V_h^{\vartheta,\alpha}(s)&\coloneq\sum_a\pi_{k,h}(a\mid s)\operatorname{Clip}\bigl(R_h^{\boldsymbol{c}}(s,a)+\boldsymbol{\phi}(s,a)^\top\boldsymbol{v}+\alpha\|\boldsymbol{\phi}(s,a)\|_{\boldsymbol{\Gamma}^{-1}},\mathcal{I}_h\bigr).
\end{align}
The value-function class is
\begin{align}
\mathcal{V}_{k,h}(\alpha)&\coloneq\bigl\{V_h^{\vartheta,\alpha}\bigm|\|\boldsymbol{c}\|_2\leq C_\phi C_{\mathbf{K}}\sqrt{NHd_\phi},\ \|\boldsymbol{v}\|_2\leq(H-h)G\sqrt N,\ I\preceq\boldsymbol{\Gamma}\preceq(1+NC_\phi^2d_\phi)I\bigr\}. \label{eq:on_value_class_set}
\end{align}
At $h=H$, take $\boldsymbol{v}=\boldsymbol{0}$ and $\alpha=0$. Hence, the matrix parameter plays no role.

Every step-$h$ value estimate returned by the on-policy call to \OPE{} is $V_h^{\vartheta,\alpha_h^{\mathrm{On}}}$ with $\boldsymbol{c}=\mathbf{M}_k^\top\widehat{\boldsymbol{\omega}}_{k,h-1}$, $\boldsymbol{v}=\widehat{\boldsymbol{v}}_{k,h}$, and $\boldsymbol{\Gamma}=\boldsymbol{\Gamma}_{k,h}$, as can be read off from~\cref{eq:bilinear_occupancy_reconstruction,eq:pseudo_reward_estimate} and \cref{alg:ope}. These parameters satisfy the constraints in~\cref{eq:on_value_class_set}. First, since $\mathbf{M}_k$ minimizes the ridge objective in~\cref{eq:bilinear_tme_estimator}, its objective value is no larger than the value at $\mathbf{M}=\boldsymbol{0}$. Dropping the nonnegative squared-loss term and using condition (iii) of \cref{def:bilienar_mdp} for each of the $NH$ regression targets, we obtain
\begin{align}
\lVert \mathbf{M}_k \rVert_F^2\leq\sum_{i=1}^N\sum_{h=1}^H\bigl\lVert \boldsymbol{\psi}(s_{i,h+1})^\top\mathbf{K}_\psi^{-1} \bigr\rVert_2^2\leq NHC_{\mathbf{K}}^2. \label{eq:core_frobenius_bound}
\end{align}
Since $\widehat{\boldsymbol{\omega}}_{k,h-1}$ is an average of features with norm at most $C_\phi\sqrt{d_\phi}$, we obtain $\|\boldsymbol{c}\|_2\leq\|\mathbf{M}_k\|_F\|\widehat{\boldsymbol{\omega}}_{k,h-1}\|_2\leq C_\phi C_{\mathbf{K}}\sqrt{NHd_\phi}$. Second, comparing the ridge objective in~\cref{eq:ope_ridge_objective} with its value at $\boldsymbol{v}=\boldsymbol{0}$ and using the clipping range $|\widehat{V}_{k,h+1}|\leq(H-h)G$ of \cref{alg:ope}, we obtain $\|\widehat{\boldsymbol{v}}_{k,h}\|_2^2\leq N(H-h)^2G^2$. Finally, $\boldsymbol{\Gamma}_{k,h}=I+\sum_{i=1}^N\boldsymbol{\phi}_{i,h}\boldsymbol{\phi}_{i,h}^\top$ satisfies $I\preceq\boldsymbol{\Gamma}_{k,h}\preceq(1+NC_\phi^2d_\phi)I$, since $\boldsymbol{\phi}\boldsymbol{\phi}^\top\preceq C_\phi^2d_\phi I$.

\begin{lemma}[Cover of the Value Functions] \label{lem:on_value_covering_formal} \label{lem:value_class_covering}
For every $k\in[K]$, $h\in[H]$, $\alpha\geq0$, and $\varepsilon>0$, the class in~\cref{eq:on_value_class_set} contains every possible step-$h$ value estimate returned by \OPE{} when $\alpha=\alpha_h^{\mathrm{On}}$, and its covering number satisfies
\begin{align}
\log\mathcal{N}_\varepsilon(\mathcal{V}_{k,h}(\alpha),\lVert \cdot \rVert_\infty) &\leq d_\psi\log\left(1+\frac{8LC_\psi C_\phi C_{\mathbf{K}}\sqrt{NHd_\phi}}{\varepsilon}\right)+d_\phi\log\left(1+\frac{8C_\phi\sqrt{d_\phi}(H-h)G\sqrt N}{\varepsilon}\right)\\
&\quad+\frac{d_\phi(d_\phi+1)}2\log\left(1+\frac{32\alpha^2C_\phi^2d_\phi^{3/2}(1+NC_\phi^2d_\phi)}{\varepsilon^2}\right)\eqcolon\mathcal{H}_h^{\mathrm{On}}(\alpha,\varepsilon). \label{eq:on_cover_explicit}
\end{align}
In particular, if $1/\varepsilon$ and $\alpha$ are bounded by a polynomial in $N$, $H$, $d_\phi$, $d_\psi$, $L$, $G$, $G^{-1}$, $1/\delta$, and the feature constants, then the right-hand side is $\widetilde{\mathcal{O}}(d_\phi^2+d_\psi)$.
\end{lemma}

\begin{proof}
We first recall the relevant notation. Each member $V_h^{\vartheta,\alpha}$ of the class in~\cref{eq:on_value_class_set} averages, under $\pi_{k,h}$, the clipped backup $R_h^{\boldsymbol{c}}+\boldsymbol{\phi}^\top\boldsymbol{v}+\alpha\|\boldsymbol{\phi}\|_{\boldsymbol{\Gamma}^{-1}}$, where the pseudo-reward $R_h^{\boldsymbol{c}}=\nabla f_h(\mu_h^{\boldsymbol{c}})$ is induced by the clipped occupancy $\mu_h^{\boldsymbol{c}}(s,a)=\operatorname{Clip}(\pi_{k,h}(a\mid s)\boldsymbol{c}^\top\boldsymbol{\psi}(s),[0,1])$.

The containment was shown above. For the covering bound, consider $\vartheta=(\boldsymbol{c},\boldsymbol{v},\boldsymbol{\Gamma})$ and $\vartheta^\prime=(\boldsymbol{c}^\prime,\boldsymbol{v}^\prime,\boldsymbol{\Gamma}^\prime)$ in the parameter set of~\cref{eq:on_value_class_set}. Clipping onto an interval is nonexpansive, and averaging with respect to $\pi_{k,h}(\cdot\mid s)$ cannot increase the sup-norm difference. Applying the triangle inequality to the three parts of the unclipped backup, we obtain
\begin{align}
\bigl\lVert V_h^{\vartheta,\alpha}-V_h^{\vartheta^\prime,\alpha} \bigr\rVert_\infty &\leq\underbrace{\bigl\lVert R_h^{\boldsymbol{c}}-R_h^{\boldsymbol{c}^\prime} \bigr\rVert_\infty}_{\text{(I)}}+\underbrace{\bigl\lVert \boldsymbol{\phi}(\cdot,\cdot)^\top(\boldsymbol{v}-\boldsymbol{v}^\prime) \bigr\rVert_\infty}_{\text{(II)}}\\
&\quad+\underbrace{\alpha\sup_{(s,a)\in\mathcal{S}\times\mathcal{A}}\Bigl|\sqrt{\boldsymbol{\phi}(s,a)^\top\boldsymbol{\Gamma}^{-1}\boldsymbol{\phi}(s,a)}-\sqrt{\boldsymbol{\phi}(s,a)^\top{\boldsymbol{\Gamma}^\prime}^{-1}\boldsymbol{\phi}(s,a)}\Bigr|}_{\text{(III)}}. \label{eq:value_class_terms}
\end{align}
We bound these three terms in turn.

\paragraph{Bounding Term (I)}
At $h=1$, Term (I) is zero. For $h\geq2$, consider a specific queried coordinate $(s,a)$ and define $\boldsymbol{u}\coloneq\mu_h^{\boldsymbol{c}}$ and $\boldsymbol{u}'\coloneq\mu_h^{\boldsymbol{c}'}$. To apply smoothness without paying for the entire state-action space, form the vector $\widetilde{\boldsymbol{u}}$ that equals $\boldsymbol{u}'$ at $(s,a)$ and equals $\boldsymbol{u}$ elsewhere. It belongs to the utility domain because every coordinate lies in $[0,1]$. By coordinate separability (\cref{asm:coordinate_separability}), $[\nabla f_h(\widetilde{\boldsymbol{u}})](s,a)=[\nabla f_h(\boldsymbol{u}')](s,a)$, since $\widetilde{\boldsymbol{u}}(s,a)=\boldsymbol{u}'(s,a)$. Hence, by smoothness (\cref{asm:lipschitzness_and_smoothness}),
\begin{align}
\bigl|R_h^{\boldsymbol{c}}(s,a)-R_h^{\boldsymbol{c}^\prime}(s,a)\bigr|&=\bigl|[\nabla f_h(\boldsymbol{u})](s,a)-[\nabla f_h(\widetilde{\boldsymbol{u}})](s,a)\bigr|\\
&\leq L\|\boldsymbol{u}-\widetilde{\boldsymbol{u}}\|_1= L\bigl|\mu_h^{\boldsymbol{c}}(s,a)-\mu_h^{\boldsymbol{c}^\prime}(s,a)\bigr|.
\end{align}
Since clipping onto $[0,1]$ is nonexpansive and $\pi_{k,h}(a\mid s)\leq1$,
\begin{align}
\bigl|\mu_h^{\boldsymbol{c}}(s,a)-\mu_h^{\boldsymbol{c}^\prime}(s,a)\bigr|&\leq\pi_{k,h}(a\mid s)\bigl|(\boldsymbol{c}-\boldsymbol{c}^\prime)^\top\boldsymbol{\psi}(s)\bigr| \leq C_\psi\lVert \boldsymbol{c}-\boldsymbol{c}^\prime \rVert_2.
\end{align}
Here $\|\boldsymbol{\psi}(s)\|_2\leq C_\psi$ follows by applying condition (iii) of \cref{def:bilienar_mdp} to the indicator $g=\mathbf{1}\{\cdot=s\}$. Taking the supremum over $(s,a)$ gives
\begin{align}
\text{Term (I)}=\|R_h^{\boldsymbol{c}}-R_h^{\boldsymbol{c}'}\|_\infty&\leq LC_\psi\|\boldsymbol{c}-\boldsymbol{c}'\|_2. \label{eq:on_pseudo_reward_cover}
\end{align}

\paragraph{Bounding Term (II)}
By the Cauchy--Schwarz inequality and condition (ii) of \cref{def:bilienar_mdp},
\begin{align}
\text{Term (II)}&\leq C_\phi\sqrt{d_\phi}\lVert \boldsymbol{v}-\boldsymbol{v}^\prime \rVert_2. \label{eq:value_cover_term_II}
\end{align}

\paragraph{Bounding Term (III)}
Define $\boldsymbol{A}\coloneq\boldsymbol{\Gamma}^{-1}$ and $\boldsymbol{A}'\coloneq{\boldsymbol{\Gamma}'}^{-1}$. Since $|\sqrt{x}-\sqrt{y}|\leq\sqrt{|x-y|}$ for $x,y\geq0$, we have
\begin{align}
\bigl|\sqrt{\boldsymbol{\phi}^\top\boldsymbol{A}\boldsymbol{\phi}}-\sqrt{\boldsymbol{\phi}^\top\boldsymbol{A}'\boldsymbol{\phi}}\bigr|\leq\sqrt{|\boldsymbol{\phi}^\top(\boldsymbol{A}-\boldsymbol{A}')\boldsymbol{\phi}|}\leq\|\boldsymbol{\phi}\|_2\|\boldsymbol{A}-\boldsymbol{A}'\|_{\mathrm{op}}^{1/2}.
\end{align}
The identity $\boldsymbol{A}-\boldsymbol{A}'=\boldsymbol{\Gamma}^{-1}(\boldsymbol{\Gamma}'-\boldsymbol{\Gamma}){\boldsymbol{\Gamma}'}^{-1}$, together with $\|\boldsymbol{\Gamma}^{-1}\|_{\mathrm{op}},\|{\boldsymbol{\Gamma}'}^{-1}\|_{\mathrm{op}}\leq1$ (because $\boldsymbol{\Gamma},\boldsymbol{\Gamma}'\succeq I$), gives $\|\boldsymbol{A}-\boldsymbol{A}'\|_{\mathrm{op}}\leq\|\boldsymbol{\Gamma}-\boldsymbol{\Gamma}'\|_{\mathrm{op}}\leq\|\boldsymbol{\Gamma}-\boldsymbol{\Gamma}'\|_F$. With $\|\boldsymbol{\phi}\|_2\leq C_\phi\sqrt{d_\phi}$,
\begin{align}
\text{Term (III)}&\leq\alpha C_\phi\sqrt{d_\phi}\lVert \boldsymbol{\Gamma}-\boldsymbol{\Gamma}^\prime \rVert_F^{1/2}. \label{eq:value_cover_term_III}
\end{align}

\paragraph{Putting Everything Together}
Combining~\cref{eq:on_pseudo_reward_cover,eq:value_cover_term_II,eq:value_cover_term_III} with~\cref{eq:value_class_terms} gives
\begin{align}
\bigl\lVert V_h^{\vartheta,\alpha}-V_h^{\vartheta^\prime,\alpha} \bigr\rVert_\infty \leq LC_\psi\lVert \boldsymbol{c}-\boldsymbol{c}^\prime \rVert_2+C_\phi\sqrt{d_\phi}\lVert \boldsymbol{v}-\boldsymbol{v}^\prime \rVert_2+\alpha C_\phi\sqrt{d_\phi}\lVert \boldsymbol{\Gamma}-\boldsymbol{\Gamma}^\prime \rVert_F^{1/2}. \label{eq:value_cover_master}
\end{align}
If the coefficient of a parameter block is zero, that block contributes no covering error and can be omitted. For the remaining blocks, we assign an approximation-error budget of $\varepsilon/4$ to each term in~\cref{eq:value_cover_master}. Accordingly, we approximate $\boldsymbol{c}$ to Euclidean accuracy $\varepsilon/(4LC_\psi)$, $\boldsymbol{v}$ to Euclidean accuracy $\varepsilon/(4C_\phi\sqrt{d_\phi})$, and $\boldsymbol{\Gamma}$ to Frobenius accuracy $\varepsilon^2/(16\alpha^2C_\phi^2d_\phi)$.

Let $\mathcal{C}_{\boldsymbol{c}}$, $\mathcal{C}_{\boldsymbol{v}}$, and $\mathcal{C}_{\boldsymbol{\Gamma}}$ be internal covers of the three parameter sets at these accuracies. For any admissible $\vartheta=(\boldsymbol{c},\boldsymbol{v},\boldsymbol{\Gamma})$, select $\boldsymbol{c}^\circ\in\mathcal{C}_{\boldsymbol{c}}$, $\boldsymbol{v}^\circ\in\mathcal{C}_{\boldsymbol{v}}$, and $\boldsymbol{\Gamma}^\circ\in\mathcal{C}_{\boldsymbol{\Gamma}}$ within these accuracies, and set $\vartheta^\circ=(\boldsymbol{c}^\circ,\boldsymbol{v}^\circ,\boldsymbol{\Gamma}^\circ)$, which is again admissible. By~\cref{eq:value_cover_master},
\begin{align}
\bigl\lVert V_h^{\vartheta,\alpha}-V_h^{\vartheta^\circ,\alpha} \bigr\rVert_\infty\leq\tfrac34\varepsilon\leq\varepsilon. \label{eq:value_representative_error}
\end{align}

The parameter sets lie in a Euclidean ball of radius $C_\phi C_{\mathbf{K}}\sqrt{NHd_\phi}$ in $\mathbb{R}^{d_\psi}$, a Euclidean ball of radius $(H-h)G\sqrt N$ in $\mathbb{R}^{d_\phi}$, and a Frobenius ball of radius $\sqrt{d_\phi}(1+NC_\phi^2d_\phi)$ in the $d_\phi(d_\phi+1)/2$-dimensional space of symmetric matrices, respectively. The last radius holds because every eigenvalue of an admissible $\boldsymbol{\Gamma}$ is at most $1+NC_\phi^2d_\phi$. By \cref{lem:euclidean_cover}, the covers can be chosen with
\begin{align}
|\mathcal{C}_{\boldsymbol{c}}|&\leq\left(1+\frac{8LC_\psi C_\phi C_{\mathbf{K}}\sqrt{NHd_\phi}}{\varepsilon}\right)^{d_\psi}, \label{eq:value_occupancy_cover_size}\\
|\mathcal{C}_{\boldsymbol{v}}|&\leq\left(1+\frac{8C_\phi\sqrt{d_\phi}(H-h)G\sqrt N}{\varepsilon}\right)^{d_\phi}, \label{eq:value_coefficient_cover_size}\\
|\mathcal{C}_{\boldsymbol{\Gamma}}|&\leq\left(1+\frac{32\alpha^2C_\phi^2d_\phi^{3/2}(1+NC_\phi^2d_\phi)}{\varepsilon^2}\right)^{d_\phi(d_\phi+1)/2}. \label{eq:value_gram_cover_size}
\end{align}
These parameter representatives generate the finite collection $\{V_h^{\vartheta,\alpha}\mid\boldsymbol{c}\in\mathcal{C}_{\boldsymbol{c}},\ \boldsymbol{v}\in\mathcal{C}_{\boldsymbol{v}},\ \boldsymbol{\Gamma}\in\mathcal{C}_{\boldsymbol{\Gamma}}\}\subseteq\mathcal{V}_{k,h}(\alpha)$, which by~\cref{eq:value_representative_error} is an internal $\varepsilon$-cover. Its cardinality is at most $|\mathcal{C}_{\boldsymbol{c}}|\,|\mathcal{C}_{\boldsymbol{v}}|\,|\mathcal{C}_{\boldsymbol{\Gamma}}|$, since different tuples may generate the same function. Taking logarithms and substituting~\cref{eq:value_occupancy_cover_size,eq:value_coefficient_cover_size,eq:value_gram_cover_size} gives~\cref{eq:on_cover_explicit}. At $h=H$, the class contains neither $\boldsymbol{v}$ nor Term (III). Hence, it is smaller and satisfies the same bound. Finally, under the stated polynomial bounds, each logarithm in~\cref{eq:on_cover_explicit} is logarithmic in the listed quantities, and the coefficients sum to at most $2(d_\phi^2+d_\psi)$, which gives the order $\widetilde{\mathcal{O}}(d_\phi^2+d_\psi)$.
\end{proof}

\subsection{Theoretical Guarantees} \label{app:on_theoretical_guarantees}

\subsubsection{The Good Event} \label{app:on_good_events}

We use the good event $\mathcal{E}=\mathcal{E}_{\mathrm{P}}\cap\mathcal{E}_{\mathrm{R}}\cap\mathcal{E}_{\mathrm{V}}$ in~\cref{def:good_event} with the radii $\beta^{\mathrm{On}}$, $\Delta_{k,h}^{\mathrm{On}}$, and $\alpha_h^{\mathrm{On}}$. The radii $\beta^{\mathrm{On}}$ and $\alpha_h^{\mathrm{On}}$ are chosen in \cref{lem:good_estimation_event_P,lem:good_estimation_event_V}, respectively, and $\Delta_{k,h}^{\mathrm{On}}$ is defined in~\cref{lem:on_policy_reward_error}. \cref{lem:good_estimation_event_P} shows that $\mathcal{E}_{\mathrm{P}}$ holds with probability at least $1-\delta/4$, \cref{lem:on_policy_reward_error} shows that, under the event $\mathcal{E}_{\mathrm{P}}$, $\mathcal{E}_{\mathrm{R}}$ holds with probability at least $1-\delta/4$, and \cref{lem:good_estimation_event_V} shows that $\mathcal{E}_{\mathrm{V}}$ holds with probability at least $1-\delta/4$. These failure probabilities are combined with that of~\cref{lem:on_martingale} in the proof of \cref{thm:on_policy}.

\subsubsection{Transition Model Estimation} \label{app:on_tme_guarantees}

\begin{lemma}[Transition-Core Confidence] \label{lem:on_model_confidence_formal} \label{lem:good_estimation_event_P}
If we choose $\beta^{\mathrm{On}}=\mathcal{O}(d_\phi\log(NHK/\delta))$ with a sufficiently large constant, then $\mathcal{E}_{\mathrm{P}}$ holds with probability at least $1-\delta/4$.
\end{lemma}

\begin{proof}
We first recall the relevant notation. \BTE{} computes the ridge estimate $\mathbf{M}_k$ in~\cref{line:tme_model} of \cref{alg:tme_bilinear} from the $N$ fresh trajectories, with the Gram matrix $\boldsymbol{\Lambda}_k=I+\sum_{i=1}^N\sum_{h=1}^H\boldsymbol{\phi}_{i,h}\boldsymbol{\phi}_{i,h}^\top$. The event $\mathcal{E}_{\mathrm{P}}$ in~\cref{def:good_event} requires $\|\boldsymbol{\Lambda}_k^{1/2}(\mathbf{M}_k-\mathbf{M}^\star)\|_F\leq\sqrt{\beta^{\mathrm{On}}}$ for every $k\in[K]$.

We take, for a sufficiently large numerical constant $c$,
\begin{align}
\beta^{\mathrm{On}}&=c\bigl\{C_{\mathbf{M}}^2d_\phi+C_{\mathbf{K}}^2[d_\phi\log(1+NHC_\phi^2)+\log(4K/\delta)]\bigr\}. \label{eq:on_beta}
\end{align}
Each term in the braces is at most a constant multiple of $d_\phi\log(NHK/\delta)$. Hence, this choice has the stated order. Consider a specific iteration $k$ and condition on the history before its batch is collected. We first identify the centered regression noise, then bound its self-normalized sum and the ridge bias.

\paragraph{Conditional Regression Model}
For an observed transition, define $\boldsymbol{y}_{i,h}=\mathbf{K}_\psi^{-1}\boldsymbol{\psi}(s_{i,h+1})$, the transpose of the regression target in \cref{lem:bilinear_regression_response}. Conditional on the history through $(s_{i,h},a_{i,h})$, the successor $s_{i,h+1}$ has law $P^\star(\cdot\mid s_{i,h},a_{i,h})$ by the Markov property. Transposing the identity of \cref{lem:bilinear_regression_response}, and using that $\mathbf{K}_\psi$ is symmetric, gives
\begin{align}
\mathbb{E}[\boldsymbol{y}_{i,h}\mid s_{i,h},a_{i,h}]&=(\mathbf{M}^\star)^\top\boldsymbol{\phi}_{i,h}.
\end{align}
Thus $\boldsymbol{\eta}_{i,h}\coloneq\boldsymbol{y}_{i,h}-(\mathbf{M}^\star)^\top\boldsymbol{\phi}_{i,h}$ is conditionally centered. Condition (iii) of \cref{def:bilienar_mdp} gives $\|\boldsymbol{y}_{i,h}\|_2=\|\boldsymbol{\psi}(s_{i,h+1})^\top\mathbf{K}_\psi^{-1}\|_2\leq C_{\mathbf{K}}$. Jensen's inequality bounds the norm of its conditional expectation $(\mathbf{M}^\star)^\top\boldsymbol{\phi}_{i,h}$ by the same constant. By the triangle inequality, $\|\boldsymbol{\eta}_{i,h}\|_2\leq2C_{\mathbf{K}}$.

Order observations first by trajectory and then by step, and reveal each current state and action before its successor. In this order, each covariate is known before its noise is revealed. Transitions within a trajectory need not be independent: the conditional centering just proved is the property needed by the matrix martingale bound.

\paragraph{Noise and Regularization Error}
Recall the normal equations $\boldsymbol{\Lambda}_k\mathbf{M}_k=\sum_{i=1}^N\sum_{h=1}^H\boldsymbol{\phi}_{i,h}\boldsymbol{y}_{i,h}^\top$ derived in \cref{app:on_tme_details}. Substituting $\boldsymbol{y}_{i,h}=(\mathbf{M}^\star)^\top\boldsymbol{\phi}_{i,h}+\boldsymbol{\eta}_{i,h}$ and using $\sum_{i=1}^N\sum_{h=1}^H\boldsymbol{\phi}_{i,h}\boldsymbol{\phi}_{i,h}^\top=\boldsymbol{\Lambda}_k-I$, we obtain
\begin{align}
\boldsymbol{\Lambda}_k\mathbf{M}_k&=\sum_{i=1}^N\sum_{h=1}^H\boldsymbol{\phi}_{i,h}\boldsymbol{\phi}_{i,h}^\top\mathbf{M}^\star+\sum_{i=1}^N\sum_{h=1}^H\boldsymbol{\phi}_{i,h}\boldsymbol{\eta}_{i,h}^\top=(\boldsymbol{\Lambda}_k-I)\mathbf{M}^\star+\sum_{i=1}^N\sum_{h=1}^H\boldsymbol{\phi}_{i,h}\boldsymbol{\eta}_{i,h}^\top.
\end{align}
Multiplying both sides by $\boldsymbol{\Lambda}_k^{-1}$ and subtracting $\mathbf{M}^\star$ gives
\begin{align}
\mathbf{M}_k-\mathbf{M}^\star&=\boldsymbol{\Lambda}_k^{-1}\biggl(\sum_{i=1}^N\sum_{h=1}^H\boldsymbol{\phi}_{i,h}\boldsymbol{\eta}_{i,h}^\top-\mathbf{M}^\star\biggr).
\end{align}
The $-\mathbf{M}^\star$ term comes from the identity regularizer. Multiplying by $\boldsymbol{\Lambda}_k^{1/2}$ and applying the triangle inequality for the Frobenius norm gives
\begin{align}
\|\boldsymbol{\Lambda}_k^{1/2}(\mathbf{M}_k-\mathbf{M}^\star)\|_F&\leq\biggl\|\boldsymbol{\Lambda}_k^{-1/2}\sum_{i=1}^N\sum_{h=1}^H\boldsymbol{\phi}_{i,h}\boldsymbol{\eta}_{i,h}^\top\biggr\|_F+\|\boldsymbol{\Lambda}_k^{-1/2}\mathbf{M}^\star\|_F.
\end{align}
Since $\boldsymbol{\Lambda}_k\succeq I$, its inverse square root has operator norm at most one. Hence, condition (i) of \cref{def:bilienar_mdp} bounds the last term by $\|\mathbf{M}^\star\|_F\leq C_{\mathbf{M}}\sqrt{d_\phi}$. Apply \cref{lem:lin_matrix_self_normalized} to the first term, with $n=NH$ observations, covariate radius $R_x=C_\phi\sqrt{d_\phi}$ (condition (ii) of \cref{def:bilienar_mdp}), noise radius $R=2C_{\mathbf{K}}$, and failure probability $\zeta=\delta/(4K)$. Since $nR_x^2/d_\phi=NHC_\phi^2$, we obtain
\begin{align}
\|\boldsymbol{\Lambda}_k^{1/2}(\mathbf{M}_k-\mathbf{M}^\star)\|_F&\leq2C_{\mathbf{K}}\sqrt{d_\phi\log(1+NHC_\phi^2)+2\log(4K/\delta)}+C_{\mathbf{M}}\sqrt{d_\phi}.
\end{align}
Squaring and using $(a+b)^2\leq2a^2+2b^2$ shows that~\cref{eq:on_beta} dominates the square of the right-hand side for a sufficiently large numerical $c$.

The probability bound holds conditional on every pre-batch history. Taking expectations over that history leaves the same failure probability $\delta/(4K)$ for iteration $k$. Summing over $k\in[K]$ gives total failure probability at most $\delta/4$. Hence, $\mathcal{E}_{\mathrm{P}}$ holds with probability at least $1-\delta/4$. The matrix inequality controls $\boldsymbol{y}_{i,h}$ as a whole. Hence, no union bound over its $d_\psi$ coordinates is needed.
\end{proof}

\subsubsection{Pseudo-Reward Estimation} \label{app:on_pre_guarantees}

\begin{lemma}[Pseudo-Reward Error] \label{lem:on_policy_reward_error}
Let $\widehat{e}_{k,h}\coloneq N^{-1}\sum_{i=1}^N\|\boldsymbol{\phi}_{i,h}\|_{\boldsymbol{\Lambda}_k^{-1}}$ and let $\beta^{\mathrm{On}}$ be as in \cref{lem:good_estimation_event_P}. Under the event $\mathcal{E}_{\mathrm{P}}$, with probability at least $1-\delta/4$, for every $k\in[K]$ and $h\in[H]$,
\begin{align}
\MoveEqLeft\|\widehat{R}_{k,h}-R_{k,h}\|_\infty\leq\Delta_{k,h}^{\mathrm{On}}\nonumber\\
&\coloneq\begin{cases}0,&h=1,\\[2pt] LC_\psi\left(\sqrt{\beta^{\mathrm{On}}}\,\widehat{e}_{k,h-1}+C_{\mathbf{M}}C_\phi d_\phi\sqrt{\dfrac{8\log(8KH/\delta)}{N}}\right),&h\in\{2,\ldots,H\}.\end{cases} \label{eq:Delta_on}
\end{align}
\end{lemma}

\begin{proof}
We first recall the relevant notation. For $h\geq2$, \PRE{} estimates the occupancy measure by $\widehat{\mu}_{k,h}(s,a)=\operatorname{Clip}(\pi_{k,h}(a\mid s)\widehat{\boldsymbol{\omega}}_{k,h-1}^\top\mathbf{M}_k\boldsymbol{\psi}(s),[0,1])$ in~\cref{eq:bilinear_occupancy_reconstruction}, where $\widehat{\boldsymbol{\omega}}_{k,h-1}$ is the empirical feature expectation over the $N$ fresh trajectories, and sets $\widehat{R}_{k,h}=\nabla f_h(\widehat{\mu}_{k,h})$. By~\cref{eq:bilinear_occupancy_decomposition}, the true occupancy measure satisfies $\mu_h^{\pi_k}(s,a)=\pi_{k,h}(a\mid s)(\boldsymbol{\omega}_{h-1}^{\pi_k})^\top\mathbf{M}^\star\boldsymbol{\psi}(s)$, and $R_{k,h}=\nabla f_h(\mu_h^{\pi_k})$.

We first reduce the pseudo-reward error to an occupancy error. Both $\widehat{\mu}_{k,h}$ and $\mu_h^{\pi_k}$ lie in the utility domain $[0,1]^{\mathcal{S}\times\mathcal{A}}$. Hence, $L$-smoothness of $f_h$ under the $\ell_1/\ell_\infty$ norm pair (\cref{asm:lipschitzness_and_smoothness}) gives
\begin{align}
\bigl\lVert \widehat{R}_{k,h}-R_{k,h} \bigr\rVert_\infty=\bigl\lVert \nabla f_h(\widehat{\mu}_{k,h})-\nabla f_h(\mu_h^{\pi_k}) \bigr\rVert_\infty\leq L\bigl\lVert \widehat{\mu}_{k,h}-\mu_h^{\pi_k} \bigr\rVert_1.
\end{align}
It therefore suffices to bound the occupancy error $\|\widehat{\mu}_{k,h}-\mu_h^{\pi_k}\|_1$.

When $h=1$,~\cref{alg:pre} sets $\widehat{\mu}_{k,1}(s,a)=\rho(s)\pi_{k,1}(a\mid s)=\mu_1^{\pi_k}(s,a)$ for every $(s,a)\in\mathcal{S}\times\mathcal{A}$. Hence, the occupancy estimation error is zero, matching $\Delta_{k,1}^{\mathrm{On}}=0$.

\paragraph{Splitting the Occupancy Error}
Now consider a specific $h\in\{2,\ldots,H\}$, and let $\widetilde{\mu}_{k,h}(s,a)\coloneq\pi_{k,h}(a\mid s)\widehat{\boldsymbol{\omega}}_{k,h-1}^\top\mathbf{M}_k\boldsymbol{\psi}(s)$ denote the unclipped estimate in~\cref{alg:pre}. Since $\mu_h^{\pi_k}(s,a)\in[0,1]$ and $|\operatorname{Clip}(x,[0,1])-y|\leq|x-y|$ for every $x\in\mathbb{R}$ and $y\in[0,1]$, we have $|\widehat{\mu}_{k,h}(s,a)-\mu_h^{\pi_k}(s,a)|\leq|\widetilde{\mu}_{k,h}(s,a)-\mu_h^{\pi_k}(s,a)|$ for all $(s,a)$. Consequently, clipping cannot increase the error, and it suffices to analyze $\widetilde{\mu}_{k,h}$. By \cref{lem:occupancy_measure}, $\mu_h^{\pi_k}(s,a)=\pi_{k,h}(a\mid s)(\boldsymbol{\omega}_{h-1}^{\pi_k})^\top\mathbf{M}^\star\boldsymbol{\psi}(s)$. Therefore
\begin{align}
\bigl\lVert \widehat{\mu}_{k,h}-\mu_h^{\pi_k} \bigr\rVert_1 &\leq\sum_{s\in\mathcal{S}}\sum_{a\in\mathcal{A}}\bigl|\pi_{k,h}(a\mid s)\widehat{\boldsymbol{\omega}}_{k,h-1}^\top\mathbf{M}_k\boldsymbol{\psi}(s)-\pi_{k,h}(a\mid s)(\boldsymbol{\omega}_{h-1}^{\pi_k})^\top\mathbf{M}^\star\boldsymbol{\psi}(s)\bigr|\\
&=\sum_{s\in\mathcal{S}}\bigl|\widehat{\boldsymbol{\omega}}_{k,h-1}^\top\mathbf{M}_k\boldsymbol{\psi}(s)-(\boldsymbol{\omega}_{h-1}^{\pi_k})^\top\mathbf{M}^\star\boldsymbol{\psi}(s)\bigr|\\
&\leq\underbrace{\sum_{s\in\mathcal{S}}\bigl|\widehat{\boldsymbol{\omega}}_{k,h-1}^\top(\mathbf{M}_k-\mathbf{M}^\star)\boldsymbol{\psi}(s)\bigr|}_{\text{(I)}}+\underbrace{\sum_{s\in\mathcal{S}}\bigl|(\widehat{\boldsymbol{\omega}}_{k,h-1}-\boldsymbol{\omega}_{h-1}^{\pi_k})^\top\mathbf{M}^\star\boldsymbol{\psi}(s)\bigr|}_{\text{(II)}}.
\end{align}
The equality uses $\sum_a\pi_{k,h}(a\mid s)=1$, and the last inequality adds and subtracts $\widehat{\boldsymbol{\omega}}_{k,h-1}^\top\mathbf{M}^\star\boldsymbol{\psi}(s)$.

\paragraph{Bounding Term (I)}
Define $g(s)\coloneq\operatorname{sign}(\widehat{\boldsymbol{\omega}}_{k,h-1}^\top(\mathbf{M}_k-\mathbf{M}^\star)\boldsymbol{\psi}(s))$, so that $|g|\leq1$ and each absolute value in Term (I) equals the signed term multiplied by $g(s)$. Then,
\begin{align}
\text{Term (I)} &=\widehat{\boldsymbol{\omega}}_{k,h-1}^\top(\mathbf{M}_k-\mathbf{M}^\star)\sum_{s\in\mathcal{S}}\boldsymbol{\psi}(s)g(s)\\
&\leq\bigl\lVert \widehat{\boldsymbol{\omega}}_{k,h-1} \bigr\rVert_{\boldsymbol{\Lambda}_k^{-1}}\bigl\lVert \boldsymbol{\Lambda}_k^{1/2}(\mathbf{M}_k-\mathbf{M}^\star) \bigr\rVert_F\biggl\lVert \sum_{s\in\mathcal{S}}\boldsymbol{\psi}(s)g(s) \biggr\rVert_2\\
&\leq C_\psi\sqrt{\beta^{\mathrm{On}}}\bigl\lVert \widehat{\boldsymbol{\omega}}_{k,h-1} \bigr\rVert_{\boldsymbol{\Lambda}_k^{-1}}\\
&\leq C_\psi\sqrt{\beta^{\mathrm{On}}}\biggl(\frac1N\sum_{i=1}^N\lVert \boldsymbol{\phi}_{i,h-1} \rVert_{\boldsymbol{\Lambda}_k^{-1}}\biggr)=C_\psi\sqrt{\beta^{\mathrm{On}}}\,\widehat{e}_{k,h-1}.
\end{align}
The first inequality is the Cauchy--Schwarz inequality. The second inequality uses $\mathcal{E}_{\mathrm{P}}$ and condition (iii) of \cref{def:bilienar_mdp} with $\|g\|_\infty\leq1$. The third inequality substitutes $\widehat{\boldsymbol{\omega}}_{k,h-1}=N^{-1}\sum_i\boldsymbol{\phi}_{i,h-1}$ and applies the triangle inequality for the norm induced by $\boldsymbol{\Lambda}_k^{-1}$.

\paragraph{Bounding Term (II)}
Similarly, define $g^\prime(s)\coloneq\operatorname{sign}((\widehat{\boldsymbol{\omega}}_{k,h-1}-\boldsymbol{\omega}_{h-1}^{\pi_k})^\top\mathbf{M}^\star\boldsymbol{\psi}(s))$. Then,
\begin{align}
\text{Term (II)} &=(\widehat{\boldsymbol{\omega}}_{k,h-1}-\boldsymbol{\omega}_{h-1}^{\pi_k})^\top\mathbf{M}^\star\sum_{s\in\mathcal{S}}\boldsymbol{\psi}(s)g^\prime(s)\leq C_{\mathbf{M}}C_\psi\sqrt{d_\phi}\,\bigl\lVert \widehat{\boldsymbol{\omega}}_{k,h-1}-\boldsymbol{\omega}_{h-1}^{\pi_k} \bigr\rVert_2,
\end{align}
where the inequality applies the Cauchy--Schwarz inequality, $\|\mathbf{M}^\star\|_{\mathrm{op}}\leq\|\mathbf{M}^\star\|_F\leq C_{\mathbf{M}}\sqrt{d_\phi}$ (condition (i) of \cref{def:bilienar_mdp}), and condition (iii) with $\|g'\|_\infty\leq1$.

It remains to control the feature-average error. Condition on the history before batch $k$ and put $\boldsymbol{X}_i\coloneq\boldsymbol{\phi}_{i,h-1}-\boldsymbol{\omega}_{h-1}^{\pi_k}$ for $i\in[N]$. Since each step-$(h-1)$ pair has distribution $\mu_{h-1}^{\pi_k}$ and the $N$ trajectories are independent under this conditioning, the $\boldsymbol{X}_i$ are independent and centered. Hence, they form a vector martingale difference sequence. Their norms are at most $2C_\phi\sqrt{d_\phi}$, since both the observed feature and its mean have norm at most $C_\phi\sqrt{d_\phi}$. Applying \cref{lem:vector_azuma_hoeffding} with $R=2C_\phi\sqrt{d_\phi}$ and setting its right-hand side to $\delta/(4KH)$ shows that $\|\widehat{\boldsymbol{\omega}}_{k,h-1}-\boldsymbol{\omega}_{h-1}^{\pi_k}\|_2=\|N^{-1}\sum_i\boldsymbol{X}_i\|_2$ exceeds $C_\phi\sqrt{8d_\phi\log(8KH/\delta)/N}$ with conditional probability at most $\delta/(4KH)$. Averaging over the conditioning history and summing over the at most $KH$ pairs $(k,h)$ shows that, with probability at least $1-\delta/4$, for every $k\in[K]$ and $h\in\{2,\ldots,H\}$,
\begin{align}
\bigl\lVert \widehat{\boldsymbol{\omega}}_{k,h-1}-\boldsymbol{\omega}_{h-1}^{\pi_k} \bigr\rVert_2\leq C_\phi\sqrt{\frac{8d_\phi\log(8KH/\delta)}{N}}. \label{eq:on_feature_average_bound}
\end{align}

\paragraph{Putting Everything Together}
On the event $\mathcal{E}_{\mathrm{P}}$, whenever~\cref{eq:on_feature_average_bound} holds, combining the reduction to the occupancy error with the bounds on Terms (I) and (II) yields, for every $k$ and $h\geq2$,
\begin{align}
\bigl\lVert \widehat{R}_{k,h}-R_{k,h} \bigr\rVert_\infty&\leq L(\text{Term (I)}+\text{Term (II)})\\
&\leq LC_\psi\left(\sqrt{\beta^{\mathrm{On}}}\,\widehat{e}_{k,h-1}+C_{\mathbf{M}}C_\phi d_\phi\sqrt{\frac{8\log(8KH/\delta)}{N}}\right)=\Delta_{k,h}^{\mathrm{On}}.
\end{align}
Hence, under the event $\mathcal{E}_{\mathrm{P}}$, $\mathcal{E}_{\mathrm{R}}$ holds whenever~\cref{eq:on_feature_average_bound} holds, which happens with probability at least $1-\delta/4$. This proves the lemma.
\end{proof}

\subsubsection{Optimistic Policy Evaluation} \label{app:on_ope_guarantees}

Although $\mathcal{D}_k$ is a fresh on-policy batch, freshness alone does not make the value-regression target independent of the samples used to fit it. Conditional on the history before batch $k$, the policy $\pi_k$ is fixed, but the backward recursion in~\cref{alg:ope} constructs $\widehat{V}_{k,h+1}$ only after collecting $\mathcal{D}_k$ from the environment. In particular, $\widehat{V}_{k,h+1}$ depends on the same next states that appear in the step-$h$ regression. Hence, a concentration inequality proved for one fixed deterministic value function cannot be applied directly after substituting this data-dependent choice. We therefore follow the value-aware concentration argument of~\citet{jin2023provably}: first cover the full class of possible values, then prove a regression bound simultaneously over that class.

The main result of this subsubsection is the Bellman-error bound in \cref{lem:on_bellman_error}, which the sample-complexity proof uses directly. We prove it for general radii $\Delta_{k,h}$ and $\alpha_{k,h}$. Hence, it also applies to $\OCUPOOFF$ in~\cref{app:ucgu_off} and to linear MDPs in~\cref{app:linear_off_proofs}. We then establish the value part of the good event in \cref{lem:good_estimation_event_V}, which chooses the \OPE{} radii, using the cover of the value functions in \cref{app:on_covering_arguments}.

\begin{proof}[Proof of \cref{lem:on_bellman_error}]
Consider specific $k\in[K]$, $h\in[H]$, and $(s,a)\in\mathcal{S}\times\mathcal{A}$, and first let $h<H$. We recall the relevant notation. \cref{line:optimistic_Q,line:clipped_Q} of \cref{alg:ope} compute the unclipped and clipped action values
\begin{align}
\widetilde{Q}_{k,h}(s,a)&=\widehat{R}_{k,h}(s,a)+\boldsymbol{\phi}(s,a)^\top\widehat{\boldsymbol{v}}_{k,h}+\alpha_{k,h}\|\boldsymbol{\phi}(s,a)\|_{\boldsymbol{\Gamma}_{k,h}^{-1}},\\
\widehat{Q}_{k,h}(s,a)&=\operatorname{Clip}\bigl(\widetilde{Q}_{k,h}(s,a),[-(H-h+1)G,(H-h+1)G]\bigr).
\end{align}
Let $\boldsymbol{v}_{k,h}^\star\coloneq\boldsymbol{v}^\star(\widehat{V}_{k,h+1})$, with $\boldsymbol{v}^\star$ as in \cref{lem:bilinear_next_state_values}, which gives $P^\star\widehat{V}_{k,h+1}(s,a)=\boldsymbol{\phi}(s,a)^\top\boldsymbol{v}_{k,h}^\star$. In a linear MDP, the same identity holds with $\boldsymbol{v}^\star$ as in \cref{lem:linear_value_response}. On the events $\mathcal{E}_{\mathrm{R}}$ and $\mathcal{E}_{\mathrm{V}}$ in~\cref{def:good_event}, $\|\widehat{R}_{k,h}-R_{k,h}\|_\infty\leq\Delta_{k,h}$ and $\|\widehat{\boldsymbol{v}}_{k,h}-\boldsymbol{v}_{k,h}^\star\|_{\boldsymbol{\Gamma}_{k,h}}\leq\alpha_{k,h}$.

Under the good event $\mathcal{E}$, we obtain that
\begin{align}
\xi_{k,h}(s,a)&=R_{k,h}(s,a)+P^\star\widehat{V}_{k,h+1}(s,a)-\widehat{Q}_{k,h}(s,a)\\
&\leq\max\bigl\{R_{k,h}(s,a)+P^\star\widehat{V}_{k,h+1}(s,a)-\widetilde{Q}_{k,h}(s,a),0\bigr\}\\
&=\max\bigl\{(R_{k,h}-\widehat{R}_{k,h})(s,a)-\boldsymbol{\phi}(s,a)^\top(\widehat{\boldsymbol{v}}_{k,h}-\boldsymbol{v}_{k,h}^\star)-\alpha_{k,h}\|\boldsymbol{\phi}(s,a)\|_{\boldsymbol{\Gamma}_{k,h}^{-1}},0\bigr\}\\
&\leq\max\bigl\{\Delta_{k,h}+\alpha_{k,h}\|\boldsymbol{\phi}(s,a)\|_{\boldsymbol{\Gamma}_{k,h}^{-1}}-\alpha_{k,h}\|\boldsymbol{\phi}(s,a)\|_{\boldsymbol{\Gamma}_{k,h}^{-1}},0\bigr\}=\Delta_{k,h},
\end{align}
and, symmetrically,
\begin{align}
-\xi_{k,h}(s,a)&=\widehat{Q}_{k,h}(s,a)-R_{k,h}(s,a)-P^\star\widehat{V}_{k,h+1}(s,a)\\
&\leq\max\bigl\{\widetilde{Q}_{k,h}(s,a)-R_{k,h}(s,a)-P^\star\widehat{V}_{k,h+1}(s,a),0\bigr\}\\
&=\max\bigl\{(\widehat{R}_{k,h}-R_{k,h})(s,a)+\boldsymbol{\phi}(s,a)^\top(\widehat{\boldsymbol{v}}_{k,h}-\boldsymbol{v}_{k,h}^\star)+\alpha_{k,h}\|\boldsymbol{\phi}(s,a)\|_{\boldsymbol{\Gamma}_{k,h}^{-1}},0\bigr\}\\
&\leq\Delta_{k,h}+2\alpha_{k,h}\|\boldsymbol{\phi}(s,a)\|_{\boldsymbol{\Gamma}_{k,h}^{-1}}.
\end{align}
In both derivations, the first inequality follows from the non-expansiveness of clipping. Since $\|R_{k,h}\|_\infty\leq G$ and $\|\widehat{V}_{k,h+1}\|_\infty\leq(H-h)G$, the true backup $R_{k,h}(s,a)+P^\star\widehat{V}_{k,h+1}(s,a)$ lies in the clipping interval and is left unchanged by clipping. Since clipping onto an interval is nondecreasing and non-expansive, the difference between $\widehat{Q}_{k,h}(s,a)$ and the true backup lies between zero and the corresponding difference for $\widetilde{Q}_{k,h}(s,a)$. The equalities substitute $\widetilde{Q}_{k,h}(s,a)$ and $P^\star\widehat{V}_{k,h+1}(s,a)=\boldsymbol{\phi}(s,a)^\top\boldsymbol{v}_{k,h}^\star$. The last inequalities bound $|(R_{k,h}-\widehat{R}_{k,h})(s,a)|$ by $\Delta_{k,h}$ on $\mathcal{E}_{\mathrm{R}}$, and bound $|\boldsymbol{\phi}(s,a)^\top(\widehat{\boldsymbol{v}}_{k,h}-\boldsymbol{v}_{k,h}^\star)|$ by the Cauchy--Schwarz inequality and $\mathcal{E}_{\mathrm{V}}$ as
\begin{align}
|\boldsymbol{\phi}(s,a)^\top(\widehat{\boldsymbol{v}}_{k,h}-\boldsymbol{v}_{k,h}^\star)|\leq\|\boldsymbol{\phi}(s,a)\|_{\boldsymbol{\Gamma}_{k,h}^{-1}}\|\widehat{\boldsymbol{v}}_{k,h}-\boldsymbol{v}_{k,h}^\star\|_{\boldsymbol{\Gamma}_{k,h}}\leq\alpha_{k,h}\|\boldsymbol{\phi}(s,a)\|_{\boldsymbol{\Gamma}_{k,h}^{-1}}.
\end{align}
In the first derivation, the bonus $\alpha_{k,h}\|\boldsymbol{\phi}(s,a)\|_{\boldsymbol{\Gamma}_{k,h}^{-1}}$ offsets this bound, which leaves only $\Delta_{k,h}$. In the second, the bonus adds to this bound, which gives the factor two.

At $h=H$, we have $\widehat{V}_{k,H+1}\equiv0$ and $\alpha_{k,H}=0$. Hence, the same derivations apply with $\widetilde{Q}_{k,H}=\widehat{R}_{k,H}$, without the terms involving $\widehat{\boldsymbol{v}}_{k,H}$, and with the clipping interval $[-G,G]$.
\end{proof}

The first inequality of \cref{lem:on_bellman_error} is uniform over state-action pairs. Hence, it bounds Term (II) by $\sum_{k,h}\Delta_{k,h}^{\mathrm{On}}$ without requiring the fresh batches to cover $\pi^\star$. The bound on Term (III) contains the same $\Delta_{k,h}^{\mathrm{On}}$ together with the bonus $2\alpha_h^{\mathrm{On}}\|\boldsymbol{\phi}(s,a)\|_{\boldsymbol{\Gamma}_{k,h}^{-1}}$. Hence, the bounds on the two terms together contain at most $2\sum_{k,h}\Delta_{k,h}^{\mathrm{On}}$. Adding $\Delta_{k,h}^{\mathrm{On}}$ to the backup as an additional bonus would make Term (II) nonpositive but would double the contribution of $\Delta_{k,h}^{\mathrm{On}}$ to Term (III), leaving this total unchanged.

The following lemma establishes the value part of the good event.

\begin{lemma}[Value-Regression Confidence] \label{lem:on_value_regression_formal} \label{lem:good_estimation_event_V}
With the radii $\alpha_h^{\mathrm{On}}$ specified in the proof, which satisfy
\begin{align}
\alpha_h^{\mathrm{On}}=\begin{cases}\widetilde{\mathcal{O}}\Bigl((H-h)G\sqrt{d_\phi^2+d_\psi}\Bigr),&h<H,\\ 0,&h=H,\end{cases}
\end{align}
$\mathcal{E}_{\mathrm{V}}$ holds with probability at least $1-\delta/4$.
\end{lemma}

\begin{proof}[Proof of \cref{lem:on_value_regression_formal}]
We first recall the relevant notation. \cref{line:ope_gram_matrix,line:vtr_estimator} of \cref{alg:ope} compute $\boldsymbol{\Gamma}_{k,h}=I+\sum_{i=1}^N\boldsymbol{\phi}_{i,h}\boldsymbol{\phi}_{i,h}^\top$ and $\widehat{\boldsymbol{v}}_{k,h}=\boldsymbol{\Gamma}_{k,h}^{-1}\sum_{i=1}^N\boldsymbol{\phi}_{i,h}\widehat{V}_{k,h+1}(s_{i,h+1})$ from the $N$ fresh trajectories. By~\cref{lem:bilinear_next_state_values}, $P^\star V=\boldsymbol{\phi}^\top\boldsymbol{v}^\star(V)$ for every bounded $V\colon\mathcal{S}\to\mathbb{R}$, and the event $\mathcal{E}_{\mathrm{V}}$ in~\cref{def:good_event} requires $\|\widehat{\boldsymbol{v}}_{k,h}-\boldsymbol{v}^\star(\widehat{V}_{k,h+1})\|_{\boldsymbol{\Gamma}_{k,h}}\leq\alpha_h^{\mathrm{On}}$ for every $k\in[K]$ and $h<H$. The radii $\alpha_h^{\mathrm{On}}$ are chosen at the end of the proof.

Consider specific $k$ and $h<H$, and condition on the history before batch $k$. Define $\boldsymbol{v}_{k,h}^\star\coloneq\boldsymbol{v}^\star(\widehat{V}_{k,h+1})$. For a bounded function $V\colon \mathcal{S}\to\mathbb{R}$, define the regression residual $\epsilon_i(V)\coloneq V(s_{i,h+1})-P^\star V(s_{i,h},a_{i,h})$ for $i\in[N]$. The proof has four steps: an error decomposition, concentration on a fixed cover, extension to the fitted target, and the choice of $\alpha_h^{\mathrm{On}}$.

\paragraph{Error Decomposition}
By \cref{lem:bilinear_next_state_values}, $P^\star\widehat{V}_{k,h+1}(s_{i,h},a_{i,h})=\boldsymbol{\phi}_{i,h}^\top\boldsymbol{v}_{k,h}^\star$. Hence, the definition of the residual gives $\widehat{V}_{k,h+1}(s_{i,h+1})=\boldsymbol{\phi}_{i,h}^\top\boldsymbol{v}_{k,h}^\star+\epsilon_i(\widehat{V}_{k,h+1})$. Substituting this into the closed form of $\widehat{\boldsymbol{v}}_{k,h}$ in \cref{app:on_ope_details} and using $\sum_{i=1}^N\boldsymbol{\phi}_{i,h}\boldsymbol{\phi}_{i,h}^\top=\boldsymbol{\Gamma}_{k,h}-I$, we obtain
\begin{align}
\widehat{\boldsymbol{v}}_{k,h} &=\boldsymbol{\Gamma}_{k,h}^{-1}\sum_{i=1}^N\boldsymbol{\phi}_{i,h}\widehat{V}_{k,h+1}(s_{i,h+1})\\
&=\boldsymbol{\Gamma}_{k,h}^{-1}\biggl((\boldsymbol{\Gamma}_{k,h}-I)\boldsymbol{v}_{k,h}^\star+\sum_{i=1}^N\boldsymbol{\phi}_{i,h}\epsilon_i(\widehat{V}_{k,h+1})\biggr)\\
&=\boldsymbol{v}_{k,h}^\star+\boldsymbol{\Gamma}_{k,h}^{-1}\biggl(\sum_{i=1}^N\boldsymbol{\phi}_{i,h}\epsilon_i(\widehat{V}_{k,h+1})-\boldsymbol{v}_{k,h}^\star\biggr).
\end{align}
Consequently,
\begin{align}
\bigl\lVert \widehat{\boldsymbol{v}}_{k,h}-\boldsymbol{v}_{k,h}^\star \bigr\rVert_{\boldsymbol{\Gamma}_{k,h}} &=\Biggl\lVert \boldsymbol{\Gamma}_{k,h}^{-1}\biggl(\sum_{i=1}^N\boldsymbol{\phi}_{i,h}\epsilon_i(\widehat{V}_{k,h+1})-\boldsymbol{v}_{k,h}^\star\biggr) \Biggr\rVert_{\boldsymbol{\Gamma}_{k,h}}\\
&=\biggl\lVert \sum_{i=1}^N\boldsymbol{\phi}_{i,h}\epsilon_i(\widehat{V}_{k,h+1})-\boldsymbol{v}_{k,h}^\star \biggr\rVert_{\boldsymbol{\Gamma}_{k,h}^{-1}}\\
&\leq\biggl\lVert \sum_{i=1}^N\boldsymbol{\phi}_{i,h}\epsilon_i(\widehat{V}_{k,h+1}) \biggr\rVert_{\boldsymbol{\Gamma}_{k,h}^{-1}}+\bigl\lVert \boldsymbol{v}_{k,h}^\star \bigr\rVert_{\boldsymbol{\Gamma}_{k,h}^{-1}}\\
&\leq\biggl\lVert \sum_{i=1}^N\boldsymbol{\phi}_{i,h}\epsilon_i(\widehat{V}_{k,h+1}) \biggr\rVert_{\boldsymbol{\Gamma}_{k,h}^{-1}}+C_{\mathbf{M}}C_\psi\sqrt{d_\phi}\,(H-h)G. \label{eq:value_regression_error_decomposition}
\end{align}
For the last inequality, $\boldsymbol{\Gamma}_{k,h}\succeq I$ gives $\|\boldsymbol{v}_{k,h}^\star\|_{\boldsymbol{\Gamma}_{k,h}^{-1}}\leq\|\boldsymbol{v}_{k,h}^\star\|_2$, and conditions (i) and (iii) of \cref{def:bilienar_mdp} together with $\|\widehat{V}_{k,h+1}\|_\infty\leq (H-h)G$ give
\begin{align}
\bigl\lVert \boldsymbol{v}_{k,h}^\star \bigr\rVert_2 =\biggl\lVert \mathbf{M}^\star\sum_{s^\prime\in\mathcal{S}}\boldsymbol{\psi}(s^\prime)\widehat{V}_{k,h+1}(s^\prime) \biggr\rVert_2\leq\lVert \mathbf{M}^\star \rVert_F\biggl\lVert \sum_{s^\prime\in\mathcal{S}}\boldsymbol{\psi}(s^\prime)\widehat{V}_{k,h+1}(s^\prime) \biggr\rVert_2\leq C_{\mathbf{M}}\sqrt{d_\phi}\,C_\psi (H-h)G.
\end{align}
Thus, in~\cref{eq:value_regression_error_decomposition}, the first term comes from the noise $\epsilon_i(\widehat{V}_{k,h+1})$ and the second from the regularization.

\paragraph{Concentration on a Fixed Cover}
Under the conditioning, $\pi_k$ is fixed. Hence, the class defined in \cref{app:on_covering_arguments}, $\mathcal{V}_{k,h+1}(\alpha_{h+1}^{\mathrm{On}})$, is a fixed class, and by \cref{lem:on_value_covering_formal} it contains $\widehat{V}_{k,h+1}$. Let $\varepsilon\coloneq (H-h)G/(N+1)$, and let $\mathcal{C}_\varepsilon$ be an internal $\varepsilon$-cover of this class in $\|\cdot\|_\infty$ of minimal cardinality. For $i\in\{0,\ldots,N\}$, let $\mathcal{F}_i$ be generated by the pre-batch history, the first $i$ trajectories of the batch, and the step-$h$ pair of trajectory $i+1$ (if $i<N$). Then $\boldsymbol{\phi}_{i,h}$ is $\mathcal{F}_{i-1}$-measurable and $\epsilon_i(V_0)$ is $\mathcal{F}_i$-measurable. Consider a specific representative $V_0\in\mathcal{C}_\varepsilon$. Given $\mathcal{F}_{i-1}$, the successor $s_{i,h+1}$ has law $P^\star(\cdot\mid s_{i,h},a_{i,h})$, because trajectory $i$ is independent of the earlier trajectories under the conditioning. Hence \cref{lem:bilinear_value_response} gives $\mathbb{E}[\epsilon_i(V_0)\mid\mathcal{F}_{i-1}]=0$. Since $V_0$ takes values in $[-(H-h)G,(H-h)G]$, $\epsilon_i(V_0)$ lies in an interval of length $2(H-h)G$, and Hoeffding's lemma gives $\mathbb{E}[\exp(\lambda\epsilon_i(V_0))\mid\mathcal{F}_{i-1}]\leq\exp(\lambda^2(H-h)^2G^2/2)$ for every $\lambda\in\mathbb{R}$. Applying \cref{lem:self_normalized_concentration} with $\boldsymbol{\Lambda}_0=I$, $\sigma=(H-h)G$, and failure probability $\delta/(4KH|\mathcal{C}_\varepsilon|)$ gives
\begin{align}
\biggl\|\sum_{i=1}^N\boldsymbol{\phi}_{i,h}\epsilon_i(V_0)\biggr\|_{\boldsymbol{\Gamma}_{k,h}^{-1}}^2&\leq2(H-h)^2G^2\log\frac{\det(\boldsymbol{\Gamma}_{k,h})^{1/2}}{\delta/(4KH|\mathcal{C}_\varepsilon|)}\\
&=(H-h)^2G^2\bigl[\log\det\boldsymbol{\Gamma}_{k,h}+2\log|\mathcal{C}_\varepsilon|+2\log(4KH/\delta)\bigr].
\end{align}
By the arithmetic--geometric mean inequality applied to the eigenvalues of $\boldsymbol{\Gamma}_{k,h}$, $\det\boldsymbol{\Gamma}_{k,h}\leq(\operatorname{tr}\boldsymbol{\Gamma}_{k,h}/d_\phi)^{d_\phi}$, and $\operatorname{tr}\boldsymbol{\Gamma}_{k,h}=d_\phi+\sum_i\|\boldsymbol{\phi}_{i,h}\|_2^2\leq d_\phi+NC_\phi^2d_\phi$. Hence, $\log\det\boldsymbol{\Gamma}_{k,h}\leq d_\phi\log(1+NC_\phi^2)$. A union bound over the $|\mathcal{C}_\varepsilon|$ representatives shows that, with conditional failure probability at most $\delta/(4KH)$,
\begin{align}
\max_{V_0\in\mathcal{C}_\varepsilon}\biggl\|\sum_{i=1}^N\boldsymbol{\phi}_{i,h}\epsilon_i(V_0)\biggr\|_{\boldsymbol{\Gamma}_{k,h}^{-1}}&\leq (H-h)G\sqrt{d_\phi\log(1+NC_\phi^2)+2\log|\mathcal{C}_\varepsilon|+2\log(4KH/\delta)}. \label{eq:on_fixed_cover_concentration}
\end{align}

\paragraph{Extension to the Fitted Target}
Choose $V_0\in\mathcal{C}_\varepsilon$ with $\|\widehat{V}_{k,h+1}-V_0\|_\infty\leq\varepsilon$. Since $\epsilon_i(\widehat{V}_{k,h+1})-\epsilon_i(V_0)=(\widehat{V}_{k,h+1}-V_0)(s_{i,h+1})-P^\star(\widehat{V}_{k,h+1}-V_0)(s_{i,h},a_{i,h})$, \cref{lem:lin_target_perturbation} with $\boldsymbol{x}_i=\boldsymbol{\phi}_{i,h}$, $y_i=s_{i,h+1}$, and $P_i=P^\star(\cdot\mid s_{i,h},a_{i,h})$ gives
\begin{align}
\biggl\|\sum_{i=1}^N\boldsymbol{\phi}_{i,h}\bigl[\epsilon_i(\widehat{V}_{k,h+1})-\epsilon_i(V_0)\bigr]\biggr\|_{\boldsymbol{\Gamma}_{k,h}^{-1}}\leq2\sqrt N\varepsilon=\frac{2\sqrt N(H-h)G}{N+1}\leq2(H-h)G.
\end{align}
This argument is deterministic and therefore remains valid for the data-dependent choice of $V_0$. Combining it with~\cref{eq:on_fixed_cover_concentration} by the triangle inequality, and substituting into~\cref{eq:value_regression_error_decomposition}, we obtain, outside the failure event of~\cref{eq:on_fixed_cover_concentration},
\begin{align}
\MoveEqLeft\|\widehat{\boldsymbol{v}}_{k,h}-\boldsymbol{v}^\star(\widehat{V}_{k,h+1})\|_{\boldsymbol{\Gamma}_{k,h}}\\&\leq (H-h)G\Bigl[\sqrt{d_\phi\log(1+NC_\phi^2)+2\log|\mathcal{C}_\varepsilon|+2\log(4KH/\delta)}+2+C_{\mathbf{M}}C_\psi\sqrt{d_\phi}\Bigr]. \label{eq:on_value_error_explicit}
\end{align}

\paragraph{Choice of Radii}
By \cref{lem:on_value_covering_formal}, $\log|\mathcal{C}_\varepsilon|\leq\mathcal{H}_{h+1}^{\mathrm{On}}(\alpha_{h+1}^{\mathrm{On}},\varepsilon)$ with $\varepsilon=(H-h)G/(N+1)$. We therefore define the radii backward from $h=H$ by
\begin{align}
\alpha_h^{\mathrm{On}}&\coloneq\begin{cases}(H-h)G\Bigl[\sqrt{d_\phi\log(1+NC_\phi^2)+2\mathcal{H}_{h+1}^{\mathrm{On}}(\alpha_{h+1}^{\mathrm{On}},\varepsilon)+2\log(4KH/\delta)}\\ \qquad\qquad+2+C_{\mathbf{M}}C_\psi\sqrt{d_\phi}\Bigr],&h<H,\\ 0,&h=H.\end{cases} \label{eq:on_value_radius}
\end{align}
Since $\mathcal{H}_{h+1}^{\mathrm{On}}(\alpha_{h+1}^{\mathrm{On}},\varepsilon)$ depends only on $\alpha_{h+1}^{\mathrm{On}}$ and known quantities, these radii are explicit and are computed before learning. By~\cref{eq:on_value_error_explicit}, $\|\widehat{\boldsymbol{v}}_{k,h}-\boldsymbol{v}^\star(\widehat{V}_{k,h+1})\|_{\boldsymbol{\Gamma}_{k,h}}\leq\alpha_h^{\mathrm{On}}$ outside the failure event of~\cref{eq:on_fixed_cover_concentration}.

It remains to verify the order of the radii. For $h\in[H-1]$, let $r_h\coloneq\alpha_h^{\mathrm{On}}/((H-h)G)$, and let $r_H\coloneq0$. Since $\alpha_{h+1}^{\mathrm{On}}\leq(H-h)G\,r_{h+1}$, every logarithm in $\mathcal{H}_{h+1}^{\mathrm{On}}(\alpha_{h+1}^{\mathrm{On}},\varepsilon)$ has an argument bounded by a polynomial in $K$, $N$, $H$, $d_\phi$, $d_\psi$, $L$, $G$, $G^{-1}$, $1/\delta$, the feature constants, and $r_{h+1}$, and the coefficients in front of these logarithms sum to at most $2(d_\phi^2+d_\psi)$. Hence, there exist a numerical constant $c$ and the logarithm $\ell$ of a polynomial in these quantities other than $r_{h+1}$, both independent of $h$, such that
\begin{align}
r_h\leq\sqrt{c(d_\phi^2+d_\psi)\bigl[\ell+\log(1+r_{h+1})\bigr]}+2+C_{\mathbf{M}}C_\psi\sqrt{d_\phi}.
\end{align}
The right-hand side is nondecreasing in $r_{h+1}$ and grows only logarithmically in it. Hence, some $\overline{r}=\widetilde{\mathcal{O}}(\sqrt{d_\phi^2+d_\psi})$ satisfies this inequality with both $r_h$ and $r_{h+1}$ replaced by $\overline{r}$, and backward induction from $r_H=0$ gives $r_h\leq\overline{r}$ for every $h$. Therefore, $\alpha_h^{\mathrm{On}}=\widetilde{\mathcal{O}}((H-h)G\sqrt{d_\phi^2+d_\psi})$.

\paragraph{Failure Probability}
For each pair $(k,h)$, the conditional failure probability is at most $\delta/(4KH)$. Averaging over the pre-batch history leaves that bound unchanged. Summing over the $K(H-1)$ pairs gives probability at most $\delta/4$ that some value-confidence bound fails. At $h=H$ no value regression is performed, and $\mathcal{E}_{\mathrm{V}}$ only concerns $h<H$. Hence $\mathcal{E}_{\mathrm{V}}$ holds with probability at least $1-\delta/4$.
\end{proof}

\subsection{Analysis of Sample Complexity: Proof of \texorpdfstring{\cref{thm:on_policy}}{Theorem \ref*{thm:on_policy}}} \label{app:regret_analysis_on}

\begin{lemma}[Regret Decomposition for CURL] \label{lem:curl_regret_decomposition}
Suppose the utility functions are differentiable and concave. For any policy sequence $\{\pi_k\}_{k=1}^K$ and fitted action values $\widehat{Q}_{k,h}$, let $\widehat{V}_{k,h}(s)=\sum_a\pi_{k,h}(a\mid s)\widehat{Q}_{k,h}(s,a)$ and $\widehat{V}_{k,H+1}=0$. Set $R_{k,h}=\nabla f_h(\mu_h^{\pi_k})$ and $\xi_{k,h}=R_{k,h}+P^\star\widehat{V}_{k,h+1}-\widehat{Q}_{k,h}$. Then
\begin{align}
\operatorname{Reg}_K&\leq\underbrace{\sum_{k=1}^K\sum_{h=1}^H\mathbb{E}_{d_h^{\pi^\star}}\bigl\langle \widehat{Q}_{k,h},\pi_h^\star-\pi_{k,h} \bigr\rangle}_{\textnormal{(I)}}+\underbrace{\sum_{k=1}^K\sum_{h=1}^H\mathbb{E}_{\mu_h^{\pi^\star}}[\xi_{k,h}]}_{\textnormal{(II)}}+\underbrace{\sum_{k=1}^K\sum_{h=1}^H\mathbb{E}_{\mu_h^{\pi_k}}[-\xi_{k,h}]}_{\textnormal{(III)}}.
\end{align}
The bound holds for every realization of the fitted functions, without any independence requirement. For step-dependent transitions, the same statement holds with $P^\star$ replaced by $P_h^\star$ in $\xi_{k,h}$.
\end{lemma}

\begin{proof}
Consider a specific iteration $k$, and recall that $R_{k,h}=\nabla f_h(\mu_h^{\pi_k})$. Concavity of each $f_h$ gives
\begin{align}
F(\mu^{\pi^\star})-F(\mu^{\pi_k})&=\sum_{h=1}^H\bigl[f_h(\mu_h^{\pi^\star})-f_h(\mu_h^{\pi_k})\bigr]\\
&\leq\sum_{h=1}^H\bigl\langle R_{k,h},\mu_h^{\pi^\star}-\mu_h^{\pi_k} \bigr\rangle\\
&=\mathbb{E}_\rho[V_1^{\pi^\star,R_k}-V_1^{\pi_k,R_k}]. \label{eq:concavity_regret_reduction}
\end{align}
The inequality applies the first-order concavity bound $f_h(\boldsymbol{u})-f_h(\boldsymbol{u}')\leq\langle\nabla f_h(\boldsymbol{u}'),\boldsymbol{u}-\boldsymbol{u}'\rangle$ with $\boldsymbol{u}'=\mu_h^{\pi_k}$. For the equality, $\langle R_{k,h},\mu_h^\pi\rangle=\mathbb{E}_{\pi,P^\star}[R_{k,h}(s_h,a_h)]$ for every policy $\pi$, and summing over steps gives $\mathbb{E}_\rho[V_1^{\pi,R_k}]$.

The right-hand side of~\cref{eq:concavity_regret_reduction} is a value difference under the fixed reward $R_k$. We decompose it with the extended performance difference lemma (\cref{lem:extended_pdl}). Applying it with $\pi=\pi^\star$, $\pi'=\pi_k$, and $\widehat{Q}_h=\widehat{Q}_{k,h}$ gives
\begin{align}
\mathbb{E}_\rho[V_1^{\pi^\star,R_k}-\widehat{V}_{k,1}]&=\sum_{h=1}^H\mathbb{E}_{d_h^{\pi^\star}}\bigl\langle \widehat{Q}_{k,h},\pi_h^\star-\pi_{k,h} \bigr\rangle+\sum_{h=1}^H\mathbb{E}_{\mu_h^{\pi^\star}}[\xi_{k,h}].
\end{align}
Applying it with $\pi=\pi'=\pi_k$, the first sum vanishes, and we obtain
\begin{align}
\mathbb{E}_\rho[V_1^{\pi_k,R_k}-\widehat{V}_{k,1}]&=\sum_{h=1}^H\mathbb{E}_{\mu_h^{\pi_k}}[\xi_{k,h}].
\end{align}
Subtracting the second identity from the first cancels $\widehat{V}_{k,1}$, and~\cref{eq:concavity_regret_reduction} gives
\begin{align}
F(\mu^{\pi^\star})-F(\mu^{\pi_k})&\leq\sum_{h=1}^H\mathbb{E}_{d_h^{\pi^\star}}\bigl\langle \widehat{Q}_{k,h},\pi_h^\star-\pi_{k,h} \bigr\rangle+\sum_{h=1}^H\mathbb{E}_{\mu_h^{\pi^\star}}[\xi_{k,h}]+\sum_{h=1}^H\mathbb{E}_{\mu_h^{\pi_k}}[-\xi_{k,h}].
\end{align}
Summing over $k$ proves the lemma. For step-dependent transitions,~\cref{lem:extended_pdl} holds with $P_h^\star$. Hence, the same argument applies.
\end{proof}

\begin{proof}[Proof of \cref{thm:on_policy}]
We first recall the relevant notation. The regret decomposition in~\cref{eq:regret_decomposition} bounds $\operatorname{Reg}_K$ by three terms. Term (I) is the regret of the policies $\pi_1,\ldots,\pi_K$ with respect to the value estimates $\widehat{Q}_1,\ldots,\widehat{Q}_K$, Term (II) sums the expected Bellman errors $\mathbb{E}_{\mu_h^{\pi^\star}}[\xi_{k,h}]$ under $\pi^\star$, and Term (III) sums the expected negative Bellman errors $\mathbb{E}_{\mu_h^{\pi_k}}[-\xi_{k,h}]$. The radii $\Delta_{k,h}^{\mathrm{On}}$ and $\alpha_h^{\mathrm{On}}$ of the good event are given in~\cref{lem:on_policy_reward_error,lem:good_estimation_event_V}, and $\operatorname{Sub\text{-}Opt}(\overline{\pi}_K)\leq\operatorname{Reg}_K/K$.

The confidence parameters required by \cref{thm:on_policy} are the value radii $\alpha_h^{\mathrm{On}}$ in~\cref{eq:on_value_radius}. All other radii are used only in the analysis. We first apply the shared regret decomposition, then bound Term (I) and Terms (II) and (III). A martingale argument (\cref{lem:on_martingale}) controls $\mathbb{E}_{\mu_h^{\pi_k}}\|\boldsymbol{\phi}\|_{\boldsymbol{\Gamma}_{k,h}^{-1}}$, after which we sum the bounds on Terms (II) and (III) and choose the batch size.

\paragraph{Regret Decomposition}
The fitted values satisfy $\widehat{V}_{k,h}=\langle \widehat{Q}_{k,h},\pi_{k,h}\rangle$ and $\widehat{V}_{k,H+1}=0$. Applying~\cref{lem:curl_regret_decomposition} therefore gives~\cref{eq:regret_decomposition}, with $R_{k,h}=\nabla f_h(\mu_h^{\pi_k})$ and $\xi_{k,h}=R_{k,h}+P^\star\widehat{V}_{k,h+1}-\widehat{Q}_{k,h}$. The reduction is deterministic. The high-probability events are needed only to bound Terms (II) and (III).

\paragraph{Bounding Term (I)}
Since \OPE{} clips the fitted values,~\cref{thm:npg_regret} applies to every sequence of value estimates generated by the framework, including the fitted values in this run. With $u=\pi^\star$ and the step size in~\cref{thm:on_policy}, it gives
\begin{align}
\mathrm{Term~(I)}&\leq\mathcal{O}\Bigl(H^2G\sqrt{K\log|\mathcal{A}|}\Bigr). \label{eq:Term_I}
\end{align}
The remaining argument uses no property of the \NPG{} implementation beyond~\cref{thm:npg_regret}. The concentration arguments below condition on the history before each fresh batch, when $\pi_k$ is fixed. They therefore require no policy-class cover.

\paragraph{Bounding Terms (II) and (III)}
On the good event $\mathcal{E}$, \cref{lem:on_bellman_error} bounds the Bellman error uniformly over the state-action space. Integrating its first inequality under $\mu_h^{\pi^\star}$ and its second under the current-policy occupancy $\mu_h^{\pi_k}$ yields
\begin{align}
\mathrm{Term~(II)}+\mathrm{Term~(III)}&\leq2\sum_{k=1}^K\sum_{h=1}^H\Delta_{k,h}^{\mathrm{On}}+2\sum_{k=1}^K\sum_{h=1}^{H-1}\alpha_h^{\mathrm{On}}\mathbb{E}_{\mu_h^{\pi_k}}\|\boldsymbol{\phi}\|_{\boldsymbol{\Gamma}_{k,h}^{-1}}. \label{eq:on_bellman_sum_proof}
\end{align}
The step-$H$ bonus vanishes because $\alpha_H^{\mathrm{On}}=0$. Uniformity of the first bound is essential: no samples from $\pi^\star$ are required to control Term (II).

\paragraph{Bounding the Expected Feature Norms}
Consider specific $k$ and $h<H$, order the $N$ step-$h$ features of the batch as $\boldsymbol{\phi}_{1,h},\ldots,\boldsymbol{\phi}_{N,h}$, and define $\boldsymbol{\Gamma}_i=I+\sum_{r=1}^i\boldsymbol{\phi}_{r,h}\boldsymbol{\phi}_{r,h}^\top$, with $\boldsymbol{\Gamma}_0=I$, and $q_i=\|\boldsymbol{\phi}_{i,h}\|_{\boldsymbol{\Gamma}_{i-1}^{-1}}^2$. By \cref{lem:on_martingale}, with probability at least $1-\delta/4$, for every $k\in[K]$ and $h<H$,
\begin{align}
\mathbb{E}_{\mu_h^{\pi_k}}\|\boldsymbol{\phi}\|_{\boldsymbol{\Gamma}_{k,h}^{-1}}&\leq\frac1N\sum_{i=1}^N\sqrt{q_i}+C_\phi\sqrt{\frac{2d_\phi\log(8KH/\delta)}{N}}. \label{eq:on_leverage_martingale}
\end{align}

\paragraph{Cumulative Estimation Error}
The definition of $\widehat{e}_{k,h}$ and Cauchy--Schwarz give, for each fixed $k$,
\begin{align}
\sum_{h=1}^{H-1}\widehat{e}_{k,h}&\leq\frac1N\sum_{i=1}^N\sum_{h=1}^H\|\boldsymbol{\phi}_{i,h}\|_{\boldsymbol{\Lambda}_k^{-1}}\\
&\leq\sqrt{\frac HN\sum_{i=1}^N\sum_{h=1}^H\|\boldsymbol{\phi}_{i,h}\|_{\boldsymbol{\Lambda}_k^{-1}}^2}\\
&=\sqrt{\frac HN\operatorname{tr}\biggl(\boldsymbol{\Lambda}_k^{-1}\sum_{i=1}^N\sum_{h=1}^H\boldsymbol{\phi}_{i,h}\boldsymbol{\phi}_{i,h}^\top\biggr)}\\
&=\sqrt{\frac HN\operatorname{tr}\bigl(I-\boldsymbol{\Lambda}_k^{-1}\bigr)}\leq\sqrt{\frac{Hd_\phi}{N}}.
\end{align}
The second equality uses the definition of $\boldsymbol{\Lambda}_k$, and the last inequality uses $0\preceq I-\boldsymbol{\Lambda}_k^{-1}\preceq I$. These steps are deterministic even though $\boldsymbol{\Lambda}_k$ uses the entire batch.

Because $\Delta_{k,1}^{\mathrm{On}}=0$, substituting~\cref{eq:Delta_on} and summing the feature-average error over the remaining $H-1$ steps yields
\begin{align}
\sum_{k=1}^K\sum_{h=1}^H\Delta_{k,h}^{\mathrm{On}}&\leq LC_\psi K\left[\sqrt{\frac{\beta^{\mathrm{On}}Hd_\phi}{N}}+C_{\mathbf{M}}C_\phi H d_\phi\sqrt{\frac{8\log(8KH/\delta)}{N}}\right]. \label{eq:on_reward_sum}
\end{align}
This bounds the cumulative pseudo-reward error appearing under both policies. We next bound $\sum_{i=1}^N\sqrt{q_i}$ in~\cref{eq:on_leverage_martingale}.

Since $\|\boldsymbol{\phi}_{i,h}\|_2\leq C_\phi\sqrt{d_\phi}$, we apply the elliptical potential lemma (\cref{lem:elliptical_potential}) with $\boldsymbol{x}_i=\boldsymbol{\phi}_{i,h}$, $R_x=C_\phi\sqrt{d_\phi}$, and $n=N$ to obtain that
\begin{align}
\sum_{i=1}^N\min\{1,q_i\}&\leq2d_\phi\log(1+NC_\phi^2).
\end{align}

The features need not have unit norm. Hence, $q_i$ can exceed one. We split the sum accordingly:
\begin{align}
\sum_{i=1}^N\sqrt{q_i}&=\sum_{i:q_i\leq1}\sqrt{q_i}+\sum_{i:q_i>1}\sqrt{q_i}\\
&\leq\sqrt{N\sum_{i:q_i\leq1}q_i}+C_\phi\sqrt{d_\phi}\,|\{i\mid q_i>1\}|\\
&\leq\sqrt{N\sum_{i=1}^N\min\{1,q_i\}}+C_\phi\sqrt{d_\phi}\sum_{i=1}^N\min\{1,q_i\}\\
&\leq\sqrt{2Nd_\phi\log(1+NC_\phi^2)}+2C_\phi d_\phi^{3/2}\log(1+NC_\phi^2).
\end{align}
The first inequality uses Cauchy--Schwarz on the first group and $\sqrt{q_i}\leq C_\phi\sqrt{d_\phi}$ on the second. The second inequality uses the fact that every index with $q_i>1$ contributes one to the truncated sum. Combining these bounds proves
\begin{align}
\mathbb{E}_{\mu_h^{\pi_k}}\|\boldsymbol{\phi}\|_{\boldsymbol{\Gamma}_{k,h}^{-1}}&\leq\sqrt{\frac{2d_\phi\log(1+NC_\phi^2)}N}+\frac{2C_\phi d_\phi^{3/2}\log(1+NC_\phi^2)}N+C_\phi\sqrt{\frac{2d_\phi\log(8KH/\delta)}N}.
\end{align}
Here we combined~\cref{eq:on_leverage_martingale} with the bound on $\sum_i\sqrt{q_i}$. In particular, the right-hand side is $\widetilde{\mathcal{O}}(\sqrt{d_\phi/N}+d_\phi^{3/2}/N)$. By \cref{lem:good_estimation_event_V}, $\alpha_h^{\mathrm{On}}=\widetilde{\mathcal{O}}((H-h)G\sqrt{d_\phi^2+d_\psi})$. Summing over $h$ gives $\sum_{h<H}\alpha_h^{\mathrm{On}}=\widetilde{\mathcal{O}}(H^2G\sqrt{d_\phi^2+d_\psi})$, where the factor $H^2$ follows from $\sum_{h<H}(H-h)=H(H-1)/2$. Thus
\begin{align}
2\sum_{k=1}^K\sum_{h=1}^{H-1}\alpha_h^{\mathrm{On}}\mathbb{E}_{\mu_h^{\pi_k}}\|\boldsymbol{\phi}\|_{\boldsymbol{\Gamma}_{k,h}^{-1}}&\leq\widetilde{\mathcal{O}}\left(KH^2G\sqrt{\frac{d_\phi(d_\phi^2+d_\psi)}N}+\frac{KH^2Gd_\phi^{3/2}\sqrt{d_\phi^2+d_\psi}}N\right). \label{eq:on_value_uncertainty_sum}
\end{align}

\paragraph{Putting Everything Together}
\cref{lem:good_estimation_event_P} shows that $\mathcal{E}_{\mathrm{P}}$ fails with probability at most $\delta/4$, \cref{lem:on_policy_reward_error} gives $\Pr(\mathcal{E}_{\mathrm{P}}\cap\mathcal{E}_{\mathrm{R}}^c)\leq\delta/4$, \cref{lem:good_estimation_event_V} shows that $\mathcal{E}_{\mathrm{V}}$ fails with probability at most $\delta/4$, and \cref{lem:on_martingale} shows that~\cref{eq:on_leverage_martingale} fails with probability at most $\delta/4$. Since $\mathcal{E}^c\subseteq\mathcal{E}_{\mathrm{P}}^c\cup(\mathcal{E}_{\mathrm{P}}\cap\mathcal{E}_{\mathrm{R}}^c)\cup\mathcal{E}_{\mathrm{V}}^c$, the union bound shows that $\mathcal{E}$ and~\cref{eq:on_leverage_martingale} hold simultaneously with probability at least $1-\delta$. No independence between these events is required.

On this event, combine~\cref{eq:Term_I,eq:on_reward_sum,eq:on_value_uncertainty_sum} with~\cref{eq:on_bellman_sum_proof} and substitute into~\cref{eq:regret_decomposition}. Dividing by $K$ gives
\begin{align}
\frac{\operatorname{Reg}_K}{K}&\leq\mathcal{O}\left(H^2G\sqrt{\frac{\log|\mathcal{A}|}{K}}\right)+2LC_\psi\left[\sqrt{\frac{\beta^{\mathrm{On}}Hd_\phi}{N}}+C_{\mathbf{M}}C_\phi H d_\phi\sqrt{\frac{8\log(8KH/\delta)}N}\right]\\
&\quad+\widetilde{\mathcal{O}}\left(H^2G\sqrt{\frac{d_\phi(d_\phi^2+d_\psi)}N}+\frac{H^2Gd_\phi^{3/2}\sqrt{d_\phi^2+d_\psi}}N\right).
\end{align}
Using $\sqrt{\beta^{\mathrm{On}}}=\widetilde{\mathcal{O}}(\sqrt{d_\phi})$ and suppressing fixed feature constants, this becomes
\begin{align}
\frac{\operatorname{Reg}_K}{K}&\leq\widetilde{\mathcal{O}}\left(\frac{H^2G}{\sqrt K}+\frac{LH d_\phi}{\sqrt N}+H^2G\sqrt{\frac{d_\phi(d_\phi^2+d_\psi)}N}+\frac{H^2Gd_\phi^{3/2}\sqrt{d_\phi^2+d_\psi}}N\right).
\end{align}
In particular, the \NPG{} regret term remains present independently of the batch size. Choose $N=\widetilde{\mathcal{O}}(d_\phi(d_\phi^2+d_\psi)K)$ with a sufficiently large multiplier. The second term is at most $\widetilde{\mathcal{O}}(LH/\sqrt K)$, the third is $\widetilde{\mathcal{O}}(H^2G/\sqrt K)$, and the fourth is at most $\widetilde{\mathcal{O}}(H^2G/K)$. These comparisons use $d_\phi^2+d_\psi\geq d_\phi^2$ and $K\geq1$. Since $H\geq1$, we conclude that $\operatorname{Reg}_K/K\leq\widetilde{\mathcal{O}}(H^2(L+G)/\sqrt K)$.

The episode-wise mixture has the average occupancy. By concavity, $F(\mu^{\overline{\pi}_K})\geq K^{-1}\sum_kF(\mu^{\pi_k})$, and hence $\operatorname{Sub\text{-}Opt}(\overline{\pi}_K)\leq\operatorname{Reg}_K/K$. Choosing $K=\widetilde{\mathcal{O}}(H^4(L+G)^2/\epsilon^2)$ makes the latter at most $\epsilon$. With this choice, $N=\widetilde{\mathcal{O}}(H^4d_\phi(d_\phi^2+d_\psi)(L+G)^2/\epsilon^2)$ fresh trajectories are collected per iteration. The total sample complexity is therefore $KN=\widetilde{\mathcal{O}}(d_\phi(d_\phi^2+d_\psi)H^8(L+G)^4/\epsilon^4)$, and the number of observed transitions is $HKN$.
\end{proof}

The following lemma controls the expected feature norms used in the proof above.

\begin{lemma}[Martingale Concentration] \label{lem:on_martingale}
For every $k\in[K]$ and $h<H$, order the $N$ step-$h$ features of the batch at iteration $k$ as $\boldsymbol{\phi}_{1,h},\ldots,\boldsymbol{\phi}_{N,h}$, and let $\boldsymbol{\Gamma}_i=I+\sum_{r=1}^i\boldsymbol{\phi}_{r,h}\boldsymbol{\phi}_{r,h}^\top$ for $i\in\{0,\ldots,N\}$. Then, with probability at least $1-\delta/4$, for every $k\in[K]$ and $h<H$,
\begin{align}
\mathbb{E}_{\mu_h^{\pi_k}}\|\boldsymbol{\phi}\|_{\boldsymbol{\Gamma}_{k,h}^{-1}}&\leq\frac1N\sum_{i=1}^N\|\boldsymbol{\phi}_{i,h}\|_{\boldsymbol{\Gamma}_{i-1}^{-1}}+C_\phi\sqrt{\frac{2d_\phi\log(8KH/\delta)}{N}}.
\end{align}
\end{lemma}

\begin{proof}
Consider specific $k$ and $h<H$, and condition on the history before batch $k$. Let $\mathcal{F}_i$ contain this history and the first $i$ step-$h$ features, and put $q_i=\|\boldsymbol{\phi}_{i,h}\|_{\boldsymbol{\Gamma}_{i-1}^{-1}}^2$ and $m_i=\mathbb{E}[\sqrt{q_i}\mid\mathcal{F}_{i-1}]=\mathbb{E}_{\mu_h^{\pi_k}}\|\boldsymbol{\phi}\|_{\boldsymbol{\Gamma}_{i-1}^{-1}}$. The conditional-mean equality holds because the trajectories are independent given the history before the batch and all follow the fixed policy $\pi_k$. Hence $x_i\coloneq m_i-\sqrt{q_i}$ is a martingale difference with respect to $\{\mathcal{F}_i\}$. Both $m_i$ and $\sqrt{q_i}$ lie in $[0,C_\phi\sqrt{d_\phi}]$, since $\boldsymbol{\Gamma}_{i-1}\succeq I$ gives $\sqrt{q_i}\leq\|\boldsymbol{\phi}_{i,h}\|_2\leq C_\phi\sqrt{d_\phi}$. Thus $|x_i|\leq C_\phi\sqrt{d_\phi}$. Applying \cref{lem:azuma_hoeffding} with $R=C_\phi\sqrt{d_\phi}$ and setting its right-hand side to $\delta/(4KH)$, and then taking a union bound over the at most $KH$ pairs $(k,h)$, gives, with probability at least $1-\delta/4$,
\begin{align}
\sum_{i=1}^N m_i&\leq\sum_{i=1}^N\sqrt{q_i}+C_\phi\sqrt{2Nd_\phi\log(8KH/\delta)}
\end{align}
simultaneously for every $k\in[K]$ and $h<H$. The conditional bound is uniform over prior histories. Hence, averaging over them preserves this failure probability. Finally, $\boldsymbol{\Gamma}_{k,h}=\boldsymbol{\Gamma}_N\succeq\boldsymbol{\Gamma}_{i-1}$ implies $\boldsymbol{\Gamma}_{k,h}^{-1}\preceq\boldsymbol{\Gamma}_{i-1}^{-1}$, and hence $\mathbb{E}_{\mu_h^{\pi_k}}\|\boldsymbol{\phi}\|_{\boldsymbol{\Gamma}_{k,h}^{-1}}\leq m_i$ for every $i$. Averaging over $i$ and dividing the display by $N$ proves the lemma.
\end{proof}

\section{Deferred Details and Proofs for \texorpdfstring{\cref{section:ocupo-off}}{Section \ref*{section:ocupo-off}}} \label{app:ucgu_off}

We first describe the three estimation components and collect the covering arguments in \cref{app:off_covering_arguments}. We then define the good events used in the replay analysis and prove their guarantees, and the final subsection combines these bounds into the sample-complexity proof. As in \cref{app:ucgu_on}, each guarantee states the order of its confidence radius, the exact value is given in its proof, and $\widetilde{\mathcal{O}}(\cdot)$ suppresses logarithmic factors and the fixed feature-regularity constants and the constants in~\cref{asm:pnpg,asm:policy_class}. The policies collecting the replay data may depend on all preceding trajectories. Each transition still has conditional law $P^\star$ after its current state and action are fixed. We exploit this conditional centering only for fixed functions, and use covers chosen before data collection to handle the fitted targets.

\subsection{Algorithm Design} \label{app:off_algorithm_design}

\subsubsection{Transition Model Estimation} \label{app:model_off}

We use the same \BTE{} routine as in \cref{subsec:on_policy_algorithm}, now with the replay dataset $\mathcal{D}_k=\{\tau_i\}_{i=1}^k$. \cref{lem:bilinear_regression_response} applies to every replay transition, since the conditional successor law is unchanged. Thus $\boldsymbol{\Lambda}_k=I+\sum_{i=1}^k\sum_{h=1}^H\boldsymbol{\phi}_{i,h}\boldsymbol{\phi}_{i,h}^\top$. The policy update is \PNPG{} satisfying~\cref{asm:pnpg}.

\subsubsection{Pseudo-Reward Estimation} \label{app:reward_off} \label{app:off_bfr_details}

The pseudo-reward $R_{k,h}=\nabla f_h(\mu_h^{\pi_k})$ requires the current-policy occupancy. Recall the expected feature $\boldsymbol{\omega}_h^\pi=\mathbb{E}_{\mu_h^\pi}\boldsymbol{\phi}$. This vector suffices to reconstruct the occupancy: by \cref{lem:occupancy_measure},
\begin{align}
\mu_h^{\pi_k}(s,a)&=\pi_{k,h}(a\mid s)(\boldsymbol{\omega}_{h-1}^{\pi_k})^\top\mathbf{M}^\star\boldsymbol{\psi}(s),\qquad h\geq2.
\end{align}

A direct average of the features in $\mathcal{D}_k$ mixes the occupancies of the behavior policies $\pi_1,\ldots,\pi_k$ and therefore does not estimate $\boldsymbol{\omega}_h^{\pi_k}$. We instead propose pseudo-reward estimation by backward feature regression (\BFR{}), which evaluates the current policy using the stored transitions.

For a target step $h$, the function $\boldsymbol{g}_{h,t}^{\pi}$ gives the expected step-$h$ feature vector from a state at an earlier step $t$. The following recursion explains why backward regression can recover the feature expectation.

\begin{lemma}[Backward Feature Recursion] \label{lem:backward_feature_recursion}
Consider a specific policy $\pi$ and target step $h$. For $t\leq h$, let $\boldsymbol{g}_{h,t}^{\pi}(s)=\mathbb{E}_{\pi,P^\star}[\boldsymbol{\phi}(s_h,a_h)\mid s_t=s]$ and $\overline{\boldsymbol{\phi}}_{\pi,t}(s)=\sum_a\pi_t(a\mid s)\boldsymbol{\phi}(s,a)$. Then
\begin{align}
\boldsymbol{g}_{h,h}^{\pi}(s)&=\overline{\boldsymbol{\phi}}_{\pi,h}(s), \label{eq:off_population_feature_base}\\
\boldsymbol{g}_{h,t}^{\pi}(s)&=\sum_a\pi_t(a\mid s)P^\star\boldsymbol{g}_{h,t+1}^{\pi}(s,a),\qquad t<h. \label{eq:off_population_feature_recursion}
\end{align}
Here $P^\star\boldsymbol{g}(s,a)=\sum_{s'}P^\star(s'\mid s,a)\boldsymbol{g}(s')$. Moreover, $\boldsymbol{\omega}_h^\pi=\mathbb{E}_\rho[\boldsymbol{g}_{h,1}^{\pi}(s_1)]$.
\end{lemma}

\begin{proof}
At $t=h$, only the action $a_h\sim\pi_h(\cdot\mid s)$ remains to be averaged. Hence, $\boldsymbol{g}_{h,h}^{\pi}(s)=\sum_a\pi_h(a\mid s)\boldsymbol{\phi}(s,a)=\overline{\boldsymbol{\phi}}_{\pi,h}(s)$, which is~\cref{eq:off_population_feature_base}. For $t<h$ and $s\in\mathcal{S}$, conditioning first on the current action and then on the next state gives
\begin{align}
\boldsymbol{g}_{h,t}^{\pi}(s) &=\mathbb{E}_{\pi,P^\star}\bigl[ \boldsymbol{\phi}(s_h,a_h)\mid s_t=s \bigr]\\
&=\sum_{a\in\mathcal{A}}\pi_t(a\mid s)\mathbb{E}_{\pi,P^\star}\bigl[ \boldsymbol{\phi}(s_h,a_h)\mid s_t=s,a_t=a \bigr]\\
&=\sum_{a\in\mathcal{A}}\pi_t(a\mid s)\sum_{s^\prime\in\mathcal{S}}P^\star(s^\prime\mid s,a) \mathbb{E}_{\pi,P^\star}\bigl[ \boldsymbol{\phi}(s_h,a_h)\mid s_{t+1}=s^\prime \bigr]\\
&=\sum_{a\in\mathcal{A}}\pi_t(a\mid s)\sum_{s^\prime\in\mathcal{S}}P^\star(s^\prime\mid s,a)\boldsymbol{g}_{h,t+1}^{\pi}(s^\prime).
\end{align}
The second equality is the law of total expectation over $a_t\sim\pi_t(\cdot\mid s)$. For the third, conditional on $(s_t,a_t)=(s,a)$ the next state is distributed as $P^\star(\cdot\mid s,a)$, and the Markov property makes the conditional distribution of the remaining trajectory depend on the past only through $s_{t+1}$. The fourth equality uses the definition of $\boldsymbol{g}_{h,t+1}^{\pi}$. This proves~\cref{eq:off_population_feature_recursion}. Finally, average $\boldsymbol{g}_{h,1}^{\pi}$ over $s_1\sim\rho$ and apply the law of total expectation. The last equality below is the definition of $\boldsymbol{\omega}_h^\pi$ in \cref{sec:preliminaries}:
\begin{align}
\mathbb{E}_\rho[\boldsymbol{g}_{h,1}^{\pi}(s_1)]&=\mathbb{E}_{\pi,P^\star}[\boldsymbol{\phi}(s_h,a_h)]=\boldsymbol{\omega}_h^\pi.
\end{align}
This establishes the feature-expectation identity.
\end{proof}

Here $h$ remains fixed: $\boldsymbol{g}_{h,t}^{\pi}$ predicts the feature at step $h$ from a state at step $t$. To fit each backward update from observed transitions, we use its linear conditional mean.

\begin{lemma}[Linearity of Expected Next-State Features] \label{lem:bilinear_bfr_linearity}
Suppose that the MDP is a bilinear MDP (\cref{def:bilienar_mdp}). For any bounded function $\boldsymbol{g}\colon \mathcal{S}\to\mathbb{R}^{d_\phi}$, let $\mathbf{B}^\star(\boldsymbol{g})\coloneq\mathbf{M}^\star\sum_{s'}\boldsymbol{\psi}(s')\boldsymbol{g}(s')^\top\in\mathbb{R}^{d_\phi\times d_\phi}$. Then, for every $(s,a)\in\mathcal{S}\times\mathcal{A}$,
\begin{align}
P^\star\boldsymbol{g}(s,a)&=\mathbf{B}^\star(\boldsymbol{g})^\top\boldsymbol{\phi}(s,a). \label{eq:bfr_linearity}
\end{align}
\end{lemma}

\begin{proof}
By the definition of $P^\star\boldsymbol{g}$ and the bilinear transition in \cref{def:bilienar_mdp},
\begin{align}
P^\star\boldsymbol{g}(s,a)&=\sum_{s'}\boldsymbol{g}(s')P^\star(s'\mid s,a)\\
&=\biggl(\sum_{s'}\boldsymbol{g}(s')\boldsymbol{\psi}(s')^\top\biggr)(\mathbf{M}^\star)^\top\boldsymbol{\phi}(s,a)=\mathbf{B}^\star(\boldsymbol{g})^\top\boldsymbol{\phi}(s,a).
\end{align}
The second equality substitutes $P^\star(s'\mid s,a)=\boldsymbol{\psi}(s')^\top(\mathbf{M}^\star)^\top\boldsymbol{\phi}(s,a)$ and pulls $(\mathbf{M}^\star)^\top\boldsymbol{\phi}(s,a)$ out of the sum, since it does not depend on $s'$. The last equality is the definition of $\mathbf{B}^\star(\boldsymbol{g})$.
\end{proof}

\begin{lemma}[Unbiased Backward Feature Regression] \label{lem:bilinear_bfr_response}
Under \cref{def:bilienar_mdp}, let $\boldsymbol{g}\colon \mathcal{S}\to\mathbb{R}^{d_\phi}$ be a bounded function, and let $\mathbf{B}^\star(\boldsymbol{g})$ be as in \cref{lem:bilinear_bfr_linearity}. Let $\mathcal{H}$ contain the history through a current state-action pair $(s,a)$, and suppose $\boldsymbol{g}$ is fixed conditional on $\mathcal{H}$. If $S'$ has conditional law $P^\star(\cdot\mid s,a)$, then
\begin{align}
\mathbb{E}[\boldsymbol{g}(S')\mid\mathcal{H}]&=P^\star\boldsymbol{g}(s,a)=\mathbf{B}^\star(\boldsymbol{g})^\top\boldsymbol{\phi}(s,a). \label{eq:rfe_population_prediction}
\end{align}
For a fixed $\boldsymbol{g}$ and independent initial states $s_{i,1}\sim\rho$,
\begin{align}
\mathbb{E}\biggl[\frac1k\sum_{i=1}^k\boldsymbol{g}(s_{i,1})\biggr]&=\mathbb{E}_{s_1\sim\rho}[\boldsymbol{g}(s_1)].
\end{align}
\end{lemma}

\begin{proof}
Since $\boldsymbol{g}$ and $(s,a)$ are fixed under the conditioning and $S'$ has law $P^\star(\cdot\mid s,a)$, the definition of the conditional expectation gives $\mathbb{E}[\boldsymbol{g}(S')\mid\mathcal{H}]=\sum_{s'}\boldsymbol{g}(s')P^\star(s'\mid s,a)=P^\star\boldsymbol{g}(s,a)$. \cref{lem:bilinear_bfr_linearity} then gives the second equality in~\cref{eq:rfe_population_prediction}. The behavior policy does not enter this conditional mean. For the initial-state average, every summand has expectation $\sum_s\rho(s)\boldsymbol{g}(s)$. Hence, taking the expectation of the average proves the claim. Both assertions concern a fixed function. Substituting a function fitted from these same observations requires the uniform concentration argument below.
\end{proof}

Combining~\cref{eq:bfr_linearity,eq:off_population_feature_recursion} with $\boldsymbol{g}=\boldsymbol{g}_{h,t+1}^{\pi}$ gives the explicit connection
\begin{align}
\boldsymbol{g}_{h,t}^{\pi}(s)=\mathbf{B}^\star(\boldsymbol{g}_{h,t+1}^{\pi})^\top\overline{\boldsymbol{\phi}}_{\pi,t}(s),\qquad t<h.
\end{align}
Thus $\mathbf{B}^\star(\boldsymbol{g})$ contains the linear coefficients of the one-step conditional expectation of $\boldsymbol{g}$. The policy average turns this prediction into the feature expectation at the preceding step. The coefficient depends on the feature-regression target. Hence, different target steps $h$ generally require different matrices.

\BFR{} estimates these coefficients by backward regression. In \cref{line:rfe_policy_feature}, it computes the policy-averaged features and, for each $h$, initializes $\widehat{\boldsymbol{g}}_{h,h}^{\pi_k}=\overline{\boldsymbol{\phi}}_{\pi_k,h}$ in \cref{line:rfe_base_target}. At step $t<h$, the successor function $\widehat{\boldsymbol{g}}_{h,t+1}^{\pi_k}$ has already been constructed. Its evaluations at the stored successor states, $\{\widehat{\boldsymbol{g}}_{h,t+1}^{\pi_k}(s_{i,t+1})\}_{i=1}^k$, are the targets of the step-$t$ regression. Hence the regression uses the replay transitions together with evaluations of $\widehat{\boldsymbol{g}}_{h,t+1}^{\pi_k}$.

\cref{lem:bilinear_bfr_linearity,lem:bilinear_bfr_response} establish the population regression model. The learner computes neither the state-space sum defining $\mathbf{B}^\star(\boldsymbol{g})$ nor the unknown matrix $\mathbf{M}^\star$.

Writing $\boldsymbol{\phi}_{i,t}=\boldsymbol{\phi}(s_{i,t},a_{i,t})$, \cref{line:rfe_regression} solves
\begin{align}
\widehat{\mathbf{B}}_{h,t}^{\pi_k}\coloneq\operatorname*{arg\,min}_{\mathbf{B}\in\mathbb{R}^{d_\phi\times d_\phi}}\biggl\{\sum_{i=1}^k\bigl\lVert \mathbf{B}^\top\boldsymbol{\phi}_{i,t}-\widehat{\boldsymbol{g}}_{h,t+1}^{\pi_k}(s_{i,t+1}) \bigr\rVert_2^2+\lVert \mathbf{B} \rVert_F^2\biggr\}. \label{eq:rfe_ridge_objective}
\end{align}
Both the target and the prediction lie in $\mathbb{R}^{d_\phi}$. Each column of $\mathbf{B}$ fits one target coordinate. The objective in~\cref{eq:rfe_ridge_objective} is a strictly convex quadratic in $\mathbf{B}$ with gradient $2\sum_{i=1}^k\boldsymbol{\phi}_{i,t}(\boldsymbol{\phi}_{i,t}^\top\mathbf{B}-\widehat{\boldsymbol{g}}_{h,t+1}^{\pi_k}(s_{i,t+1})^\top)+2\mathbf{B}$. Setting it to zero gives the normal equations
\begin{align}
\boldsymbol{\Gamma}_{k,t}\widehat{\mathbf{B}}_{h,t}^{\pi_k}
=\sum_{i=1}^k\boldsymbol{\phi}_{i,t}\widehat{\boldsymbol{g}}_{h,t+1}^{\pi_k}(s_{i,t+1})^\top.
\end{align}
Since $\boldsymbol{\Gamma}_{k,t}=I+\sum_{i=1}^k\boldsymbol{\phi}_{i,t}\boldsymbol{\phi}_{i,t}^\top\succeq I$, the unique solution is
\begin{align}
\widehat{\mathbf{B}}_{h,t}^{\pi_k}=\boldsymbol{\Gamma}_{k,t}^{-1}\sum_{i=1}^k\boldsymbol{\phi}_{i,t}\widehat{\boldsymbol{g}}_{h,t+1}^{\pi_k}(s_{i,t+1})^\top. \label{eq:rfe_ridge_solution}
\end{align}
Thus the computation uses $k$ observed feature-target pairs and a $d_\phi\times d_\phi$ matrix solve with $d_\phi$ target coordinates. For $t<h$, evaluating the fitted function $\widehat{\boldsymbol{g}}_{h,t}^{\pi_k}$ uses its stored regression coefficient and a finite action sum. Neither fitting nor evaluating the prediction requires enumeration of $\mathcal{S}$ or an integral oracle.

For a fixed target, the behavior policies change the covariate distribution but preserve the conditional mean above. Here, unbiasedness concerns the sampled target $\boldsymbol{g}(s')$. The regularized fitted matrix itself can be biased. In \BFR{}, the selected $\widehat{\boldsymbol{g}}_{h,t+1}^{\pi_k}$ depends on the same replay data. Hence, we do not apply fixed-target centering directly to its selected residuals. The analysis first proves concentration uniformly over the deterministic class of feature-regression targets $\mathcal{W}$, defined in~\cref{app:off_covering_arguments}, and then substitutes $\widehat{\boldsymbol{g}}_{h,t+1}^{\pi_k}$. This accounts for target reuse without collecting additional trajectories.

\cref{line:rfe_projection} uses the Euclidean projection $\operatorname{Proj}_{\mathbb{B}_2(R_\phi)}(\boldsymbol{v})=\boldsymbol{v}\min\{1,R_\phi/\|\boldsymbol{v}\|_2\}$ with $R_\phi=C_\phi\sqrt{d_\phi}$, and with multiplier one at $\boldsymbol{v}=\boldsymbol{0}$. The feature target at $t=h$ has norm at most $C_\phi\sqrt{d_\phi}$, and projection followed by policy averaging preserves this bound at every preceding step. Consequently, $P^\star\widehat{\boldsymbol{g}}_{h,t+1}^{\pi_k}(s,a)$, an average of vectors in the convex ball $\mathbb{B}_2(C_\phi\sqrt{d_\phi})$, lies in the same ball. Euclidean projection onto a closed convex set is nonexpansive and fixes the points of the set. Hence, $\|\operatorname{Proj}(\boldsymbol{v})-\boldsymbol{w}\|_2\leq\|\boldsymbol{v}-\boldsymbol{w}\|_2$ for every $\boldsymbol{w}$ in the ball. In particular, projection cannot increase the prediction error relative to this expectation.

The outer loop selects the target feature step $h$. The inner loop alternates regression, projection, and policy averaging to estimate $\boldsymbol{g}_{h,t}^{\pi_k}$ at each preceding step down to $t=1$. \cref{line:rfe_omega} then averages $\widehat{\boldsymbol{g}}_{h,1}^{\pi_k}$ over the stored initial states to estimate $\boldsymbol{\omega}_h^{\pi_k}$. All targets reuse the same transitions and per-step Gram matrices, without collecting additional trajectories.

Finally, \BFR{} uses the known initial distribution at $h=1$ and combines $\widehat{\boldsymbol{\omega}}_{k,h-1}$ with $\mathbf{M}_k$ for $h\geq2$. Clipping keeps each occupancy coordinate in $[0,1]$, after which the coordinatewise utility gradient gives $\widehat{R}_{k,h}$. The state-space sum defining $\mathbf{B}^\star(\boldsymbol{g})$ only characterizes the population coefficient (see~\cref{lem:bilinear_bfr_linearity}). The algorithm estimates it through~\cref{eq:rfe_ridge_solution}, using stored transitions and finite action sums.

\subsubsection{Optimistic Policy Evaluation} \label{app:ope_off}

The policy evaluation step uses the same \OPE{} routine (\cref{alg:ope}) with the estimated pseudo-reward $\widehat{R}_k$, the replay dataset $\mathcal{D}_k$, and the value-regression radii $\{\alpha_h^{\mathrm{Off}}\}_{h=1}^{H-1}$. The fixed-target conditional mean remains that of \cref{lem:bilinear_value_response}. The radii account for the larger function class induced by replay-dependent policy selection. For every $h<H$, with $\boldsymbol{\Gamma}_{k,h}=I+\sum_{i=1}^k\boldsymbol{\phi}_{i,h}\boldsymbol{\phi}_{i,h}^\top$, it fits
\begin{align}
\widehat{\boldsymbol{v}}_{k,h}=\boldsymbol{\Gamma}_{k,h}^{-1}\sum_{i=1}^k\boldsymbol{\phi}(s_{i,h},a_{i,h})\widehat{V}_{k,h+1}(s_{i,h+1}),
\end{align}
and forms the backup
\begin{align}
\widetilde{Q}_{k,h}(s,a)=\widehat{R}_{k,h}(s,a)+\boldsymbol{\phi}(s,a)^\top\widehat{\boldsymbol{v}}_{k,h}+\alpha_h^{\mathrm{Off}}\bigl\lVert \boldsymbol{\phi}(s,a) \bigr\rVert_{\boldsymbol{\Gamma}_{k,h}^{-1}}.
\end{align}
The routine clips this backup to $[-(H-h+1)G,(H-h+1)G]$ and averages it under $\pi_{k,h}$. At step $H$, it simply clips $\widehat{R}_{k,H}$ to $[-G,G]$.

The value-regression bonus compensates for error in the predicted expected next-state value. The pseudo-reward estimation error is handled separately in the regret analysis: its uniform norm bounds the possible positive Bellman error under the comparator policy, and the same norm also enters the prediction-error bound under the current policy. \BFR{} and transition-core concentration (\cref{lem:lin_matrix_self_normalized}) make the sum of these errors $\widetilde{\mathcal{O}}(\sqrt K)$. Thus the algorithm does not need to compute a pseudo-reward confidence radius or evaluate expectations under the occupancy of the current policy. Those expectations are used only in the proof.

\subsection{Covering Arguments} \label{app:off_covering_arguments}

This subsection collects the covering bounds used in the proofs of \cref{lem:off_reward_event_main,lem:off_value_event_main}. In contrast to \cref{app:on_covering_arguments}, the classes also vary over the policies, since the policies are selected using the replay data. Let $\overline{\Pi}\coloneq\Pi\cup\{\pi_{\mathrm{Unif}}\}$. Every policy generated by $\OCUPOOFF$ lies in $\overline{\Pi}$, since $\pi_1=\pi_{\mathrm{Unif}}$ and $\pi_k=\operatorname{Proj}_\Pi(\pi_{k-1/2})\in\Pi$ for $k\geq2$. Adding one policy increases the size of any cover by at most one. Hence, $\log\mathcal{N}(\varepsilon,\overline{\Pi},\mathrm{dist}_\Pi)\leq\log\mathcal{N}(\varepsilon,\Pi,\mathrm{dist}_\Pi)+\log2$ for every $\varepsilon\in(0,1]$, and \cref{asm:policy_class} also holds for $\overline{\Pi}$ with the same $d_\Pi$. Below, we apply \cref{asm:policy_class} to $\overline{\Pi}$.

\paragraph{Feature-Regression Targets}
For $\pi\in\overline{\Pi}$, $t\in[H]$, and $\mathbf{B}\in\mathbb{R}^{d_\phi\times d_\phi}$, define
\begin{align}
\boldsymbol{g}_{\mathbf{B},\pi,t}(s)&\coloneq\sum_a\pi_t(a\mid s)\operatorname{Proj}_{\mathbb{B}_2(C_\phi\sqrt{d_\phi})}(\mathbf{B}^\top\boldsymbol{\phi}(s,a)).
\end{align}
The base target $\overline{\boldsymbol{\phi}}_{\pi_k,t}$ averages features with norm at most $C_\phi\sqrt{d_\phi}$, and each fitted target $\widehat{\boldsymbol{g}}_{h,t}^{\pi_k}$ averages projections onto $\mathbb{B}_2(C_\phi\sqrt{d_\phi})$ (\cref{line:rfe_projection} of \cref{alg:rfe}). Hence, every feature-regression target has uniform Euclidean norm at most $C_\phi\sqrt{d_\phi}$. Comparing the objective in~\cref{eq:rfe_ridge_objective} at its minimizer with its value at $\mathbf{B}=\boldsymbol{0}$, we obtain $\|\widehat{\mathbf{B}}_{h,t}^{\pi_k}\|_F^2\leq\sum_{i=1}^k\|\widehat{\boldsymbol{g}}_{h,t+1}^{\pi_k}(s_{i,t+1})\|_2^2\leq KC_\phi^2d_\phi$. Hence, every fitted feature-regression target, together with every base target $\overline{\boldsymbol{\phi}}_{\pi_k,t}$, belongs to the fixed class
\begin{align}
\mathcal{W}&\coloneq\bigcup_{t=1}^H\bigl(\{\boldsymbol{g}_{\mathbf{B},\pi,t}\mid\|\mathbf{B}\|_F\leq C_\phi\sqrt{d_\phi K},\ \pi\in\overline{\Pi}\}\cup\{\overline{\boldsymbol{\phi}}_{\pi,t}\mid\pi\in\overline{\Pi}\}\bigr). \label{eq:off_feature_target_class}
\end{align}

\begin{lemma}[Cover of the Feature Regression Targets] \label{lem:off_feature_cover}
The class $\mathcal{W}$ in~\cref{eq:off_feature_target_class} contains every feature-regression target produced by \BFR{}, and each member has uniform Euclidean norm at most $C_\phi\sqrt{d_\phi}$. For every $\varepsilon>0$,
\begin{align}
\log\mathcal{N}(\varepsilon,\mathcal{W},\|\cdot\|_{\infty,2})&\leq\log(2H)+d_\phi^2\log\left(1+\frac{8C_\phi^2d_\phi\sqrt K}{\varepsilon}\right)+\log\mathcal{N}\left(\min\left\{1,\frac{\varepsilon}{4C_\phi\sqrt{d_\phi}}\right\},\overline{\Pi},\mathrm{dist}_\Pi\right). \label{eq:off_feature_cover_explicit}
\end{align}
Under \cref{asm:policy_class}, its logarithmic covering number at inverse-polynomial radii is $\widetilde{\mathcal{O}}(d_\phi^2+d_\Pi)$.
\end{lemma}

\begin{proof}
The containment and the norm bound are shown above. For two parameter pairs $(\mathbf{B},\pi)$ and $(\mathbf{B}',\pi')$ at the same step $t$, add and subtract $\boldsymbol{g}_{\mathbf{B}',\pi,t}$, the function using $\pi_t$ and $\mathbf{B}'$ together. The coefficient difference is bounded, at each state, by $\sum_a\pi_t(a\mid s)\|(\mathbf{B}-\mathbf{B}')^\top\boldsymbol{\phi}(s,a)\|_2\leq C_\phi\sqrt{d_\phi}\|\mathbf{B}-\mathbf{B}'\|_F$, using nonexpansiveness of projection and $\|\boldsymbol{A}^\top\boldsymbol{u}\|_2\leq\|\boldsymbol{A}\|_F\|\boldsymbol{u}\|_2$. Since every projected vector has norm at most $C_\phi\sqrt{d_\phi}$, the policy difference is bounded by $\sum_a|\pi_t(a\mid s)-\pi_t'(a\mid s)|C_\phi\sqrt{d_\phi}\leq C_\phi\sqrt{d_\phi}\,\mathrm{dist}_\Pi(\pi,\pi')$. Therefore
\begin{align}
\|\boldsymbol{g}_{\mathbf{B},\pi,t}-\boldsymbol{g}_{\mathbf{B}',\pi',t}\|_{\infty,2}&\leq C_\phi\sqrt{d_\phi}\|\mathbf{B}-\mathbf{B}'\|_F+C_\phi\sqrt{d_\phi}\mathrm{dist}_\Pi(\pi,\pi'),\\
\|\overline{\boldsymbol{\phi}}_{\pi,t}-\overline{\boldsymbol{\phi}}_{\pi',t}\|_{\infty,2}&\leq C_\phi\sqrt{d_\phi}\mathrm{dist}_\Pi(\pi,\pi').
\end{align}
Choose an internal matrix cover at Frobenius radius $\varepsilon/(4C_\phi\sqrt{d_\phi})$. For the policies, use an internal cover at radius $\min\{1,\varepsilon/(4C_\phi\sqrt{d_\phi})\}$. The cap at one keeps the radius in the range of \cref{asm:policy_class}. Each varying block then contributes at most $\varepsilon/4$. \cref{lem:euclidean_cover}, applied in the $d_\phi^2$-dimensional matrix space, bounds the matrix-cover size by $(1+8C_\phi^2d_\phi\sqrt K/\varepsilon)^{d_\phi^2}$. For each step, the first family is covered by pairs of representatives and the second by policy representatives alone. Hence, each step needs at most twice the product of the two cover sizes. Over $H$ steps, this contributes a factor $2H$. Taking logarithms gives~\cref{eq:off_feature_cover_explicit}. All cover representatives are taken from the defining parameter sets. Hence, they retain the norm bound $C_\phi\sqrt{d_\phi}$. Substituting \cref{asm:policy_class} into~\cref{eq:off_feature_cover_explicit} gives the stated logarithmic order.
\end{proof}

\paragraph{Value-Function Class}
For $h\in[H]$, $\pi\in\overline{\Pi}$, and $\boldsymbol{c}\in\mathbb{R}^{d_\psi}$, define $r_{\boldsymbol{c},h}^{\pi}\coloneq\nabla f_h(\mu_{\boldsymbol{c},h}^{\pi})$, where
\begin{align}
\mu_{\boldsymbol{c},h}^{\pi}(s,a)\coloneq\begin{cases}\rho(s)\pi_1(a\mid s),&h=1,\\
\operatorname{Clip}(\pi_h(a\mid s)\boldsymbol{c}^\top\boldsymbol{\psi}(s),[0,1]),&h\in\{2,\ldots,H\}.\end{cases}
\end{align} For a bonus radius $\alpha\geq0$ and the clipping interval $\mathcal{I}_h\coloneq[-(H-h+1)G,(H-h+1)G]$, let $\mathcal{V}_h(\alpha)$ consist of the functions
\begin{align}
V_{\boldsymbol{c},\boldsymbol{v},\boldsymbol{A},h}^{\pi}(s)&=\sum_a\pi_h(a\mid s)\operatorname{Clip}\bigl(r_{\boldsymbol{c},h}^{\pi}(s,a)+\boldsymbol{\phi}(s,a)^\top\boldsymbol{v}+\alpha\|\boldsymbol{\phi}(s,a)\|_{\boldsymbol{A}},\mathcal{I}_h\bigr), \label{eq:off_value_class}
\end{align}
where $\pi\in\overline{\Pi}$, $\|\boldsymbol{c}\|_2\leq C_{\mathbf{K}}C_\phi\sqrt{d_\phi KH}$, $\|\boldsymbol{v}\|_2\leq(H-h)G\sqrt K$, and $0\preceq\boldsymbol{A}\preceq I$. At $h=H$, take $\boldsymbol{v}=\boldsymbol{0}$ and $\alpha=0$.

\begin{lemma}[Cover of the Value Functions] \label{lem:off_covering_numbers_formal} \label{lem:off_covering_numbers}
For every $h\in[H]$, the class $\mathcal{V}_h(\alpha_h^{\mathrm{Off}})$ contains every step-$h$ value estimate produced by \OPE{} on replay data. For every $\alpha\geq0$ and $\varepsilon>0$, with $a_c\coloneq LC_\psi$ and $a_{\pi,h}\coloneq(H-h+1)G+L\max\{1,C_\psi C_{\mathbf{K}}C_\phi\sqrt{d_\phi KH}\}$,
\begin{align}
\log\mathcal{N}(\varepsilon,\mathcal{V}_h(\alpha),\|\cdot\|_\infty)&\leq d_\psi\log\left(1+\frac{8a_cC_{\mathbf{K}}C_\phi\sqrt{d_\phi KH}}{\varepsilon}\right)+d_\phi\log\left(1+\frac{8C_\phi\sqrt{d_\phi} (H-h)G\sqrt K}{\varepsilon}\right)\\
&\quad+d_\phi^2\log\left(1+\frac{32\alpha^2C_\phi^2d_\phi^{3/2}}{\varepsilon^2}\right)+\log\mathcal{N}\left(\min\left\{1,\frac{\varepsilon}{4a_{\pi,h}}\right\},\overline{\Pi},\mathrm{dist}_\Pi\right)\eqcolon\mathcal{H}_h^{\mathrm{Off}}(\alpha,\varepsilon). \label{eq:off_value_cover_explicit}
\end{align}
Under \cref{asm:policy_class}, if $1/\varepsilon$ and $\alpha$ are bounded by a polynomial in the problem parameters, the right-hand side is $\widetilde{\mathcal{O}}(d_\phi^2+d_\psi+d_\Pi)$.
\end{lemma}

\begin{proof}
We first show that the class contains the value estimates. The step-$h$ value estimate is $V_{\boldsymbol{c},\boldsymbol{v},\boldsymbol{A},h}^{\pi}$ with $\pi=\pi_k\in\overline{\Pi}$, $\boldsymbol{c}=\mathbf{M}_k^\top\widehat{\boldsymbol{\omega}}_{k,h-1}$, $\boldsymbol{v}=\widehat{\boldsymbol{v}}_{k,h}$, $\boldsymbol{A}=\boldsymbol{\Gamma}_{k,h}^{-1}$, and $\alpha=\alpha_h^{\mathrm{Off}}$, as can be read off from the occupancy reconstruction in \cref{alg:rfe} and from \cref{alg:ope}. These parameters satisfy the constraints of the class. The argument for~\cref{eq:core_frobenius_bound}, applied to the $kH\leq KH$ regression targets on replay data, shows that $\|\mathbf{M}_k\|_F\leq C_{\mathbf{K}}\sqrt{KH}$. Since $\widehat{\boldsymbol{\omega}}_{k,h-1}$ is an average of values of $\widehat{\boldsymbol{g}}_{h-1,1}^{\pi_k}$, its norm is at most $C_\phi\sqrt{d_\phi}$, and $\|\boldsymbol{c}\|_2\leq\|\mathbf{M}_k\|_F\|\widehat{\boldsymbol{\omega}}_{k,h-1}\|_2\leq C_{\mathbf{K}}C_\phi\sqrt{d_\phi KH}$. Comparing the objective in~\cref{eq:ope_ridge_objective} with its value at zero and using $|\widehat{V}_{k,h+1}|\leq(H-h)G$, we obtain $\|\widehat{\boldsymbol{v}}_{k,h}\|_2\leq(H-h)G\sqrt k\leq (H-h)G\sqrt K$. Finally, $\boldsymbol{\Gamma}_{k,h}\succeq I$ implies $0\preceq\boldsymbol{\Gamma}_{k,h}^{-1}\preceq I$.

We next show that the value functions are Lipschitz in their parameters. Let $r\coloneq\mathrm{dist}_\Pi(\pi,\pi')$. For $h\geq2$, the occupancy coordinates used in the two pseudo-rewards differ by at most $C_\psi(\|\boldsymbol{c}-\boldsymbol{c}'\|_2+C_{\mathbf{K}}C_\phi\sqrt{d_\phi KH} r)$. Indeed, before clipping, add and subtract $\pi_h(a\mid s)\boldsymbol{c}'^\top\boldsymbol{\psi}(s)$. The first difference is at most $\pi_h(a\mid s)|(\boldsymbol{c}-\boldsymbol{c}')^\top\boldsymbol{\psi}(s)|\leq C_\psi\|\boldsymbol{c}-\boldsymbol{c}'\|_2$, and the second at most $|\pi_h(a\mid s)-\pi_h'(a\mid s)|\,|\boldsymbol{c}'^\top\boldsymbol{\psi}(s)|\leq C_\psi C_{\mathbf{K}}C_\phi\sqrt{d_\phi KH}\,r$, using $\|\boldsymbol{\psi}(s)\|_2\leq C_\psi$. Clipping cannot increase their sum. For $h=1$, the occupancy is $\rho(s)\pi_1(a\mid s)$. Hence, its coordinates differ by at most $r$, since $\rho(s)\leq1$. The argument behind~\cref{eq:on_pseudo_reward_cover}, applied to the queried coordinate $(s,a)$, therefore bounds the pseudo-reward difference by $a_c\|\boldsymbol{c}-\boldsymbol{c}'\|_2+L\max\{1,C_\psi C_{\mathbf{K}}C_\phi\sqrt{d_\phi KH}\}r$. The linear and bonus parts are bounded as in Terms (II) and (III) of \cref{lem:on_value_covering_formal}. For the bonus, $|\sqrt{\boldsymbol{\phi}^\top\boldsymbol{A}\boldsymbol{\phi}}-\sqrt{\boldsymbol{\phi}^\top\boldsymbol{A}'\boldsymbol{\phi}}|\leq\|\boldsymbol{\phi}\|_2\|\boldsymbol{A}-\boldsymbol{A}'\|_{\mathrm{op}}^{1/2}\leq C_\phi\sqrt{d_\phi}\|\boldsymbol{A}-\boldsymbol{A}'\|_F^{1/2}$ directly. Finally, the outer policy average contributes at most $(H-h+1)Gr$, because $|\sum_a\pi_h(a\mid s)x_a-\sum_a\pi_h'(a\mid s)x_a'|\leq\max_a|x_a-x_a'|+\sum_a|\pi_h(a\mid s)-\pi_h'(a\mid s)||x_a'|$ and every clipped action value has absolute value at most $(H-h+1)G$. Thus
\begin{align}
\|V_{\boldsymbol{c},\boldsymbol{v},\boldsymbol{A},h}^{\pi}-V_{\boldsymbol{c}',\boldsymbol{v}',\boldsymbol{A}',h}^{\pi'}\|_\infty&\leq a_c\|\boldsymbol{c}-\boldsymbol{c}'\|_2+C_\phi\sqrt{d_\phi}\|\boldsymbol{v}-\boldsymbol{v}'\|_2+\alpha C_\phi\sqrt{d_\phi}\|\boldsymbol{A}-\boldsymbol{A}'\|_F^{1/2}+a_{\pi,h}r. \label{eq:off_value_cover_lipschitz}
\end{align}

Finally, we construct the cover. Allocate accuracy $\varepsilon/4$ to each parameter block. More precisely, use Euclidean radii $\varepsilon/(4a_c)$ and $\varepsilon/(4C_\phi\sqrt{d_\phi})$ for the two vectors, Frobenius radius $\varepsilon^2/(16\alpha^2C_\phi^2d_\phi)$ for $\boldsymbol{A}$, and policy radius $\min\{1,\varepsilon/(4a_{\pi,h})\}$. The matrix radius is squared because its term in~\cref{eq:off_value_cover_lipschitz} is a square root. The parameter balls have radii $C_{\mathbf{K}}C_\phi\sqrt{d_\phi KH}$, $(H-h)G\sqrt K$, and $\sqrt{d_\phi}$, respectively, the last because every eigenvalue of an admissible $\boldsymbol{A}$ is at most one. Apply \cref{lem:euclidean_cover} to each. For the matrix block, using dimension $d_\phi^2$ only enlarges the symmetric-matrix bound. Taking products and then logarithms gives~\cref{eq:off_value_cover_explicit}. Zero coefficients have no covering cost. Substituting \cref{asm:policy_class} into~\cref{eq:off_value_cover_explicit} proves the leading dimension $d_\phi^2+d_\psi+d_\Pi$, and all remaining dependence is logarithmic under the stated polynomial bounds. The construction uses no property of a particular policy optimization algorithm.
\end{proof}

\subsection{Theoretical Guarantees} \label{app:off_theoretical_guarantees}

\subsubsection{The Good Event} \label{app:off_good_events}

Throughout this appendix, $\widehat{\ell}_{k,h}=\mathbb{E}_{(s,a)\sim\mu_h^{\pi_k}}\|\boldsymbol{\phi}(s,a)\|_{\boldsymbol{\Gamma}_{k-1,h}^{-1}}$, where $\boldsymbol{\Gamma}_{0,h}=I$. The matrix and policy are fixed before trajectory $k$ is collected. Hence, $\widehat{\ell}_{k,h}$ is determined by the history before trajectory $k$, although it is not evaluated by the learner.

The fixed class $\mathcal{W}$ of feature-regression targets is defined in~\cref{app:off_covering_arguments}. For a candidate target $\boldsymbol{g}\in\mathcal{W}$, define $\widehat{\mathbf{B}}_{k,h}(\boldsymbol{g})\coloneq\boldsymbol{\Gamma}_{k,h}^{-1}\sum_{i=1}^k\boldsymbol{\phi}_{i,h}\boldsymbol{g}(s_{i,h+1})^\top$ for its ridge fit, and recall $\mathbf{B}^\star(\boldsymbol{g})$ from \cref{lem:bilinear_bfr_linearity} and $\boldsymbol{v}^\star(V)$ from \cref{lem:bilinear_next_state_values}.

We use the good event $\mathcal{E}=\mathcal{E}_{\mathrm{P}}\cap\mathcal{E}_{\mathrm{R}}\cap\mathcal{E}_{\mathrm{V}}$ in~\cref{def:good_event} with the radii $\beta_k^{\mathrm{Off}}$, $\Delta_{k,h}^{\mathrm{Off}}$, and $\alpha_h^{\mathrm{Off}}$. The radii $\beta_k^{\mathrm{Off}}$, $(\kappa,\gamma_\rho)$, and $\alpha_h^{\mathrm{Off}}$ are chosen in \cref{lem:off_model_event_main,lem:off_reward_event_main,lem:off_value_event_main}, respectively, and $\Delta_{k,h}^{\mathrm{Off}}$ is defined in~\cref{lem:off_reward_error_main}. \cref{lem:off_model_event_main} shows that $\mathcal{E}_{\mathrm{P}}$ holds with probability at least $1-\delta/4$, \cref{lem:off_reward_error_main} shows that, under the event $\mathcal{E}_{\mathrm{P}}$, $\mathcal{E}_{\mathrm{R}}$ holds with probability at least $1-\delta/4$, and \cref{lem:off_value_event_main} shows that $\mathcal{E}_{\mathrm{V}}$ holds with probability at least $1-\delta/4$. Only the value radii $\alpha_h^{\mathrm{Off}}$ are used by the algorithm. The other radii are used only in the analysis. These failure probabilities are combined with that of~\cref{lem:off_martingale} in the proof of \cref{thm:off_policy}.

\subsubsection{Transition Model Estimation} \label{app:off_tme_guarantees}

\begin{lemma}[Transition-Core Confidence] \label{lem:off_model_confidence} \label{lem:off_model_event_main}
If we choose $\beta_k^{\mathrm{Off}}=\mathcal{O}(d_\phi\log(KH/\delta))$, $k\in[K]$, with a sufficiently large constant, then $\mathcal{E}_{\mathrm{P}}$ holds with probability at least $1-\delta/4$.
\end{lemma}

\begin{proof}
We first recall the relevant notation. In the off-policy setting, \BTE{} computes $\mathbf{M}_k$ from the replay dataset $\mathcal{D}_k$ of $k$ trajectories, with the Gram matrix $\boldsymbol{\Lambda}_k=I+\sum_{i=1}^k\sum_{h=1}^H\boldsymbol{\phi}_{i,h}\boldsymbol{\phi}_{i,h}^\top$. The event $\mathcal{E}_{\mathrm{P}}$ requires $\|\boldsymbol{\Lambda}_k^{1/2}(\mathbf{M}_k-\mathbf{M}^\star)\|_F\leq\sqrt{\beta_k^{\mathrm{Off}}}$ for every $k\in[K]$.

We take, for a sufficiently large numerical constant $c$,
\begin{align}
\beta_k^{\mathrm{Off}}&=c\bigl\{C_{\mathbf{M}}^2d_\phi+C_{\mathbf{K}}^2[d_\phi\log(1+kHC_\phi^2)+\log(4K/\delta)]\bigr\}, \label{eq:off_beta}
\end{align}
which has the stated order because $k\leq K$. We also take $c\geq1$, so that $\sqrt{\beta_k^{\mathrm{Off}}}\geq C_{\mathbf{M}}\sqrt{d_\phi}$. The argument parallels \cref{lem:good_estimation_event_P}, with the adaptively collected replay data in place of a fresh batch.

We first set up the conditional regression model. Order the transitions lexicographically by episode and step. For a transition $(s_{i,h},a_{i,h},s_{i,h+1})$, set $\boldsymbol{x}_{i,h}\coloneq\boldsymbol{\phi}(s_{i,h},a_{i,h})$ and $\boldsymbol{y}_{i,h}\coloneq\mathbf{K}_\psi^{-1}\boldsymbol{\psi}(s_{i,h+1})$. The covariate $\boldsymbol{x}_{i,h}$ is measurable before $s_{i,h+1}$ is sampled, and, conditional on the history through $(s_{i,h},a_{i,h})$, the successor has law $P^\star(\cdot\mid s_{i,h},a_{i,h})$ whatever behavior policy generated the current pair. Transposing the identity of \cref{lem:bilinear_regression_response}, and using that $\mathbf{K}_\psi$ is symmetric, gives
\begin{align}
\mathbb{E}[\boldsymbol{y}_{i,h}\mid s_{i,h},a_{i,h}]&=(\mathbf{M}^\star)^\top\boldsymbol{x}_{i,h}.
\end{align}
Thus $\boldsymbol{\varepsilon}_{i,h}\coloneq\boldsymbol{y}_{i,h}-(\mathbf{M}^\star)^\top\boldsymbol{x}_{i,h}$ is a conditionally mean-zero vector. Condition (iii) of \cref{def:bilienar_mdp} gives $\|\boldsymbol{y}_{i,h}\|_2\leq C_{\mathbf{K}}$. By Jensen's inequality its conditional mean obeys the same bound. Hence, $\|\boldsymbol{\varepsilon}_{i,h}\|_2\leq2C_{\mathbf{K}}$.

We next bound the noise and regularization errors. The normal equations of the ridge regression (\cref{app:on_tme_details}) give $\boldsymbol{\Lambda}_k\mathbf{M}_k=\sum_{i=1}^k\sum_{h=1}^H\boldsymbol{x}_{i,h}\boldsymbol{y}_{i,h}^\top$, with $\boldsymbol{\Lambda}_k=I+\sum_{i\leq k,h}\boldsymbol{x}_{i,h}\boldsymbol{x}_{i,h}^\top$. Substituting $\boldsymbol{y}_{i,h}=(\mathbf{M}^\star)^\top\boldsymbol{x}_{i,h}+\boldsymbol{\varepsilon}_{i,h}$ gives $\boldsymbol{\Lambda}_k\mathbf{M}_k=(\boldsymbol{\Lambda}_k-I)\mathbf{M}^\star+\sum_{i\leq k,h}\boldsymbol{x}_{i,h}\boldsymbol{\varepsilon}_{i,h}^\top$. Multiplying by $\boldsymbol{\Lambda}_k^{-1}$ and subtracting $\mathbf{M}^\star$ gives the ridge decomposition
\begin{align}
\mathbf{M}_k-\mathbf{M}^\star =\boldsymbol{\Lambda}_k^{-1}\sum_{i=1}^k\sum_{h=1}^H\boldsymbol{x}_{i,h}\boldsymbol{\varepsilon}_{i,h}^\top -\boldsymbol{\Lambda}_k^{-1}\mathbf{M}^\star.
\end{align}
Multiplying by $\boldsymbol{\Lambda}_k^{1/2}$ and applying the triangle inequality for the Frobenius norm yields
\begin{align}
\bigl\lVert \boldsymbol{\Lambda}_k^{1/2}(\mathbf{M}_k-\mathbf{M}^\star) \bigr\rVert_F &\leq\biggl\lVert \boldsymbol{\Lambda}_k^{-1/2}\sum_{i=1}^k\sum_{h=1}^H\boldsymbol{x}_{i,h}\boldsymbol{\varepsilon}_{i,h}^\top \biggr\rVert_F +\bigl\lVert \boldsymbol{\Lambda}_k^{-1/2}\mathbf{M}^\star \bigr\rVert_F.
\end{align}
The second term is at most $\|\mathbf{M}^\star\|_F\leq C_{\mathbf{M}}\sqrt{d_\phi}$ because $\boldsymbol{\Lambda}_k\succeq I$. For the first term, apply \cref{lem:lin_matrix_self_normalized} with $n=kH$ observations, covariate radius $R_x=C_\phi\sqrt{d_\phi}$, noise radius $R=2C_{\mathbf{K}}$, and failure probability $\delta/(4K)$. Adaptivity causes no difficulty because every covariate is determined before its successor is drawn and every residual is conditionally centered. Since $nR_x^2/d_\phi=kHC_\phi^2$, for each prefix $k$, outside an event of probability $\delta/(4K)$,
\begin{align}
\bigl\lVert \boldsymbol{\Lambda}_k^{1/2}(\mathbf{M}_k-\mathbf{M}^\star) \bigr\rVert_F\leq2C_{\mathbf{K}}\sqrt{d_\phi\log(1+kHC_\phi^2)+2\log(4K/\delta)}+C_{\mathbf{M}}\sqrt{d_\phi}.
\end{align}
Squaring and applying $(a+b)^2\leq2a^2+2b^2$ shows that a numerical $c$ in~\cref{eq:off_beta} dominates the square of the right-hand side. A union bound over $k\in[K]$ shows that $\mathcal{E}_{\mathrm{P}}$ holds with probability at least $1-\delta/4$. The argument uses $kH$ adaptively collected transitions and their Gram matrix. No lower bound on the buffer size or fresh batch of $N$ trajectories is required for this self-normalized statement.
\end{proof}

\subsubsection{Pseudo-Reward Estimation} \label{app:off_pre_guarantees}

The feature-regression targets range over the fixed class $\mathcal{W}$ in~\cref{eq:off_feature_target_class}. By \cref{lem:off_feature_cover}, it contains every target produced by \BFR{}, and its members have uniform Euclidean norm at most $C_\phi\sqrt{d_\phi}$. The main result of this subsubsection is the following lemma, which establishes the pseudo-reward part of the good event. Its proof uses the backward-regression and initial-state confidence bounds in \cref{lem:off_reward_event_main}.

\begin{lemma}[Pseudo-Reward Error] \label{lem:off_reward_error_main} \label{lem:off_reward_error_formal}
Let $\beta_k^{\mathrm{Off}}$ be as in \cref{lem:off_model_event_main}, and let $\kappa$ and $\gamma_\rho$ be as in \cref{lem:off_reward_event_main}. Under the event $\mathcal{E}_{\mathrm{P}}$, with probability at least $1-\delta/4$, for every $k\in[K]$ and $h\in[H]$,
\begin{align}
\MoveEqLeft\|\widehat{R}_{k,h}-R_{k,h}\|_\infty\leq\Delta_{k,h}^{\mathrm{Off}}\nonumber\\
&\coloneq\begin{cases}0,&h=1,\\[2pt] LC_\psi\Biggl[\sqrt{\beta_k^{\mathrm{Off}}}\,\widehat{\ell}_{k,h-1}+(\sqrt{\beta_k^{\mathrm{Off}}}+C_{\mathbf{M}}\sqrt{d_\phi})\biggl(\dfrac{\gamma_\rho}{\sqrt k}+\kappa\displaystyle\sum_{t=1}^{h-2}\widehat{\ell}_{k,t}\biggr)\Biggr],&h\in\{2,\ldots,H\}.\end{cases} \label{eq:Delta_off}
\end{align}
\end{lemma}

\begin{proof}[Proof of \cref{lem:off_reward_error_main}]
We first recall the relevant notation. For $h\geq2$, $\OCUPOOFF$ estimates the occupancy measure by $\widehat{\mu}_{k,h}(s,a)=\operatorname{Clip}(\pi_{k,h}(a\mid s)\widehat{\boldsymbol{\omega}}_{k,h-1}^\top\mathbf{M}_k\boldsymbol{\psi}(s),[0,1])$, where $\widehat{\boldsymbol{\omega}}_{k,h-1}$ is returned by \BFR{} (\cref{alg:rfe}). For a target $\boldsymbol{g}\in\mathcal{W}$, $\widehat{\mathbf{B}}_{k,h}(\boldsymbol{g})$ denotes the ridge fit of the backward regression and $\mathbf{B}^\star(\boldsymbol{g})$ its population counterpart in~\cref{lem:bilinear_bfr_linearity}. The expected feature norm $\widehat{\ell}_{k,t}=\mathbb{E}_{(s,a)\sim\mu_t^{\pi_k}}\|\boldsymbol{\phi}(s,a)\|_{\boldsymbol{\Gamma}_{k-1,t}^{-1}}$ is defined in~\cref{subsec:off_policy_proof_outline}.

By \cref{lem:off_reward_event_main}, with probability at least $1-\delta/4$, for every $k\in[K]$, $h\in[H-1]$, and $\boldsymbol{g}\in\mathcal{W}$,
\begin{align}
\bigl\lVert \boldsymbol{\Gamma}_{k,h}^{1/2}(\widehat{\mathbf{B}}_{k,h}(\boldsymbol{g})-\mathbf{B}^\star(\boldsymbol{g})) \bigr\rVert_F&\leq\kappa, \label{eq:off_bfr_regression_confidence}\\
\bigl\lVert k^{-1}\textstyle\sum_{i=1}^k\boldsymbol{g}(s_{i,1})-\mathbb{E}_\rho[\boldsymbol{g}(s_1)] \bigr\rVert_2&\leq\gamma_\rho/\sqrt k. \label{eq:off_bfr_initial_confidence}
\end{align}
We work on the intersection of this event with $\mathcal{E}_{\mathrm{P}}$ and consider a specific $k\in[K]$. At $h=1$, the true and estimated occupancies are both $\widehat{\mu}_{k,1}(s,a)=\mu_1^{\pi_k}(s,a)=\rho(s)\pi_{k,1}(a\mid s)$. Therefore $\widehat{R}_{k,1}=R_{k,1}$, matching $\Delta_{k,1}^{\mathrm{Off}}=0$. Now consider a specific $h\in\{2,\ldots,H\}$.

\paragraph{Splitting the Occupancy Error}
Both $\widehat{\mu}_{k,h}$ and $\mu_h^{\pi_k}$ lie in the utility domain. Hence, $L$-smoothness of $f_h$ (\cref{asm:lipschitzness_and_smoothness}) gives $\|\widehat{R}_{k,h}-R_{k,h}\|_\infty\leq L\|\widehat{\mu}_{k,h}-\mu_h^{\pi_k}\|_1$. Coordinatewise clipping cannot increase the distance to the true coordinate, because every true occupancy coordinate lies in $[0,1]$. By \cref{lem:occupancy_measure}, $\mu_h^{\pi_k}(s,a)=\pi_{k,h}(a\mid s)(\boldsymbol{\omega}_{h-1}^{\pi_k})^\top\mathbf{M}^\star\boldsymbol{\psi}(s)$. Therefore
\begin{align}
\bigl\lVert \widehat{\mu}_{k,h}-\mu_h^{\pi_k} \bigr\rVert_1 &\leq\sum_{s\in\mathcal{S}}\sum_{a\in\mathcal{A}}\pi_{k,h}(a\mid s) \bigl|\widehat{\boldsymbol{\omega}}_{k,h-1}^\top\mathbf{M}_k\boldsymbol{\psi}(s) -(\boldsymbol{\omega}_{h-1}^{\pi_k})^\top\mathbf{M}^\star\boldsymbol{\psi}(s)\bigr|\\
&\leq\underbrace{\sum_{s\in\mathcal{S}}\bigl|\widehat{\boldsymbol{\omega}}_{k,h-1}^\top(\mathbf{M}_k-\mathbf{M}^\star)\boldsymbol{\psi}(s)\bigr|}_{\text{(I)}}+\underbrace{\sum_{s\in\mathcal{S}}\bigl|(\widehat{\boldsymbol{\omega}}_{k,h-1}-\boldsymbol{\omega}_{h-1}^{\pi_k})^\top\mathbf{M}^\star\boldsymbol{\psi}(s)\bigr|}_{\text{(II)}}. \label{eq:off_occupancy_split}
\end{align}
The second inequality uses $\sum_a\pi_{k,h}(a\mid s)=1$ and adds and subtracts $\widehat{\boldsymbol{\omega}}_{k,h-1}^\top\mathbf{M}^\star\boldsymbol{\psi}(s)$. Both terms are bounded through the following consequence of condition (iii) of \cref{def:bilienar_mdp}. For any $\boldsymbol{v}\in\mathbb{R}^{d_\psi}$, choosing the sign function $g(s)=\operatorname{sign}(\boldsymbol{v}^\top\boldsymbol{\psi}(s))$ and applying the Cauchy--Schwarz inequality to $\boldsymbol{v}^\top\sum_s\boldsymbol{\psi}(s)g(s)$ gives $\sum_s|\boldsymbol{v}^\top\boldsymbol{\psi}(s)|\leq C_\psi\|\boldsymbol{v}\|_2$.

\paragraph{Bounding Term (I)}
We apply this bound with the vector $\boldsymbol{v}=(\mathbf{M}_k-\mathbf{M}^\star)^\top\widehat{\boldsymbol{\omega}}_{k,h-1}$. The Cauchy--Schwarz inequality and $\mathcal{E}_{\mathrm{P}}$ then give
\begin{align}
\text{Term (I)}\leq C_\psi\bigl\lVert (\mathbf{M}_k-\mathbf{M}^\star)^\top\widehat{\boldsymbol{\omega}}_{k,h-1} \bigr\rVert_2\leq C_\psi\sqrt{\beta_k^{\mathrm{Off}}}\bigl\lVert \widehat{\boldsymbol{\omega}}_{k,h-1} \bigr\rVert_{\boldsymbol{\Lambda}_k^{-1}}.
\end{align}
By the triangle inequality and $\boldsymbol{\Lambda}_k\succeq I$, $\|\widehat{\boldsymbol{\omega}}_{k,h-1}\|_{\boldsymbol{\Lambda}_k^{-1}}\leq\|\boldsymbol{\omega}_{h-1}^{\pi_k}\|_{\boldsymbol{\Lambda}_k^{-1}}+\|\widehat{\boldsymbol{\omega}}_{k,h-1}-\boldsymbol{\omega}_{h-1}^{\pi_k}\|_2$. For the first norm, Jensen's inequality gives
\begin{align}
\bigl\lVert \boldsymbol{\omega}_{h-1}^{\pi_k} \bigr\rVert_{\boldsymbol{\Lambda}_k^{-1}} =\bigl\lVert \mathbb{E}_{\mu_{h-1}^{\pi_k}}[\boldsymbol{\phi}(s,a)] \bigr\rVert_{\boldsymbol{\Lambda}_k^{-1}}\leq\mathbb{E}_{\mu_{h-1}^{\pi_k}}\Bigl[ \bigl\lVert \boldsymbol{\phi}(s,a) \bigr\rVert_{\boldsymbol{\Lambda}_k^{-1}} \Bigr] \leq\widehat{\ell}_{k,h-1}.
\end{align}
The final inequality uses $\boldsymbol{\Lambda}_k\succeq\boldsymbol{\Gamma}_{k-1,h-1}$: the pooled model Gram matrix includes every earlier step-$(h-1)$ observation. Matrix inversion reverses this order. Hence, $\|\boldsymbol{\phi}(s,a)\|_{\boldsymbol{\Lambda}_k^{-1}}\leq\|\boldsymbol{\phi}(s,a)\|_{\boldsymbol{\Gamma}_{k-1,h-1}^{-1}}$. Hence
\begin{align}
\text{Term (I)}\leq C_\psi\sqrt{\beta_k^{\mathrm{Off}}}\Bigl(\widehat{\ell}_{k,h-1}+\bigl\lVert \widehat{\boldsymbol{\omega}}_{k,h-1}-\boldsymbol{\omega}_{h-1}^{\pi_k} \bigr\rVert_2\Bigr). \label{eq:off_term_I_bound}
\end{align}

\paragraph{Bounding Term (II)}
We apply the same bound with the vector $\boldsymbol{v}=(\mathbf{M}^\star)^\top(\widehat{\boldsymbol{\omega}}_{k,h-1}-\boldsymbol{\omega}_{h-1}^{\pi_k})$. Since $\|\mathbf{M}^\star\|_{\mathrm{op}}\leq\|\mathbf{M}^\star\|_F\leq C_{\mathbf{M}}\sqrt{d_\phi}$, we obtain
\begin{align}
\text{Term (II)}\leq C_{\mathbf{M}}C_\psi\sqrt{d_\phi}\,\bigl\lVert \widehat{\boldsymbol{\omega}}_{k,h-1}-\boldsymbol{\omega}_{h-1}^{\pi_k} \bigr\rVert_2. \label{eq:off_term_II_bound}
\end{align}

\paragraph{Bounding the Feature Estimation Error}
Both terms are controlled by the feature estimation error $\|\widehat{\boldsymbol{\omega}}_{k,j}-\boldsymbol{\omega}_j^{\pi_k}\|_2$ with $j=h-1$. We bound it for a fixed target step $j<H$. Let $\boldsymbol{g}_{j,t}^{\pi_k}$ denote the true expected step-$j$ feature vectors defined in~\cref{eq:off_population_feature_base,eq:off_population_feature_recursion}. By construction, $\widehat{\boldsymbol{g}}_{j,j}^{\pi_k}(s)=\boldsymbol{g}_{j,j}^{\pi_k}(s)=\overline{\boldsymbol{\phi}}_{\pi_k,j}(s)$. Hence, the feature estimation error is zero at step $j$. For every $t<j$, define the one-step fitted residual and the feature estimation error by
\begin{align}
\boldsymbol{\chi}_{j,t}^{\pi_k}(s,a) \coloneq\widehat{\boldsymbol{q}}_{j,t}^{\pi_k}(s,a)-P^\star\widehat{\boldsymbol{g}}_{j,t+1}^{\pi_k}(s,a),\qquad\boldsymbol{\delta}_{j,t}^{\pi_k}(s) \coloneq\widehat{\boldsymbol{g}}_{j,t}^{\pi_k}(s)-\boldsymbol{g}_{j,t}^{\pi_k}(s),
\end{align}
with $\boldsymbol{\delta}_{j,j}^{\pi_k}=\boldsymbol{0}$. By \cref{lem:bilinear_bfr_linearity}, $P^\star\widehat{\boldsymbol{g}}_{j,t+1}^{\pi_k}(s,a)=\mathbf{B}^\star(\widehat{\boldsymbol{g}}_{j,t+1}^{\pi_k})^\top\boldsymbol{\phi}(s,a)$. This is a deterministic identity for every function. Hence, it applies to the fitted target. This vector belongs to $\mathbb{B}_2(C_\phi\sqrt{d_\phi})$, as shown in \cref{app:off_bfr_details}, and Euclidean projection onto this ball is nonexpansive and fixes every point in the ball. Moreover, $\widehat{\mathbf{B}}_{j,t}^{\pi_k}=\widehat{\mathbf{B}}_{k,t}(\widehat{\boldsymbol{g}}_{j,t+1}^{\pi_k})$ with $\widehat{\boldsymbol{g}}_{j,t+1}^{\pi_k}\in\mathcal{W}$. Therefore,
\begin{align}
\bigl\lVert \boldsymbol{\chi}_{j,t}^{\pi_k}(s,a) \bigr\rVert_2 &\leq\bigl\lVert (\widehat{\mathbf{B}}_{j,t}^{\pi_k}-\mathbf{B}^\star(\widehat{\boldsymbol{g}}_{j,t+1}^{\pi_k}))^\top\boldsymbol{\phi}(s,a) \bigr\rVert_2\\
&\leq\bigl\lVert \boldsymbol{\Gamma}_{k,t}^{1/2}(\widehat{\mathbf{B}}_{j,t}^{\pi_k}-\mathbf{B}^\star(\widehat{\boldsymbol{g}}_{j,t+1}^{\pi_k})) \bigr\rVert_F \bigl\lVert \boldsymbol{\phi}(s,a) \bigr\rVert_{\boldsymbol{\Gamma}_{k,t}^{-1}}\\
&\leq\kappa\bigl\lVert \boldsymbol{\phi}(s,a) \bigr\rVert_{\boldsymbol{\Gamma}_{k-1,t}^{-1}}. \label{eq:off_forward_bellman_residual}
\end{align}
The first inequality is the projection property. The second is the Cauchy--Schwarz inequality. The third uses~\cref{eq:off_bfr_regression_confidence} and $\boldsymbol{\Gamma}_{k,t}\succeq\boldsymbol{\Gamma}_{k-1,t}$, which implies $\boldsymbol{\Gamma}_{k,t}^{-1}\preceq\boldsymbol{\Gamma}_{k-1,t}^{-1}$.

From the definition $\widehat{\boldsymbol{g}}_{j,t}^{\pi_k}(s)=\sum_a\pi_{k,t}(a\mid s)\widehat{\boldsymbol{q}}_{j,t}^{\pi_k}(s,a)$ in \cref{alg:rfe} and the recursion~\cref{eq:off_population_feature_recursion}, adding and subtracting $P^\star\widehat{\boldsymbol{g}}_{j,t+1}^{\pi_k}(s,a)$ gives
\begin{align}
\boldsymbol{\delta}_{j,t}^{\pi_k}(s) &=\sum_a\pi_{k,t}(a\mid s) \bigl[\boldsymbol{\chi}_{j,t}^{\pi_k}(s,a)+P^\star\boldsymbol{\delta}_{j,t+1}^{\pi_k}(s,a)\bigr].
\end{align}
For this fixed data realization the fitted residuals are ordinary functions. Averaging over $s\sim d_t^{\pi_k}$ turns the policy average into an expectation under $\mu_t^{\pi_k}$, and the occupancy flow identity gives $\mathbb{E}_{\mu_t^{\pi_k}}P^\star\boldsymbol{\delta}_{j,t+1}^{\pi_k}=\mathbb{E}_{d_{t+1}^{\pi_k}}\boldsymbol{\delta}_{j,t+1}^{\pi_k}$. Taking norms and using the triangle inequality and Jensen's inequality gives
\begin{align}
\bigl\lVert \mathbb{E}_{d_t^{\pi_k}}[\boldsymbol{\delta}_{j,t}^{\pi_k}(s_t)] \bigr\rVert_2 &\leq\mathbb{E}_{\mu_t^{\pi_k}}\Bigl[ \bigl\lVert \boldsymbol{\chi}_{j,t}^{\pi_k}(s_t,a_t) \bigr\rVert_2 \Bigr] +\bigl\lVert \mathbb{E}_{d_{t+1}^{\pi_k}}[\boldsymbol{\delta}_{j,t+1}^{\pi_k}(s_{t+1})] \bigr\rVert_2.
\end{align}
Iterating this inequality from $t=1$ to $j-1$, and using $\boldsymbol{\delta}_{j,j}^{\pi_k}=\boldsymbol{0}$, $d_1^{\pi_k}=\rho$, and $\mathbb{E}_\rho[\boldsymbol{g}_{j,1}^{\pi_k}(s_1)]=\boldsymbol{\omega}_j^{\pi_k}$ from \cref{lem:backward_feature_recursion}, yields
\begin{align}
\bigl\lVert \mathbb{E}_\rho[\widehat{\boldsymbol{g}}_{j,1}^{\pi_k}(s_1)]-\boldsymbol{\omega}_j^{\pi_k} \bigr\rVert_2 \leq\sum_{t=1}^{j-1}\mathbb{E}_{\mu_t^{\pi_k}}\Bigl[ \bigl\lVert \boldsymbol{\chi}_{j,t}^{\pi_k}(s,a) \bigr\rVert_2 \Bigr]\leq\kappa\sum_{t=1}^{j-1}\widehat{\ell}_{k,t}. \label{eq:off_feature_telescope}
\end{align}
The last inequality averages~\cref{eq:off_forward_bellman_residual} under $\mu_t^{\pi_k}$ and uses the definition of $\widehat{\ell}_{k,t}$.

\cref{eq:off_bfr_initial_confidence} applies to $\widehat{\boldsymbol{g}}_{j,1}^{\pi_k}\in\mathcal{W}$. Since $\widehat{\boldsymbol{\omega}}_{k,j}=k^{-1}\sum_{i=1}^k\widehat{\boldsymbol{g}}_{j,1}^{\pi_k}(s_{i,1})$, adding and subtracting $\mathbb{E}_\rho[\widehat{\boldsymbol{g}}_{j,1}^{\pi_k}(s_1)]$, applying the triangle inequality, and using~\cref{eq:off_bfr_initial_confidence,eq:off_feature_telescope} gives
\begin{align}
\bigl\lVert \widehat{\boldsymbol{\omega}}_{k,j}-\boldsymbol{\omega}_j^{\pi_k} \bigr\rVert_2 &\leq\frac{\gamma_\rho}{\sqrt k}+\kappa\sum_{t=1}^{j-1}\widehat{\ell}_{k,t}. \label{eq:off_feature_error_internal}
\end{align}

\paragraph{Putting Everything Together}
Substituting~\cref{eq:off_term_I_bound,eq:off_term_II_bound} into~\cref{eq:off_occupancy_split}, applying~\cref{eq:off_feature_error_internal} with $j=h-1$, and multiplying by $L$ gives
\begin{align}
\|\widehat{R}_{k,h}-R_{k,h}\|_\infty\leq LC_\psi\Biggl[\sqrt{\beta_k^{\mathrm{Off}}}\,\widehat{\ell}_{k,h-1}+(\sqrt{\beta_k^{\mathrm{Off}}}+C_{\mathbf{M}}\sqrt{d_\phi})\biggl(\frac{\gamma_\rho}{\sqrt k}+\kappa\sum_{t=1}^{h-2}\widehat{\ell}_{k,t}\biggr)\Biggr]=\Delta_{k,h}^{\mathrm{Off}}.
\end{align}
The expectation defining $\widehat{\ell}_{k,t}$ is used only in the analysis and is not evaluated by the algorithm.
\end{proof}

The following lemma establishes the backward-regression and initial-state confidence bounds used in the proof of \cref{lem:off_reward_error_main}.

\begin{lemma}[Backward-Regression and Initial-State Confidence] \label{lem:off_reward_event_formal} \label{lem:off_reward_event_main}
If we choose
\begin{align}
\kappa=\mathcal{O}\Bigl(\sqrt{d_\phi(d_\phi^2+d_\Pi)\log(d_\phi KH/\delta)}\Bigr)\qquad\text{and}\qquad\gamma_\rho=\mathcal{O}\Bigl(\sqrt{d_\phi(d_\phi^2+d_\Pi)\log(d_\phi KH/\delta)}\Bigr)
\end{align}
with sufficiently large constants, then, with probability at least $1-\delta/4$,~\cref{eq:off_bfr_regression_confidence,eq:off_bfr_initial_confidence} hold for every $k\in[K]$, $h\in[H-1]$, and $\boldsymbol{g}\in\mathcal{W}$.
\end{lemma}

\begin{proof}[Proof of \cref{lem:off_reward_event_main}]
We first recall the relevant notation. For a target $\boldsymbol{g}\in\mathcal{W}$, the step-$h$ backward regression fits $\widehat{\mathbf{B}}_{k,h}(\boldsymbol{g})=\boldsymbol{\Gamma}_{k,h}^{-1}\sum_{i=1}^k\boldsymbol{\phi}_{i,h}\boldsymbol{g}(s_{i,h+1})^\top$ on the replay data, with $\boldsymbol{\Gamma}_{k,h}=I+\sum_{i=1}^k\boldsymbol{\phi}_{i,h}\boldsymbol{\phi}_{i,h}^\top$, and~\cref{lem:bilinear_bfr_linearity} gives $P^\star\boldsymbol{g}(s,a)=\mathbf{B}^\star(\boldsymbol{g})^\top\boldsymbol{\phi}(s,a)$. The initial-state average $k^{-1}\sum_{i=1}^k\boldsymbol{g}(s_{i,1})$ estimates $\mathbb{E}_\rho[\boldsymbol{g}(s_1)]$.

The event has two parts: accuracy of every backward regression, and accuracy of the initial-state average for every candidate feature-regression target. Both arguments use a fixed internal cover $\mathcal{C}_{\mathrm{W}}$ of $\mathcal{W}$ in $\|\cdot\|_{\infty,2}$ at radius $C_\phi\sqrt{d_\phi}/(K+1)^2$, which exists by \cref{lem:off_feature_cover} and is fixed before data collection. By~\cref{eq:off_feature_cover_explicit}, $\log|\mathcal{C}_{\mathrm{W}}|$ is at most the sum of the matrix term $d_\phi^2\log(1+8C_\phi\sqrt{d_\phi K}(K+1)^2)$ and the policy term, which \cref{asm:policy_class} bounds by a constant multiple of $d_\Pi\log(1+4K(K+1)^2)$. Hence, $\log|\mathcal{C}_{\mathrm{W}}|=\mathcal{O}((d_\phi^2+d_\Pi)\log(d_\phi KH))$. We take
\begin{align}
\kappa&\coloneq C_\phi\sqrt{d_\phi}\biggl[2\sqrt{d_\phi\log(1+KC_\phi^2)+2\bigl(\log|\mathcal{C}_{\mathrm{W}}|+\log(8KH/\delta)\bigr)}+2\biggr]+C_{\mathbf{M}}C_\psi C_\phi d_\phi^{3/2}, \label{eq:off_bfr_kappa}\\
\gamma_\rho&\coloneq4C_\phi\sqrt{d_\phi}\sqrt{\log(1+K)+2\bigl(\log|\mathcal{C}_{\mathrm{W}}|+\log(8K/\delta)\bigr)+1}. \label{eq:off_bfr_initial_radius}
\end{align}
Substituting the bound on $\log|\mathcal{C}_{\mathrm{W}}|$ and using $d_\phi^{3/2}\leq\sqrt{d_\phi(d_\phi^2+d_\Pi)}$ gives the stated orders. We give each part of the event failure probability $\delta/8$.

\paragraph{Uniform Backward Regression}
Consider a specific step $t<H$. Order the replay trajectories by episode, and let $\mathcal{F}_{i-1}$ contain the first $i-1$ trajectories together with the first $t$ steps of trajectory $i$. Then $\boldsymbol{\phi}_{i,t}$ is $\mathcal{F}_{i-1}$-measurable, and given $\mathcal{F}_{i-1}$, the next state $s_{i,t+1}$ has law $P^\star(\cdot\mid s_{i,t},a_{i,t})$. The class $\mathcal{W}$ is fixed before data collection, and each of its members has uniform Euclidean norm at most $C_\phi\sqrt{d_\phi}$ by \cref{lem:off_feature_cover}. Since $(K+1)^{-2}\leq K^{-1}$, $\mathcal{C}_{\mathrm{W}}$ is also an internal $(C_\phi\sqrt{d_\phi}/K)$-cover of $\mathcal{W}$. Consider a specific $k\in[K]$. We apply \cref{lem:lin_uniform_matrix_regression} with $\mathcal{F}=\mathcal{W}$, $R_g=R_x=C_\phi\sqrt{d_\phi}$, $\boldsymbol{x}_i=\boldsymbol{\phi}_{i,t}$, $Y_i=s_{i,t+1}$, $P_i=P^\star(\cdot\mid s_{i,t},a_{i,t})$, $E=\log|\mathcal{C}_{\mathrm{W}}|$, and $\zeta=\delta/(8KH)$ to obtain that, with probability at least $1-\delta/(8KH)$, for every $\boldsymbol{g}\in\mathcal{W}$ simultaneously,
\begin{align}
\MoveEqLeft\biggl\|\boldsymbol{\Gamma}_{k,t}^{-1/2}\sum_{i=1}^k\boldsymbol{\phi}_{i,t}[\boldsymbol{g}(s_{i,t+1})-P^\star\boldsymbol{g}(s_{i,t},a_{i,t})]^\top\biggr\|_F\nonumber\\*
&\leq C_\phi\sqrt{d_\phi}\biggl[2\sqrt{d_\phi\log(1+KC_\phi^2)+2\bigl(\log|\mathcal{C}_{\mathrm{W}}|+\log(8KH/\delta)\bigr)}+2\biggr], \label{eq:off_bfr_noise}
\end{align}
where we use $KR_x^2/d_\phi=KC_\phi^2$. We take a union bound over $k\in[K]$ and $t<H$ to show that~\cref{eq:off_bfr_noise} holds for all $k$ and $t$ with probability at least $1-\delta/8$. Since \cref{lem:lin_uniform_matrix_regression} holds simultaneously over $\mathcal{W}$, this bound also applies to targets selected after observing the replay data.

For each coordinate $j$, the corresponding column of $\mathbf{B}^\star(\boldsymbol{g})$ is $\mathbf{M}^\star\sum_s\boldsymbol{\psi}(s)g_j(s)$. By conditions (i) and (iii) of \cref{def:bilienar_mdp}, its norm is at most $C_{\mathbf{M}}\sqrt{d_\phi}\,C_\psi\|g_j\|_\infty\leq C_{\mathbf{M}}C_\psi C_\phi d_\phi$. Summing squared column norms gives
\begin{align}
\|\mathbf{B}^\star(\boldsymbol{g})\|_F^2&=\sum_{j=1}^{d_\phi}\biggl\|\mathbf{M}^\star\sum_s\boldsymbol{\psi}(s)g_j(s)\biggr\|_2^2\leq C_{\mathbf{M}}^2C_\psi^2C_\phi^2d_\phi^3.
\end{align}
By \cref{lem:bilinear_bfr_linearity}, $\boldsymbol{g}(s_{i,t+1})=\mathbf{B}^\star(\boldsymbol{g})^\top\boldsymbol{\phi}_{i,t}+[\boldsymbol{g}(s_{i,t+1})-P^\star\boldsymbol{g}(s_{i,t},a_{i,t})]$. Substituting this into the normal equations $\boldsymbol{\Gamma}_{k,t}\widehat{\mathbf{B}}_{k,t}(\boldsymbol{g})=\sum_i\boldsymbol{\phi}_{i,t}\boldsymbol{g}(s_{i,t+1})^\top$ and using $\sum_i\boldsymbol{\phi}_{i,t}\boldsymbol{\phi}_{i,t}^\top=\boldsymbol{\Gamma}_{k,t}-I$ gives
\begin{align}
\boldsymbol{\Gamma}_{k,t}\widehat{\mathbf{B}}_{k,t}(\boldsymbol{g})&=(\boldsymbol{\Gamma}_{k,t}-I)\mathbf{B}^\star(\boldsymbol{g})+\sum_{i=1}^k\boldsymbol{\phi}_{i,t}[\boldsymbol{g}(s_{i,t+1})-P^\star\boldsymbol{g}(s_{i,t},a_{i,t})]^\top.
\end{align}
Multiplying by $\boldsymbol{\Gamma}_{k,t}^{-1/2}$ and rearranging,
\begin{align}
\boldsymbol{\Gamma}_{k,t}^{1/2}(\widehat{\mathbf{B}}_{k,t}(\boldsymbol{g})-\mathbf{B}^\star(\boldsymbol{g}))&=\boldsymbol{\Gamma}_{k,t}^{-1/2}\sum_{i=1}^k\boldsymbol{\phi}_{i,t}[\boldsymbol{g}(s_{i,t+1})-P^\star\boldsymbol{g}(s_{i,t},a_{i,t})]^\top-\boldsymbol{\Gamma}_{k,t}^{-1/2}\mathbf{B}^\star(\boldsymbol{g}).
\end{align}
We take Frobenius norms, apply the triangle inequality, and use $\|\boldsymbol{\Gamma}_{k,t}^{-1/2}\|_{\mathrm{op}}\leq1$ to bound the final term by $C_{\mathbf{M}}C_\psi C_\phi d_\phi^{3/2}$. Together with~\cref{eq:off_bfr_noise}, the two terms are exactly the two summands in $\kappa$ in~\cref{eq:off_bfr_kappa}, which proves~\cref{eq:off_bfr_regression_confidence}.

\paragraph{Uniform Initial-State Averaging}
Each initial state is an independent draw from $\rho$, regardless of the subsequent behavior policy. Consider specific $k\in[K]$ and a representative $\boldsymbol{g}_0\in\mathcal{C}_{\mathrm{W}}$. The vectors $\boldsymbol{g}_0(s_{i,1})-\mathbb{E}_\rho\boldsymbol{g}_0$ have conditional mean zero and Euclidean norm at most $2C_\phi\sqrt{d_\phi}$. We apply \cref{lem:lin_matrix_self_normalized} with the scalar covariates $x_i=1$, so that $d=1$, $R_x=1$, $R=2C_\phi\sqrt{d_\phi}$, $n=k$, and the Gram matrix is $k+1$, and with failure probability $\delta/(8K|\mathcal{C}_{\mathrm{W}}|)$ to obtain that
\begin{align}
\frac{1}{\sqrt{k+1}}\biggl\|\sum_{i=1}^k[\boldsymbol{g}_0(s_{i,1})-\mathbb{E}_\rho\boldsymbol{g}_0]\biggr\|_2\leq2C_\phi\sqrt{d_\phi}\sqrt{\log(1+K)+2\bigl(\log|\mathcal{C}_{\mathrm{W}}|+\log(8K/\delta)\bigr)}.
\end{align}
We take a union bound over all representatives and $k\in[K]$, which costs at most $\delta/8$. For an arbitrary $\boldsymbol{g}\in\mathcal{W}$, let $\boldsymbol{g}_0$ be its representative. Both the empirical and population averages change by at most $C_\phi\sqrt{d_\phi}/(K+1)^2$ when $\boldsymbol{g}$ is replaced by $\boldsymbol{g}_0$:
\begin{align}
\biggl\|\frac1k\sum_{i=1}^k(\boldsymbol{g}-\boldsymbol{g}_0)(s_{i,1})\biggr\|_2&\leq\frac{C_\phi\sqrt{d_\phi}}{(K+1)^2},\qquad \|\mathbb{E}_\rho(\boldsymbol{g}-\boldsymbol{g}_0)\|_2\leq\frac{C_\phi\sqrt{d_\phi}}{(K+1)^2}.
\end{align}
We multiply the preceding bound by $\sqrt{k+1}/k\leq\sqrt{2/k}$ and use $(K+1)^{-2}\leq k^{-1/2}$ and $2\sqrt2\sqrt x+2\leq4\sqrt{x+1}$ for $x\geq0$ to obtain that, uniformly over $\mathcal{W}$,
\begin{align}
\biggl\|\frac1k\sum_{i=1}^k\boldsymbol{g}(s_{i,1})-\mathbb{E}_\rho\boldsymbol{g}\biggr\|_2\leq\frac{\gamma_\rho}{\sqrt k},
\end{align}
with $\gamma_\rho$ in~\cref{eq:off_bfr_initial_radius}.

The intersection of the two events has probability at least $1-\delta/4$, and on it both~\cref{eq:off_bfr_regression_confidence,eq:off_bfr_initial_confidence} hold.
\end{proof}

\subsubsection{Optimistic Policy Evaluation} \label{app:off_ope_guarantees}

Since the proof of \cref{lem:on_bellman_error} in \cref{app:on_ope_guarantees} holds for general radii, the Bellman-error bound for $\OCUPOOFF$ is \cref{lem:on_bellman_error} with $\Delta_{k,h}=\Delta_{k,h}^{\mathrm{Off}}$ and $\alpha_{k,h}=\alpha_h^{\mathrm{Off}}$. In this subsubsection, we establish the value part of the good event in \cref{lem:off_value_event_main}, using the cover of the value functions in \cref{app:off_covering_arguments}.

\begin{lemma}[Value-Regression Confidence] \label{lem:off_value_regression} \label{lem:off_value_event_main}
With the radii $\alpha_h^{\mathrm{Off}}$ specified in the proof, which satisfy
\begin{align}
\alpha_h^{\mathrm{Off}}=\begin{cases}\widetilde{\mathcal{O}}\Bigl((H-h)G\sqrt{d_\phi^2+d_\psi+d_\Pi}\Bigr),&h<H,\\ 0,&h=H,\end{cases}
\end{align}
$\mathcal{E}_{\mathrm{V}}$ holds with probability at least $1-\delta/4$.
\end{lemma}

\begin{proof}[Proof of \cref{lem:off_value_event_main}]
We first recall the relevant notation. In the off-policy setting, \OPE{} fits $\widehat{\boldsymbol{v}}_{k,h}=\boldsymbol{\Gamma}_{k,h}^{-1}\sum_{i=1}^k\boldsymbol{\phi}_{i,h}\widehat{V}_{k,h+1}(s_{i,h+1})$ on the $k$ replay trajectories. The event $\mathcal{E}_{\mathrm{V}}$ requires $\|\widehat{\boldsymbol{v}}_{k,h}-\boldsymbol{v}^\star(\widehat{V}_{k,h+1})\|_{\boldsymbol{\Gamma}_{k,h}}\leq\alpha_h^{\mathrm{Off}}$ for every $k\in[K]$ and $h<H$, where $\boldsymbol{v}^\star$ is as in~\cref{lem:bilinear_next_state_values}. The radii $\alpha_h^{\mathrm{Off}}$ are chosen at the end of the proof.

Consider a specific step $h<H$, and for a bounded function $V$ define the residual $\epsilon_i(V)\coloneq V(s_{i,h+1})-P^\star V(s_{i,h},a_{i,h})$.

\paragraph{Error Decomposition}
The error decomposition~\cref{eq:value_regression_error_decomposition} in the proof of \cref{lem:good_estimation_event_V} is deterministic: it uses only the closed form of $\widehat{\boldsymbol{v}}_{k,h}$, \cref{lem:bilinear_next_state_values}, and \cref{def:bilienar_mdp}. Applying it with the $k$ replay trajectories in place of the $N$ fresh ones gives
\begin{align}
\|\widehat{\boldsymbol{v}}_{k,h}-\boldsymbol{v}^\star(\widehat{V}_{k,h+1})\|_{\boldsymbol{\Gamma}_{k,h}}\leq\biggl\|\sum_{i=1}^k\boldsymbol{\phi}_{i,h}\epsilon_i(\widehat{V}_{k,h+1})\biggr\|_{\boldsymbol{\Gamma}_{k,h}^{-1}}+C_{\mathbf{M}}C_\psi\sqrt{d_\phi}\,(H-h)G. \label{eq:off_value_error_decomposition}
\end{align}

\paragraph{Concentration on a Fixed Cover}
Let $\varepsilon\coloneq (H-h)G/(K+1)$, and let $\mathcal{C}_h$ be an internal $\varepsilon$-cover of the class $\mathcal{V}_{h+1}(\alpha_{h+1}^{\mathrm{Off}})$ of \cref{app:off_covering_arguments}. The class and its cover are fixed before data collection. In particular, the representative functions do not change when the algorithm selects a policy from the replay data. Consider a specific $V_0\in\mathcal{C}_h$. Order the replay trajectories by episode, and let $\mathcal{F}_{i-1}$ contain the first $i-1$ trajectories together with the first $h$ steps of trajectory $i$. The step-$h$ feature $\boldsymbol{\phi}_{i,h}$ is $\mathcal{F}_{i-1}$-measurable, and given $\mathcal{F}_{i-1}$ the successor $s_{i,h+1}$ has law $P^\star(\cdot\mid s_{i,h},a_{i,h})$, even though the behavior policy $\pi_i$ depends on the preceding trajectories. Hence \cref{lem:bilinear_value_response} gives $\mathbb{E}[\epsilon_i(V_0)\mid\mathcal{F}_{i-1}]=0$, and since $V_0$ takes values in $[-(H-h)G,(H-h)G]$, Hoeffding's lemma gives $\mathbb{E}[\exp(\lambda\epsilon_i(V_0))\mid\mathcal{F}_{i-1}]\leq\exp(\lambda^2(H-h)^2G^2/2)$. Apply \cref{lem:self_normalized_concentration} with $\boldsymbol{\Lambda}_0=I$, $\sigma=(H-h)G$, and failure probability $\delta/(4KH|\mathcal{C}_h|)$ for each fixed representative and prefix, and bound $\log\det\boldsymbol{\Gamma}_{k,h}\leq d_\phi\log(1+kC_\phi^2)\leq d_\phi\log(1+KC_\phi^2)$ by the trace and the arithmetic--geometric mean inequality. A union bound over the $|\mathcal{C}_h|$ representatives, $K$ prefixes, and $H-1$ steps gives a single event of probability at least $1-\delta/4$ on which
\begin{align}
\max_{V_0\in\mathcal{C}_h}\biggl\|\sum_{i=1}^k\boldsymbol{\phi}_{i,h}\epsilon_i(V_0)\biggr\|_{\boldsymbol{\Gamma}_{k,h}^{-1}}&\leq (H-h)G\sqrt{d_\phi\log(1+KC_\phi^2)+2\log|\mathcal{C}_h|+2\log(4KH/\delta)} \label{eq:off_fixed_cover_concentration}
\end{align}
for all $k\in[K]$ and $h<H$.

\paragraph{Extension to the Fitted Target}
By \cref{lem:off_covering_numbers_formal}, $\widehat{V}_{k,h+1}\in\mathcal{V}_{h+1}(\alpha_{h+1}^{\mathrm{Off}})$. Select $V_0\in\mathcal{C}_h$ with $\|\widehat{V}_{k,h+1}-V_0\|_\infty\leq\varepsilon$. \cref{lem:lin_target_perturbation}, applied as in the proof of \cref{lem:good_estimation_event_V}, gives
\begin{align}
\biggl\|\sum_{i=1}^k\boldsymbol{\phi}_{i,h}\epsilon_i(\widehat{V}_{k,h+1})\biggr\|_{\boldsymbol{\Gamma}_{k,h}^{-1}}&\leq\biggl\|\sum_{i=1}^k\boldsymbol{\phi}_{i,h}\epsilon_i(V_0)\biggr\|_{\boldsymbol{\Gamma}_{k,h}^{-1}}+2\sqrt{k}\varepsilon,
\end{align}
and $2\sqrt k\varepsilon\leq2\sqrt K(H-h)G/(K+1)\leq2(H-h)G$. This extension is deterministic and holds for all targets simultaneously. Combining it with~\cref{eq:off_value_error_decomposition,eq:off_fixed_cover_concentration} yields, on the event of~\cref{eq:off_fixed_cover_concentration},
\begin{align}
\MoveEqLeft\|\widehat{\boldsymbol{v}}_{k,h}-\boldsymbol{v}^\star(\widehat{V}_{k,h+1})\|_{\boldsymbol{\Gamma}_{k,h}}\\&\leq (H-h)G\Bigl[\sqrt{d_\phi\log(1+KC_\phi^2)+2\log|\mathcal{C}_h|+2\log(4KH/\delta)}+2+C_{\mathbf{M}}C_\psi\sqrt{d_\phi}\Bigr]. \label{eq:off_value_error_explicit}
\end{align}

\paragraph{Choice of Radii}
By \cref{lem:off_covering_numbers_formal}, the cover $\mathcal{C}_h$ can be chosen so that $\log|\mathcal{C}_h|\leq\mathcal{H}_{h+1}^{\mathrm{Off}}(\alpha_{h+1}^{\mathrm{Off}},\varepsilon)$. As in the proof of \cref{lem:good_estimation_event_V}, we define the radii backward from $h=H$ by
\begin{align}
\alpha_h^{\mathrm{Off}}&\coloneq\begin{cases}(H-h)G\Bigl[\sqrt{d_\phi\log(1+KC_\phi^2)+2\mathcal{H}_{h+1}^{\mathrm{Off}}(\alpha_{h+1}^{\mathrm{Off}},\varepsilon)+2\log(4KH/\delta)}\\ \qquad\qquad+2+C_{\mathbf{M}}C_\psi\sqrt{d_\phi}\Bigr],&h<H,\\ 0,&h=H.\end{cases} \label{eq:off_value_radius}
\end{align}
These radii depend only on the policy class and known quantities, and are computed before data collection.

To verify their order, let $r_h\coloneq\alpha_h^{\mathrm{Off}}/((H-h)G)$ for $h\in[H-1]$, and let $r_H\coloneq0$. As in the proof of \cref{lem:good_estimation_event_V}, every logarithm in the first three terms of~\cref{eq:off_value_cover_explicit} has an argument bounded by a polynomial in the problem parameters, $1/\delta$, and $r_{h+1}$. For the policy term, the radius $\min\{1,\varepsilon/(4a_{\pi,h+1})\}$ is at least the reciprocal of a polynomial in the problem parameters, and \cref{asm:policy_class} bounds that term by a constant multiple of $d_\Pi$ times the logarithm of such a polynomial. Hence, there exist a constant $c$, depending only on the constant in \cref{asm:policy_class}, and the logarithm $\ell$ of a polynomial in the problem parameters and $1/\delta$, both independent of $h$, such that
\begin{align}
r_h\leq\sqrt{c(d_\phi^2+d_\psi+d_\Pi)\bigl[\ell+\log(1+r_{h+1})\bigr]}+2+C_{\mathbf{M}}C_\psi\sqrt{d_\phi}.
\end{align}
The same fixed-point argument gives $\alpha_h^{\mathrm{Off}}=\widetilde{\mathcal{O}}((H-h)G\sqrt{d_\phi^2+d_\psi+d_\Pi})$. By~\cref{eq:off_value_error_explicit}, on the event of~\cref{eq:off_fixed_cover_concentration}, which has probability at least $1-\delta/4$, the regression error is at most $\alpha_h^{\mathrm{Off}}$ for every $k$ and $h<H$, that is, $\mathcal{E}_{\mathrm{V}}$ holds.
\end{proof}

\subsection{Analysis of Sample Complexity: Proof of \texorpdfstring{\cref{thm:off_policy}}{Theorem \ref*{thm:off_policy}}} \label{app:regret_analysis_off}

\begin{proof}[Proof of \cref{thm:off_policy}]
We first recall the relevant notation. As in the proof of~\cref{thm:on_policy}, Terms (I)--(III) refer to the regret decomposition in~\cref{eq:regret_decomposition}. The radii $\beta_k^{\mathrm{Off}}$, $\Delta_{k,h}^{\mathrm{Off}}$, and $\alpha_h^{\mathrm{Off}}$ of the good event are given in~\cref{lem:off_model_event_main,lem:off_reward_error_main,lem:off_value_event_main}, and $\widehat{\ell}_{k,h}=\mathbb{E}_{(s,a)\sim\mu_h^{\pi_k}}\|\boldsymbol{\phi}(s,a)\|_{\boldsymbol{\Gamma}_{k-1,h}^{-1}}$ denotes the expected feature norm of $\pi_k$ under the Gram matrix of the previous iteration.

The parameters required by \cref{thm:off_policy} are the value radii $\alpha_h^{\mathrm{Off}}$ in~\cref{eq:off_value_radius}. All other radii are used only in the analysis. We use the same deterministic decomposition as in the fresh-batch analysis. The replay-specific work is to control $\sum_k\widehat{\ell}_{k,h}$ along the adaptively collected trajectories and to sum the errors accumulated by backward feature regression. When $H=1$, occupancies and pseudo-rewards are known exactly and Terms (II) and (III) vanish. \cref{asm:pnpg} directly gives the theorem. We henceforth consider $H\geq2$.

\paragraph{Regret Decomposition}
Applying~\cref{lem:curl_regret_decomposition} to the replay estimates gives~\cref{eq:regret_decomposition}, with $R_{k,h}=\nabla f_h(\mu_h^{\pi_k})$ and $\xi_{k,h}=R_{k,h}+P^\star\widehat{V}_{k,h+1}-\widehat{Q}_{k,h}$. This decomposition holds for every realized set of fitted functions, even though the policies and regression targets depend on the replay data.

\paragraph{Bounding Term (I)}
Apply~\cref{asm:pnpg} with $u=\pi^\star$. Because the assumption covers every sequence of fitted values generated by the framework, it applies to the replay-dependent estimates and gives
\begin{align}
\mathrm{Term~(I)}&\leq\mathcal{O}(H^2G\sqrt{K\log|\mathcal{A}|}+K\varepsilon_{\mathrm{proj}}). \label{eq:off_oracle_regret}
\end{align}
For log-linear policies, \cref{prop:pnpg_oracle} bounds $\varepsilon_{\mathrm{proj}}$. \cref{asm:policy_class} is used in the confidence lemmas, not in this deterministic regret bound.

\paragraph{Bounding Terms (II) and (III)}
Work for now on the good event $\mathcal{E}$. We integrate the first inequality of~\cref{lem:on_bellman_error}, with $\Delta_{k,h}=\Delta_{k,h}^{\mathrm{Off}}$ and $\alpha_{k,h}=\alpha_h^{\mathrm{Off}}$, under $\mu_h^{\pi^\star}$ and the second under $\mu_h^{\pi_k}$. The first bound is uniform in $(s,a)$, and Term (II) therefore requires no trajectories from $\pi^\star$. The step-$H$ bonus is zero. To control the remaining terms, the matrix inequality $\boldsymbol{\Gamma}_{k,h}\succeq\boldsymbol{\Gamma}_{k-1,h}$ gives
\begin{align}
\mathbb{E}_{\mu_h^{\pi_k}}\|\boldsymbol{\phi}\|_{\boldsymbol{\Gamma}_{k,h}^{-1}}&\leq\mathbb{E}_{\mu_h^{\pi_k}}\|\boldsymbol{\phi}\|_{\boldsymbol{\Gamma}_{k-1,h}^{-1}}=\widehat{\ell}_{k,h}.
\end{align}
Hence, writing $\alpha_{\max}^{\mathrm{Off}}=\max_{h<H}\alpha_h^{\mathrm{Off}}$,
\begin{align}
\mathrm{Term~(II)}+\mathrm{Term~(III)}&\leq2\sum_{k=1}^K\sum_{h=1}^H\Delta_{k,h}^{\mathrm{Off}}+2\alpha_{\max}^{\mathrm{Off}}\sum_{k=1}^K\sum_{h=1}^{H-1}\widehat{\ell}_{k,h}. \label{eq:off_bellman_sum_proof}
\end{align}
The factor two arises because $\Delta_{k,h}^{\mathrm{Off}}$ appears in the bounds on both Term (II) and Term (III).

\paragraph{Bounding the Sum of $\widehat{\ell}_{k,h}$}
For $h<H$, define $q_{k,h}=\|\boldsymbol{\phi}(s_{k,h},a_{k,h})\|_{\boldsymbol{\Gamma}_{k-1,h}^{-1}}^2$. By \cref{lem:off_martingale}, with probability at least $1-\delta/4$, for every $h<H$,
\begin{align}
\sum_{k=1}^K\widehat{\ell}_{k,h}&\leq\sum_{k=1}^K\sqrt{q_{k,h}}+C_\phi\sqrt{2Kd_\phi\log(8H/\delta)}. \label{eq:off_leverage_transfer}
\end{align}

\paragraph{Cumulative Estimation Error}
We first bound $\sum_{k=1}^K\sqrt{q_{k,h}}$ in~\cref{eq:off_leverage_transfer}. The Gram matrix $\boldsymbol{\Gamma}_{k,h}=I+\sum_{i=1}^k\boldsymbol{\phi}(s_{i,h},a_{i,h})\boldsymbol{\phi}(s_{i,h},a_{i,h})^\top$ increases by the outer product of the current feature, and $\|\boldsymbol{\phi}(s_{k,h},a_{k,h})\|_2\leq C_\phi\sqrt{d_\phi}$. We apply the elliptical potential lemma (\cref{lem:elliptical_potential}) with $\boldsymbol{x}_k=\boldsymbol{\phi}(s_{k,h},a_{k,h})$, $R_x=C_\phi\sqrt{d_\phi}$, and $n=K$ to obtain that
\begin{align}
\sum_{k=1}^K\min\{1,q_{k,h}\}&\leq2d_\phi\log(1+KC_\phi^2).
\end{align}
This inequality holds for every realized trajectory sequence.

Split the indices into $q_{k,h}\leq1$ and $q_{k,h}>1$. Cauchy--Schwarz bounds the first sum of square roots, while each term in the second group is at most $C_\phi\sqrt{d_\phi}$. Every index in that second group contributes one to $\sum_k\min\{1,q_{k,h}\}$. Hence,
\begin{align}
\sum_{k=1}^K\sqrt{q_{k,h}}&\leq\sqrt{K\sum_{k=1}^K\min\{1,q_{k,h}\}}+C_\phi\sqrt{d_\phi}\sum_{k=1}^K\min\{1,q_{k,h}\}\\
&\leq\sqrt{2Kd_\phi\log(1+KC_\phi^2)}+2C_\phi d_\phi^{3/2}\log(1+KC_\phi^2).
\end{align}
Summing~\cref{eq:off_leverage_transfer} over steps now gives
\begin{align}
\sum_{k=1}^K\sum_{h=1}^{H-1}\widehat{\ell}_{k,h}&\leq H\Bigl[\sqrt{2Kd_\phi\log(1+KC_\phi^2)}+2C_\phi d_\phi^{3/2}\log(1+KC_\phi^2)+C_\phi\sqrt{2Kd_\phi\log(8H/\delta)}\Bigr]\\
&=\widetilde{\mathcal{O}}(H\sqrt{d_\phi K}+Hd_\phi^{3/2}).
\end{align}
For $K\geq d_\phi^2$, $d_\phi^{3/2}\leq\sqrt{d_\phi K}$. Hence, the second term is absorbed into the first.

To sum $\Delta_{k,h}^{\mathrm{Off}}$, let $\beta_{\max}^{\mathrm{Off}}=\max_{k\in[K]}\beta_k^{\mathrm{Off}}$ and use its explicit expression in~\cref{eq:Delta_off}. Since $\beta_k^{\mathrm{Off}}\leq\beta_{\max}^{\mathrm{Off}}$, the terms $LC_\psi\sqrt{\beta_k^{\mathrm{Off}}}\,\widehat{\ell}_{k,h-1}$ sum to at most $LC_\psi\sqrt{\beta_{\max}^{\mathrm{Off}}}\sum_{k=1}^K\sum_{h=1}^{H-1}\widehat{\ell}_{k,h}$. The term $\gamma_\rho/\sqrt k$ is repeated at $H-1$ steps. For the terms $\kappa\sum_{t=1}^{h-2}\widehat{\ell}_{k,t}$, interchange the two finite step sums:
\begin{align}
\sum_{h=2}^H\sum_{t=1}^{h-2}\widehat{\ell}_{k,t}&=\sum_{t=1}^{H-2}(H-t-1)\widehat{\ell}_{k,t}\leq H\sum_{t=1}^{H-1}\widehat{\ell}_{k,t}.
\end{align}
This counts how often an error at step $t$ affects a later occupancy. Combining these three bounds yields
\begin{align}
\sum_{k=1}^K\sum_{h=1}^H\Delta_{k,h}^{\mathrm{Off}}&\leq LC_\psi\bigl[\sqrt{\beta_{\max}^{\mathrm{Off}}}+H\kappa(\sqrt{\beta_{\max}^{\mathrm{Off}}}+C_{\mathbf{M}}\sqrt{d_\phi})\bigr]\sum_{k=1}^K\sum_{h=1}^{H-1}\widehat{\ell}_{k,h}\\
&\quad+LC_\psi H(\sqrt{\beta_{\max}^{\mathrm{Off}}}+C_{\mathbf{M}}\sqrt{d_\phi})\gamma_\rho\sum_{k=1}^Kk^{-1/2}. \label{eq:off_reward_sum_proof}
\end{align}

It remains to bound $\sum_{k=1}^Kk^{-1/2}$. Since $x\mapsto x^{-1/2}$ is decreasing, integral comparison gives
\begin{align}
\sum_{k=1}^Kk^{-1/2}&\leq1+\int_1^Kx^{-1/2}\,\mathrm{d}x=2\sqrt K-1\leq2\sqrt K.
\end{align}
By \cref{lem:off_model_event_main,lem:off_reward_event_main,lem:off_value_event_main}, and suppressing fixed feature constants, the radii satisfy
\begin{align}
\sqrt{\beta_{\max}^{\mathrm{Off}}}&=\widetilde{\mathcal{O}}(\sqrt{d_\phi}),\qquad\kappa,\gamma_\rho=\widetilde{\mathcal{O}}(\sqrt{d_\phi(d_\phi^2+d_\Pi)}),\qquad\alpha_{\max}^{\mathrm{Off}}=\widetilde{\mathcal{O}}(HG\sqrt{d_\phi^2+d_\psi+d_\Pi}).
\end{align}
For $K\geq d_\phi^2$, the last term in~\cref{eq:off_bellman_sum_proof} is thus $\widetilde{\mathcal{O}}(H^2G\sqrt{d_\phi(d_\phi^2+d_\psi+d_\Pi)K})$. After expanding the bracket, the three terms in~\cref{eq:off_reward_sum_proof} have respective orders
\begin{align}
LC_\psi\sqrt{\beta_{\max}^{\mathrm{Off}}}\sum_{k=1}^K\sum_{h=1}^{H-1}\widehat{\ell}_{k,h}&=\widetilde{\mathcal{O}}(LH d_\phi\sqrt K),\\
LC_\psi H\kappa(\sqrt{\beta_{\max}^{\mathrm{Off}}}+C_{\mathbf{M}}\sqrt{d_\phi})\sum_{k=1}^K\sum_{h=1}^{H-1}\widehat{\ell}_{k,h}&=\widetilde{\mathcal{O}}(LH^2d_\phi^{3/2}\sqrt{d_\phi^2+d_\Pi}\sqrt K),\\
LC_\psi H(\sqrt{\beta_{\max}^{\mathrm{Off}}}+C_{\mathbf{M}}\sqrt{d_\phi})\gamma_\rho\sum_{k=1}^Kk^{-1/2}&=\widetilde{\mathcal{O}}(LH d_\phi\sqrt{d_\phi^2+d_\Pi}\sqrt K).
\end{align}
Here the first order uses $\sum_{k=1}^K\sum_{h=1}^{H-1}\widehat{\ell}_{k,h}=\widetilde{\mathcal{O}}(H\sqrt{d_\phi K})$, the second also uses $C_{\mathbf{M}}\sqrt{d_\phi}\leq\sqrt{\beta_{\max}^{\mathrm{Off}}}$, and the third uses $\sum_kk^{-1/2}\leq2\sqrt K$. The middle order dominates the other two because $H,d_\phi\geq1$.

\paragraph{Putting Everything Together}
\cref{lem:off_model_event_main} shows that $\mathcal{E}_{\mathrm{P}}$ fails with probability at most $\delta/4$, \cref{lem:off_reward_error_main} gives $\Pr(\mathcal{E}_{\mathrm{P}}\cap\mathcal{E}_{\mathrm{R}}^c)\leq\delta/4$, \cref{lem:off_value_event_main} shows that $\mathcal{E}_{\mathrm{V}}$ fails with probability at most $\delta/4$, and \cref{lem:off_martingale} shows that~\cref{eq:off_leverage_transfer} fails with probability at most $\delta/4$. Since $\mathcal{E}^c\subseteq\mathcal{E}_{\mathrm{P}}^c\cup(\mathcal{E}_{\mathrm{P}}\cap\mathcal{E}_{\mathrm{R}}^c)\cup\mathcal{E}_{\mathrm{V}}^c$, the union bound shows that $\mathcal{E}$ and~\cref{eq:off_leverage_transfer} hold simultaneously with probability at least $1-\delta$, with no independence assumption.

On this event and for $K\geq d_\phi^2$, combine~\cref{eq:off_oracle_regret,eq:off_bellman_sum_proof,eq:off_reward_sum_proof} with the bounds on $\sum_{k=1}^K\sum_{h=1}^{H-1}\widehat{\ell}_{k,h}$ and $\sum_{k=1}^Kk^{-1/2}$ above and substitute into~\cref{eq:regret_decomposition}. This gives
\begin{align}
\operatorname{Reg}_K&\leq\mathcal{O}(H^2G\sqrt{K\log|\mathcal{A}|}+K\varepsilon_{\mathrm{proj}})+\widetilde{\mathcal{O}}\biggl(H^2\Bigl[G\sqrt{d_\phi(d_\phi^2+d_\psi+d_\Pi)}+Ld_\phi^{3/2}\sqrt{d_\phi^2+d_\Pi}\Bigr]\sqrt K\biggr). \label{eq:off_regret_proof}
\end{align}

For $K<d_\phi^2$, use a deterministic bound instead of absorbing the second term of the bound on $\sum_{k=1}^K\sum_{h=1}^{H-1}\widehat{\ell}_{k,h}$. Each true occupancy is a probability distribution. Hence, its $\ell_1$ distance to another occupancy is at most two. Lipschitzness gives $F(\mu^{\pi^\star})-F(\mu^{\pi_k})\leq2HG$. Meanwhile $H^2G\sqrt{d_\phi(d_\phi^2+d_\psi+d_\Pi)/K}\geq HG$. Thus a numerical constant makes the stated rate dominate $2HG$ as well, and $\varepsilon_{\mathrm{proj}}\geq0$ only adds to the bound. This covers every $K\geq1$.

Finally, concavity of $F$ at the average occupancy gives $\operatorname{Sub\text{-}Opt}(\overline{\pi}_K)\leq\operatorname{Reg}_K/K$. Dividing~\cref{eq:off_regret_proof} by $K$ gives the claimed suboptimality bound. The first term of this bound is at most $\epsilon$, and the policy is therefore $\mathcal{O}(\epsilon+\varepsilon_{\mathrm{proj}})$-optimal, once
\begin{align}
K&=\widetilde{\mathcal{O}}\left(\frac{H^4[G\sqrt{d_\phi(d_\phi^2+d_\psi+d_\Pi)}+Ld_\phi^{3/2}\sqrt{d_\phi^2+d_\Pi}]^2}{\epsilon^2}\right).
\end{align} Since exactly one trajectory is collected per iteration, this is also the sample complexity, and the number of observed transitions is $HK$.
\end{proof}

The following lemma controls the sum of $\widehat{\ell}_{k,h}$ used in the proof above.

\begin{lemma}[Martingale Concentration] \label{lem:off_martingale}
With probability at least $1-\delta/4$, for every $h<H$,
\begin{align}
\sum_{k=1}^K\widehat{\ell}_{k,h}&\leq\sum_{k=1}^K\|\boldsymbol{\phi}(s_{k,h},a_{k,h})\|_{\boldsymbol{\Gamma}_{k-1,h}^{-1}}+C_\phi\sqrt{2Kd_\phi\log(8H/\delta)}.
\end{align}
\end{lemma}

\begin{proof}
Consider a specific $h<H$, and let $\mathcal{F}_{k-1}$ be the history before collecting trajectory $k$. Both $\pi_k$ and $\boldsymbol{\Gamma}_{k-1,h}$ are determined by this history. Set $q_{k,h}=\|\boldsymbol{\phi}(s_{k,h},a_{k,h})\|_{\boldsymbol{\Gamma}_{k-1,h}^{-1}}^2$. The current step-$h$ state-action pair has conditional distribution $\mu_h^{\pi_k}$. Hence, $\mathbb{E}[\sqrt{q_{k,h}}\mid\mathcal{F}_{k-1}]=\widehat{\ell}_{k,h}$. Using $\boldsymbol{\Gamma}_{k-1,h}$ rather than $\boldsymbol{\Gamma}_{k,h}$ is necessary for this equality. Thus $\widehat{\ell}_{k,h}-\sqrt{q_{k,h}}$ is a martingale difference with respect to the episode histories. It has absolute value at most $C_\phi\sqrt{d_\phi}$, since both terms lie between zero and that value: $\boldsymbol{\Gamma}_{k-1,h}\succeq I$ gives $\sqrt{q_{k,h}}\leq\|\boldsymbol{\phi}(s_{k,h},a_{k,h})\|_2\leq C_\phi\sqrt{d_\phi}$. Applying \cref{lem:azuma_hoeffding} with $R=C_\phi\sqrt{d_\phi}$ and setting its right-hand side to $\delta/(4H)$, and then taking a union bound over steps, proves the lemma.
\end{proof}

\section{Deferred Details and Proofs for \texorpdfstring{\cref{section:ocupo-lin}}{Section \ref*{section:ocupo-lin}}} \label{app:linear_off_proofs}

We first describe the changes to the estimation components and collect the covering arguments in \cref{app:lin_covering_arguments}. We then define the good event for linear MDPs and prove its guarantees, and the final subsection combines these bounds into the proof of \cref{thm:lin_off}. As in \cref{app:ucgu_off}, each guarantee states the order of its confidence radius, the exact value is given in its proof, and $\widetilde{\mathcal{O}}(\cdot)$ suppresses logarithmic factors, the feature regularity constants $C_\phi$ and $C_{\mathbf{T}}$ in~\cref{def:linear_mdp}, and the constants in~\cref{asm:pnpg,asm:policy_class}. Throughout, we assume $H\geq2$. When $H=1$, the occupancy measure $\rho(s)\pi_{k,1}(a\mid s)$ and its pseudo-reward are known exactly, the Bellman errors vanish, and we apply~\cref{asm:pnpg} directly to obtain the claimed rate.

\subsection{Algorithm Design} \label{app:lin_algorithm_design}

\subsubsection{Transition Model Estimation} \label{app:lin_off_transition} \label{app:lin_tme_details}

The occupancy decomposition in~\cref{eq:lin_occupancy_decomposition} requires the column $\mathbf{T}^\star(s)$ of $\mathbf{T}^\star$ at each state $s$ where the occupancy measure is evaluated. Maintaining an estimate of every column would require memory proportional to $|\mathcal{S}|$. Instead, \STE{} regresses the one-hot vector of the observed next state on the features of the current state-action pair. The following lemma shows that this one-hot vector is an unbiased estimator of the transition probabilities under any behavior policy.

\begin{lemma}[Unbiased One-Hot Next-State Regression] \label{lem:linear_regression_response}
Under \cref{def:linear_mdp}, consider a specific history $\mathcal{H}$ through a state-action pair $(s,a)$, and let $S'\sim P^\star(\cdot\mid s,a)$. Then $\boldsymbol{e}_{S'}$ has unit Euclidean norm and
\begin{align}
\mathbb{E}[\boldsymbol{e}_{S'}\mid\mathcal{H}]&=P^\star(\cdot\mid s,a)=(\mathbf{T}^\star)^\top\boldsymbol{\phi}(s,a).
\end{align}
\end{lemma}

\begin{proof}
For each state $x$, the coordinate of $\boldsymbol{e}_{S'}$ indexed by $x$ is $\mathbf{1}\{S'=x\}$. Its conditional expectation is $P^\star(x\mid s,a)=\mathbf{T}^\star(x)^\top\boldsymbol{\phi}(s,a)$. Collecting these coordinates, we obtain the claimed identity. Exactly one coordinate of $\boldsymbol{e}_{S'}$ equals one, and all others are zero. Hence, its Euclidean norm is one. Neither argument depends on how the pair $(s,a)$ was selected.
\end{proof}

Recall that $\boldsymbol{\Lambda}_k=I+\sum_{i=1}^k\sum_{h=1}^H\boldsymbol{\phi}_{i,h}\boldsymbol{\phi}_{i,h}^\top$, which is also the Gram matrix of \BTE{} on replay data. The objective of \STE{} in~\cref{eq:ste_estimator} is a sum of quadratic losses over the columns of $\mathbf{T}$ plus the squared Frobenius norm. Hence, its unique minimizer satisfies $\boldsymbol{\Lambda}_k\widehat{\mathbf{T}}_k=\sum_{i=1}^k\sum_{h=1}^H\boldsymbol{\phi}_{i,h}\boldsymbol{e}_{s_{i,h+1}}^\top$. The column of the right-hand side indexed by $s$ is $\sum_{(i,h):s_{i,h+1}=s}\boldsymbol{\phi}_{i,h}$, which is zero unless $s\in\mathcal{S}_k^+$. Therefore, $\widehat{\mathbf{T}}_k(s)$ is given by~\cref{eq:ste_columns}. Since $|\mathcal{S}_k^+|\leq kH$, the learner stores at most $kH$ vectors in $\mathbb{R}^{d_\phi}$ and evaluates $\widehat{\mathbf{T}}_k(s)$ at any other state by returning zero. We present the pseudocode in~\cref{alg:tme_linear}.

\begin{algorithm2e}[H]
\caption{Sparse Transition Estimation (\STE{})} \label{alg:tme_linear} \label{alg:ste}
\begin{algorithmic}[1]
\State \textbf{Input:} replay dataset $\mathcal{D}_k=\{\tau_i\}_{i=1}^k$.
\State $\boldsymbol{\Lambda}_k\gets I+\sum_{i=1}^k\sum_{h=1}^H\boldsymbol{\phi}_{i,h}\boldsymbol{\phi}_{i,h}^\top$. \label{line:lin_tme_gram}
\State $\mathcal{S}_k^{+}\gets\{s_{i,h+1}\mid i\in[k],h\in[H]\}$. \label{line:lin_tme_support}
\State $\widehat{\mathbf{T}}_k(s)\gets\boldsymbol{\Lambda}_k^{-1}\sum_{(i,h):s_{i,h+1}=s}\boldsymbol{\phi}_{i,h}$, for every $s\in\mathcal{S}_k^{+}$. \label{line:lin_tme_column}
\State $\widehat{P}_k(s'\mid s,a)\gets\boldsymbol{\phi}(s,a)^\top\widehat{\mathbf{T}}_k(s')$, for every queried $(s'\mid s,a)$, with $\widehat{\mathbf{T}}_k(s')=\boldsymbol{0}$ for $s'\notin\mathcal{S}_k^{+}$.
\State \Return\quad $\widehat{P}_k$, represented by $\{\widehat{\mathbf{T}}_k(s)\}_{s\in\mathcal{S}_k^{+}}$.
\end{algorithmic}
\end{algorithm2e}

Assigning zero to a column that does not appear in $\mathcal{D}_k$ does not assert that the corresponding transition probability is zero. \cref{lem:lin_off_model_event} bounds the error of all columns together in Frobenius norm, and \cref{lem:lin_column_error} shows that this bound controls the error of every column. Since every one-hot vector has unit norm regardless of $|\mathcal{S}|$, this bound does not depend on $|\mathcal{S}|$.

\subsubsection{Pseudo-Reward Estimation} \label{app:lin_off_reward} \label{app:lin_bfr_details}

The backward recursion in \cref{lem:backward_feature_recursion} holds for any MDP. The following lemma shows that, in a linear MDP, $P^\star\boldsymbol{g}(s,a)$ is linear in $\boldsymbol{\phi}(s,a)$ for every bounded function $\boldsymbol{g}\colon \mathcal{S}\to\mathbb{R}^{d_\phi}$. Hence, \BFR{} (\cref{alg:rfe}) applies without change to estimate $\widehat{\boldsymbol{\omega}}_{k,h}$, with the same projection radius $R_\phi=C_\phi\sqrt{d_\phi}$.

\begin{lemma}[Unbiased Backward Feature Regression] \label{lem:linear_bfr_response}
Under \cref{def:linear_mdp}, let $\boldsymbol{g}\colon \mathcal{S}\to\mathbb{R}^{d_\phi}$ be bounded and fixed before the next state is drawn, and let $\mathbf{B}^\star(\boldsymbol{g})\coloneq\sum_{s'}\mathbf{T}^\star(s')\boldsymbol{g}(s')^\top\in\mathbb{R}^{d_\phi\times d_\phi}$. Consider a specific history $\mathcal{H}$ through a state-action pair $(s,a)$, and let $S'\sim P^\star(\cdot\mid s,a)$. Then
\begin{align}
\mathbb{E}[\boldsymbol{g}(S')\mid\mathcal{H}]=P^\star\boldsymbol{g}(s,a)=\mathbf{B}^\star(\boldsymbol{g})^\top\boldsymbol{\phi}(s,a). \label{eq:lin_off_forward_coefficient}
\end{align}
\end{lemma}

\begin{proof}
Since $\boldsymbol{g}$ and $(s,a)$ are fixed under the conditioning, we apply the definition of the conditional expectation and~\cref{def:linear_mdp} to obtain that
\begin{align}
\mathbb{E}[\boldsymbol{g}(S')\mid\mathcal{H}]&=\sum_{s'}P^\star(s'\mid s,a)\boldsymbol{g}(s')=\biggl(\sum_{s'}\boldsymbol{g}(s')\mathbf{T}^\star(s')^\top\biggr)\boldsymbol{\phi}(s,a)=\mathbf{B}^\star(\boldsymbol{g})^\top\boldsymbol{\phi}(s,a).
\end{align}
The first expression equals $P^\star\boldsymbol{g}(s,a)$ by definition.
\end{proof}

\cref{line:rfe_reward} of \cref{alg:rfe} uses~\cref{eq:lin_occupancy_reconstruction} in place of~\cref{eq:bilinear_occupancy_reconstruction}, with the decomposition in~\cref{eq:lin_occupancy_decomposition} proved in \cref{lem:linear_occupancy_measure}. The occupancy measure at step one remains exact. At later steps, clipping keeps each coordinate in $[0,1]$, even though $\widehat{\mathbf{T}}_k$ need not define a transition kernel. By \cref{asm:coordinate_separability}, each coordinate of $\widehat{R}_{k,h}$ requires only the corresponding coordinate of $\widehat{\mu}_{k,h}$, which requires a single column of $\widehat{\mathbf{T}}_k$. The following deterministic bound shows how the error of each occupancy coordinate controls the error of the corresponding pseudo-reward coordinate.

\begin{lemma}[Coordinatewise Gradient Perturbation] \label{lem:coordinatewise_gradient_perturbation}
Suppose \cref{asm:lipschitzness_and_smoothness,asm:coordinate_separability} hold. For any $\boldsymbol{u},\boldsymbol{u}'\in[0,1]^{\mathcal{S}\times\mathcal{A}}$ and $(s,a)\in\mathcal{S}\times\mathcal{A}$,
\begin{align}
|[\nabla f_h(\boldsymbol{u})-\nabla f_h(\boldsymbol{u}')](s,a)|&\leq L|\boldsymbol{u}(s,a)-\boldsymbol{u}'(s,a)|\leq L\|\boldsymbol{u}-\boldsymbol{u}'\|_\infty. \label{eq:coordinatewise_gradient_perturbation}
\end{align}
Moreover, every gradient coordinate has absolute value at most $G$.
\end{lemma}

\begin{proof}
Consider specific $h$, $s$, and $a$, and form a vector $\boldsymbol{w}$ by replacing the coordinate $\boldsymbol{u}(s,a)$ with $\boldsymbol{u}'(s,a)$, leaving all other coordinates unchanged. The vector $\boldsymbol{w}$ lies in $[0,1]^{\mathcal{S}\times\mathcal{A}}$, which is the domain of $f_h$, even though it need not be an occupancy measure. By~\cref{asm:coordinate_separability}, $[\nabla f_h(\boldsymbol{w})](s,a)=[\nabla f_h(\boldsymbol{u}')](s,a)$. We apply the smoothness bound in~\cref{asm:lipschitzness_and_smoothness} between $\boldsymbol{u}$ and $\boldsymbol{w}$ to obtain that
\begin{align}
|[\nabla f_h(\boldsymbol{u})-\nabla f_h(\boldsymbol{u}')](s,a)|&=|[\nabla f_h(\boldsymbol{u})-\nabla f_h(\boldsymbol{w})](s,a)|\\
&\leq L\|\boldsymbol{u}-\boldsymbol{w}\|_1=L|\boldsymbol{u}(s,a)-\boldsymbol{u}'(s,a)|.
\end{align}
To bound a gradient coordinate, we vary only that coordinate inside $[0,1]^{\mathcal{S}\times\mathcal{A}}$. The Lipschitz bound in~\cref{asm:lipschitzness_and_smoothness} bounds every resulting difference quotient by $G$ in absolute value, and continuity of the gradient extends the bound to the boundary.
\end{proof}

\subsubsection{Optimistic Policy Evaluation} \label{app:lin_off_ope} \label{app:lin_ope_details}

As in the bilinear case, \OPE{} regresses $\widehat{V}_{k,h+1}(s_{i,h+1})$ on $\boldsymbol{\phi}_{i,h}$. The following identity justifies this regression in a linear MDP.

\begin{lemma}[Unbiased Value Regression] \label{lem:linear_value_response}
Under \cref{def:linear_mdp}, let $V\colon \mathcal{S}\to\mathbb{R}$ be bounded and fixed before the next state is drawn, and let $\boldsymbol{v}^\star(V)\coloneq\sum_{s'}\mathbf{T}^\star(s')V(s')$. Consider a specific history $\mathcal{H}$ through a state-action pair $(s,a)$, and let $S'\sim P^\star(\cdot\mid s,a)$. Then
\begin{align}
\mathbb{E}[V(S')\mid\mathcal{H}]&=P^\star V(s,a)=\boldsymbol{\phi}(s,a)^\top\boldsymbol{v}^\star(V).
\end{align}
\end{lemma}

\begin{proof}
We apply the definition of the conditional expectation and~\cref{def:linear_mdp} to obtain that $\mathbb{E}[V(S')\mid\mathcal{H}]=\sum_{s'}P^\star(s'\mid s,a)V(s')=\boldsymbol{\phi}(s,a)^\top\sum_{s'}\mathbf{T}^\star(s')V(s')=\boldsymbol{\phi}(s,a)^\top\boldsymbol{v}^\star(V)$.
\end{proof}

\OPE{} is invoked with $\widehat{R}_k$, the replay dataset $\mathcal{D}_k$, and the radii $\{\alpha_{k,h}^{\mathrm{Lin}}\}_{h=1}^{H-1}$ specified in~\cref{eq:lin_value_radius_outline}. For every $h<H$, with $\boldsymbol{\Gamma}_{k,h}=I+\sum_{i=1}^k\boldsymbol{\phi}_{i,h}\boldsymbol{\phi}_{i,h}^\top$, it computes
\begin{align}
\widehat{\boldsymbol{v}}_{k,h}&=\boldsymbol{\Gamma}_{k,h}^{-1}\sum_{i=1}^k\boldsymbol{\phi}_{i,h}\widehat{V}_{k,h+1}(s_{i,h+1}),\\
\widetilde{Q}_{k,h}(s,a)&=\widehat{R}_{k,h}(s,a)+\boldsymbol{\phi}(s,a)^\top\widehat{\boldsymbol{v}}_{k,h}+\alpha_{k,h}^{\mathrm{Lin}}\bigl\lVert \boldsymbol{\phi}(s,a) \bigr\rVert_{\boldsymbol{\Gamma}_{k,h}^{-1}}.
\end{align}
Compared with $\OCUPOOFF$, only the radii change. The values $\widehat{V}_{k,h+1}(s_{i,h+1})$ used in this regression contain $\widehat{R}_{k,h+1}$, which depends on the columns of $\widehat{\mathbf{T}}_k$. The number of nonzero columns grows with $k$, and covering them would introduce a dependence on the number of next states in $\mathcal{D}_k$. Instead, the analysis compares each value function with the one that uses the true columns $\mathbf{T}^\star(s)$ in its pseudo-reward, and the radius $\alpha_{k,h}^{\mathrm{Lin}}$ accounts for this replacement. The radius is computable from $\boldsymbol{\Lambda}_k$ and $\widehat{\boldsymbol{\omega}}_{k,h}$.

\subsection{Covering Arguments} \label{app:lin_covering_arguments} \label{app:lin_off_covering}

This subsection collects the covering bounds used in the proofs of \cref{lem:lin_off_reward_event,lem:lin_off_value_event}. As in \cref{app:off_covering_arguments}, every class below is fixed before data collection and varies over the policy class $\overline{\Pi}=\Pi\cup\{\pi_{\mathrm{Unif}}\}$, which contains every generated policy and satisfies \cref{asm:policy_class}.

\paragraph{Functions Produced by Backward Feature Regression}
For each step $t<H$, let $\mathcal{W}_t$ be the union of the two families
\begin{align}
\overline{\boldsymbol{\phi}}_{\pi,t}(s)=\sum_a\pi_t(a\mid s)\boldsymbol{\phi}(s,a)\quad\text{and}\quad\boldsymbol{g}_{\mathbf{B},\pi,t}(s)=\sum_a\pi_t(a\mid s)\operatorname{Proj}_{\mathbb{B}_2(C_\phi\sqrt{d_\phi})}(\mathbf{B}^\top\boldsymbol{\phi}(s,a)),
\end{align}
where $\pi\in\overline{\Pi}$ and $\|\mathbf{B}\|_F\leq C_\phi\sqrt{d_\phi K}$. The regression targets in~\cref{eq:bfr_ridge_objective} lie in $\mathbb{B}_2(C_\phi\sqrt{d_\phi})$. Hence, comparing the objective with its value at zero shows that $\|\widehat{\mathbf{B}}_{j,t}^{\pi_k}\|_F\leq C_\phi\sqrt{d_\phi k}\leq C_\phi\sqrt{d_\phi K}$, and every function $\widehat{\boldsymbol{g}}_{j,t}^{\pi_k}$ produced by \BFR{} belongs to $\mathcal{W}_t$.

\begin{lemma}[Cover of the Function Class $\mathcal{W}_t$] \label{lem:lin_off_W_cover}
Every member of $\mathcal{W}_t$ has uniform Euclidean norm at most $C_\phi\sqrt{d_\phi}$. For $0<\varepsilon\leq1$, $\mathcal{W}_t$ has an internal $(C_\phi\sqrt{d_\phi}\varepsilon)$-cover in $\|\cdot\|_{\infty,2}$ whose log cardinality is at most
\begin{align}
\log2+\log\mathcal{N}(\varepsilon/2,\overline{\Pi},\mathrm{dist}_\Pi)+d_\phi^2\log\left(1+\frac{4C_\phi\sqrt{d_\phi K}}{\varepsilon}\right)\eqcolon\mathcal{H}_{\mathrm{W}}(\varepsilon). \label{eq:lin_off_W_entropy}
\end{align}
\end{lemma}

\begin{proof}
Both families average vectors in $\mathbb{B}_2(C_\phi\sqrt{d_\phi})$. Hence, every member of $\mathcal{W}_t$ has uniform Euclidean norm at most $C_\phi\sqrt{d_\phi}$. For two members of the second family, we add and subtract $\boldsymbol{g}_{\mathbf{B}',\pi,t}$ and apply the nonexpansiveness of the projection and $\|(\mathbf{B}-\mathbf{B}')^\top\boldsymbol{\phi}(s,a)\|_2\leq C_\phi\sqrt{d_\phi}\|\mathbf{B}-\mathbf{B}'\|_F$ to obtain that
\begin{align}
\|\boldsymbol{g}_{\mathbf{B},\pi,t}-\boldsymbol{g}_{\mathbf{B}',\pi',t}\|_{\infty,2}&\leq C_\phi\sqrt{d_\phi}\bigl(\|\mathbf{B}-\mathbf{B}'\|_F+\mathrm{dist}_\Pi(\pi,\pi')\bigr).
\end{align}
For the first family, the same calculation leaves only the policy term. We take an internal $(\varepsilon/2)$-cover of $\overline{\Pi}$ and an internal $(\varepsilon/2)$-cover of the Frobenius ball of radius $C_\phi\sqrt{d_\phi K}$, whose size is at most $(1+4C_\phi\sqrt{d_\phi K}/\varepsilon)^{d_\phi^2}$ by \cref{lem:euclidean_cover}. We pair their members to obtain a $(C_\phi\sqrt{d_\phi}\varepsilon)$-cover of the second family, and the members of the policy cover form a $(C_\phi\sqrt{d_\phi}\varepsilon)$-cover of the first family. We take logarithms of the total size to obtain~\cref{eq:lin_off_W_entropy}.
\end{proof}

\paragraph{Value Functions}
Define
\begin{align}
\nu_{\mathrm{Lin}}\coloneq\frac{K+H+d_\phi+d_\Pi+C_\phi+C_{\mathbf{T}}+G+L+HG+(HG)^{-1}+|\mathcal{A}|+2}{\delta}\qquad\text{and}\qquad Z_{\mathrm{Lin}}\coloneq c_0\nu_{\mathrm{Lin}}^{20}, \label{eq:lin_off_envelope}
\end{align}
where $c_0\geq1$ is a sufficiently large constant fixed in the proof of \cref{lem:lin_off_value_event}. Note that $\nu_{\mathrm{Lin}}\geq2$.
The constant $c_0$ may depend on the constant in~\cref{asm:policy_class} and does not depend on $K$ or $|\mathcal{S}|$. Define $a_{\mathbf{T}}\coloneq LC_{\mathbf{T}}\sqrt{d_\phi}/(HG)$ and $a_\phi\coloneq C_\phi\sqrt{d_\phi}$.
For $j\in\{2,\ldots,H\}$, $\pi\in\overline{\Pi}$, and $\|\boldsymbol{\omega}\|_2\leq C_\phi\sqrt{d_\phi}$, define
\begin{align}
\mu_j^{\pi,\boldsymbol{\omega}}(s,a)\coloneq\operatorname{Clip}(\pi_j(a\mid s)\boldsymbol{\omega}^\top\mathbf{T}^\star(s),[0,1])\qquad\text{and}\qquad R_j^{\pi,\boldsymbol{\omega}}\coloneq\nabla f_j(\mu_j^{\pi,\boldsymbol{\omega}}).
\end{align}
For $j\in\{2,\ldots,H\}$ and the clipping interval $\mathcal{I}_j\coloneq[-(H-j+1)G,(H-j+1)G]$, let $\mathcal{V}_j^\circ$ consist of the functions
\begin{align}
V_j^{\pi,\boldsymbol{\omega},\boldsymbol{v},\boldsymbol{A},\alpha}(s)&\coloneq\sum_a\pi_j(a\mid s)\operatorname{Clip}\bigl(R_j^{\pi,\boldsymbol{\omega}}(s,a)+HG\boldsymbol{\phi}(s,a)^\top\boldsymbol{v}+\alpha\sqrt{\boldsymbol{\phi}(s,a)^\top\boldsymbol{A}\boldsymbol{\phi}(s,a)},\mathcal{I}_j\bigr), \label{eq:lin_off_true_value_class}
\end{align}
where $\pi\in\overline{\Pi}$, $\|\boldsymbol{\omega}\|_2\leq C_\phi\sqrt{d_\phi}$, $\|\boldsymbol{v}\|_2\leq\sqrt K$, $0\preceq\boldsymbol{A}\preceq I$, and the bonus radius satisfies $0\leq\alpha\leq HGZ_{\mathrm{Lin}}$. At $j=H$, take $\boldsymbol{v}=\boldsymbol{0}$ and $\alpha=0$. The unknown matrix $\mathbf{T}^\star$ is fixed within this class. It is neither covered nor evaluated by the learner.

\begin{lemma}[Cover of the Value Functions] \label{lem:lin_off_value_cover} \label{lem:lin_off_covering_numbers}
For $0<\varepsilon\leq1$, $\mathcal{V}_j^\circ$ has an internal $(HG\varepsilon)$-cover in sup norm whose log cardinality is at most
\begin{align}
&\log\mathcal{N}\left(\frac{\varepsilon}{5(1+a_{\mathbf{T}} a_\phi)},\overline{\Pi},\mathrm{dist}_\Pi\right)+d_\phi\log\left(1+\frac{10a_{\mathbf{T}} a_\phi}{\varepsilon}\right)+d_\phi\log\left(1+\frac{10a_\phi\sqrt K}{\varepsilon}\right)\\
&\quad+\frac{d_\phi(d_\phi+1)}2\log\left(1+\frac{50Z_{\mathrm{Lin}}^2a_\phi^2\sqrt{d_\phi}}{\varepsilon^2}\right)+\log\left(1+\frac{10Z_{\mathrm{Lin}}a_\phi}{\varepsilon}\right)\eqcolon\mathcal{H}_{\mathrm{V}}(\varepsilon). \label{eq:lin_off_V_entropy}
\end{align}
\end{lemma}

\begin{proof}
By condition (ii) of \cref{def:linear_mdp} with an indicator function, $\|\mathbf{T}^\star(s)\|_2\leq C_{\mathbf{T}}\sqrt{d_\phi}$. We add and subtract $\pi_j(a\mid s)\boldsymbol{\omega}'^\top\mathbf{T}^\star(s)$ and use $\|\boldsymbol{\omega}'\|_2\leq a_\phi$ to obtain that $|\mu_j^{\pi,\boldsymbol{\omega}}(s,a)-\mu_j^{\pi',\boldsymbol{\omega}'}(s,a)|\leq C_{\mathbf{T}}\sqrt{d_\phi}(\|\boldsymbol{\omega}-\boldsymbol{\omega}'\|_2+a_\phi\,\mathrm{dist}_\Pi(\pi,\pi'))$. By \cref{lem:coordinatewise_gradient_perturbation}, the pseudo-reward coordinates differ by at most $L$ times this amount. For the remaining terms, we use $\|\boldsymbol{\phi}(s,a)\|_2\leq a_\phi$, $\alpha\leq HGZ_{\mathrm{Lin}}$, $\boldsymbol{A}'\preceq I$, $|\sqrt x-\sqrt y|\leq\sqrt{|x-y|}$, and the fact that every clipped value has absolute value at most $HG$, to obtain that
\begin{align}
\frac{\|V_j^{\pi,\boldsymbol{\omega},\boldsymbol{v},\boldsymbol{A},\alpha}-V_j^{\pi',\boldsymbol{\omega}',\boldsymbol{v}',\boldsymbol{A}',\alpha'}\|_\infty}{HG}&\leq(1+a_{\mathbf{T}} a_\phi)\mathrm{dist}_\Pi(\pi,\pi')+a_{\mathbf{T}}\|\boldsymbol{\omega}-\boldsymbol{\omega}'\|_2+a_\phi\|\boldsymbol{v}-\boldsymbol{v}'\|_2\\
&\quad+Z_{\mathrm{Lin}}a_\phi\|\boldsymbol{A}-\boldsymbol{A}'\|_F^{1/2}+\frac{a_\phi|\alpha-\alpha'|}{HG}.
\end{align}
We allocate $\varepsilon/5$ to each of the five terms. We use an internal cover of $\overline{\Pi}$ at radius $\varepsilon/[5(1+a_{\mathbf{T}} a_\phi)]$, and Euclidean covers for $\boldsymbol{\omega}$, $\boldsymbol{v}$, $\boldsymbol{A}$, and $\alpha$ at radii $\varepsilon/(5a_{\mathbf{T}})$, $\varepsilon/(5a_\phi)$, $\varepsilon^2/(25Z_{\mathrm{Lin}}^2a_\phi^2)$, and $HG\varepsilon/(5a_\phi)$, respectively. If $L=0$, the block for $\boldsymbol{\omega}$ is inactive and its cover is omitted. The parameter sets have radii $a_\phi$, $\sqrt K$, $\sqrt{d_\phi}$, and $HGZ_{\mathrm{Lin}}$, and the matrix space has dimension $d_\phi(d_\phi+1)/2$. We apply \cref{lem:euclidean_cover}, multiply the sizes of these covers, and take logarithms to obtain~\cref{eq:lin_off_V_entropy}. At $j=H$, only the blocks for the policy and $\boldsymbol{\omega}$ are needed, and the same bound holds.
\end{proof}

Together with \cref{lem:lin_off_W_cover}, this lemma shows that the policy contributes $d_\Pi$ and the remaining blocks contribute $\mathcal{O}(d_\phi^2)$ parameters, while all radii enter logarithmically. In particular, no column of $\mathbf{T}^\star$ is covered.

\subsection{Theoretical Guarantees} \label{app:lin_estimation_guarantees}

\subsubsection{The Good Event} \label{app:lin_good_events} \label{app:lin_off_good_event}

Throughout this appendix, $\widehat{\ell}_{k,h}=\mathbb{E}_{(s,a)\sim\mu_h^{\pi_k}}\|\boldsymbol{\phi}(s,a)\|_{\boldsymbol{\Gamma}_{k-1,h}^{-1}}$, as in \cref{app:ucgu_off}. For a function $\boldsymbol{g}$ in the class $\mathcal{W}_{t+1}$ of \cref{app:lin_covering_arguments}, define $\widehat{\mathbf{B}}_{k,t}(\boldsymbol{g})\coloneq\boldsymbol{\Gamma}_{k,t}^{-1}\sum_{i=1}^k\boldsymbol{\phi}_{i,t}\boldsymbol{g}(s_{i,t+1})^\top$ for the ridge regression estimate of $\mathbf{B}^\star(\boldsymbol{g})$, and recall $\mathbf{B}^\star(\boldsymbol{g})$ from \cref{lem:linear_bfr_response} and $\boldsymbol{v}^\star(V)$ from \cref{lem:linear_value_response}. A linear MDP has no transition core. We therefore replace $\mathcal{E}_{\mathrm{P}}$ in~\cref{def:good_event} by a confidence bound on $\widehat{\mathbf{T}}_k$.

\begin{definition}[Good Event for Linear MDPs] \label{def:lin_off_good_event}
For radii $\beta_k,\Delta_{k,h},\alpha_{k,h}\geq0$, let $\mathcal{E}\coloneq\mathcal{E}_{\mathrm{P}}\cap\mathcal{E}_{\mathrm{R}}\cap\mathcal{E}_{\mathrm{V}}$, where
\begin{align}
\mathcal{E}_{\mathrm{P}}&\coloneq\Bigl\{\forall k\in[K]\colon\bigl\lVert \boldsymbol{\Lambda}_k^{1/2}(\widehat{\mathbf{T}}_k-\mathbf{T}^\star) \bigr\rVert_F\leq\sqrt{\beta_k}\Bigr\}, \label{eq:lin_off_model_event}\\
\mathcal{E}_{\mathrm{R}}&\coloneq\bigl\{\forall k\in[K],\ h\in[H]\colon\|\widehat{R}_{k,h}-R_{k,h}\|_\infty\leq\Delta_{k,h}\bigr\},\\
\mathcal{E}_{\mathrm{V}}&\coloneq\Bigl\{\forall k\in[K],\ h<H\colon\bigl\lVert \widehat{\boldsymbol{v}}_{k,h}-\boldsymbol{v}^\star(\widehat{V}_{k,h+1}) \bigr\rVert_{\boldsymbol{\Gamma}_{k,h}}\leq\alpha_{k,h}\Bigr\}.
\end{align}
\end{definition}
In particular, we denote the radii for $\mathcal{E}_{\mathrm{P}}$, $\mathcal{E}_{\mathrm{R}}$, and $\mathcal{E}_{\mathrm{V}}$ by $\beta^{\mathrm{Lin}}$, $\Delta_{k,h}^{\mathrm{Lin}}$, and $\alpha_{k,h}^{\mathrm{Lin}}$, respectively. The radii $\beta^{\mathrm{Lin}}$, $(\kappa,\gamma_\rho)$, and $\alpha_{k,h}^{\mathrm{Lin}}$ are chosen in \cref{lem:lin_off_model_event,lem:lin_off_reward_event} and~\cref{eq:lin_value_radius_outline}, respectively, and $\Delta_{k,h}^{\mathrm{Lin}}$ is defined in~\cref{lem:lin_off_reward_error}. \cref{lem:lin_off_model_event} shows that $\mathcal{E}_{\mathrm{P}}$ holds with probability at least $1-\delta/4$. \cref{lem:lin_off_reward_error,lem:lin_off_value_event} show that, under the event $\mathcal{E}_{\mathrm{P}}$, each of $\mathcal{E}_{\mathrm{R}}$ and $\mathcal{E}_{\mathrm{V}}$ holds with probability at least $1-\delta/4$. Only the radii $\alpha_{k,h}^{\mathrm{Lin}}$ are used by the algorithm.

\subsubsection{Transition Model Estimation} \label{app:lin_tme_guarantees}

\begin{lemma}[Sparse Transition Confidence] \label{lem:lin_off_model_event} \label{lem:lin_off_model_event_formal}
If we choose $\beta^{\mathrm{Lin}}=\mathcal{O}((C_{\mathbf{T}}^2+\log(1+KHC_\phi^2)+\log(K/\delta))d_\phi)$ with a sufficiently large constant, then $\mathcal{E}_{\mathrm{P}}$ holds with probability at least $1-\delta/4$.
\end{lemma}

\begin{proof}
We first recall the relevant notation. \STE{} computes $\widehat{\mathbf{T}}_k$ from the $k$ replay trajectories by the ridge regression in~\cref{eq:ste_estimator}, with the Gram matrix $\boldsymbol{\Lambda}_k=I+\sum_{i=1}^k\sum_{h=1}^H\boldsymbol{\phi}_{i,h}\boldsymbol{\phi}_{i,h}^\top$. The event $\mathcal{E}_{\mathrm{P}}$ requires $\|\boldsymbol{\Lambda}_k^{1/2}(\widehat{\mathbf{T}}_k-\mathbf{T}^\star)\|_F\leq\sqrt{\beta^{\mathrm{Lin}}}$ for every $k\in[K]$.

We take
\begin{align}
\beta^{\mathrm{Lin}}&\coloneq\Bigl(C_{\mathbf{T}}\sqrt{d_\phi}+2\sqrt{d_\phi\log(1+KHC_\phi^2)+2\log(4K/\delta)}\Bigr)^2, \label{eq:lin_model_radius}
\end{align}
which has the stated order. Consider a specific $k\in[K]$. We first decompose the matrix error into the error in the observed transitions and the contribution of regularization. We then bound the two terms separately.

\paragraph{Decomposing the Matrix Error}
By~\cref{eq:ste_estimator}, $\widehat{\mathbf{T}}_k$ satisfies the normal equations
\begin{align}
\boldsymbol{\Lambda}_k\widehat{\mathbf{T}}_k=\sum_{i=1}^k\sum_{h=1}^H\boldsymbol{\phi}_{i,h}\boldsymbol{e}_{s_{i,h+1}}^\top.
\end{align}
For every $i\in[k]$ and $h\in[H]$, let $\boldsymbol{y}_{i,h}\coloneq\boldsymbol{e}_{s_{i,h+1}}-P^\star(\cdot\mid s_{i,h},a_{i,h})\in\mathbb{R}^{|\mathcal{S}|}$.
We obtain that
\begin{align}
\bigl\lVert \boldsymbol{\Lambda}_k^{1/2}(\widehat{\mathbf{T}}_k-\mathbf{T}^\star) \bigr\rVert_F&=\Bigl\lVert \boldsymbol{\Lambda}_k^{-1/2}\bigl(\boldsymbol{\Lambda}_k\widehat{\mathbf{T}}_k-\boldsymbol{\Lambda}_k\mathbf{T}^\star\bigr) \Bigr\rVert_F\nonumber\\
&=\Biggl\lVert \boldsymbol{\Lambda}_k^{-1/2}\biggl(\sum_{i=1}^k\sum_{h=1}^H\boldsymbol{\phi}_{i,h}\boldsymbol{e}_{s_{i,h+1}}^\top-\boldsymbol{\Lambda}_k\mathbf{T}^\star\biggr) \Biggr\rVert_F\nonumber\\
&=\Biggl\lVert \boldsymbol{\Lambda}_k^{-1/2}\biggl[\sum_{i=1}^k\sum_{h=1}^H\boldsymbol{\phi}_{i,h}\bigl(\boldsymbol{e}_{s_{i,h+1}}-(\mathbf{T}^\star)^\top\boldsymbol{\phi}_{i,h}\bigr)^\top-\mathbf{T}^\star\biggr] \Biggr\rVert_F\nonumber\\
&=\biggl\lVert \boldsymbol{\Lambda}_k^{-1/2}\sum_{i=1}^k\sum_{h=1}^H\boldsymbol{\phi}_{i,h}\boldsymbol{y}_{i,h}^\top-\boldsymbol{\Lambda}_k^{-1/2}\mathbf{T}^\star \biggr\rVert_F\nonumber\\
&\leq\biggl\lVert \boldsymbol{\Lambda}_k^{-1/2}\sum_{i=1}^k\sum_{h=1}^H\boldsymbol{\phi}_{i,h}\boldsymbol{y}_{i,h}^\top \biggr\rVert_F+\bigl\lVert \boldsymbol{\Lambda}_k^{-1/2}\mathbf{T}^\star \bigr\rVert_F. \label{eq:linear_matrix_error_decomposition}
\end{align}
In the first equality, we use $\boldsymbol{\Lambda}_k^{1/2}=\boldsymbol{\Lambda}_k^{-1/2}\boldsymbol{\Lambda}_k$. In the second equality, we substitute the normal equations. In the third equality, we expand $\boldsymbol{\Lambda}_k\mathbf{T}^\star=\mathbf{T}^\star+\sum_{i=1}^k\sum_{h=1}^H\boldsymbol{\phi}_{i,h}\boldsymbol{\phi}_{i,h}^\top\mathbf{T}^\star$, which leaves the term $-\mathbf{T}^\star$ from the regularization. In the fourth equality, we use $(\mathbf{T}^\star)^\top\boldsymbol{\phi}_{i,h}=P^\star(\cdot\mid s_{i,h},a_{i,h})$ from~\cref{def:linear_mdp}. In the last step, we apply the triangle inequality. The first term in~\cref{eq:linear_matrix_error_decomposition} is the error in the observed transitions, and the second term is the contribution of regularization.

\paragraph{Bounding the Error in the Observed Transitions}
We order the $kH$ transitions in $\mathcal{D}_k$ by trajectory and then by step. Let $\mathcal{H}_{i,h}^{-}$ be the history of the earlier transitions together with $(s_{i,h},a_{i,h})$. The vector $\boldsymbol{\phi}_{i,h}$ is fixed given $\mathcal{H}_{i,h}^{-}$, and we apply \cref{lem:linear_regression_response} to obtain that $\mathbb{E}[\boldsymbol{y}_{i,h}\mid\mathcal{H}_{i,h}^{-}]=\boldsymbol{0}$. Moreover, $\boldsymbol{e}_{s_{i,h+1}}$ has a one at $s_{i,h+1}$ and zeros elsewhere. Hence,
\begin{align}
\|\boldsymbol{y}_{i,h}\|_2^2&=\bigl(1-P^\star(s_{i,h+1}\mid s_{i,h},a_{i,h})\bigr)^2+\sum_{s'\neq s_{i,h+1}}P^\star(s'\mid s_{i,h},a_{i,h})^2\nonumber\\
&=1-2P^\star(s_{i,h+1}\mid s_{i,h},a_{i,h})+\sum_{s'\in\mathcal{S}}P^\star(s'\mid s_{i,h},a_{i,h})^2\leq2,
\end{align}
where we use $\sum_{s'\in\mathcal{S}}P^\star(s'\mid s_{i,h},a_{i,h})^2\leq\sum_{s'\in\mathcal{S}}P^\star(s'\mid s_{i,h},a_{i,h})=1$ in the last inequality. In particular, $\|\boldsymbol{y}_{i,h}\|_2\leq2$ for any number of states, since each observed transition contributes a single one-hot vector. Moreover, $\|\boldsymbol{\phi}_{i,h}\|_2\leq C_\phi\sqrt{d_\phi}$ by~\cref{def:linear_mdp}, and $\boldsymbol{\Lambda}_k=I+\sum_{i=1}^k\sum_{h=1}^H\boldsymbol{\phi}_{i,h}\boldsymbol{\phi}_{i,h}^\top$. We apply \cref{lem:lin_matrix_self_normalized} with the covariates $\boldsymbol{\phi}_{i,h}$, the vectors $\boldsymbol{y}_{i,h}$, $R_x=C_\phi\sqrt{d_\phi}$, $R=2$, $n=kH$, and $\zeta=\delta/(4K)$ to obtain that, with probability at least $1-\delta/(4K)$,
\begin{align}
\biggl\lVert \boldsymbol{\Lambda}_k^{-1/2}\sum_{i=1}^k\sum_{h=1}^H\boldsymbol{\phi}_{i,h}\boldsymbol{y}_{i,h}^\top \biggr\rVert_F&\leq2\sqrt{d_\phi\log(1+kHC_\phi^2)+2\log(4K/\delta)}\nonumber\\
&\leq2\sqrt{d_\phi\log(1+KHC_\phi^2)+2\log(4K/\delta)}. \label{eq:linear_observation_error}
\end{align}
Since \cref{lem:lin_matrix_self_normalized} only requires the conditional mean of $\boldsymbol{y}_{i,h}$ to be zero, the behavior policies can depend on the history. Moreover, the bound does not depend on $|\mathcal{S}|$.

\paragraph{Bounding the Contribution of Regularization}
Let $\{\sigma_s\}_{s\in\mathcal{S}}$ be independent random signs, each equal to $1$ or $-1$ with equal probability. These signs are used only in the analysis. Since $\mathbb{E}_\sigma[\sigma_s\sigma_{s'}]=0$ for $s\neq s'$, we obtain that
\begin{align}
\lVert \mathbf{T}^\star \rVert_F^2=\sum_{s\in\mathcal{S}}\bigl\lVert \mathbf{T}^\star(s) \bigr\rVert_2^2=\mathbb{E}_\sigma\biggl\lVert \sum_{s\in\mathcal{S}}\sigma_s\mathbf{T}^\star(s) \biggr\rVert_2^2\leq C_{\mathbf{T}}^2d_\phi,
\end{align}
where we apply condition (ii) of \cref{def:linear_mdp} with $g(s)=\sigma_s$, which satisfies $\|g\|_\infty=1$, in the last inequality. Since $\boldsymbol{\Lambda}_k\succeq I$, we have $\boldsymbol{\Lambda}_k^{-1}\preceq I$. Hence,
\begin{align}
\bigl\lVert \boldsymbol{\Lambda}_k^{-1/2}\mathbf{T}^\star \bigr\rVert_F^2=\operatorname{tr}\bigl((\mathbf{T}^\star)^\top\boldsymbol{\Lambda}_k^{-1}\mathbf{T}^\star\bigr)\leq\operatorname{tr}\bigl((\mathbf{T}^\star)^\top\mathbf{T}^\star\bigr)=\lVert \mathbf{T}^\star \rVert_F^2\leq C_{\mathbf{T}}^2d_\phi. \label{eq:linear_regularization_bias}
\end{align}

\paragraph{Putting Everything Together}
We substitute~\cref{eq:linear_observation_error,eq:linear_regularization_bias} into~\cref{eq:linear_matrix_error_decomposition} to obtain that, with probability at least $1-\delta/(4K)$,
\begin{align}
\bigl\lVert \boldsymbol{\Lambda}_k^{1/2}(\widehat{\mathbf{T}}_k-\mathbf{T}^\star) \bigr\rVert_F\leq C_{\mathbf{T}}\sqrt{d_\phi}+2\sqrt{d_\phi\log(1+KHC_\phi^2)+2\log(4K/\delta)}=\sqrt{\beta^{\mathrm{Lin}}}.
\end{align}
We take a union bound over $k\in[K]$ to show that $\mathcal{E}_{\mathrm{P}}$ holds with probability at least $1-\delta/4$. We never take a union bound over states. Hence, $\beta^{\mathrm{Lin}}$ does not depend on $|\mathcal{S}|$.
\end{proof}

The following lemma shows that the matrix error in $\mathcal{E}_{\mathrm{P}}$ controls the error at every state, including the states that do not appear in $\mathcal{D}_k$.

\begin{lemma}[Transition Error at Every State] \label{lem:lin_column_error}
For every $k\in[K]$, $s\in\mathcal{S}$, and $\boldsymbol{\omega}\in\mathbb{R}^{d_\phi}$,
\begin{align}
\bigl\lVert \widehat{\mathbf{T}}_k(s)-\mathbf{T}^\star(s) \bigr\rVert_{\boldsymbol{\Lambda}_k}&\leq\bigl\lVert \boldsymbol{\Lambda}_k^{1/2}(\widehat{\mathbf{T}}_k-\mathbf{T}^\star) \bigr\rVert_F, \label{eq:lin_column_error}\\
\Bigl|\boldsymbol{\omega}^\top\big(\widehat{\mathbf{T}}_k(s)-\mathbf{T}^\star(s)\big)\Bigr|&\leq\lVert \boldsymbol{\omega} \rVert_{\boldsymbol{\Lambda}_k^{-1}}\bigl\lVert \boldsymbol{\Lambda}_k^{1/2}(\widehat{\mathbf{T}}_k-\mathbf{T}^\star) \bigr\rVert_F. \label{eq:lin_column_prediction_error}
\end{align}
\end{lemma}

\begin{proof}
Consider specific $k\in[K]$ and $s\in\mathcal{S}$. The matrices $\mathbf{T}^\star$ and $\widehat{\mathbf{T}}_k$ have the columns $\mathbf{T}^\star(s')$ and $\widehat{\mathbf{T}}_k(s')$ for $s'\in\mathcal{S}$, respectively. Hence, the column of $\boldsymbol{\Lambda}_k^{1/2}(\widehat{\mathbf{T}}_k-\mathbf{T}^\star)$ indexed by $s'$ is $\boldsymbol{\Lambda}_k^{1/2}(\widehat{\mathbf{T}}_k(s')-\mathbf{T}^\star(s'))$. Since the squared Frobenius norm of a matrix is the sum of the squared Euclidean norms of its columns, we obtain that
\begin{align}
\bigl\lVert \widehat{\mathbf{T}}_k(s)-\mathbf{T}^\star(s) \bigr\rVert_{\boldsymbol{\Lambda}_k}^2&=\Bigl\lVert \boldsymbol{\Lambda}_k^{1/2}\big(\widehat{\mathbf{T}}_k(s)-\mathbf{T}^\star(s)\big) \Bigr\rVert_2^2\nonumber\\
&\leq\sum_{s'\in\mathcal{S}}\Bigl\lVert \boldsymbol{\Lambda}_k^{1/2}\big(\widehat{\mathbf{T}}_k(s')-\mathbf{T}^\star(s')\big) \Bigr\rVert_2^2=\bigl\lVert \boldsymbol{\Lambda}_k^{1/2}(\widehat{\mathbf{T}}_k-\mathbf{T}^\star) \bigr\rVert_F^2,
\end{align}
which proves~\cref{eq:lin_column_error}. For~\cref{eq:lin_column_prediction_error}, we use the identity $\boldsymbol{\omega}^\top(\widehat{\mathbf{T}}_k(s)-\mathbf{T}^\star(s))=(\boldsymbol{\Lambda}_k^{-1/2}\boldsymbol{\omega})^\top\boldsymbol{\Lambda}_k^{1/2}(\widehat{\mathbf{T}}_k(s)-\mathbf{T}^\star(s))$ and apply the Cauchy--Schwarz inequality to obtain that
\begin{align}
\Bigl|\boldsymbol{\omega}^\top\big(\widehat{\mathbf{T}}_k(s)-\mathbf{T}^\star(s)\big)\Bigr|&\leq\bigl\lVert \boldsymbol{\Lambda}_k^{-1/2}\boldsymbol{\omega} \bigr\rVert_2\Bigl\lVert \boldsymbol{\Lambda}_k^{1/2}\big(\widehat{\mathbf{T}}_k(s)-\mathbf{T}^\star(s)\big) \Bigr\rVert_2\nonumber\\
&=\lVert \boldsymbol{\omega} \rVert_{\boldsymbol{\Lambda}_k^{-1}}\bigl\lVert \widehat{\mathbf{T}}_k(s)-\mathbf{T}^\star(s) \bigr\rVert_{\boldsymbol{\Lambda}_k}\leq\lVert \boldsymbol{\omega} \rVert_{\boldsymbol{\Lambda}_k^{-1}}\bigl\lVert \boldsymbol{\Lambda}_k^{1/2}(\widehat{\mathbf{T}}_k-\mathbf{T}^\star) \bigr\rVert_F,
\end{align}
where we apply~\cref{eq:lin_column_error} in the last inequality. Both bounds hold deterministically for every $\boldsymbol{\omega}$. Hence, they apply to $\widehat{\boldsymbol{\omega}}_{k,h}$, although $\widehat{\boldsymbol{\omega}}_{k,h}$ is computed from the same replay data.
\end{proof}

\subsubsection{Pseudo-Reward Estimation} \label{app:lin_pre_guarantees}

The main result of this subsubsection is the following lemma, which bounds the pseudo-reward error and the weighted norms of $\widehat{\boldsymbol{\omega}}_{k,j}$. Its proof uses the backward-regression and initial-state confidence bounds in \cref{lem:lin_off_reward_event}.

\begin{lemma}[Pseudo-Reward Error] \label{lem:lin_off_reward_error} \label{lem:lin_off_reward_confidence}
Let $\beta^{\mathrm{Lin}}$ be as in \cref{lem:lin_off_model_event}, and let $\kappa$ and $\gamma_\rho$ be as in \cref{lem:lin_off_reward_event}. Under the event $\mathcal{E}_{\mathrm{P}}$, with probability at least $1-\delta/4$, for all $k\in[K]$ and $h\in[H]$,
\begin{align}
\MoveEqLeft\bigl\lVert \widehat{R}_{k,h}-R_{k,h} \bigr\rVert_\infty \leq\Delta_{k,h}^{\mathrm{Lin}}\nonumber\\
&\coloneq\begin{cases}0,&h=1,\\[2pt] L\Biggl[\sqrt{\beta^{\mathrm{Lin}}}\,\widehat{\ell}_{k,h-1}+(\sqrt{\beta^{\mathrm{Lin}}}+C_{\mathbf{T}}\sqrt{d_\phi})\biggl(\dfrac{\gamma_\rho}{\sqrt k}+\kappa\displaystyle\sum_{t=1}^{h-2}\widehat{\ell}_{k,t}\biggr)\Biggr],&h\in\{2,\ldots,H\},\end{cases} \label{eq:lin_off_Delta}
\end{align}
and, for all $k\in[K]$ and $j\in[H-1]$,
\begin{align}
\bigl\lVert \widehat{\boldsymbol{\omega}}_{k,j} \bigr\rVert_{\boldsymbol{\Lambda}_k^{-1}} &\leq\widehat{\ell}_{k,j} +\frac{\gamma_\rho}{\sqrt k} +\kappa\sum_{t=1}^{j-1}\widehat{\ell}_{k,t}. \label{eq:lin_off_weighted_feature_bound}
\end{align}
\end{lemma}

\begin{proof}[Proof of \cref{lem:lin_off_reward_error}]
We first recall the relevant notation. For $h\geq2$, $\OCUPOLIN$ estimates the occupancy measure by $\widehat{\mu}_{k,h}(s,a)=\operatorname{Clip}(\pi_{k,h}(a\mid s)\widehat{\boldsymbol{\omega}}_{k,h-1}^\top\widehat{\mathbf{T}}_k(s),[0,1])$ in~\cref{eq:lin_occupancy_reconstruction}, where $\widehat{\boldsymbol{\omega}}_{k,h-1}$ is returned by \BFR{}. By~\cref{eq:lin_occupancy_decomposition}, the true occupancy measure satisfies $\mu_h^{\pi_k}(s,a)=\pi_{k,h}(a\mid s)(\boldsymbol{\omega}_{h-1}^{\pi_k})^\top\mathbf{T}^\star(s)$. The expected feature norm $\widehat{\ell}_{k,t}$ is defined in~\cref{subsec:off_policy_proof_outline}.

By \cref{lem:lin_off_reward_event}, with probability at least $1-\delta/4$, its two bounds hold for every $k\in[K]$. We work on the intersection of this event with $\mathcal{E}_{\mathrm{P}}$. At $h=1$, $\widehat{\mu}_{k,1}(s,a)=\rho(s)\pi_{k,1}(a\mid s)=\mu_1^{\pi_k}(s,a)$. Hence, $\widehat{R}_{k,1}=R_{k,1}$, matching $\Delta_{k,1}^{\mathrm{Lin}}=0$. Now consider a specific $h\in\{2,\ldots,H\}$.

\paragraph{Splitting the Error}
Both the true coordinate $\mu_h^{\pi_k}(s,a)$ and its clipped estimate lie in $[0,1]$. Hence, clipping cannot increase their distance, and multiplying by $\pi_{k,h}(a\mid s)\in[0,1]$ cannot increase it either. We apply \cref{lem:coordinatewise_gradient_perturbation}, \cref{lem:linear_occupancy_measure}, and~\cref{eq:lin_occupancy_reconstruction}, and add and subtract $\widehat{\boldsymbol{\omega}}_{k,h-1}^\top\mathbf{T}^\star(s)$, to obtain that
\begin{align}
\|\widehat{R}_{k,h}-R_{k,h}\|_\infty&\leq L\|\widehat{\mu}_{k,h}-\mu_h^{\pi_k}\|_\infty\leq L\sup_s\bigl|\widehat{\boldsymbol{\omega}}_{k,h-1}^\top\widehat{\mathbf{T}}_k(s)-(\boldsymbol{\omega}_{h-1}^{\pi_k})^\top\mathbf{T}^\star(s)\bigr|\\
&\leq L\Bigl[\underbrace{\sup_s\bigl|\widehat{\boldsymbol{\omega}}_{k,h-1}^\top(\widehat{\mathbf{T}}_k(s)-\mathbf{T}^\star(s))\bigr|}_{\text{(I)}}+\underbrace{\sup_s\bigl|(\widehat{\boldsymbol{\omega}}_{k,h-1}-\boldsymbol{\omega}_{h-1}^{\pi_k})^\top\mathbf{T}^\star(s)\bigr|}_{\text{(II)}}\Bigr]. \label{eq:linear_reward_error_decomposition}
\end{align}
Term (I) is the error in the estimated columns of $\mathbf{T}^\star$, and Term (II) is the error in estimating $\boldsymbol{\omega}_{h-1}^{\pi_k}$.

\paragraph{Bounding Term (I)}
On the event $\mathcal{E}_{\mathrm{P}}$, we apply~\cref{eq:lin_column_prediction_error} in \cref{lem:lin_column_error} with $\boldsymbol{\omega}=\widehat{\boldsymbol{\omega}}_{k,h-1}$ and the definition of $\mathcal{E}_{\mathrm{P}}$ to obtain that
\begin{align}
\mathrm{Term~(I)}&\leq\sqrt{\beta^{\mathrm{Lin}}}\|\widehat{\boldsymbol{\omega}}_{k,h-1}\|_{\boldsymbol{\Lambda}_k^{-1}}. \label{eq:linear_reward_transition_term}
\end{align}

\paragraph{Bounding Term (II)}
For a fixed state $s$, we apply condition (ii) of \cref{def:linear_mdp} with $g(s')=\mathbf{1}\{s'=s\}$ to obtain that $\|\mathbf{T}^\star(s)\|_2\leq C_{\mathbf{T}}\sqrt{d_\phi}$. We then apply the Cauchy--Schwarz inequality to obtain that
\begin{align}
\mathrm{Term~(II)}&\leq C_{\mathbf{T}}\sqrt{d_\phi}\|\widehat{\boldsymbol{\omega}}_{k,h-1}-\boldsymbol{\omega}_{h-1}^{\pi_k}\|_2. \label{eq:linear_reward_feature_term}
\end{align}

\paragraph{Bounding the Error in Estimating $\boldsymbol{\omega}_j^{\pi_k}$}
We now show that, for every $k$ and $j<H$,
\begin{align}
\bigl\lVert \widehat{\boldsymbol{\omega}}_{k,j}-\boldsymbol{\omega}_j^{\pi_k} \bigr\rVert_2 &\leq\frac{\gamma_\rho}{\sqrt k} +\kappa\sum_{t=1}^{j-1}\widehat{\ell}_{k,t}. \label{eq:lin_off_feature_propagation}
\end{align}
The argument is the same as in the proof of \cref{lem:off_reward_error_main}. For $t<j$, define
\begin{align}
\boldsymbol{\chi}_{j,t}^{\pi_k}(s,a) \coloneq\widehat{\boldsymbol{q}}_{j,t}^{\pi_k}(s,a)-P^\star\widehat{\boldsymbol{g}}_{j,t+1}^{\pi_k}(s,a)\qquad\text{and}\qquad\boldsymbol{\delta}_{j,t}^{\pi_k}(s) \coloneq\widehat{\boldsymbol{g}}_{j,t}^{\pi_k}(s)-\boldsymbol{g}_{j,t}^{\pi_k}(s),
\end{align}
with $\boldsymbol{\delta}_{j,j}^{\pi_k}=\boldsymbol{0}$. By \cref{lem:linear_bfr_response}, $P^\star\widehat{\boldsymbol{g}}_{j,t+1}^{\pi_k}(s,a)=\mathbf{B}^\star(\widehat{\boldsymbol{g}}_{j,t+1}^{\pi_k})^\top\boldsymbol{\phi}(s,a)$. This vector is an average of vectors in $\mathbb{B}_2(C_\phi\sqrt{d_\phi})$ and hence belongs to this ball. Projection onto the ball is nonexpansive and fixes this vector. Moreover, $\widehat{\mathbf{B}}_{j,t}^{\pi_k}=\widehat{\mathbf{B}}_{k,t}(\widehat{\boldsymbol{g}}_{j,t+1}^{\pi_k})$ with $\widehat{\boldsymbol{g}}_{j,t+1}^{\pi_k}\in\mathcal{W}_{t+1}$. We apply the first bound of~\cref{lem:lin_off_reward_event}, the Cauchy--Schwarz inequality, and $\boldsymbol{\Gamma}_{k,t}\succeq\boldsymbol{\Gamma}_{k-1,t}$ to obtain that
\begin{align}
\bigl\lVert \boldsymbol{\chi}_{j,t}^{\pi_k}(s,a) \bigr\rVert_2 &\leq\Bigl\lVert \bigl(\widehat{\mathbf{B}}_{j,t}^{\pi_k} -\mathbf{B}^\star(\widehat{\boldsymbol{g}}_{j,t+1}^{\pi_k})\bigr)^\top \boldsymbol{\phi}(s,a) \Bigr\rVert_2\leq\kappa \bigl\lVert \boldsymbol{\phi}(s,a) \bigr\rVert_{\boldsymbol{\Gamma}_{k,t}^{-1}} \leq\kappa \bigl\lVert \boldsymbol{\phi}(s,a) \bigr\rVert_{\boldsymbol{\Gamma}_{k-1,t}^{-1}}.
\end{align}
Using \cref{line:rfe_policy_average} of \cref{alg:rfe} and \cref{lem:backward_feature_recursion}, we add and subtract $P^\star\widehat{\boldsymbol{g}}_{j,t+1}^{\pi_k}(s,a)$ to obtain that $\boldsymbol{\delta}_{j,t}^{\pi_k}(s)=\sum_a\pi_{k,t}(a\mid s)[\boldsymbol{\chi}_{j,t}^{\pi_k}(s,a)+P^\star\boldsymbol{\delta}_{j,t+1}^{\pi_k}(s,a)]$. Averaging over $s\sim d_t^{\pi_k}$ and using the occupancy flow equation, we obtain $\mathbb{E}_{d_t^{\pi_k}}\boldsymbol{\delta}_{j,t}^{\pi_k}=\mathbb{E}_{\mu_t^{\pi_k}}\boldsymbol{\chi}_{j,t}^{\pi_k}+\mathbb{E}_{d_{t+1}^{\pi_k}}\boldsymbol{\delta}_{j,t+1}^{\pi_k}$. Iterating from $t=1$ to $j-1$ and using $d_1^{\pi_k}=\rho$ and $\mathbb{E}_\rho\boldsymbol{g}_{j,1}^{\pi_k}=\boldsymbol{\omega}_j^{\pi_k}$, we apply the triangle inequality and Jensen's inequality to obtain that
\begin{align}
\bigl\lVert \mathbb{E}_\rho[\widehat{\boldsymbol{g}}_{j,1}^{\pi_k}(s_1)]-\boldsymbol{\omega}_j^{\pi_k} \bigr\rVert_2 \leq\sum_{t=1}^{j-1}\mathbb{E}_{\mu_t^{\pi_k}} \bigl\lVert \boldsymbol{\chi}_{j,t}^{\pi_k}(s,a) \bigr\rVert_2 \leq\kappa\sum_{t=1}^{j-1}\widehat{\ell}_{k,t}.
\end{align}
Since $\widehat{\boldsymbol{\omega}}_{k,j}=k^{-1}\sum_{i=1}^k\widehat{\boldsymbol{g}}_{j,1}^{\pi_k}(s_{i,1})$ and $\widehat{\boldsymbol{g}}_{j,1}^{\pi_k}\in\mathcal{W}_1$, adding and subtracting $\mathbb{E}_\rho\widehat{\boldsymbol{g}}_{j,1}^{\pi_k}$ and applying the second bound of~\cref{lem:lin_off_reward_event} proves~\cref{eq:lin_off_feature_propagation}.

For~\cref{eq:lin_off_weighted_feature_bound}, we apply the triangle inequality and $\boldsymbol{\Lambda}_k\succeq I$ to obtain that $\|\widehat{\boldsymbol{\omega}}_{k,j}\|_{\boldsymbol{\Lambda}_k^{-1}}\leq\|\boldsymbol{\omega}_j^{\pi_k}\|_{\boldsymbol{\Lambda}_k^{-1}}+\|\widehat{\boldsymbol{\omega}}_{k,j}-\boldsymbol{\omega}_j^{\pi_k}\|_2$. The matrix $\boldsymbol{\Lambda}_k$ contains the step-$j$ features of the first $k-1$ trajectories. Hence, $\boldsymbol{\Lambda}_k\succeq\boldsymbol{\Gamma}_{k-1,j}$, and we apply Jensen's inequality to obtain that $\|\boldsymbol{\omega}_j^{\pi_k}\|_{\boldsymbol{\Lambda}_k^{-1}}\leq\mathbb{E}_{\mu_j^{\pi_k}}\|\boldsymbol{\phi}(s,a)\|_{\boldsymbol{\Gamma}_{k-1,j}^{-1}}=\widehat{\ell}_{k,j}$. Substituting~\cref{eq:lin_off_feature_propagation} proves~\cref{eq:lin_off_weighted_feature_bound}.

\paragraph{Putting Everything Together}
Substituting~\cref{eq:linear_reward_transition_term,eq:linear_reward_feature_term} into~\cref{eq:linear_reward_error_decomposition}, and then applying~\cref{eq:lin_off_weighted_feature_bound,eq:lin_off_feature_propagation} with $j=h-1$, we obtain
\begin{align}
\|\widehat{R}_{k,h}-R_{k,h}\|_\infty&\leq L\Biggl[\sqrt{\beta^{\mathrm{Lin}}}\,\widehat{\ell}_{k,h-1}+(\sqrt{\beta^{\mathrm{Lin}}}+C_{\mathbf{T}}\sqrt{d_\phi})\biggl(\frac{\gamma_\rho}{\sqrt k}+\kappa\sum_{t=1}^{h-2}\widehat{\ell}_{k,t}\biggr)\Biggr]=\Delta_{k,h}^{\mathrm{Lin}}.
\end{align}
All these bounds hold simultaneously on the event above, which proves the lemma.
\end{proof}

The following lemma establishes the backward-regression and initial-state confidence bounds used in the proof of \cref{lem:lin_off_reward_error}.

\begin{lemma}[Backward-Regression and Initial-State Confidence] \label{lem:lin_off_reward_event} \label{lem:lin_off_reward_event_formal}
If we choose
\begin{align}
\kappa=\widetilde{\mathcal{O}}\Bigl(\sqrt{d_\phi(d_\phi^2+d_\Pi)}\Bigr)\qquad\text{and}\qquad\gamma_\rho=\widetilde{\mathcal{O}}\Bigl(\sqrt{d_\phi(d_\phi^2+d_\Pi)}\Bigr)
\end{align}
with sufficiently large constants, then, with probability at least $1-\delta/4$, for every $k\in[K]$,
\begin{align}
\sup_{\boldsymbol{g}\in\mathcal{W}_{t+1}}\bigl\lVert \boldsymbol{\Gamma}_{k,t}^{1/2}(\widehat{\mathbf{B}}_{k,t}(\boldsymbol{g})-\mathbf{B}^\star(\boldsymbol{g})) \bigr\rVert_F&\leq\kappa,\qquad t\in[H-2], \label{eq:lin_bfr_regression_confidence}\\
\sup_{\boldsymbol{g}\in\mathcal{W}_1}\bigl\lVert k^{-1}\textstyle\sum_{i=1}^k\boldsymbol{g}(s_{i,1})-\mathbb{E}_\rho[\boldsymbol{g}(s_1)] \bigr\rVert_2&\leq\gamma_\rho/\sqrt k. \label{eq:lin_bfr_initial_confidence}
\end{align}
\end{lemma}

\begin{proof}[Proof of \cref{lem:lin_off_reward_event}]
We first recall the relevant notation. For a target $\boldsymbol{g}\in\mathcal{W}_{t+1}$, the step-$t$ backward regression fits $\widehat{\mathbf{B}}_{k,t}(\boldsymbol{g})=\boldsymbol{\Gamma}_{k,t}^{-1}\sum_{i=1}^k\boldsymbol{\phi}_{i,t}\boldsymbol{g}(s_{i,t+1})^\top$ on the replay data, and~\cref{lem:linear_bfr_response} gives $P^\star\boldsymbol{g}(s,a)=\mathbf{B}^\star(\boldsymbol{g})^\top\boldsymbol{\phi}(s,a)$ with $\mathbf{B}^\star(\boldsymbol{g})=\sum_{s'}\mathbf{T}^\star(s')\boldsymbol{g}(s')^\top$. The classes $\mathcal{W}_t$ are defined in~\cref{app:lin_covering_arguments}.

With $\mathcal{H}_{\mathrm{W}}$ from~\cref{eq:lin_off_W_entropy}, we take
\begin{align}
\kappa&\coloneq C_{\mathbf{T}} C_\phi d_\phi^{3/2}+C_\phi\sqrt{d_\phi}\biggl[2\sqrt{d_\phi\log(1+KC_\phi^2)+2\bigl(\mathcal{H}_{\mathrm{W}}(1/K)+\log(12KH/\delta)\bigr)}+2\biggr], \label{eq:lin_bfr_radius}\\
\gamma_\rho&\coloneq4C_\phi\sqrt{d_\phi}\sqrt{\log(1+K)+2\bigl(\mathcal{H}_{\mathrm{W}}(1/K)+\log(12KH/\delta)\bigr)+1}. \label{eq:lin_initial_radius}
\end{align}
Under \cref{asm:policy_class}, $\mathcal{H}_{\mathrm{W}}(1/K)=\widetilde{\mathcal{O}}(d_\phi^2+d_\Pi)$. Since $d_\phi^{3/2}\leq\sqrt{d_\phi(d_\phi^2+d_\Pi)}$, these choices have the stated orders. Each class $\mathcal{W}_t$ is fixed before data collection. Hence, its finite cover can be chosen before applying a concentration inequality.

\paragraph{Backward Regression}
Consider specific $k$ and $t\leq H-2$, and consider a specific $\boldsymbol{g}\in\mathcal{W}_{t+1}$. For every $i\in[k]$, let
\begin{align}
\boldsymbol{\varepsilon}_{i,t}(\boldsymbol{g})\coloneq\boldsymbol{g}(s_{i,t+1})-P^\star\boldsymbol{g}(s_{i,t},a_{i,t})\in\mathbb{R}^{d_\phi}.
\end{align}
By \cref{lem:linear_bfr_response}, $P^\star\boldsymbol{g}(s_{i,t},a_{i,t})=\mathbf{B}^\star(\boldsymbol{g})^\top\boldsymbol{\phi}_{i,t}$. We obtain that
\begin{align}
\widehat{\mathbf{B}}_{k,t}(\boldsymbol{g})-\mathbf{B}^\star(\boldsymbol{g})&=\boldsymbol{\Gamma}_{k,t}^{-1}\biggl(\sum_{i=1}^k\boldsymbol{\phi}_{i,t}\boldsymbol{g}(s_{i,t+1})^\top-\boldsymbol{\Gamma}_{k,t}\mathbf{B}^\star(\boldsymbol{g})\biggr)\nonumber\\
&=\boldsymbol{\Gamma}_{k,t}^{-1}\biggl(\sum_{i=1}^k\boldsymbol{\phi}_{i,t}\bigl(\mathbf{B}^\star(\boldsymbol{g})^\top\boldsymbol{\phi}_{i,t}+\boldsymbol{\varepsilon}_{i,t}(\boldsymbol{g})\bigr)^\top-\boldsymbol{\Gamma}_{k,t}\mathbf{B}^\star(\boldsymbol{g})\biggr)\nonumber\\
&=\boldsymbol{\Gamma}_{k,t}^{-1}\biggl(\sum_{i=1}^k\boldsymbol{\phi}_{i,t}\boldsymbol{\phi}_{i,t}^\top\mathbf{B}^\star(\boldsymbol{g})+\sum_{i=1}^k\boldsymbol{\phi}_{i,t}\boldsymbol{\varepsilon}_{i,t}(\boldsymbol{g})^\top-\mathbf{B}^\star(\boldsymbol{g})-\sum_{i=1}^k\boldsymbol{\phi}_{i,t}\boldsymbol{\phi}_{i,t}^\top\mathbf{B}^\star(\boldsymbol{g})\biggr)\nonumber\\
&=\boldsymbol{\Gamma}_{k,t}^{-1}\sum_{i=1}^k\boldsymbol{\phi}_{i,t}\boldsymbol{\varepsilon}_{i,t}(\boldsymbol{g})^\top-\boldsymbol{\Gamma}_{k,t}^{-1}\mathbf{B}^\star(\boldsymbol{g}).
\end{align}
In the first equality, we use the definition of $\widehat{\mathbf{B}}_{k,t}(\boldsymbol{g})$ and $\boldsymbol{\Gamma}_{k,t}^{-1}\boldsymbol{\Gamma}_{k,t}=I$. In the second equality, we use the decomposition $\boldsymbol{g}(s_{i,t+1})=P^\star\boldsymbol{g}(s_{i,t},a_{i,t})+\boldsymbol{\varepsilon}_{i,t}(\boldsymbol{g})$ and apply \cref{lem:linear_bfr_response}. In the third equality, we expand $\boldsymbol{\Gamma}_{k,t}=I+\sum_{i=1}^k\boldsymbol{\phi}_{i,t}\boldsymbol{\phi}_{i,t}^\top$. In the last equality, the two sums involving $\mathbf{B}^\star(\boldsymbol{g})$ cancel. We multiply both sides by $\boldsymbol{\Gamma}_{k,t}^{1/2}$ and apply the triangle inequality to obtain that
\begin{align}
\Bigl\lVert \boldsymbol{\Gamma}_{k,t}^{1/2}\big(\widehat{\mathbf{B}}_{k,t}(\boldsymbol{g})-\mathbf{B}^\star(\boldsymbol{g})\big) \Bigr\rVert_F\leq\biggl\lVert \boldsymbol{\Gamma}_{k,t}^{-1/2}\sum_{i=1}^k\boldsymbol{\phi}_{i,t}\boldsymbol{\varepsilon}_{i,t}(\boldsymbol{g})^\top \biggr\rVert_F+\bigl\lVert \boldsymbol{\Gamma}_{k,t}^{-1/2}\mathbf{B}^\star(\boldsymbol{g}) \bigr\rVert_F. \label{eq:lin_bfr_error_decomposition}
\end{align}
The first term is the error in the observed transitions, and the second term is the contribution of regularization.

For the second term, let $g_r$ be the $r$-th coordinate of $\boldsymbol{g}$. By \cref{lem:lin_off_W_cover}, $\|g_r\|_\infty\leq\|\boldsymbol{g}\|_{\infty,2}\leq C_\phi\sqrt{d_\phi}$. The $r$-th column of $\mathbf{B}^\star(\boldsymbol{g})=\sum_{s'}\mathbf{T}^\star(s')\boldsymbol{g}(s')^\top$ is $\sum_{s'}\mathbf{T}^\star(s')g_r(s')$. Since $\boldsymbol{\Gamma}_{k,t}\succeq I$, we apply condition (ii) of \cref{def:linear_mdp} to each column to obtain that
\begin{align}
\bigl\lVert \boldsymbol{\Gamma}_{k,t}^{-1/2}\mathbf{B}^\star(\boldsymbol{g}) \bigr\rVert_F^2\leq\bigl\lVert \mathbf{B}^\star(\boldsymbol{g}) \bigr\rVert_F^2=\sum_{r=1}^{d_\phi}\biggl\lVert \sum_{s'\in\mathcal{S}}\mathbf{T}^\star(s')g_r(s') \biggr\rVert_2^2\leq\sum_{r=1}^{d_\phi}C_{\mathbf{T}}^2d_\phi\|g_r\|_\infty^2\leq C_{\mathbf{T}}^2C_\phi^2d_\phi^3. \label{eq:lin_bfr_regularization}
\end{align}

For the first term, by \cref{lem:lin_off_W_cover}, $\mathcal{W}_{t+1}$ has an internal $(C_\phi\sqrt{d_\phi}/K)$-cover whose log cardinality is at most $\mathcal{H}_{\mathrm{W}}(1/K)$. We apply \cref{lem:lin_uniform_matrix_regression} with $\mathcal{F}=\mathcal{W}_{t+1}$, $R_g=R_x=C_\phi\sqrt{d_\phi}$, $\boldsymbol{x}_i=\boldsymbol{\phi}_{i,t}$, $Y_i=s_{i,t+1}$, $P_i=P^\star(\cdot\mid s_{i,t},a_{i,t})$, and failure probability $\delta/(12KH)$ to obtain that, with probability at least $1-\delta/(12KH)$, for every $\boldsymbol{g}\in\mathcal{W}_{t+1}$ simultaneously,
\begin{align}
\biggl\lVert \boldsymbol{\Gamma}_{k,t}^{-1/2}\sum_{i=1}^k\boldsymbol{\phi}_{i,t}\boldsymbol{\varepsilon}_{i,t}(\boldsymbol{g})^\top \biggr\rVert_F&\leq C_\phi\sqrt{d_\phi}\biggl[2\sqrt{d_\phi\log(1+KC_\phi^2)+2\bigl(\mathcal{H}_{\mathrm{W}}(1/K)+\log(12KH/\delta)\bigr)}+2\biggr]. \label{eq:lin_bfr_observation_error}
\end{align}
We substitute~\cref{eq:lin_bfr_regularization,eq:lin_bfr_observation_error} into~\cref{eq:lin_bfr_error_decomposition} and use the definition of $\kappa$ in~\cref{eq:lin_bfr_radius} to obtain that, with probability at least $1-\delta/(12KH)$,
\begin{align}
\sup_{\boldsymbol{g}\in\mathcal{W}_{t+1}}\|\boldsymbol{\Gamma}_{k,t}^{1/2}(\widehat{\mathbf{B}}_{k,t}(\boldsymbol{g})-\mathbf{B}^\star(\boldsymbol{g}))\|_F&\leq\kappa.
\end{align}
A union bound over at most $K(H-2)$ pairs $(k,t)$ has failure probability at most $\delta/12$.

\paragraph{Initial-State Averaging}
Consider a specific $k$ and take an internal $(C_\phi\sqrt{d_\phi}/K)$-cover of $\mathcal{W}_1$. For a fixed member $\boldsymbol{g}_0$ of the cover, the vectors $\boldsymbol{g}_0(s_{i,1})-\mathbb{E}_\rho\boldsymbol{g}_0$ have conditional mean zero and norm at most $2C_\phi\sqrt{d_\phi}$, since each initial state is drawn independently from $\rho$. We apply \cref{lem:lin_matrix_self_normalized} with the scalar covariate $x_i=1$, so that $d=1$, $R_x=1$, $R=2C_\phi\sqrt{d_\phi}$, and the Gram matrix is $k+1$. We take a union bound over the cover, with failure probability $\delta e^{-\mathcal{H}_{\mathrm{W}}(1/K)}/(12KH)$ per member, to obtain that
\begin{align}
\frac1{\sqrt{k+1}}\biggl\|\sum_{i=1}^k[\boldsymbol{g}_0(s_{i,1})-\mathbb{E}_\rho\boldsymbol{g}_0]\biggr\|_2&\leq2C_\phi\sqrt{d_\phi}\sqrt{\log(1+K)+2[\mathcal{H}_{\mathrm{W}}(1/K)+\log(12KH/\delta)]}
\end{align}
simultaneously over the cover. For a general $\boldsymbol{g}\in\mathcal{W}_1$, replacing $\boldsymbol{g}$ by its nearest member of the cover changes each of the two averages by at most $C_\phi\sqrt{d_\phi}/K$. Multiplying the preceding bound by $\sqrt{k+1}/k\leq\sqrt{2/k}$, and using $2/K\leq2/\sqrt k$ and $2\sqrt2\sqrt x+2\leq4\sqrt{x+1}$ for $x\geq0$, we obtain~\cref{eq:lin_bfr_initial_confidence} with $\gamma_\rho$ in~\cref{eq:lin_initial_radius}. A union bound over $k\in[K]$ costs at most $\delta/(12H)$. Adding the failure probability of the backward regressions leaves a total failure probability of at most $\delta/4$.
\end{proof}

\subsubsection{Optimistic Policy Evaluation} \label{app:lin_ope_guarantees}

The values regressed by \OPE{} contain $\widehat{R}_{k,h+1}$, which depends on the columns of $\widehat{\mathbf{T}}_k$. On the event $\mathcal{E}_{\mathrm{P}}$, \cref{lem:lin_column_error} shows that replacing these columns by the true columns changes each coordinate of the step-$(h+1)$ occupancy measure by at most $\sqrt{\beta^{\mathrm{Lin}}}\|\widehat{\boldsymbol{\omega}}_{k,h}\|_{\boldsymbol{\Lambda}_k^{-1}}$, and hence each coordinate of the pseudo-reward by at most $L$ times this amount. \cref{lem:lin_target_perturbation} multiplies this change by $2\sqrt k$ in the ridge regression. We therefore set
\begin{align}
\alpha_{k,h}^{\mathrm{Lin}}&\coloneq\begin{cases}\alpha_0^{\mathrm{Lin}}+2\sqrt{k}\,L\sqrt{\beta^{\mathrm{Lin}}}\bigl\lVert \widehat{\boldsymbol{\omega}}_{k,h} \bigr\rVert_{\boldsymbol{\Lambda}_k^{-1}},&h<H,\\ 0,&h=H,\end{cases} \label{eq:lin_off_value_radius_main}
\end{align}
where $\alpha_0^{\mathrm{Lin}}$ is chosen in \cref{lem:lin_off_value_event}. This radius is computable from $\boldsymbol{\Lambda}_k$ and $\widehat{\boldsymbol{\omega}}_{k,h}$, and~\cref{eq:lin_off_weighted_feature_bound} controls its growth with $k$.

Since the proof of \cref{lem:on_bellman_error} in \cref{app:on_ope_guarantees} holds for general radii and does not use $\mathcal{E}_{\mathrm{P}}$, the Bellman-error bound in linear MDPs is \cref{lem:on_bellman_error} with $\Delta_{k,h}=\Delta_{k,h}^{\mathrm{Lin}}$ and $\alpha_{k,h}=\alpha_{k,h}^{\mathrm{Lin}}$. The only change in the proof is that we apply \cref{lem:linear_value_response} in place of \cref{lem:bilinear_next_state_values} to obtain that $P^\star\widehat{V}_{k,h+1}(s,a)=\boldsymbol{\phi}(s,a)^\top\boldsymbol{v}^\star(\widehat{V}_{k,h+1})$.

The following lemma establishes $\mathcal{E}_{\mathrm{V}}$.

\begin{lemma}[Value-Regression Confidence] \label{lem:lin_off_value_event} \label{lem:lin_off_value_event_formal}
If we choose $\alpha_0^{\mathrm{Lin}}=\widetilde{\mathcal{O}}(HG\sqrt{d_\phi^2+d_\Pi})$ with a sufficiently large constant, and set $\alpha_{k,h}^{\mathrm{Lin}}$ by~\cref{eq:lin_off_value_radius_main}, then, under the event $\mathcal{E}_{\mathrm{P}}$, with probability at least $1-\delta/4$, $\mathcal{E}_{\mathrm{V}}$ holds.
\end{lemma}

\begin{proof}[Proof of \cref{lem:lin_off_value_event}]
We first recall the relevant notation. In $\OCUPOLIN$, \OPE{} uses the radius $\alpha_{k,h}^{\mathrm{Lin}}$ in~\cref{eq:lin_off_value_radius_main}. The event $\mathcal{E}_{\mathrm{V}}$ requires $\|\widehat{\boldsymbol{v}}_{k,h}-\boldsymbol{v}^\star(\widehat{V}_{k,h+1})\|_{\boldsymbol{\Gamma}_{k,h}}\leq\alpha_{k,h}^{\mathrm{Lin}}$ for every $k\in[K]$ and $h<H$, where $\boldsymbol{v}^\star$ is as in~\cref{lem:linear_value_response}. The class $\mathcal{V}_j^\circ$ in~\cref{eq:lin_off_true_value_class} replaces $\widehat{\mathbf{T}}_k(s)$ in the pseudo-reward by $\mathbf{T}^\star(s)$.

\paragraph{Choice of the Base Radius}
With $\mathcal{H}_{\mathrm{V}}$ from~\cref{eq:lin_off_V_entropy}, we take
\begin{align}
\alpha_0^{\mathrm{Lin}}&\coloneq HG\biggl[C_{\mathbf{T}}\sqrt{d_\phi}+2\sqrt{d_\phi\log(1+KC_\phi^2)+2\bigl(\mathcal{H}_{\mathrm{V}}(1/K)+\log(12KH/\delta)\bigr)}+2\biggr]. \label{eq:lin_value_radius}
\end{align}
The constant $Z_{\mathrm{Lin}}$, the bound $\mathcal{H}_{\mathrm{V}}$, and this radius are chosen in that order, and none of them uses the data. Under \cref{asm:policy_class}, $\mathcal{H}_{\mathrm{V}}(1/K)=\widetilde{\mathcal{O}}(d_\phi^2+d_\Pi)$. Hence, $\alpha_0^{\mathrm{Lin}}$ has the stated order.

\paragraph{Bounding the Radii}
We first verify, without any concentration argument, that every radius satisfies $\alpha_{k,h}^{\mathrm{Lin}}/(HG)\leq Z_{\mathrm{Lin}}$. Every dimension in~\cref{eq:lin_off_V_entropy} is at most $\nu_{\mathrm{Lin}}^2$, and every logarithm there contains a fixed power of $\nu_{\mathrm{Lin}}$ and at most the square of $Z_{\mathrm{Lin}}$. Moreover, the policy radius in~\cref{eq:lin_off_V_entropy} at $\varepsilon=1/K$ is at least the reciprocal of a fixed polynomial in $\nu_{\mathrm{Lin}}$. Hence, by~\cref{asm:policy_class}, there exist constants $c_1,c_2$, independent of $K$, $|\mathcal{S}|$, and $c_0$, such that
\begin{align}
\mathcal{H}_{\mathrm{V}}(1/K)\leq c_1\nu_{\mathrm{Lin}}^2\log Z_{\mathrm{Lin}},\qquad\sqrt{\beta^{\mathrm{Lin}}}\leq4\nu_{\mathrm{Lin}}^{3/2}\sqrt{\log Z_{\mathrm{Lin}}},\qquad\frac{\alpha_0^{\mathrm{Lin}}}{HG}\leq c_2\nu_{\mathrm{Lin}}^{3/2}\sqrt{\log Z_{\mathrm{Lin}}}.
\end{align}
Each $\widehat{\boldsymbol{g}}_{h,t}^{\pi_k}$ takes values in $\mathbb{B}_2(C_\phi\sqrt{d_\phi})$, since it averages either the features or their projections onto this ball (\cref{line:rfe_projection} of \cref{alg:rfe}). Hence, its average $\widehat{\boldsymbol{\omega}}_{k,h}$ also lies in this ball, and since $\boldsymbol{\Lambda}_k\succeq I$, we obtain $\|\widehat{\boldsymbol{\omega}}_{k,h}\|_{\boldsymbol{\Lambda}_k^{-1}}\leq\|\widehat{\boldsymbol{\omega}}_{k,h}\|_2\leq C_\phi\sqrt{d_\phi}\leq\nu_{\mathrm{Lin}}^{3/2}$. Together with $2\sqrt k\leq2\nu_{\mathrm{Lin}}^{1/2}$, $L\leq\nu_{\mathrm{Lin}}$, and $1/(HG)\leq\nu_{\mathrm{Lin}}$, we obtain $\alpha_{k,h}^{\mathrm{Lin}}/(HG)\leq(c_2+8)\nu_{\mathrm{Lin}}^{6}\sqrt{\log Z_{\mathrm{Lin}}}$. Moreover, $\log Z_{\mathrm{Lin}}=\log c_0+20\log\nu_{\mathrm{Lin}}\leq(\log c_0+20)\nu_{\mathrm{Lin}}$. Choosing $c_0\geq(c_2+8)\sqrt{\log c_0+20}$, which is possible since the left-hand side grows faster than the right-hand side, ensures $\alpha_{k,h}^{\mathrm{Lin}}/(HG)\leq(c_2+8)\sqrt{\log c_0+20}\,\nu_{\mathrm{Lin}}^{13/2}\leq c_0\nu_{\mathrm{Lin}}^{20}=Z_{\mathrm{Lin}}$.

\paragraph{Concentration on a Fixed Class}
Consider specific $k$ and $h<H$, and set $j=h+1$. We apply \cref{lem:lin_uniform_matrix_regression} with $m=1$ to the class $\mathcal{V}_j^\circ/(HG)$, with $R_g=1$, $R_x=C_\phi\sqrt{d_\phi}$, $\boldsymbol{x}_i=\boldsymbol{\phi}_{i,h}$, $Y_i=s_{i,h+1}$, $P_i=P^\star(\cdot\mid s_{i,h},a_{i,h})$, and failure probability $\delta/(12KH)$. By \cref{lem:lin_off_value_cover}, this class has an internal $(1/K)$-cover with log cardinality at most $\mathcal{H}_{\mathrm{V}}(1/K)$. The class depends on the MDP but not on the data. Hence, the bound holds for parameters chosen after all $k$ trajectories are observed. Multiplying by $HG$, we obtain that, outside an event of probability $\delta/(12KH)$, every $V\in\mathcal{V}_j^\circ$ satisfies
\begin{align}
\biggl\lVert \sum_{i=1}^k\boldsymbol{\phi}_{i,h} \bigl[V(s_{i,h+1})-P^\star V(s_{i,h},a_{i,h})\bigr] \biggr\rVert_{\boldsymbol{\Gamma}_{k,h}^{-1}} \leq\alpha_0^{\mathrm{Lin}}-C_{\mathbf{T}}\sqrt{d_\phi} HG. \label{eq:lin_off_oracle_value_concentration}
\end{align}
A union bound over the $K(H-1)$ pairs $(k,h)$ shows that~\cref{eq:lin_off_oracle_value_concentration} holds for all $k$ and $h<H$ with probability at least $1-\delta/4$.

\paragraph{Replacing the Columns of $\widehat{\mathbf{T}}_k$}
We now identify a member of $\mathcal{V}_j^\circ$ close to $\widehat{V}_{k,j}$. If $j<H$, we take in~\cref{eq:lin_off_true_value_class} the parameters $\pi=\pi_k$, $\boldsymbol{\omega}=\widehat{\boldsymbol{\omega}}_{k,h}$, $\boldsymbol{v}=\widehat{\boldsymbol{v}}_{k,j}/(HG)$, $\boldsymbol{A}=\boldsymbol{\Gamma}_{k,j}^{-1}$, and $\alpha=\alpha_{k,j}^{\mathrm{Lin}}$. If $j=H$, we take $\pi=\pi_k$ and $\boldsymbol{\omega}=\widehat{\boldsymbol{\omega}}_{k,H-1}$. These parameters lie in their domains. Indeed, $\pi_k\in\overline{\Pi}$, and $\|\widehat{\boldsymbol{\omega}}_{k,h}\|_2\leq C_\phi\sqrt{d_\phi}$ as shown above. Since the regression targets in~\cref{eq:ope_ridge_objective} are bounded by $(H-j)G$, comparing the objective with its value at zero shows that $\|\widehat{\boldsymbol{v}}_{k,j}\|_2\leq(H-j)G\sqrt k\leq HG\sqrt K$. Moreover, $\boldsymbol{\Gamma}_{k,j}\succeq I$ implies $0\prec\boldsymbol{\Gamma}_{k,j}^{-1}\preceq I$, and the bound on the radii above implies $\alpha_{k,j}^{\mathrm{Lin}}\leq HGZ_{\mathrm{Lin}}$. We denote the resulting function by $V_{k,j}^{\circ}$. It differs from $\widehat{V}_{k,j}$ only in that $\widehat{\mathbf{T}}_k(s)$ in the step-$j$ pseudo-reward is replaced by $\mathbf{T}^\star(s)$. On the event $\mathcal{E}_{\mathrm{P}}$, we apply~\cref{eq:lin_column_prediction_error} in \cref{lem:lin_column_error}, the definition of $\mathcal{E}_{\mathrm{P}}$, \cref{lem:coordinatewise_gradient_perturbation}, and the fact that clipping and averaging under the same policy cannot increase a uniform difference to obtain that
\begin{align}
\bigl\lVert \widehat{V}_{k,j}-V_{k,j}^{\circ} \bigr\rVert_\infty \leq L\sqrt{\beta^{\mathrm{Lin}}} \bigl\lVert \widehat{\boldsymbol{\omega}}_{k,h} \bigr\rVert_{\boldsymbol{\Lambda}_k^{-1}}. \label{eq:lin_off_comparison_distance}
\end{align}

\paragraph{Extension to $\widehat{V}_{k,j}$}
We apply \cref{lem:lin_target_perturbation} to $\widehat{V}_{k,j}$ and $V_{k,j}^{\circ}$, and then~\cref{eq:lin_off_oracle_value_concentration} to $V_{k,j}^{\circ}$, to obtain that, on the event $\mathcal{E}_{\mathrm{P}}$,
\begin{align}
\biggl\lVert \sum_{i=1}^k\boldsymbol{\phi}_{i,h} \bigl[\widehat{V}_{k,j}(s_{i,h+1})-P^\star \widehat{V}_{k,j}(s_{i,h},a_{i,h})\bigr] \biggr\rVert_{\boldsymbol{\Gamma}_{k,h}^{-1}}&\leq\alpha_0^{\mathrm{Lin}}-C_{\mathbf{T}}\sqrt{d_\phi} HG +2\sqrt{k}\,L\sqrt{\beta^{\mathrm{Lin}}} \bigl\lVert \widehat{\boldsymbol{\omega}}_{k,h} \bigr\rVert_{\boldsymbol{\Lambda}_k^{-1}}. \label{eq:lin_off_perturbed_value_concentration}
\end{align}
This step is deterministic and does not treat the difference between $\widehat{V}_{k,j}$ and $V_{k,j}^{\circ}$ as noise.

\paragraph{Contribution of Regularization}
Consider specific $k$ and $h<H$, and define $\boldsymbol{v}^\star\coloneq\boldsymbol{v}^\star(\widehat{V}_{k,h+1})$. For every $i\in[k]$, let
\begin{align}
\varepsilon_{i,h}\coloneq\widehat{V}_{k,h+1}(s_{i,h+1})-P^\star\widehat{V}_{k,h+1}(s_{i,h},a_{i,h}).
\end{align}
By \cref{lem:linear_value_response}, $P^\star\widehat{V}_{k,h+1}(s_{i,h},a_{i,h})=\boldsymbol{\phi}_{i,h}^\top\boldsymbol{v}^\star$. We obtain that
\begin{align}
\widehat{\boldsymbol{v}}_{k,h}-\boldsymbol{v}^\star&=\boldsymbol{\Gamma}_{k,h}^{-1}\biggl(\sum_{i=1}^k\boldsymbol{\phi}_{i,h}\widehat{V}_{k,h+1}(s_{i,h+1})-\boldsymbol{\Gamma}_{k,h}\boldsymbol{v}^\star\biggr)\nonumber\\
&=\boldsymbol{\Gamma}_{k,h}^{-1}\biggl(\sum_{i=1}^k\boldsymbol{\phi}_{i,h}\bigl(\boldsymbol{\phi}_{i,h}^\top\boldsymbol{v}^\star+\varepsilon_{i,h}\bigr)-\boldsymbol{\Gamma}_{k,h}\boldsymbol{v}^\star\biggr)\nonumber\\
&=\boldsymbol{\Gamma}_{k,h}^{-1}\biggl(\sum_{i=1}^k\boldsymbol{\phi}_{i,h}\boldsymbol{\phi}_{i,h}^\top\boldsymbol{v}^\star+\sum_{i=1}^k\boldsymbol{\phi}_{i,h}\varepsilon_{i,h}-\boldsymbol{v}^\star-\sum_{i=1}^k\boldsymbol{\phi}_{i,h}\boldsymbol{\phi}_{i,h}^\top\boldsymbol{v}^\star\biggr)\nonumber\\
&=\boldsymbol{\Gamma}_{k,h}^{-1}\biggl(\sum_{i=1}^k\boldsymbol{\phi}_{i,h}\varepsilon_{i,h}-\boldsymbol{v}^\star\biggr). \label{eq:lin_off_value_ridge_identity}
\end{align}
In the first equality, we use the closed form of $\widehat{\boldsymbol{v}}_{k,h}$ and $\boldsymbol{\Gamma}_{k,h}^{-1}\boldsymbol{\Gamma}_{k,h}=I$. In the second equality, we use the decomposition $\widehat{V}_{k,h+1}(s_{i,h+1})=P^\star\widehat{V}_{k,h+1}(s_{i,h},a_{i,h})+\varepsilon_{i,h}$ and apply \cref{lem:linear_value_response}. In the third equality, we expand $\boldsymbol{\Gamma}_{k,h}=I+\sum_{i=1}^k\boldsymbol{\phi}_{i,h}\boldsymbol{\phi}_{i,h}^\top$. In the last equality, the two sums involving $\boldsymbol{v}^\star$ cancel. The term $-\boldsymbol{v}^\star$ is the contribution of regularization. Since $\|\widehat{V}_{k,h+1}\|_\infty\leq HG$, we apply $\boldsymbol{\Gamma}_{k,h}\succeq I$ and condition (ii) of \cref{def:linear_mdp} to obtain that
\begin{align}
\lVert \boldsymbol{v}^\star \rVert_{\boldsymbol{\Gamma}_{k,h}^{-1}}\leq\lVert \boldsymbol{v}^\star \rVert_2=\biggl\lVert \sum_{s'\in\mathcal{S}}\mathbf{T}^\star(s')\widehat{V}_{k,h+1}(s') \biggr\rVert_2\leq C_{\mathbf{T}}\sqrt{d_\phi}HG. \label{eq:lin_off_value_regularization}
\end{align}
We take the $\boldsymbol{\Gamma}_{k,h}$-norm on both sides of~\cref{eq:lin_off_value_ridge_identity} to obtain that
\begin{align}
\bigl\lVert \widehat{\boldsymbol{v}}_{k,h}-\boldsymbol{v}^\star \bigr\rVert_{\boldsymbol{\Gamma}_{k,h}}&=\biggl\lVert \sum_{i=1}^k\boldsymbol{\phi}_{i,h}\varepsilon_{i,h}-\boldsymbol{v}^\star \biggr\rVert_{\boldsymbol{\Gamma}_{k,h}^{-1}}\leq\biggl\lVert \sum_{i=1}^k\boldsymbol{\phi}_{i,h}\varepsilon_{i,h} \biggr\rVert_{\boldsymbol{\Gamma}_{k,h}^{-1}}+\lVert \boldsymbol{v}^\star \rVert_{\boldsymbol{\Gamma}_{k,h}^{-1}}\nonumber\\
&\leq\alpha_0^{\mathrm{Lin}}-C_{\mathbf{T}}\sqrt{d_\phi}HG+2\sqrt{k}\,L\sqrt{\beta^{\mathrm{Lin}}}\bigl\lVert \widehat{\boldsymbol{\omega}}_{k,h} \bigr\rVert_{\boldsymbol{\Lambda}_k^{-1}}+C_{\mathbf{T}}\sqrt{d_\phi}HG=\alpha_{k,h}^{\mathrm{Lin}}.
\end{align}
In the first equality, we use $\|\boldsymbol{\Gamma}_{k,h}^{-1}\boldsymbol{u}\|_{\boldsymbol{\Gamma}_{k,h}}=\|\boldsymbol{u}\|_{\boldsymbol{\Gamma}_{k,h}^{-1}}$ for every $\boldsymbol{u}$. In the first inequality, we apply the triangle inequality. In the second inequality, we apply~\cref{eq:lin_off_perturbed_value_concentration} with $j=h+1$ and~\cref{eq:lin_off_value_regularization}. The last equality follows from~\cref{eq:lin_off_value_radius_main}. Hence, on the event $\mathcal{E}_{\mathrm{P}}$, $\mathcal{E}_{\mathrm{V}}$ holds whenever~\cref{eq:lin_off_oracle_value_concentration} holds for all $k$ and $h<H$, which happens with probability at least $1-\delta/4$.
\end{proof}

\subsection{Analysis of Sample Complexity: Proof of \texorpdfstring{\cref{thm:lin_off}}{Theorem \ref*{thm:lin_off}}} \label{app:lin_regret_analysis_off} \label{subsec:lin_off_leverage}

\begin{proof}[Proof of \cref{thm:lin_off}]
We first recall the relevant notation. As in the proof of~\cref{thm:on_policy}, Terms (I)--(III) refer to the regret decomposition in~\cref{eq:regret_decomposition}. The radii $\beta^{\mathrm{Lin}}$, $\Delta_{k,h}^{\mathrm{Lin}}$, and $\alpha_{k,h}^{\mathrm{Lin}}$ of the good event are given in~\cref{lem:lin_off_model_event,lem:lin_off_reward_error} and~\cref{eq:lin_off_value_radius_main}, and $\widehat{\ell}_{k,h}=\mathbb{E}_{(s,a)\sim\mu_h^{\pi_k}}\|\boldsymbol{\phi}(s,a)\|_{\boldsymbol{\Gamma}_{k-1,h}^{-1}}$ denotes the expected feature norm.

The only parameters required by the algorithm are the radii $\alpha_{k,h}^{\mathrm{Lin}}$ in~\cref{eq:lin_off_value_radius_main}, whose ingredients $\beta^{\mathrm{Lin}}$ and $\alpha_0^{\mathrm{Lin}}$ are specified in the proofs of \cref{lem:lin_off_model_event,lem:lin_off_value_event}. All other radii are used only in the analysis.

\paragraph{Regret Decomposition}
We apply \cref{lem:curl_regret_decomposition} to obtain~\cref{eq:regret_decomposition} with $\xi_{k,h}=R_{k,h}+P^\star\widehat{V}_{k,h+1}-\widehat{Q}_{k,h}$ and $\widehat{V}_{k,H+1}=0$.

\paragraph{Bounding Term (I)}
By~\cref{asm:pnpg} with $u=\pi^\star$,
\begin{align}
\mathrm{Term~(I)}\leq\mathcal{O}(H^2G\sqrt{K\log|\mathcal{A}|}+K\varepsilon_{\mathrm{proj}}). \label{eq:lin_off_actor_bound}
\end{align}

\paragraph{Bounding Terms (II) and (III)}
We work on the good event $\mathcal{E}$. We integrate the first inequality of \cref{lem:on_bellman_error}, with $\Delta_{k,h}=\Delta_{k,h}^{\mathrm{Lin}}$ and $\alpha_{k,h}=\alpha_{k,h}^{\mathrm{Lin}}$, under $\mu_h^{\pi^\star}$, integrate the second inequality under $\mu_h^{\pi_k}$, and use $\boldsymbol{\Gamma}_{k,h}\succeq\boldsymbol{\Gamma}_{k-1,h}$ to obtain that
\begin{align}
\mathrm{Term~(II)}+\mathrm{Term~(III)} &\leq2\sum_{k=1}^K\sum_{h=1}^H \Delta_{k,h}^{\mathrm{Lin}} +2\sum_{k=1}^K\sum_{h=1}^{H-1} \alpha_{k,h}^{\mathrm{Lin}}\widehat{\ell}_{k,h}. \label{eq:lin_off_combined_bellman_terms}
\end{align}

\paragraph{Bounding the Sums of $\widehat{\ell}_{k,h}$}
By \cref{lem:lin_off_martingale}, with probability at least $1-\delta/4$, its bounds hold for every $h\in[H-1]$.

\paragraph{Cumulative Estimation Error}
Assume that~\cref{eq:lin_off_weighted_feature_bound} holds. We expand~\cref{eq:lin_off_value_radius_main} in the second term of~\cref{eq:lin_off_combined_bellman_terms} and apply \cref{lem:lin_off_cumulative_bounds} to obtain that
\begin{align}
\mathrm{Term~(II)}+\mathrm{Term~(III)}&\leq2\sum_{k=1}^K\sum_{h=1}^H \Delta_{k,h}^{\mathrm{Lin}}+2\alpha_0^{\mathrm{Lin}}\sum_{k=1}^K\sum_{h=1}^{H-1}\widehat{\ell}_{k,h}+4L\sqrt{\beta^{\mathrm{Lin}}}\sum_{k=1}^K\sum_{h=1}^{H-1}\sqrt{k}\bigl\lVert \widehat{\boldsymbol{\omega}}_{k,h} \bigr\rVert_{\boldsymbol{\Lambda}_k^{-1}}\widehat{\ell}_{k,h}\nonumber\\
&\leq2\Bigl[L\sqrt{\beta^{\mathrm{Lin}}}(1+2\gamma_\rho)+LH(\sqrt{\beta^{\mathrm{Lin}}}+C_{\mathbf{T}}\sqrt{d_\phi})\kappa+\alpha_0^{\mathrm{Lin}}\Bigr]\sum_{k=1}^K\sum_{h=1}^{H-1}\widehat{\ell}_{k,h}\nonumber\\
&\qquad+4L\sqrt{\beta^{\mathrm{Lin}}}(1+H\kappa)\sqrt K\sum_{k=1}^K\sum_{h=1}^{H-1}\widehat{\ell}_{k,h}^2+4LH(\sqrt{\beta^{\mathrm{Lin}}}+C_{\mathbf{T}}\sqrt{d_\phi})\gamma_\rho\sqrt K. \label{eq:lin_off_estimation_term_final}
\end{align}

\paragraph{Putting Everything Together}
\cref{lem:lin_off_model_event} shows that $\mathcal{E}_{\mathrm{P}}$ fails with probability at most $\delta/4$. \cref{lem:lin_off_reward_error} shows that $\mathcal{E}_{\mathrm{P}}$ holds while $\mathcal{E}_{\mathrm{R}}$ or~\cref{eq:lin_off_weighted_feature_bound} fails with probability at most $\delta/4$. \cref{lem:lin_off_value_event} shows that $\Pr(\mathcal{E}_{\mathrm{P}}\cap\mathcal{E}_{\mathrm{V}}^c)\leq\delta/4$, and \cref{lem:lin_off_martingale} shows that its bounds fail with probability at most $\delta/4$. Hence, we take a union bound to show that $\mathcal{E}$,~\cref{eq:lin_off_weighted_feature_bound}, and the bounds of~\cref{lem:lin_off_martingale} hold simultaneously with probability at least $1-\delta$. We work on this event and assume $K\geq d_\phi^2$. Otherwise, the claimed bound is trivial, since $\operatorname{Sub\text{-}Opt}(\overline{\pi}_K)\leq2HG$. By \cref{lem:lin_off_model_event,lem:lin_off_reward_event,lem:lin_off_value_event,lem:lin_off_martingale},
\begin{align}
\sqrt{\beta^{\mathrm{Lin}}}&=\widetilde{\mathcal{O}}(\sqrt{d_\phi}),\qquad\kappa,\gamma_\rho=\widetilde{\mathcal{O}}\Bigl(\sqrt{d_\phi(d_\phi^2+d_\Pi)}\Bigr),\qquad\alpha_0^{\mathrm{Lin}}=\widetilde{\mathcal{O}}\Bigl(HG\sqrt{d_\phi^2+d_\Pi}\Bigr),\\
\sum_{k=1}^K\sum_{h=1}^{H-1}\widehat{\ell}_{k,h}&=\widetilde{\mathcal{O}}(H\sqrt{d_\phi K}),\qquad\sum_{k=1}^K\sum_{h=1}^{H-1}\widehat{\ell}_{k,h}^2=\widetilde{\mathcal{O}}(Hd_\phi^2).
\end{align}
We substitute these orders into~\cref{eq:lin_off_estimation_term_final} to obtain that
\begin{align}
\mathrm{Term~(II)}+\mathrm{Term~(III)}&\leq\widetilde{\mathcal{O}}\Bigl(H^2G\sqrt{d_\phi(d_\phi^2+d_\Pi)K}+LH^2d_\phi^{3/2}\sqrt{(d_\phi^2+d_\Pi)K}+LH^2d_\phi^3\sqrt{(d_\phi^2+d_\Pi)K}\Bigr)\\
&=\widetilde{\mathcal{O}}\Bigl(H^2\bigl[G\sqrt{d_\phi(d_\phi^2+d_\Pi)}+Ld_\phi^3\sqrt{d_\phi^2+d_\Pi}\bigr]\sqrt K\Bigr).
\end{align} We add~\cref{eq:lin_off_actor_bound} and divide by $K$ to obtain that
\begin{align}
\operatorname{Sub\text{-}Opt}(\overline{\pi}_K)\leq\frac{\operatorname{Reg}_K}{K}\leq\widetilde{\mathcal{O}}\left(\frac{H^2[G\sqrt{d_\phi(d_\phi^2+d_\Pi)}+Ld_\phi^3\sqrt{d_\phi^2+d_\Pi}]}{\sqrt K}+\varepsilon_{\mathrm{proj}}\right).
\end{align}
To make the first term at most $\epsilon$, it suffices to choose
\begin{align}
K=\widetilde{\mathcal{O}}\left(\frac{H^4\Bigl[G\sqrt{d_\phi(d_\phi^2+d_\Pi)}+Ld_\phi^3\sqrt{d_\phi^2+d_\Pi}\Bigr]^2}{\epsilon^2}\right)=\widetilde{\mathcal{O}}\left(\frac{H^4\bigl[G^2d_\phi(d_\phi^2+d_\Pi)+L^2d_\phi^6(d_\phi^2+d_\Pi)\bigr]}{\epsilon^2}\right).
\end{align}
Since one trajectory is collected per iteration, this is also the sample complexity.
\end{proof}

\begin{lemma}[Cumulative Reward and Correction Bounds]
\label{lem:lin_off_cumulative_bounds}
Suppose that~\cref{eq:lin_off_weighted_feature_bound} holds for every $k\in[K]$ and $j\in[H-1]$. Then
\begin{align}
\sum_{k=1}^K\sum_{h=1}^H \Delta_{k,h}^{\mathrm{Lin}} &\leq L\Bigl[\sqrt{\beta^{\mathrm{Lin}}}+H(\sqrt{\beta^{\mathrm{Lin}}}+C_{\mathbf{T}}\sqrt{d_\phi})\kappa\Bigr]\sum_{k=1}^K\sum_{h=1}^{H-1}\widehat{\ell}_{k,h}\nonumber\\
&\qquad+2LH(\sqrt{\beta^{\mathrm{Lin}}}+C_{\mathbf{T}}\sqrt{d_\phi})\gamma_\rho\sqrt K, \label{eq:lin_off_cumulative_reward}\\
\sum_{k=1}^K\sum_{h=1}^{H-1} \sqrt{k}\, \bigl\lVert \widehat{\boldsymbol{\omega}}_{k,h} \bigr\rVert_{\boldsymbol{\Lambda}_k^{-1}}\widehat{\ell}_{k,h} &\leq(1+H\kappa)\sqrt K\sum_{k=1}^K\sum_{h=1}^{H-1}\widehat{\ell}_{k,h}^2+\gamma_\rho\sum_{k=1}^K\sum_{h=1}^{H-1}\widehat{\ell}_{k,h}. \label{eq:lin_off_cumulative_correction}
\end{align}
\end{lemma}

\begin{proof}
We first bound the cumulative pseudo-reward error. Since $\Delta_{k,1}^{\mathrm{Lin}}=0$, we sum~\cref{eq:lin_off_Delta} over $k$ and $h\geq2$ and bound the sum of each of its three terms separately. The first sum equals $L\sqrt{\beta^{\mathrm{Lin}}}\sum_{k=1}^K\sum_{h=1}^{H-1}\widehat{\ell}_{k,h}$. Since $\sum_{k=1}^Kk^{-1/2}\leq2\sqrt K$, the second sum is at most $2LH(\sqrt{\beta^{\mathrm{Lin}}}+C_{\mathbf{T}}\sqrt{d_\phi})\gamma_\rho\sqrt K$. For the third sum, we interchange the sums over steps to obtain that
\begin{align}
\sum_{k=1}^K\sum_{h=2}^H\sum_{t=1}^{h-2}\widehat{\ell}_{k,t}=\sum_{k=1}^K\sum_{t=1}^{H-2}(H-t-1)\widehat{\ell}_{k,t}\leq H\sum_{k=1}^K\sum_{h=1}^{H-1}\widehat{\ell}_{k,h}.
\end{align}
We substitute these three bounds to obtain~\cref{eq:lin_off_cumulative_reward}.

We next bound the cumulative correction. We multiply~\cref{eq:lin_off_weighted_feature_bound} with $j=h$ by $\sqrt{k}\widehat{\ell}_{k,h}$ to obtain that
\begin{align}
\sqrt{k}\|\widehat{\boldsymbol{\omega}}_{k,h}\|_{\boldsymbol{\Lambda}_k^{-1}}\widehat{\ell}_{k,h}&\leq\sqrt{k}\widehat{\ell}_{k,h}^2+\gamma_\rho\widehat{\ell}_{k,h}+\kappa\sum_{t=1}^{h-1}\sqrt{k}\widehat{\ell}_{k,t}\widehat{\ell}_{k,h}.
\end{align}
We sum over $k\in[K]$ and $h\in[H-1]$ and use $\sqrt k\leq\sqrt K$. The first two terms give $\sqrt K\sum_{k=1}^K\sum_{h=1}^{H-1}\widehat{\ell}_{k,h}^2+\gamma_\rho\sum_{k=1}^K\sum_{h=1}^{H-1}\widehat{\ell}_{k,h}$. For the last term, we use $2\widehat{\ell}_{k,t}\widehat{\ell}_{k,h}\leq\widehat{\ell}_{k,t}^2+\widehat{\ell}_{k,h}^2$. Since each step in $[H-1]$ belongs to $H-2$ pairs $t<h$, for every $k\in[K]$,
\begin{align}
\sum_{h=1}^{H-1}\sum_{t=1}^{h-1}\widehat{\ell}_{k,t}\widehat{\ell}_{k,h}\leq\frac{H-2}2\sum_{h=1}^{H-1}\widehat{\ell}_{k,h}^2\leq H\sum_{h=1}^{H-1}\widehat{\ell}_{k,h}^2.
\end{align}
Combining the three terms proves~\cref{eq:lin_off_cumulative_correction}.
\end{proof}

The following lemma bounds the sums of $\widehat{\ell}_{k,h}$ and $\widehat{\ell}_{k,h}^2$ used in the proof of~\cref{thm:lin_off}.

\begin{lemma}[Bounds on the Expected Feature Norms] \label{lem:lin_off_martingale}
With probability at least $1-\delta/4$, for every $h\in[H-1]$,
\begin{align}
\sum_{k=1}^K\widehat{\ell}_{k,h}&\leq\sqrt{2Kd_\phi\log(1+KC_\phi^2)}+2C_\phi d_\phi^{3/2}\log(1+KC_\phi^2)+C_\phi\sqrt{2Kd_\phi\log(16H/\delta)},\nonumber\\
\sum_{k=1}^K\widehat{\ell}_{k,h}^2&\leq2(1+C_\phi^2d_\phi)d_\phi\log(1+KC_\phi^2)+2C_\phi^2d_\phi\log(8H/\delta). \label{eq:lin_off_leverage_bound}
\end{align}
In particular, for $K\geq d_\phi^2$, $\sum_{k=1}^K\sum_{h=1}^{H-1}\widehat{\ell}_{k,h}=\widetilde{\mathcal{O}}(H\sqrt{d_\phi K})$ and $\sum_{k=1}^K\sum_{h=1}^{H-1}\widehat{\ell}_{k,h}^2=\widetilde{\mathcal{O}}(Hd_\phi^2)$.
\end{lemma}

\begin{proof}
Consider a specific $h\in[H-1]$. The policy $\pi_k$ and the matrix $\boldsymbol{\Gamma}_{k-1,h}$ are fixed before trajectory $k$ is collected, and $(s_{k,h},a_{k,h})$ then has distribution $\mu_h^{\pi_k}$. For the first bound, we apply the argument in the proofs of \cref{lem:off_martingale} and \cref{thm:off_policy} in \cref{app:regret_analysis_off}, which uses only $\|\boldsymbol{\phi}(s,a)\|_2\leq C_\phi\sqrt{d_\phi}$, with failure probability $\delta/(8H)$. For the second bound, we apply \cref{lem:lin_predictable_leverage} with $R_x=C_\phi\sqrt{d_\phi}$, $n=K$, and failure probability $\delta/(8H)$. A union bound over $h\in[H-1]$ and the two bounds costs at most $\delta/4$.
\end{proof}
\section{Useful Properties and Technical Tools}\label{app:useful_lemmas}

\subsection{Useful Properties} \label{app:useful_properties}

Consider a finite-horizon MDP with state space $\mathcal{S}$, action set $\mathcal{A}$, initial distribution $\rho$, transition kernels $P_h^\star$, and policy $\pi=\{\pi_h\}_{h=1}^H$. Denote by $d_h^\pi(s)=\Pr(s_h=s)$ and $\mu_h^\pi(s,a)=\Pr(s_h=s,a_h=a)$ its state and state-action distributions. The following identities depend only on the transition representation.

\begin{lemma}[Occupancy Decomposition for Bilinear MDPs]
\label{lem:occupancy_measure}
Let $\pi$ be any policy, and suppose $P^\star(s'\mid s,a)=\boldsymbol{\phi}(s,a)^\top\mathbf{M}^\star\boldsymbol{\psi}(s')$, where $\boldsymbol{\phi}\colon \mathcal{S}\times\mathcal{A}\to\mathbb{R}^{d_\phi}$, $\boldsymbol{\psi}\colon \mathcal{S}\to\mathbb{R}^{d_\psi}$, and $\mathbf{M}^\star\in\mathbb{R}^{d_\phi\times d_\psi}$. For every $(s,a)\in\mathcal{S}\times\mathcal{A}$ and $h\in\{2,\ldots,H\}$,
\begin{align}
\mu_h^\pi(s,a) =\pi_h(a\mid s)\bigl( \boldsymbol{\omega}_{h-1}^\pi \bigr)^\top\mathbf{M}^\star\boldsymbol{\psi}(s),
\end{align}
where $\boldsymbol{\omega}_h^\pi\coloneq\mathbb{E}_{(s_h,a_h)\sim\mu_h^\pi}[\boldsymbol{\phi}(s_h,a_h)]$.
\end{lemma}

\begin{proof}
Consider a specific step $h\in\{2,\ldots,H\}$ and state $s\in\mathcal{S}$. The state at step $h$ is obtained by first drawing $(s_{h-1},a_{h-1})$ from $\mu_{h-1}^\pi$ and then drawing the successor from the transition model. The law of total probability therefore gives
\begin{align}
d_h^\pi(s) &=\sum_{(s^\prime,a^\prime)\in\mathcal{S}\times\mathcal{A}} P^\star(s\mid s^\prime,a^\prime)\mu_{h-1}^\pi(s^\prime,a^\prime).
\end{align}
Substituting the stated bilinear transition representation $P^\star(s\mid s',a')=\boldsymbol{\phi}(s',a')^\top\mathbf{M}^\star\boldsymbol{\psi}(s)$ yields
\begin{align}
d_h^\pi(s) &=\sum_{(s^\prime,a^\prime)\in\mathcal{S}\times\mathcal{A}} \boldsymbol{\phi}(s^\prime,a^\prime)^\top\mathbf{M}^\star\boldsymbol{\psi}(s) \mu_{h-1}^\pi(s^\prime,a^\prime)\\
&=\biggl( \sum_{(s^\prime,a^\prime)\in\mathcal{S}\times\mathcal{A}} \mu_{h-1}^\pi(s^\prime,a^\prime)\boldsymbol{\phi}(s^\prime,a^\prime) \biggr)^\top\mathbf{M}^\star\boldsymbol{\psi}(s)\\
&=\bigl( \boldsymbol{\omega}_{h-1}^\pi \bigr)^\top\mathbf{M}^\star\boldsymbol{\psi}(s),
\end{align}
where the last equality follows from the definition of $\boldsymbol{\omega}_{h-1}^\pi$ because $\mu_{h-1}^\pi$ is the distribution of $(s_{h-1},a_{h-1})$ under $\pi$. Finally, $\mu_h^\pi(s,a)=d_h^\pi(s)\pi_h(a\mid s)$, and substituting the preceding identity proves the claim.
At step one there is no preceding transition: $d_1^\pi=\rho$ and $\mu_1^\pi(s,a)=\rho(s)\pi_1(a\mid s)$. This is the initial case of the occupancy identity.
\end{proof}

\begin{lemma}[Occupancy Decomposition for Linear MDPs]
\label{lem:linear_occupancy_measure}
Let $\pi$ be any policy, and suppose $P^\star(s^\prime\mid s,a)=\boldsymbol{\phi}(s,a)^\top\mathbf{T}^\star(s^\prime)$, where $\boldsymbol{\phi}\colon \mathcal{S}\times\mathcal{A}\to\mathbb{R}^{d_\phi}$ and $\mathbf{T}^\star\in\mathbb{R}^{d_\phi\times|\mathcal{S}|}$. For every $(s,a)\in\mathcal{S}\times\mathcal{A}$ and $h\in\{2,\ldots,H\}$,
\begin{align}
\mu_h^\pi(s,a) =\pi_h(a\mid s)\bigl( \boldsymbol{\omega}_{h-1}^\pi \bigr)^\top\mathbf{T}^\star(s),
\end{align}
where $\boldsymbol{\omega}_h^\pi\coloneq\mathbb{E}_{(s_h,a_h)\sim\mu_h^\pi}[\boldsymbol{\phi}(s_h,a_h)]$.
\end{lemma}

\begin{proof}
Consider a specific step $h\in\{2,\ldots,H\}$ and a state $s\in\mathcal{S}$. The state at step $h$ is obtained by first drawing $(s_{h-1},a_{h-1})$ from $\mu_{h-1}^\pi$ and then drawing the next state from $P^\star$. We apply the law of total probability and~\cref{def:linear_mdp} to obtain that
\begin{align}
d_h^\pi(s) &=\sum_{(s^\prime,a^\prime)\in\mathcal{S}\times\mathcal{A}} \boldsymbol{\phi}(s^\prime,a^\prime)^\top\mathbf{T}^\star(s) \mu_{h-1}^\pi(s^\prime,a^\prime)=\bigl( \boldsymbol{\omega}_{h-1}^\pi \bigr)^\top\mathbf{T}^\star(s).
\end{align}
Multiplying by $\pi_h(a\mid s)$ completes the proof, since $\mu_h^\pi(s,a)=d_h^\pi(s)\pi_h(a\mid s)$.
\end{proof}
The next identity holds for any transition kernels. For a function $V\colon \mathcal{S}\to\mathbb{R}$, define $P_h^\star V(s,a)\coloneq\sum_{s'}P_h^\star(s'\mid s,a)V(s')$, and let $V_h^{\pi,R}$ and $Q_h^{\pi,R}$ denote the value functions of $\pi$ under a reward function $R$ and the kernels $P_h^\star$, with $V_{H+1}^{\pi,R}=0$.

\begin{lemma}[Extended Performance Difference Lemma; {\citealp{shani2020optimistic,cai2020provably}}] \label{lem:extended_pdl}
Let $\pi,\pi'$ be policies and $R$ a reward function. For arbitrary functions $\widehat{Q}_h\colon \mathcal{S}\times\mathcal{A}\to\mathbb{R}$, $h\in[H]$, let $\widehat{V}_h(s)\coloneq\sum_a\pi'_h(a\mid s)\widehat{Q}_h(s,a)$ and $\widehat{V}_{H+1}=0$. Then
\begin{align}
\mathbb{E}_{s_1\sim\rho}\bigl[V_1^{\pi,R}(s_1)-\widehat{V}_1(s_1)\bigr]&=\sum_{h=1}^H\mathbb{E}_{s\sim d_h^{\pi}}\bigl\langle \widehat{Q}_h(s,\cdot),\pi_h(\cdot\mid s)-\pi'_h(\cdot\mid s) \bigr\rangle\\
&\quad+\sum_{h=1}^H\mathbb{E}_{(s,a)\sim\mu_h^{\pi}}\bigl[R_h(s,a)+P_h^\star\widehat{V}_{h+1}(s,a)-\widehat{Q}_h(s,a)\bigr].
\end{align}
\end{lemma}

\begin{proof}
Consider specific $h\in[H]$ and $s\in\mathcal{S}$. The Bellman equation gives $V_h^{\pi,R}(s)=\sum_a\pi_h(a\mid s)[R_h(s,a)+P_h^\star V_{h+1}^{\pi,R}(s,a)]$. Subtracting $\widehat{V}_h(s)=\sum_a\pi'_h(a\mid s)\widehat{Q}_h(s,a)$, and adding and subtracting $\sum_a\pi_h(a\mid s)\widehat{Q}_h(s,a)$ and $\sum_a\pi_h(a\mid s)P_h^\star\widehat{V}_{h+1}(s,a)$, we obtain
\begin{align}
V_h^{\pi,R}(s)-\widehat{V}_h(s)&=\bigl\langle \widehat{Q}_h(s,\cdot),\pi_h(\cdot\mid s)-\pi'_h(\cdot\mid s) \bigr\rangle+\sum_a\pi_h(a\mid s)\bigl[R_h+P_h^\star\widehat{V}_{h+1}-\widehat{Q}_h\bigr](s,a)\\
&\quad+\sum_a\pi_h(a\mid s)P_h^\star\bigl(V_{h+1}^{\pi,R}-\widehat{V}_{h+1}\bigr)(s,a).
\end{align}
Let $D_h\coloneq\mathbb{E}_{s\sim d_h^\pi}[V_h^{\pi,R}(s)-\widehat{V}_h(s)]$. We average the identity over $s\sim d_h^\pi$. Since $\mu_h^\pi(s,a)=d_h^\pi(s)\pi_h(a\mid s)$, the second term becomes an expectation under $\mu_h^\pi$. For $h<H$, the occupancy flow equation $\sum_sd_h^\pi(s)\sum_a\pi_h(a\mid s)P_h^\star(s'\mid s,a)=d_{h+1}^\pi(s')$ turns the last term into $D_{h+1}$. At $h=H$, the last term is zero because $V_{H+1}^{\pi,R}=\widehat{V}_{H+1}=0$. Hence, with $D_{H+1}\coloneq0$,
\begin{align}
D_h&=\mathbb{E}_{s\sim d_h^{\pi}}\bigl\langle \widehat{Q}_h(s,\cdot),\pi_h(\cdot\mid s)-\pi'_h(\cdot\mid s) \bigr\rangle+\mathbb{E}_{(s,a)\sim\mu_h^{\pi}}\bigl[R_h(s,a)+P_h^\star\widehat{V}_{h+1}(s,a)-\widehat{Q}_h(s,a)\bigr]+D_{h+1}.
\end{align}
Summing over $h\in[H]$ and using $d_1^\pi=\rho$ proves the lemma.
\end{proof}

\subsection{Technical Tools} \label{app:technical_tools}

These results concern Euclidean covers, martingale concentration, and least-squares prediction. For a positive definite matrix $\boldsymbol{A}$, define $\|\boldsymbol{x}\|_{\boldsymbol{A}}=(\boldsymbol{x}^\top\boldsymbol{A}\boldsymbol{x})^{1/2}$. For a bounded vector-valued function $\boldsymbol{g}$ on a set $\mathcal{Y}$, define $\|\boldsymbol{g}\|_{\infty,2}=\sup_{y\in\mathcal{Y}}\|\boldsymbol{g}(y)\|_2$. For scalar functions, this is $\|g\|_\infty$.

\begin{lemma}[Euclidean Covering Bound; {\citealp[Lemmas 5.5 and 5.7]{wainwright2019high}}] \label{lem:euclidean_cover}
Let $\mathcal{B}$ be a subset of a Euclidean ball of radius $R$ in dimension $m$. For every $r>0$, it admits an internal $r$-cover with at most $(1+2R/r)^m$ points. The same statement applies to a matrix set in Frobenius norm, using the dimension of its ambient linear space.
\end{lemma}

\begin{lemma}[Self-Normalized Concentration]
    \label{lem:self_normalized_concentration}
This is Theorem 1 of~\citet{abbasi2011improved}.
    Let $\{x_t\}_{t=1}^\infty$ be a real-valued stochastic process adapted to the filtration $\{\mathcal{F}_t\}_{t=0}^\infty$ such that $x_t$ is conditionally $\sigma$-sub-Gaussian for some $\sigma > 0$, i.e.,
    \begin{align}
\forall \lambda \in \mathbb{R}, \quad \mathbb{E} \bigl[ \exp(\lambda \, x_t) \mid \mathcal{F}_{t-1} \bigr] \leq \exp(\lambda^2 \, \sigma^2 / 2) \,.
\end{align}
    Let $\{\boldsymbol{\phi}_t\}_{t=1}^\infty$ be an $\mathbb{R}^{d_\phi}$-valued stochastic process such that $\boldsymbol{\phi}_t$ is $\mathcal{F}_{t-1}$-measurable. Assume $\boldsymbol{\Lambda}_0$ is a $d_\phi \times d_\phi$ positive definite matrix, and let $\boldsymbol{\Lambda}_t = \boldsymbol{\Lambda}_0 + \sum_{i=1}^t \boldsymbol{\phi}_i \, \boldsymbol{\phi}_i^\top$. Then, for any $\delta > 0$, with probability at least $1 - \delta$, simultaneously for all $t \geq 0$,
    \begin{align}
\biggl\lVert \sum_{i=1}^t \boldsymbol{\phi}_i x_i \biggr\rVert^2_{\boldsymbol{\Lambda}_t^{-1}} \leq 2 \, \sigma^2 \log \left[ \frac{\det(\boldsymbol{\Lambda}_t)^{1/2} \, \det(\boldsymbol{\Lambda}_0)^{-1/2}}{\delta} \right] \,.
\end{align}
\end{lemma}

\begin{lemma}[Elliptical Potential Lemma; {\citealp[Lemma 11]{abbasi2011improved}}] \label{lem:elliptical_potential}
Let $\{\boldsymbol{x}_i\}_{i=1}^n$ be a sequence in $\mathbb{R}^d$ with $\|\boldsymbol{x}_i\|_2\leq R_x$ for every $i\in[n]$. Let $\boldsymbol{\Gamma}_0=I$ and $\boldsymbol{\Gamma}_i=I+\sum_{r=1}^i\boldsymbol{x}_r\boldsymbol{x}_r^\top$ for $i\in[n]$. Then
\begin{align}
\sum_{i=1}^n\min\bigl\{1,\lVert \boldsymbol{x}_i \rVert_{\boldsymbol{\Gamma}_{i-1}^{-1}}^2\bigr\}\leq2\log\det\boldsymbol{\Gamma}_n\leq2d\log\left(1+\frac{nR_x^2}{d}\right).
\end{align}
\end{lemma}

\begin{lemma}[Azuma--Hoeffding Inequality; {\citealp[Chapter 2]{wainwright2019high}}] \label{lem:azuma_hoeffding}
    Consider a scalar martingale difference sequence $\{x_t, \mathcal{F}_t\}_{t=1}^N$ where $x_t \in \mathbb{R}$ and $\mathbb{E}[x_t | \mathcal{F}_{t-1}] = 0$ for all $t \in [N]$. If each $x_t$ is bounded such that $|x_t| \leq R$, then for any $\epsilon > 0$, it holds that
    \begin{align}
\Pr \Biggl[ \biggl\lvert \frac{1}{N} \, \sum_{t=1}^N x_t \biggr\rvert \geq \epsilon \Biggr] \leq 2 \exp \left( -\frac{N\, \epsilon^2}{2 \, R^2} \right) \,.
\end{align}
\end{lemma}

\begin{lemma}[Vector Azuma--Hoeffding Inequality; {\citealp[Theorem 3.5]{pinelis1994optimum}}] \label{lem:vector_azuma_hoeffding}
    Consider a vector martingale difference sequence $\{\boldsymbol{x}_t,\mathcal{F}_t\}_{t=1}^N$ where $\boldsymbol{x}_t \in \mathbb{R}^{d_\phi}$ and $\mathbb{E}[\boldsymbol{x}_t\mid\mathcal{F}_{t-1}] = \boldsymbol{0}$ for all $t \in [N]$. If each $\boldsymbol{x}_t$ is bounded such that $\|\boldsymbol{x}_t\|_2 \leq R$, then for any $\epsilon > 0$, it holds that
    \begin{align}
\Pr \Biggl[ \biggl\lVert \frac{1}{N} \, \sum_{t=1}^N \boldsymbol{x}_t \biggr\rVert_2 \geq \epsilon \Biggr] \leq 2 \exp \left( -\frac{N\, \epsilon^2}{2 \, R^2} \right) \,.
\end{align}
\end{lemma}
This follows from~\citet[Theorem 3.5]{pinelis1994optimum} for martingales in $(2,1)$-smooth spaces, such as $\mathbb{R}^{d_\phi}$ with the Euclidean norm, applied with $\sum_{t=1}^N\|\boldsymbol{x}_t\|_2^2\leq NR^2$ and threshold $N\epsilon$.

\begin{lemma}[Deterministic Target Perturbation]
\label{lem:lin_target_perturbation}
Let $\boldsymbol{X}\in\mathbb{R}^{n\times d}$ have rows $\boldsymbol{x}_i^\top$, and set $\boldsymbol{\Gamma}=I+\boldsymbol{X}^\top\boldsymbol{X}$. Let $y_1,\ldots,y_n$ be points of a measurable space $\mathcal{Y}$, and let $P_1,\ldots,P_n$ be probability measures on that space, with $P_i V\coloneq\int V(y)P_i(\mathrm{d}y)$. For any bounded scalar functions $V,V'\colon \mathcal{Y}\to\mathbb{R}$,
\begin{align}
\biggl\|\sum_{i=1}^n\boldsymbol{x}_i\bigl[(V-V')(y_i)-P_i(V-V')\bigr]\biggr\|_{\boldsymbol{\Gamma}^{-1}}&\leq2\sqrt n\,\|V-V'\|_\infty.
\end{align}
For bounded vector functions $\boldsymbol{g},\boldsymbol{g}'\colon \mathcal{Y}\to\mathbb{R}^m$, the corresponding bound is
\begin{align}
\biggl\|\boldsymbol{\Gamma}^{-1/2}\sum_{i=1}^n\boldsymbol{x}_i\bigl[(\boldsymbol{g}-\boldsymbol{g}')(y_i)-P_i(\boldsymbol{g}-\boldsymbol{g}')\bigr]^\top\biggr\|_F&\leq2\sqrt n\,\|\boldsymbol{g}-\boldsymbol{g}'\|_{\infty,2}.
\end{align}
Both inequalities are deterministic and thus also hold for functions selected using these points and measures.
\end{lemma}

\begin{proof}
We prove the vector statement, which includes the scalar case as output dimension one. Let $\boldsymbol{D}=\boldsymbol{g}-\boldsymbol{g}'$ and let $\boldsymbol{A}$ have $i$th row $[\boldsymbol{D}(y_i)-P_i\boldsymbol{D}]^\top$. For each row, Jensen's inequality and the triangle inequality give
\begin{align}
\|\boldsymbol{D}(y_i)-P_i\boldsymbol{D}\|_2&\leq\|\boldsymbol{D}(y_i)\|_2+\int\|\boldsymbol{D}(y)\|_2P_i(\mathrm{d}y)\leq2\|\boldsymbol{D}\|_{\infty,2}.
\end{align}
Consequently, $\|\boldsymbol{A}\|_F^2$, the sum of squared row norms, is at most $4n\|\boldsymbol{D}\|_{\infty,2}^2$.

The normalized residual difference equals $\boldsymbol{\Gamma}^{-1/2}\boldsymbol{X}^\top\boldsymbol{A}$. To control this matrix, take a singular-value decomposition $\boldsymbol{X}=\boldsymbol{U}\boldsymbol{\Sigma}\boldsymbol{V}^\top$. The nonzero eigenvalues of $\boldsymbol{X}(I+\boldsymbol{X}^\top\boldsymbol{X})^{-1}\boldsymbol{X}^\top$ are $\sigma_r^2/(1+\sigma_r^2)$, each at most one. Thus $0\preceq\boldsymbol{X}\boldsymbol{\Gamma}^{-1}\boldsymbol{X}^\top\preceq I_n$, including when $\boldsymbol{X}$ is rank deficient. It follows that
\begin{align}
\|\boldsymbol{\Gamma}^{-1/2}\boldsymbol{X}^\top\boldsymbol{A}\|_F^2&=\operatorname{tr}(\boldsymbol{A}^\top\boldsymbol{X}\boldsymbol{\Gamma}^{-1}\boldsymbol{X}^\top\boldsymbol{A})\\
&\leq\operatorname{tr}(\boldsymbol{A}^\top\boldsymbol{A})=\|\boldsymbol{A}\|_F^2\leq4n\|\boldsymbol{D}\|_{\infty,2}^2.
\end{align}
The trace inequality follows by applying $\boldsymbol{X}\boldsymbol{\Gamma}^{-1}\boldsymbol{X}^\top\preceq I_n$ to each column of $\boldsymbol{A}$. Taking square roots proves the vector claim. For a scalar target, the Frobenius norm of the resulting column vector is exactly the weighted norm in the statement. Every step is deterministic for the observed data and chosen functions. Hence, the argument remains valid when both targets were selected from those data.
\end{proof}

The following lemma extends~\cref{lem:elliptical_potential} from the observed norms $\|\boldsymbol{x}_i\|_{\boldsymbol{\Gamma}_{i-1}^{-1}}$ to their expectations.

\begin{lemma}[Expected Elliptical Potential]
\label{lem:lin_predictable_leverage}
Let $R_x>0$. For every $i$, suppose that a distribution $\mathsf{D}_i$ on $\mathbb{B}_2(R_x)\subseteq\mathbb{R}^{d}$ is fixed by the history before $\boldsymbol{x}_i\sim\mathsf{D}_i$ is drawn. Let $\boldsymbol{\Gamma}_i=I+\sum_{r=1}^i\boldsymbol{x}_r\boldsymbol{x}_r^\top$ and $\ell_i\coloneq\mathbb{E}_{\boldsymbol{x}\sim\mathsf{D}_i}[\|\boldsymbol{x}\|_{\boldsymbol{\Gamma}_{i-1}^{-1}}]$.
With probability at least $1-\zeta$,
\begin{align}
\sum_{i=1}^n\ell_i^2 &\leq2(1+R_x^2)d\log(1+nR_x^2/d)+2R_x^2\log(1/\zeta). \label{eq:lin_squared_leverage}
\end{align}
\end{lemma}

\begin{proof}
Let $\mathcal{F}_{i-1}$ contain the history that determines $\mathsf{D}_i$ and $\boldsymbol{\Gamma}_{i-1}$. Define $Z_i=\|\boldsymbol{x}_i\|_{\boldsymbol{\Gamma}_{i-1}^{-1}}^2$ and $m_i=\mathbb{E}[Z_i\mid\mathcal{F}_{i-1}]$. Since $\boldsymbol{\Gamma}_{i-1}\succeq I$, we have $0\leq Z_i,m_i\leq R_x^2$.

We first relate the conditional expectations to the observations. We apply the convexity of $z\to e^{-z}$ to obtain that $e^{-z}\leq1-cz$ for $0\leq z\leq1$, with $c=1-e^{-1}\geq1/2$. Applying this to $Z_i/R_x^2$, and using that $m_i$ is fixed before the draw, we obtain
\begin{align}
\mathbb{E}[\exp(cm_i/R_x^2-Z_i/R_x^2)\mid\mathcal{F}_{i-1}]&\leq e^{cm_i/R_x^2}(1-cm_i/R_x^2)\leq1.
\end{align}
Iterating conditional expectations and applying Markov's inequality, we obtain that, with probability at least $1-\zeta$, $\sum_i m_i\leq2\sum_iZ_i+2R_x^2\log(1/\zeta)$.

We then bound the resulting sum by a log-determinant. For $0\leq z\leq R_x^2$, $\log(1+z)\geq z/(1+z)\geq z/(1+R_x^2)$. We then apply the matrix determinant lemma and $\boldsymbol{\Gamma}_0=I$ to obtain that
\begin{align}
\sum_{i=1}^n Z_i&\leq(1+R_x^2)\sum_{i=1}^n\log(1+Z_i)=(1+R_x^2)\log\det\boldsymbol{\Gamma}_n\leq(1+R_x^2)d\log(1+nR_x^2/d),
\end{align}
where the last step applies the arithmetic--geometric mean inequality to the eigenvalues of $\boldsymbol{\Gamma}_n$ and uses $\operatorname{tr}\boldsymbol{\Gamma}_n\leq d+nR_x^2$. Finally, we apply the conditional Cauchy--Schwarz inequality to obtain that $\ell_i^2\leq m_i$, which proves~\cref{eq:lin_squared_leverage}.
\end{proof}
\begin{lemma}[Self-Normalized Matrix Martingale]
\label{lem:lin_matrix_self_normalized}
Let $\{\mathcal{F}_i\}_{i\geq0}$ be a filtration. Suppose $\boldsymbol{x}_i\in\mathbb{R}^d$ is $\mathcal{F}_{i-1}$-measurable with $\|\boldsymbol{x}_i\|_2\leq R_x$, and $\boldsymbol{y}_i\in\mathbb{R}^m$ is $\mathcal{F}_i$-measurable with $\mathbb{E}[\boldsymbol{y}_i\mid\mathcal{F}_{i-1}]=\boldsymbol{0}$ and $\|\boldsymbol{y}_i\|_2\leq R$. Define $\boldsymbol{\Gamma}_n=I+\sum_{i=1}^n\boldsymbol{x}_i\boldsymbol{x}_i^\top$ and $\boldsymbol{S}_n=\sum_{i=1}^n\boldsymbol{x}_i\boldsymbol{y}_i^\top$. For every fixed $n\geq1$ and $\zeta\in(0,1)$, with probability at least $1-\zeta$,
\begin{align}
\bigl\lVert \boldsymbol{\Gamma}_n^{-1/2}\boldsymbol{S}_n \bigr\rVert_F \leq R\sqrt{d\log(1+nR_x^2/d)+2\log(1/\zeta)}. \label{eq:lin_matrix_self_normalized}
\end{align}
The bound does not depend on $m$.
\end{lemma}

\begin{proof}
When $R=0$, every $\boldsymbol{y}_i$ is zero almost surely and the conclusion is immediate. Suppose $R>0$. We derive a recursion for the squared normalized norm, construct an exponential supermartingale from that recursion, and bound the accumulated drift by a log determinant.

We first derive a recursion for the squared norm. For $i\geq0$, let $\boldsymbol{S}_i=\sum_{r=1}^i\boldsymbol{x}_r\boldsymbol{y}_r^\top$ and $\boldsymbol{\Gamma}_i=I+\sum_{r=1}^i\boldsymbol{x}_r\boldsymbol{x}_r^\top$. Define $q_i=\operatorname{tr}(\boldsymbol{S}_i^\top\boldsymbol{\Gamma}_i^{-1}\boldsymbol{S}_i)$, $a_i=\boldsymbol{x}_i^\top\boldsymbol{\Gamma}_{i-1}^{-1}\boldsymbol{x}_i$, and $\boldsymbol{z}_i=\boldsymbol{S}_{i-1}^\top\boldsymbol{\Gamma}_{i-1}^{-1}\boldsymbol{x}_i$. Here $q_i$ is precisely the squared Frobenius norm we wish to bound, and $a_i,\boldsymbol{z}_i$ are measurable before $\boldsymbol{y}_i$ is observed. Sherman--Morrison gives
\begin{align}
\boldsymbol{\Gamma}_i^{-1}&=\boldsymbol{\Gamma}_{i-1}^{-1}-\frac{\boldsymbol{\Gamma}_{i-1}^{-1}\boldsymbol{x}_i\boldsymbol{x}_i^\top\boldsymbol{\Gamma}_{i-1}^{-1}}{1+a_i}.
\end{align}
First expanding $\boldsymbol{S}_i=\boldsymbol{S}_{i-1}+\boldsymbol{x}_i\boldsymbol{y}_i^\top$ against $\boldsymbol{\Gamma}_{i-1}^{-1}$ gives $q_{i-1}+2\langle \boldsymbol{z}_i,\boldsymbol{y}_i\rangle+a_i\|\boldsymbol{y}_i\|_2^2$. The rank-one correction subtracts $\|\boldsymbol{z}_i+a_i\boldsymbol{y}_i\|_2^2/(1+a_i)$. Consequently,
\begin{align}
q_i-q_{i-1}&=2\langle \boldsymbol{z}_i,\boldsymbol{y}_i \rangle+a_i\|\boldsymbol{y}_i\|_2^2-\frac{\|\boldsymbol{z}_i+a_i\boldsymbol{y}_i\|_2^2}{1+a_i}\\
&=\frac{2\langle \boldsymbol{z}_i,\boldsymbol{y}_i \rangle+a_i\|\boldsymbol{y}_i\|_2^2-\|\boldsymbol{z}_i\|_2^2}{1+a_i}. \label{eq:matrix_norm_recursion}
\end{align}
This calculation treats $\boldsymbol{y}_i$ as a whole, rather than bounding its coordinates separately.

We next derive an exponential bound. The scalar $\langle \boldsymbol{z}_i,\boldsymbol{y}_i\rangle$ has mean zero conditional on $\mathcal{F}_{i-1}$. Cauchy--Schwarz bounds its absolute value by $R\|\boldsymbol{z}_i\|_2$. For every real $t$, Hoeffding's exponential bound therefore gives
\begin{align}
\mathbb{E}[\exp(t\langle \boldsymbol{z}_i,\boldsymbol{y}_i \rangle)\mid\mathcal{F}_{i-1}]&\leq\exp(t^2R^2\|\boldsymbol{z}_i\|_2^2/2).
\end{align} Since $\|\boldsymbol{y}_i\|_2^2\leq R^2$,~\cref{eq:matrix_norm_recursion} implies, for $\lambda=1/(2R^2)$,
\begin{align}
\mathbb{E}\left[\exp\left(\lambda\left[q_i-q_{i-1}-\frac{R^2a_i}{1+a_i}\right]\right)\middle|\mathcal{F}_{i-1}\right]&\leq\exp\left(-\frac{\lambda\|\boldsymbol{z}_i\|_2^2}{1+a_i}+\frac{2\lambda^2R^2\|\boldsymbol{z}_i\|_2^2}{(1+a_i)^2}\right)\\
&=\exp\left(-\frac{a_i\|\boldsymbol{z}_i\|_2^2}{2R^2(1+a_i)^2}\right)\leq1.
\end{align}
The equality substitutes the value of $\lambda$, and $a_i\geq0$ gives the last inequality. Since $q_0=0$, repeated conditioning proves
\begin{align}
\mathbb{E}\exp\Biggl(\lambda\biggl[q_n-R^2\sum_{i=1}^n\frac{a_i}{1+a_i}\biggr]\Biggr)&\leq1.
\end{align}
Markov's inequality now shows that, outside an event of probability at most $\zeta$,
\begin{align}
q_n&\leq R^2\sum_{i=1}^n\frac{a_i}{1+a_i}+2R^2\log(1/\zeta).
\end{align}

Finally, we bound the log-determinant. For $a\geq0$, $a/(1+a)\leq\log(1+a)$. This follows, for example, by integrating $1/(1+x)\geq1/(1+a)$ over $x\in[0,a]$. The determinant lemma then gives
\begin{align}
\sum_{i=1}^n\frac{a_i}{1+a_i}&\leq\sum_{i=1}^n\log(1+a_i)=\log\det\boldsymbol{\Gamma}_n\\
&\leq d\log\left(\frac{\operatorname{tr}\boldsymbol{\Gamma}_n}{d}\right)\leq d\log(1+nR_x^2/d).
\end{align}
The eigenvalue arithmetic--geometric mean inequality gives the penultimate step, and $\sum_i\|\boldsymbol{x}_i\|_2^2\leq nR_x^2$ gives the last. Substitution into the bound on $q_n$ and taking square roots proves~\cref{eq:lin_matrix_self_normalized}. The dimension $m$ never entered the calculation, which explains its absence from the result.
\end{proof}

\begin{lemma}[Uniform Vector Regression Over a Fixed Class]
\label{lem:lin_uniform_matrix_regression}
Let $R_g,R_x>0$, and let $\mathcal{F}$ be a fixed class of measurable functions $\boldsymbol{g}\colon \mathcal{Y}\to\mathbb{R}^m$ with $\|\boldsymbol{g}\|_{\infty,2}\leq R_g$. Suppose it has an internal $(R_g/K)$-cover in $\|\cdot\|_{\infty,2}$ of log cardinality at most $E$. Let $\{\mathcal{H}_i\}_{i\geq0}$ be a filtration. For $i\leq k\leq K$, let $\boldsymbol{x}_i\in\mathbb{R}^d$ be $\mathcal{H}_{i-1}$-measurable with $\|\boldsymbol{x}_i\|_2\leq R_x$, and let $Y_i$ be $\mathcal{H}_i$-measurable with conditional distribution $P_i$ given $\mathcal{H}_{i-1}$. Define $\boldsymbol{\Gamma}_k=I+\sum_{i=1}^k\boldsymbol{x}_i\boldsymbol{x}_i^\top$ and $P_i\boldsymbol{g}=\int\boldsymbol{g}(y)P_i(\mathrm{d}y)$. For any $\zeta\in(0,1)$, with probability at least $1-\zeta$, simultaneously for every $\boldsymbol{g}\in\mathcal{F}$,
\begin{align}
\biggl\|\boldsymbol{\Gamma}_k^{-1/2}\sum_{i=1}^k\boldsymbol{x}_i[\boldsymbol{g}(Y_i)-P_i\boldsymbol{g}]^\top\biggr\|_F&\leq R_g\Bigl[2\sqrt{d\log(1+KR_x^2/d)+2(E+\log(1/\zeta))}+2\Bigr]. \label{eq:lin_uniform_matrix_regression}
\end{align}
The same statement holds conditionally on an initial history relative to which $\mathcal{F}$ is fixed.
\end{lemma}

\begin{proof}
Choose an internal $(R_g/K)$-cover $\mathcal{C}\subseteq\mathcal{F}$ with $|\mathcal{C}|\leq e^E$. For a fixed $\boldsymbol{g}_0\in\mathcal{C}$, the vector $\boldsymbol{g}_0(Y_i)-P_i\boldsymbol{g}_0$ has conditional mean zero and Euclidean norm at most $2R_g$. We apply \cref{lem:lin_matrix_self_normalized} with $R=2R_g$ and failure probability $\zeta e^{-E}$ to each member of $\mathcal{C}$, and take a union bound, to obtain that, with probability at least $1-\zeta$,
\begin{align}
\max_{\boldsymbol{g}_0\in\mathcal{C}}\biggl\|\boldsymbol{\Gamma}_k^{-1/2}\sum_{i=1}^k\boldsymbol{x}_i[\boldsymbol{g}_0(Y_i)-P_i\boldsymbol{g}_0]^\top\biggr\|_F&\leq2R_g\sqrt{d\log(1+KR_x^2/d)+2(E+\log(1/\zeta))}.
\end{align}
Now take any $\boldsymbol{g}\in\mathcal{F}$, possibly selected using the observations, and choose $\boldsymbol{g}_0\in\mathcal{C}$ with $\|\boldsymbol{g}-\boldsymbol{g}_0\|_{\infty,2}\leq R_g/K$. The vector form of \cref{lem:lin_target_perturbation} bounds the difference of the two normalized sums by $2\sqrt{k}R_g/K\leq2R_g$. Adding this deterministic error proves~\cref{eq:lin_uniform_matrix_regression}.
\end{proof}

\clearpage

\section{Experiments} \label{sec:experiments}

\addtocontents{toc}{\protect\setcounter{tocdepth}{1}}

We complement our theoretical results with controlled experiments designed to address five questions:
(i) Optimization: Can $\OCUPOON$ and $\OCUPOOFF$ effectively optimize nonlinear occupancy utilities?
(ii) Sample efficiency: How do the algorithms compare under matched real-environment interaction budgets?
(iii) Policy representation: How does policy-feature dimension affect off-policy optimization?
(iv) Compact dynamics: Do the algorithms remain effective with low-dimensional bilinear transition representations?
(v) Linear-MDP extension: Can $\OCUPOLIN$ optimize occupancy utilities in full-rank and compact linear-MDP instances?

The experiments use gridworld environments that permit exact evaluation and visualization of the learned policy occupancies. Hyperparameters are selected through separate tuning runs and held fixed during evaluation on five independent seeds. These experiments are controlled illustrations of the optimization and estimation behavior of our algorithms on small, exactly known models. They do not test the regime of large state spaces and compact policy classes targeted by our theory, and they do not compare against other CURL methods.

\subsection{Experimental Setup}\label{subsec:exp-setup}

\begin{figure}[t]
    \centering
        \begin{minipage}[t]{0.264\linewidth}
            \centering
            \includegraphics[width=\linewidth]{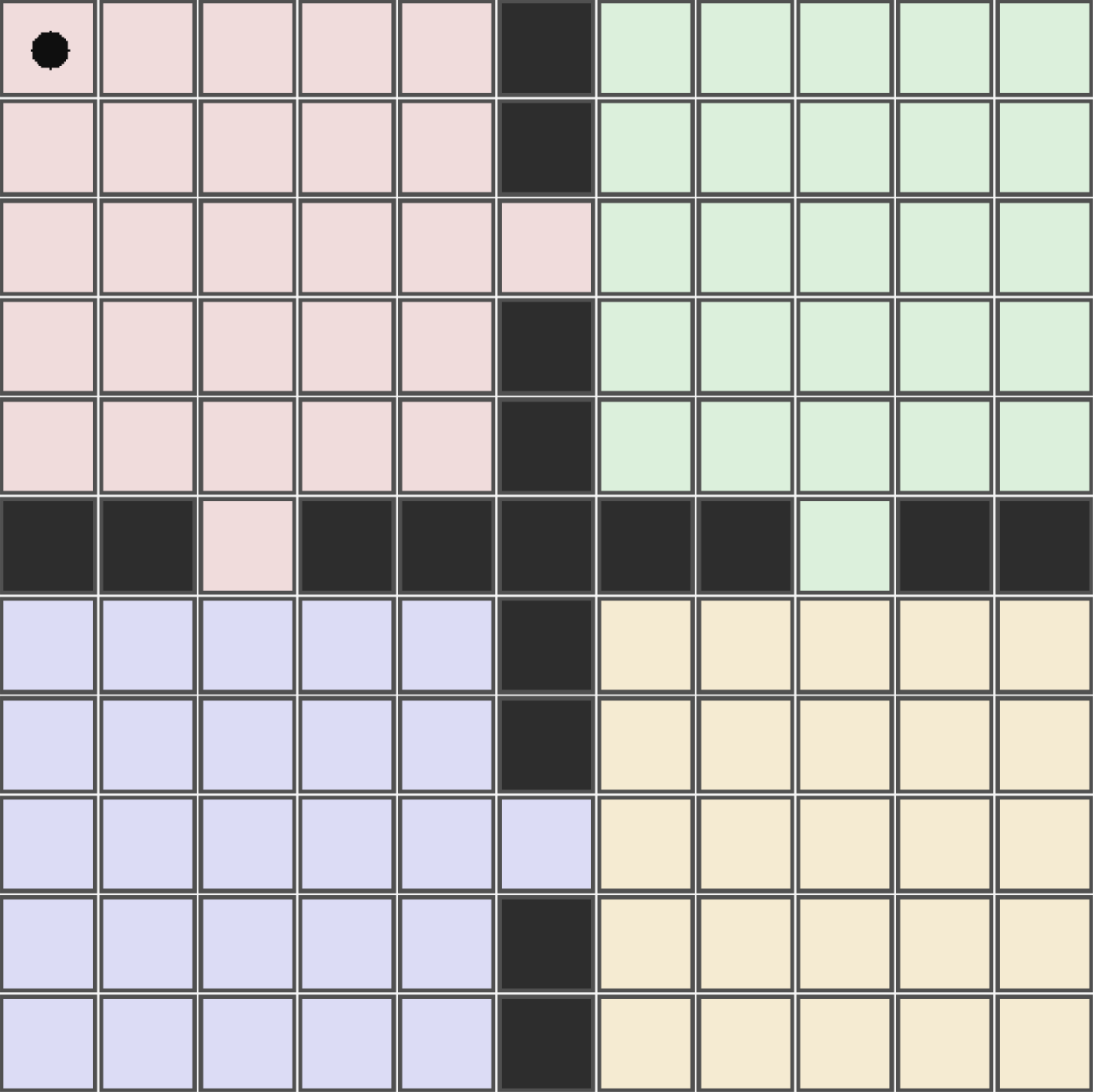}
            \par\smallskip{\footnotesize (a) Four-room gridworld\par}
        \end{minipage}\hspace{0.081\linewidth}%
        \begin{minipage}[t]{0.264\linewidth}
            \centering
            \includegraphics[width=\linewidth]{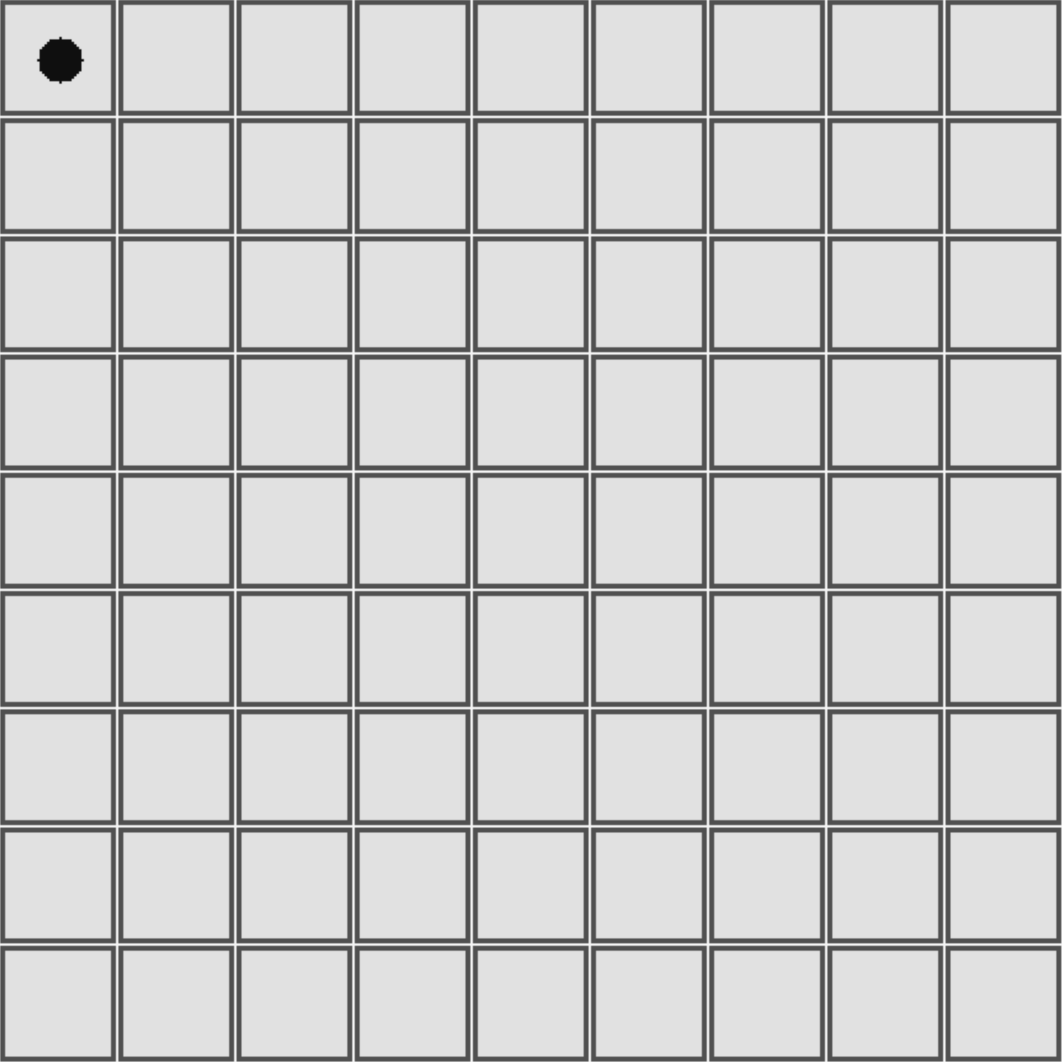}
            \par\smallskip{\footnotesize (b) Empty grid\par}
        \end{minipage}

    \par\vspace{0.5em}
    \begin{minipage}[t]{0.31\linewidth}
        \centering
        \includegraphics[width=\linewidth]{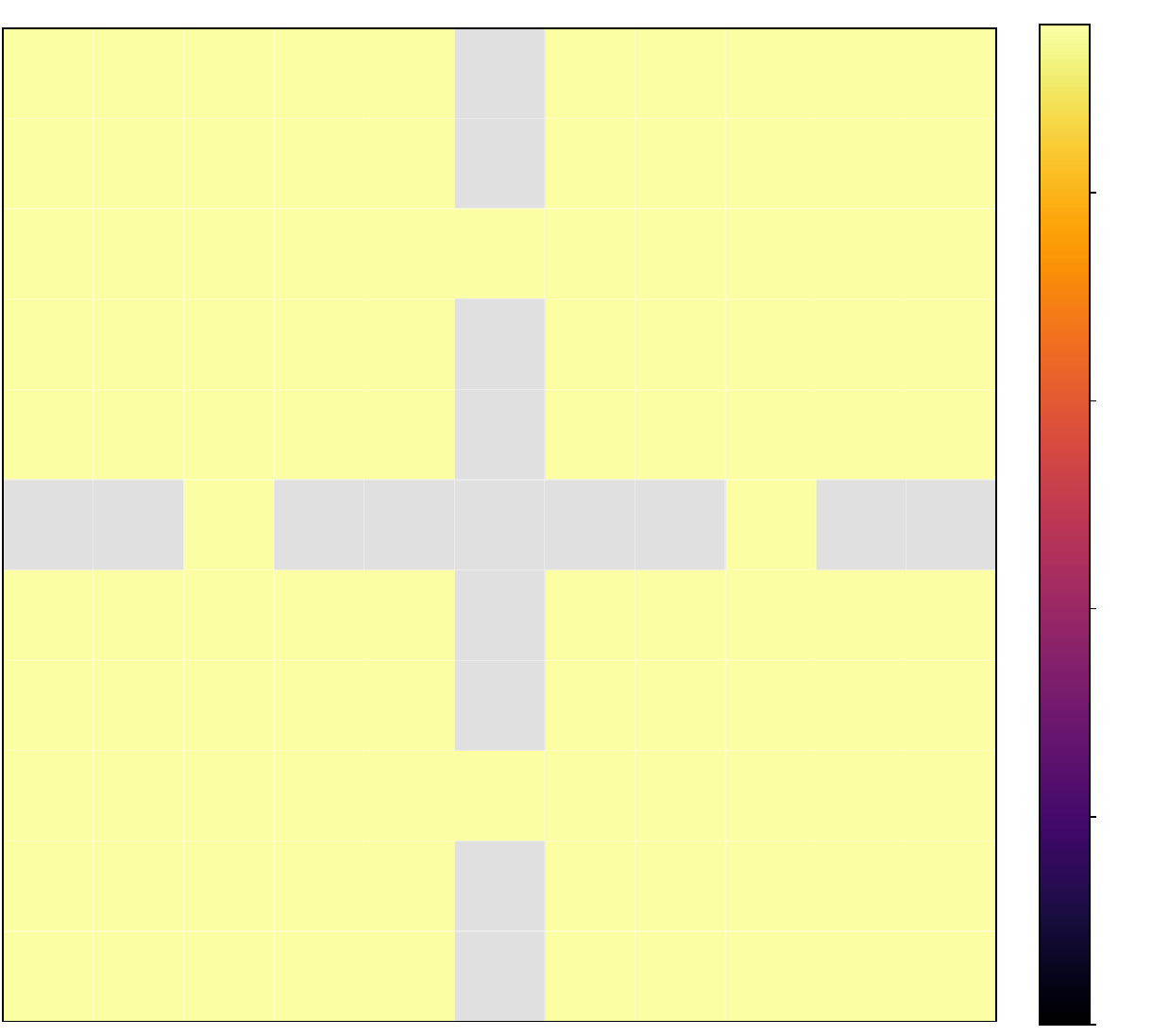}
        \par\smallskip{\footnotesize (c) Maximum entropy\par}
    \end{minipage}\hfill
    \begin{minipage}[t]{0.31\linewidth}
        \centering
        \includegraphics[width=\linewidth]{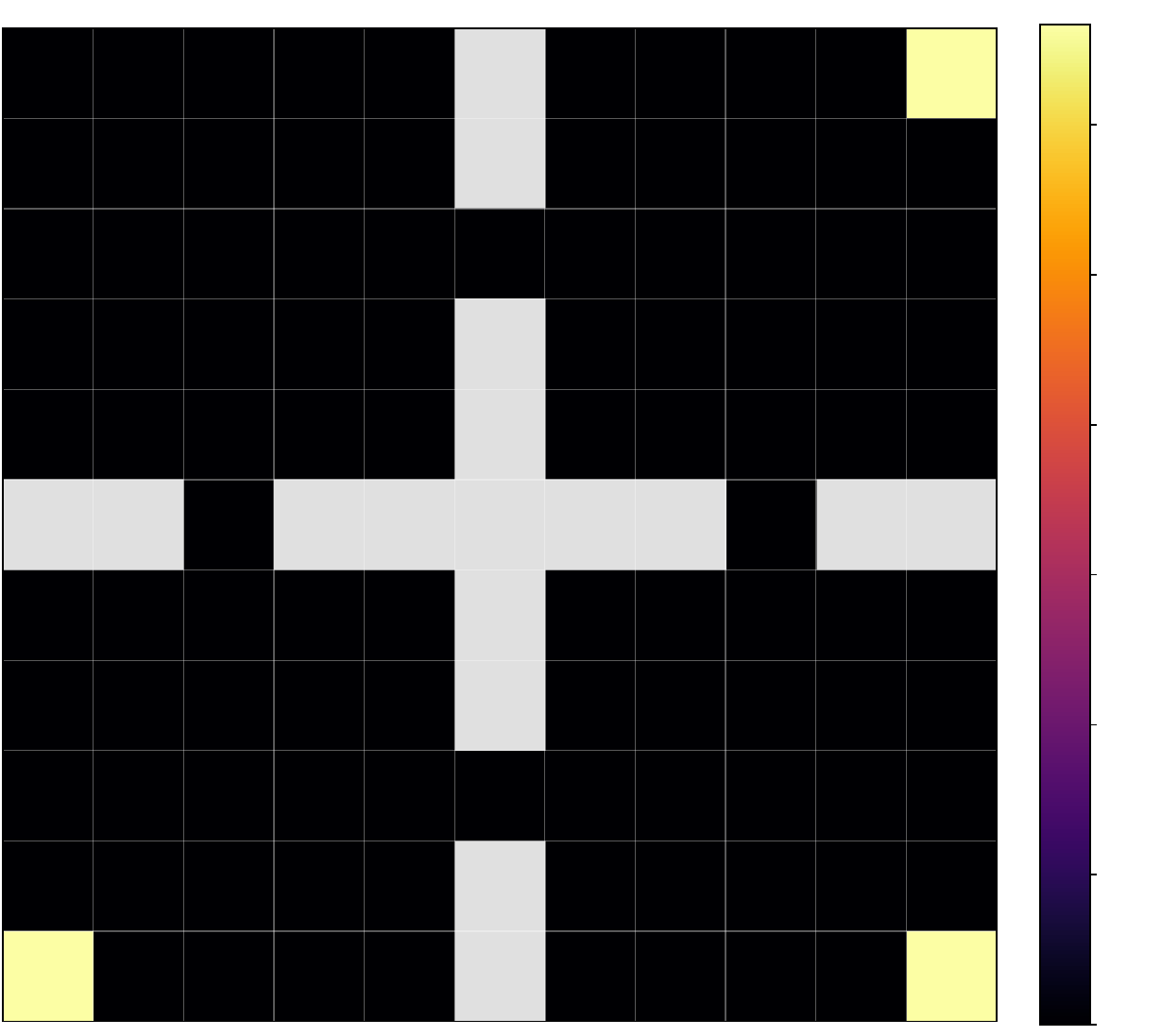}
        \par\smallskip{\footnotesize (d) Three targets\par}
    \end{minipage}\hfill
    \begin{minipage}[t]{0.31\linewidth}
        \centering
        \includegraphics[width=\linewidth]{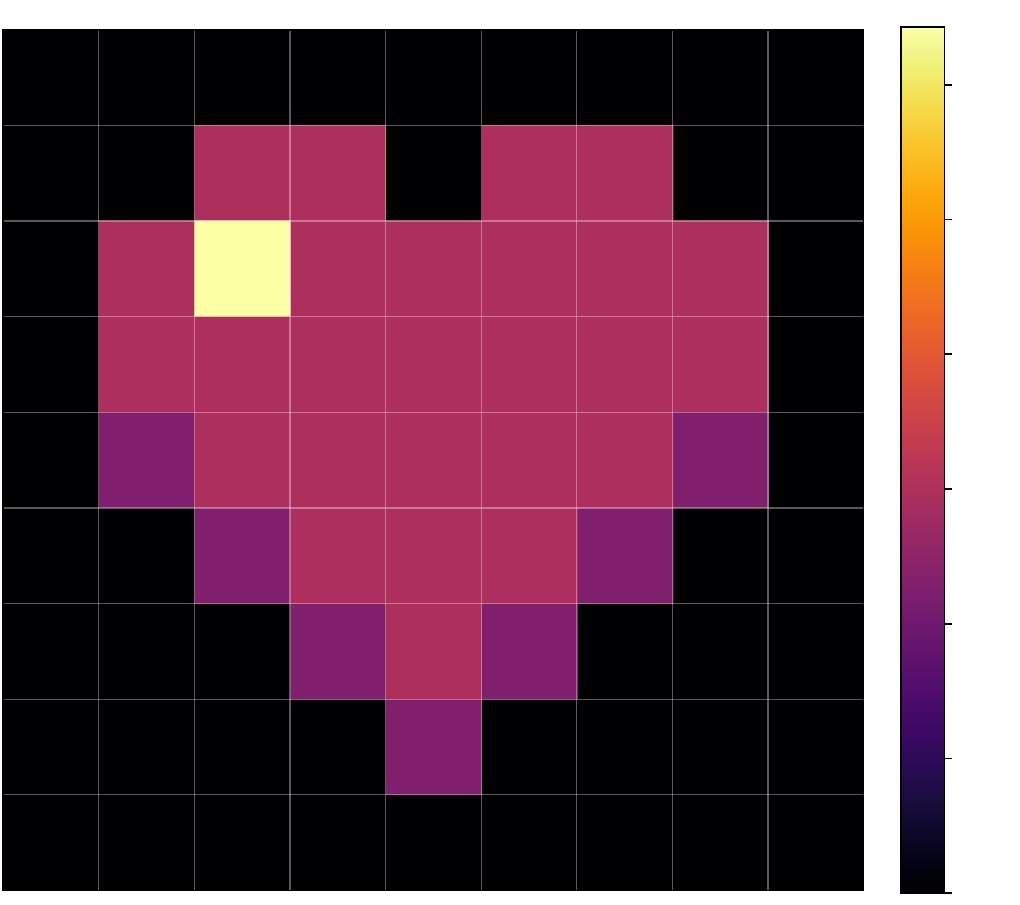}
        \par\smallskip{\footnotesize (e) KL imitation\par}
    \end{minipage}
    \caption{Experimental environments and occupancy objectives. The four-room domain is used for entropy maximization and three-target optimization; the empty grid is used for imitation of a heart-shaped terminal distribution. The bottom row illustrates the corresponding desired state occupancies.}
    \label{fig:exp-setup}
\end{figure}

\paragraph{Environments and Bilinear Dynamics}
The experiments consider an \(11\times 11\) four-room gridworld,
with narrow doorways connecting the four rooms, and an empty
\(9\times 9\) grid for imitation learning. Both environments have
five actions, \(\{\mathrm{stay},\mathrm{up},\mathrm{down},
\mathrm{left},\mathrm{right}\}\); attempted moves into walls or
beyond the grid boundary leave the agent in its current cell.
The state space contains every grid cell, including the
\(17\) wall cells of the four-room grid. Hence,
\(|\mathcal S|=121\), of which \(104\) cells are free. Wall cells are
unreachable, and every action at a wall cell leads to the
nearest free cell. Coordinates \((x,y)\) increase to the right
and downward. Episodes start from the top-left cell \((0,0)\),
the initial policy is uniform, and the horizon is \(H=40\).
\Cref{fig:exp-setup} depicts the environments and the three
optimization objectives.

Transitions are represented in bilinear form as
\(P^\star(s'\mid s,a)=\boldsymbol{\phi}(s,a)^\top
\mathbf M^\star\boldsymbol{\psi}(s')\), with known features
and an unknown core matrix (\cref{def:bilienar_mdp}).
The four-room domain admits an exact representation with
\(d_\phi=d_\psi=104\), while the corresponding dimension
in the empty grid is \(81\).

To construct these representations, we factorize the
transition matrix \(W\in\mathbb R^{|\mathcal S||\mathcal A|
\times|\mathcal S|}\) using its rank-\(r\) singular value
decomposition \(W=U_r\Sigma_r V_r^\top\). We use \(U_r\)
and \(V_r\) as base features, with \(\Sigma_r\) as the
initial core, and transform them by random invertible
matrices \(C,D\in\mathbb R^{r\times r}\). The resulting features \(\Phi=U_rC\) and
\(\Psi=V_rD\), with core
\(\mathbf M^\star=C^{-1}\Sigma_rD^{-\top}\), preserve the
transition kernel, since \(\Phi\mathbf M^\star\Psi^\top=W\).
The resulting feature maps are held fixed throughout each run and
provided to the algorithms, while the core matrix remains
unknown. Since the features are derived from the true
transition matrix, this setup follows the known-feature
model of~\cref{def:bilienar_mdp}. For $\OCUPOLIN$, we use only
the known state--action features $\boldsymbol{\phi}$ and
estimate the transition model through sparse regression.

\paragraph{Concave Occupancy Objectives}
Let \(d_h^\pi(s)=\sum_{a\in\mathcal A}\mu_h^\pi(s,a)\) denote the state occupancy at time \(h\). We consider the following three objectives:
\begin{itemize}

    \item \textbf{Maximum entropy} (\cref{fig:exp-setup}(c)):
    \(
    F_{\mathrm{ent}}(\mu^\pi)=-\sum_{h=1}^{H}\sum_{s\in\mathcal S}\bigl(d_h^\pi(s)+\tau\bigr)\log\bigl(d_h^\pi(s)+\tau\bigr).
    \)
    This objective encourages visitation of a broad range of reachable states.

    \item \textbf{Three-target optimization} (\cref{fig:exp-setup}(d)):
    \(
    F_{\mathrm{tar}}(\mu^\pi)=-\sum_{g\in\mathcal G}(1-d_H^\pi(g))^2,
    \)
    where \(\mathcal G=\{(0,10),\allowbreak(10,10),\allowbreak(10,0)\}\) contains the three designated targets. This objective rewards allocating terminal occupancy to three separated locations.

    \item \textbf{KL imitation} (\cref{fig:exp-setup}(e)):
    \(
    F_{\mathrm{imit}}(\mu^\pi)=-\sum_{s\in\mathcal S}d_H^\pi(s)\log\frac{d_H^\pi(s)}{d^{\mathrm{exp}}(s)},
    \)
    where the heart-shaped terminal-state distribution \(d^{\mathrm{exp}}\) is obtained by sampling a grayscale heart image on the \(9\times9\) grid and normalizing the intensities to sum to one. This objective encourages matching a structured expert distribution on the empty grid.

\end{itemize}
For numerical stability, we use additive smoothing for maximum entropy and lower-bound the probabilities inside the logarithms for KL imitation, with \(\tau=10^{-8}\).

All three objectives depend on \(\mu_h^\pi\) only through the state occupancy \(d_h^\pi\). Hence, they satisfy~\cref{asm:gradient_locality} with coordinate groups \(\{s\}\times\mathcal A\) and \(D_{\mathcal G}=|\mathcal A|\), rather than~\cref{asm:coordinate_separability}. The algorithms form the estimated state occupancy by summing the coordinatewise-clipped estimates \(\widehat\mu_{k,h}(s,a)\) over actions, yielding values in \([0,|\mathcal A|]\). On this domain, \OPE{} clips \(\widehat Q_{k,h}\) with \(G=\max\{|1+\log\tau|,1+\log(|\mathcal A|+\tau)\}\) for maximum entropy, \(G=2\max\{1,|\mathcal A|-1\}\) for three-target optimization, and \(G=1+\log((|\mathcal A|+\tau)/\tau)\) for KL imitation. The floored KL objective is not smooth and therefore lies outside~\cref{asm:lipschitzness_and_smoothness}.

\paragraph{Policy Classes and Actor Coresets}
Unless otherwise stated, both algorithms use tabular policy features. The off-policy algorithm performs projected natural policy gradient updates in a log-linear policy class specified by features $\boldsymbol{x}(s,a)\in\mathbb R^{d_x}$. To investigate the effect of policy representation, we compare dimensions $d_x\in\{104,605\}$ in the full-rank four-room domain. The tabular representation has $d_x=|\mathcal S||\mathcal A|=605$, while the $104$-dimensional representation uses the full bilinear transition features, whose dimension corresponds to the number of traversable states. Both configurations share the same underlying transition kernel and bilinear dynamics representation, allowing us to compare structured transition features against a tabular policy representation.

For weighted actor regression, $\OCUPOOFF$ uses a fixed
state--action coreset constructed using the greedy
experimental-design procedure of
\citet[Algorithm 4]{lin2026optimistic} with threshold $0.9$.
We augment the
selected set with linearly independent feature directions
to ensure a full-rank actor Gram matrix $\boldsymbol{G}$,
and assign uniform positive weights. The coreset is
constructed before learning and held fixed throughout
training.

\paragraph{Hyperparameter Tuning}
We tune the policy step size $\eta$, optimism coefficient $\alpha$, and ridge parameter $\lambda$ by grid search. In all three algorithms, \OPE{} uses the confidence radius $\alpha(H-h)G$ at step $h$. For $\OCUPOON$ and $\OCUPOOFF$, this matches the radii in our analysis up to the tuned constant $\alpha$. The ridge parameter controls the $\lambda I$ regularization in the least-squares regressions used for transition-model estimation, optimistic policy evaluation, and backward feature regression, which share the same $\lambda$. Tuning uses one run per configuration with seed $0$ and $500$ updates, collecting $2{,}000$ trajectories per update for $\OCUPOON$ and one for $\OCUPOOFF$ and $\OCUPOLIN$. The configuration with the highest final exact utility is selected separately for each objective, algorithm, and MDP. The search grids and the selected values for the full-rank, compact-dynamics, and linear-MDP experiments are listed in \cref{tab:exp-grids,tab:exp-fullrank-hparams,tab:exp-lowrank-hparams} at the end of this section. All reported comparisons use five held-out seeds, $10$--$14$, which are shared across methods and differ from the tuning seed.

\paragraph{Evaluation}
We report average regret as $\mathbb{E}_{k\sim\operatorname{Unif}([K])}[F(\mu^{\pi^\star})-F(\mu^{\pi_k})]=\operatorname{Reg}_K/K$, which, by the concavity of $F$, upper-bounds the suboptimality gap of the mixture policy. For each objective and environment, we compute a reference for the optimal utility by running exact \NPG{} with the known transition dynamics from four initializations, for at most $3{,}500$ updates each. For a policy $\pi$ with pseudo-reward $R=\nabla F(\mu^\pi)$, the concavity of $F$ implies that $F(\mu^\pi)\leq F(\mu^{\pi^\star})\leq F(\mu^\pi)+\Delta$, where $\Delta$ is the gap between the optimal return under the fixed reward $R$, computed by dynamic programming, and the return of $\pi$. The KL objective is not differentiable where \(d_H^\pi(s)=\tau\), but by concavity this bound remains valid when any value between the two one-sided derivatives replaces the gradient at such points. We stop once $\Delta$ falls below a tolerance between $0.002$ and $0.025$, and use the upper bound $F(\mu^\pi)+\Delta$ as the reference. Hence, the reported average regret overestimates the true average regret by at most this tolerance. \Cref{tab:exp-reference} lists the resulting bounds. Policy utilities and occupancies are evaluated exactly under the true transition dynamics. Learning curves report the empirical mean across the five seeds, with shaded regions representing pointwise $95\%$ Student's $t$ confidence intervals for the mean, $\pm t_{0.975,4}\,s/\sqrt5$, where $s$ is the sample standard deviation across seeds.

\subsection{Convergence and Occupancy Optimization}\label{subsec:exp-main}

We first examine the optimization behavior of \(\OCUPOON\) and \(\OCUPOOFF\) on all three objectives. Both algorithms run for \(K=1{,}000\) iterations using the exact bilinear representations described above and tabular policies. At each iteration, \(\OCUPOON\) gathers \(2{,}000\) fresh trajectories, whereas \(\OCUPOOFF\) gathers one trajectory and reuses the accumulated experience to estimate occupancies through backward feature regression. Consequently, the two runs collect \(2\times10^6\) and \(10^3\) real trajectories, respectively. This experiment compares optimization progress by policy update, not sample efficiency at a matched interaction budget.

\begin{figure}[t]
    \centering
    \includegraphics[width=\linewidth]{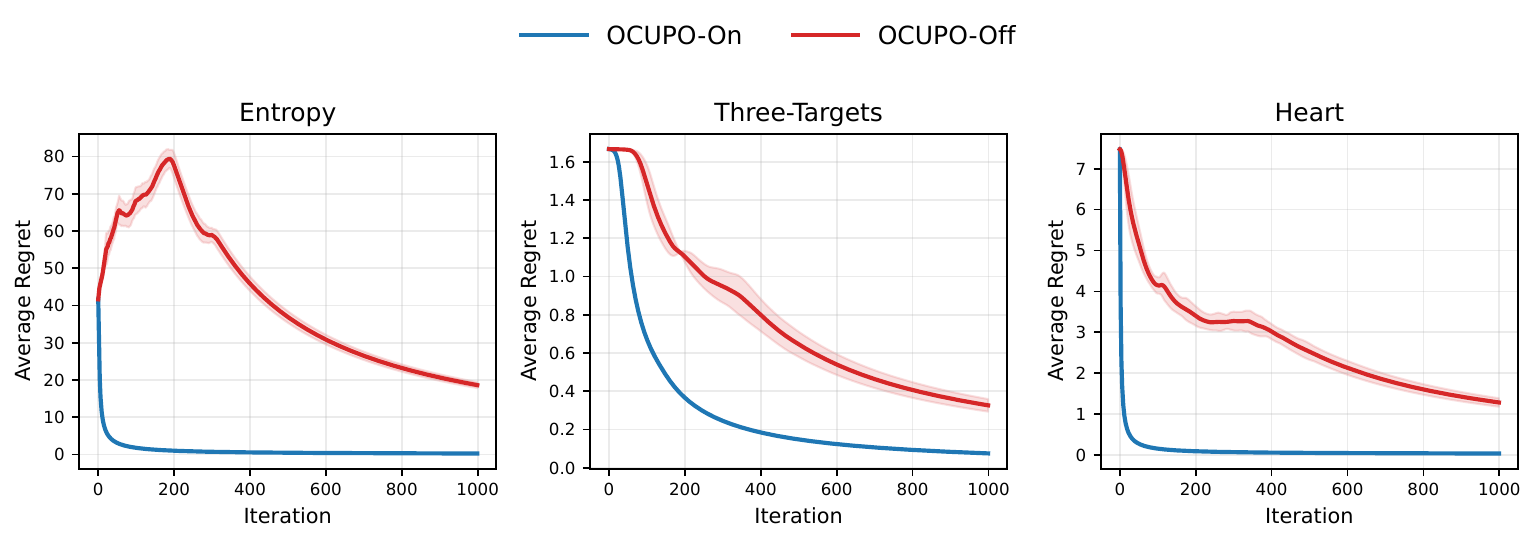}
    \caption{Average regret on maximum entropy, three-target optimization, and KL imitation (panels titled Entropy, Three-Targets, and Heart, respectively). Curves show the mean and pointwise $95\%$ confidence intervals across five held-out seeds. The algorithms are compared by policy update rather than by matched real interactions.}
    \label{fig:exp-main-subopt}
\end{figure}

\paragraph{Learning Curves}
\Cref{fig:exp-main-subopt} reveals a consistent pattern across the three objectives: $\OCUPOON$ achieves rapid improvements in the early iterations, whereas $\OCUPOOFF$ generally converges more gradually. Despite this difference in convergence speed, both algorithms learn policies that exhibit the desired occupancy behavior. In the maximum-entropy task, both methods achieve broad coverage of the reachable states, while in the three-target task, they approach a balanced allocation of terminal occupancy across the designated goals. A similar pattern emerges in the imitation task, where the on-policy method improves rapidly and the off-policy method progressively recovers the structure of the expert distribution. These results compare optimization progress per policy update rather than interaction efficiency, which is examined under matched trajectory budgets in \cref{subsec:exp-budget}.

\begin{figure}[p]
    \centering
    \includegraphics[width=0.8\linewidth]{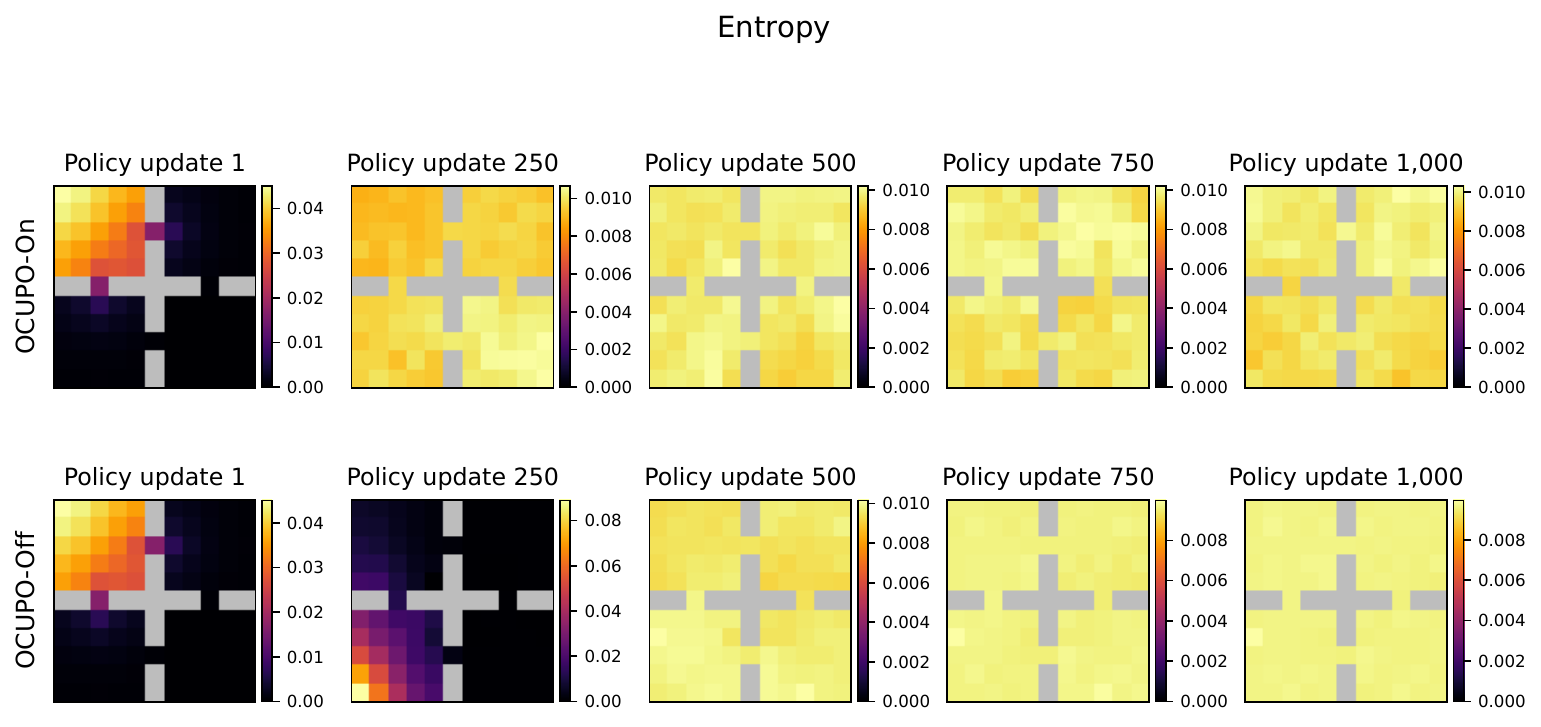}
    \par\bigskip
    \includegraphics[width=0.8\linewidth]{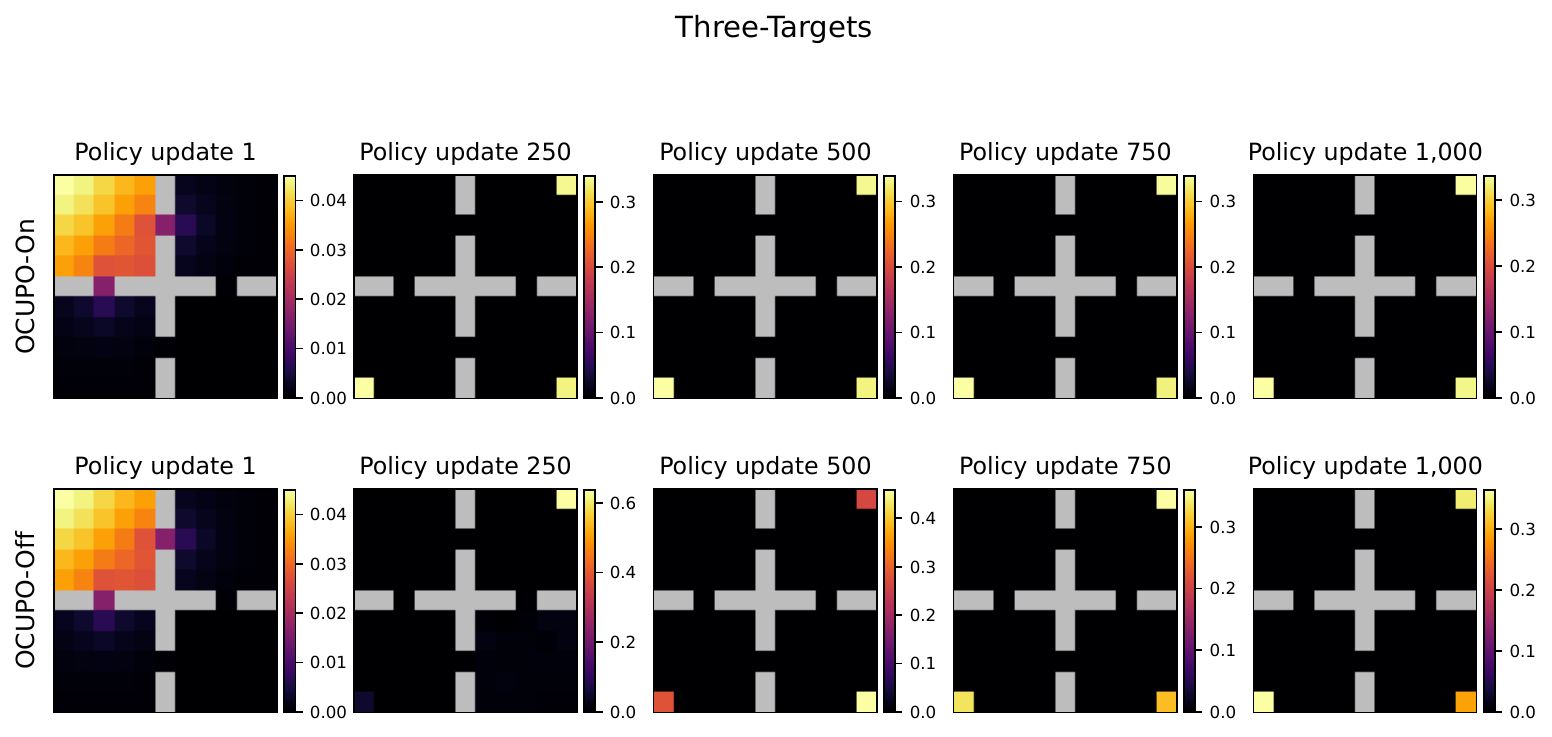}
    \par\bigskip
    \includegraphics[width=0.8\linewidth]{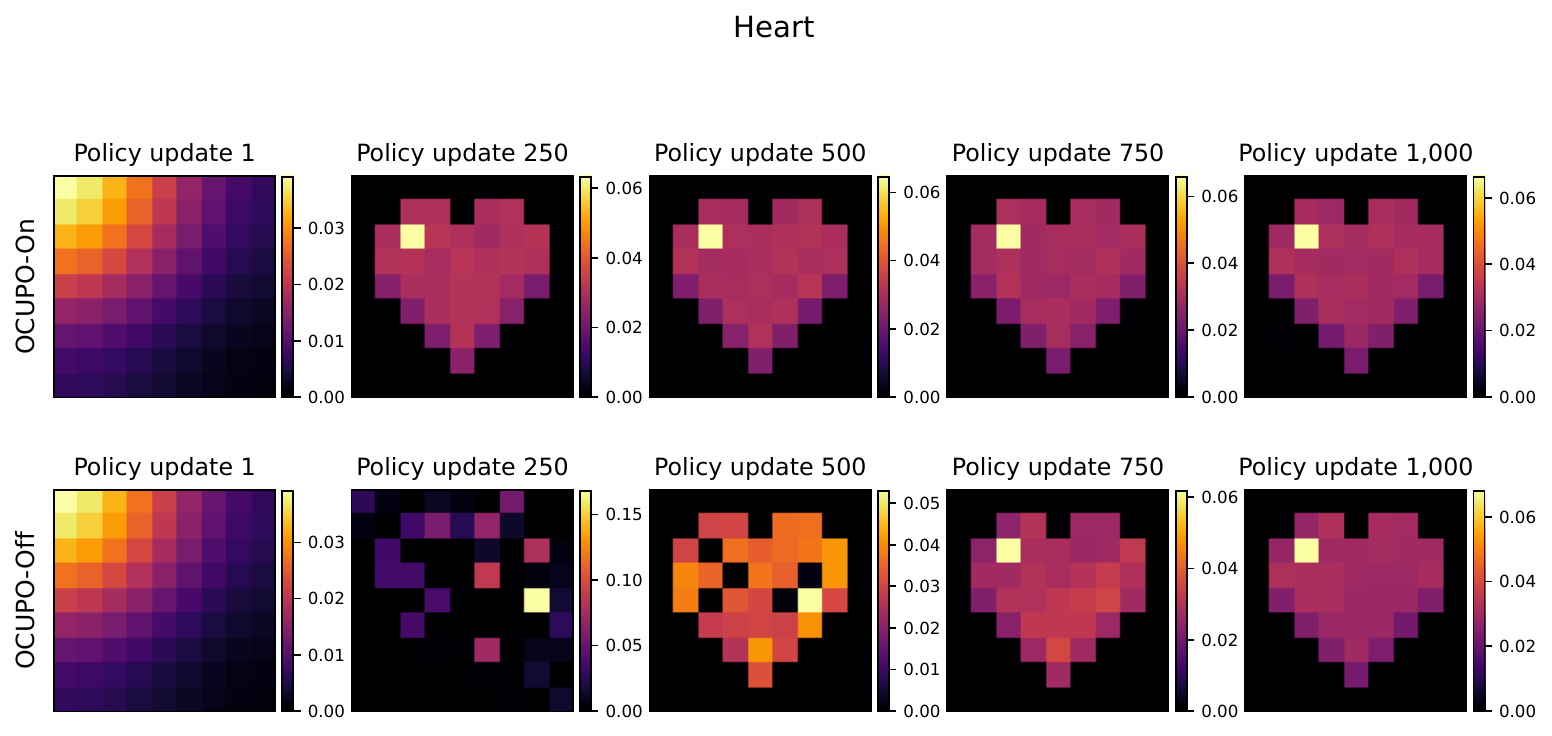}
    \caption{Exact terminal-state occupancy as training progresses. From top to bottom, the panels titled Entropy, Three-Targets, and Heart correspond to maximum entropy, three-target optimization, and KL imitation, respectively. Within each panel, the top and bottom rows show the occupancies of the policies learned by \OCUPOON\ and \OCUPOOFF, respectively, after \(1\), \(250\), \(500\), \(750\), and \(1{,}000\) policy updates. All heatmaps show the policy of a single seed, \(10\).}
    \label{fig:exp-main-occupancy}
\end{figure}

\paragraph{Occupancy Evolution}
\Cref{fig:exp-main-occupancy} shows how the exact terminal-state occupancies evolve during training. Consistent with the learning curves, both algorithms progressively develop the desired spatial patterns: broader state coverage for entropy maximization, increasing concentration around the three targets, and closer alignment with the heart-shaped expert distribution for imitation. These visualizations provide a qualitative view of the learned policies beyond their average-regret values.

\subsection{Real-Interaction Sample Efficiency}\label{subsec:exp-budget}

\begin{figure}[p]
    \centering
    \includegraphics[width=0.8\linewidth]{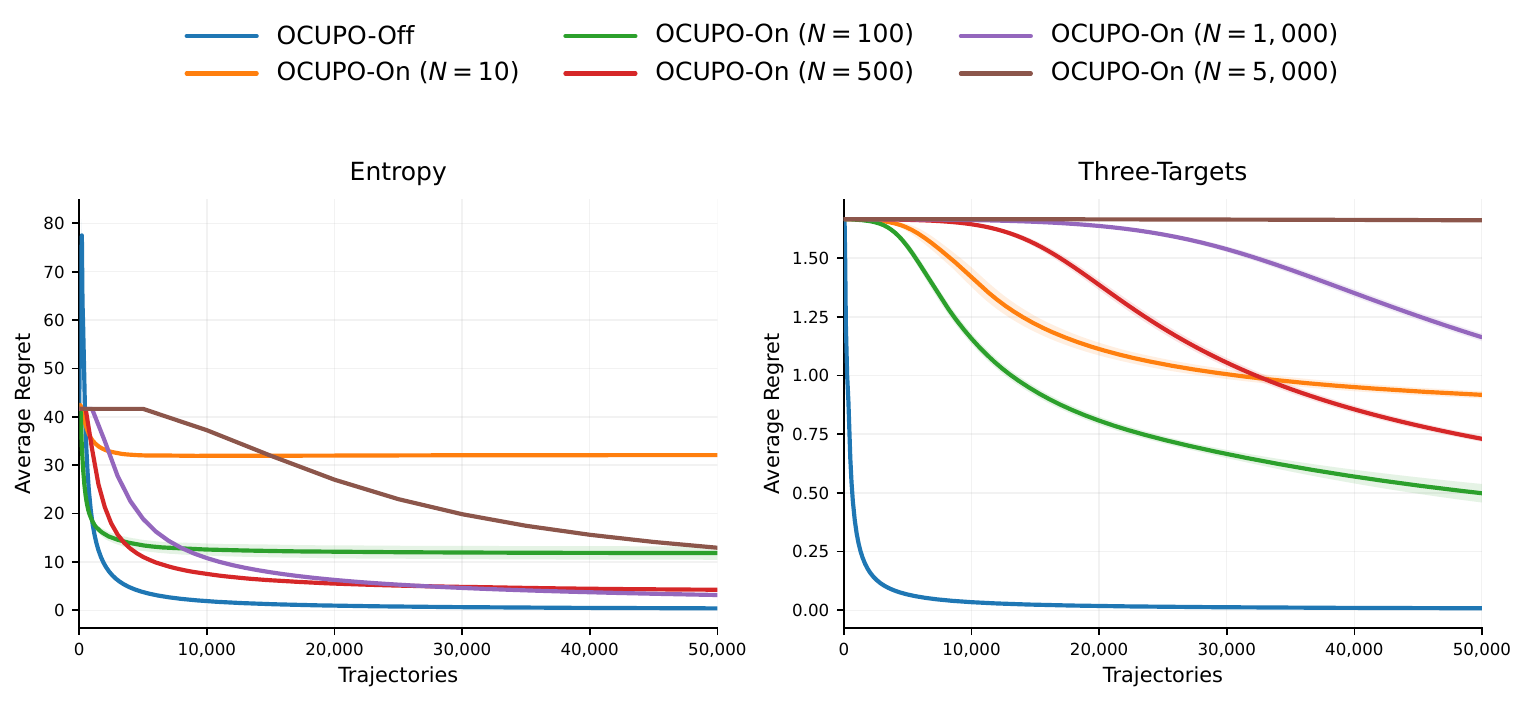}
    \caption{Average regret under a common budget of $50{,}000$ trajectories for maximum entropy (left) and three-target optimization (right). Both panels compare $\OCUPOOFF$ with five $\OCUPOON$ batch sizes.}
    \label{fig:exp-budget}

    \vspace{\floatsep}
    \includegraphics[width=0.8\linewidth]{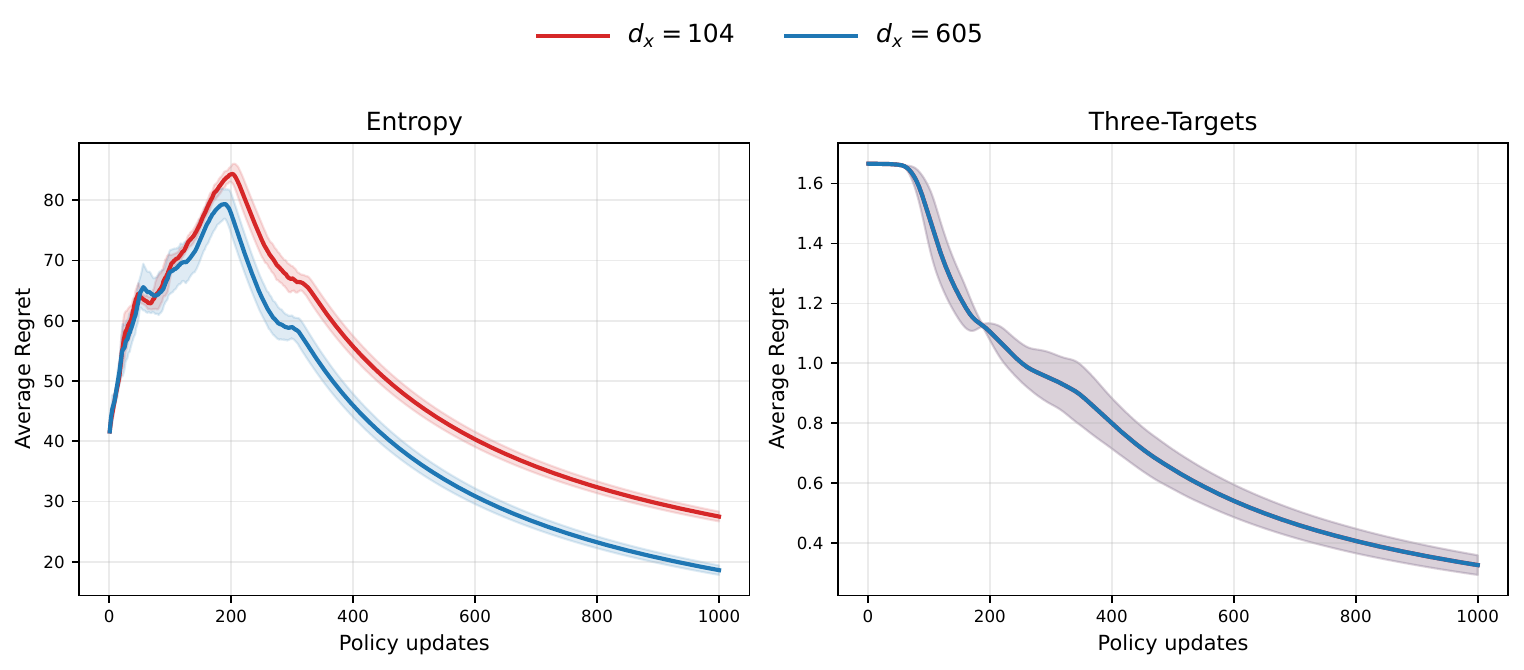}
    \caption{Average-regret learning curves for policy-feature dimensions $d_x\in\{104,605\}$ on entropy maximization (left) and three-target optimization (right).}
    \label{fig:exp-actor-learning}

    \vspace{\floatsep}
    \begin{minipage}[t]{0.3\linewidth}
        \centering
        \includegraphics[width=\linewidth]{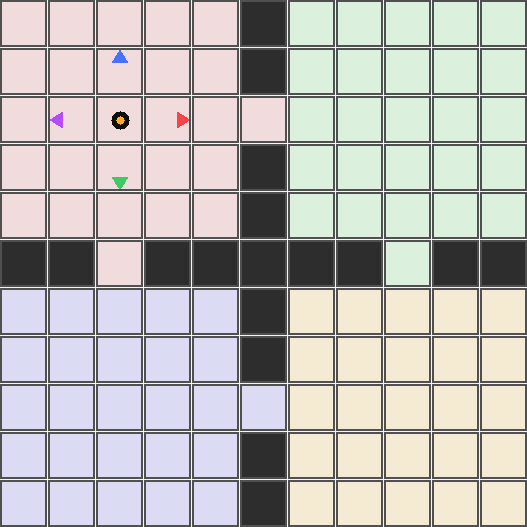}
        \par\smallskip{\footnotesize (a) Full-rank dynamics\par}
    \end{minipage}\hspace{0.081\linewidth}%
    \begin{minipage}[t]{0.3\linewidth}
        \centering
        \includegraphics[width=\linewidth]{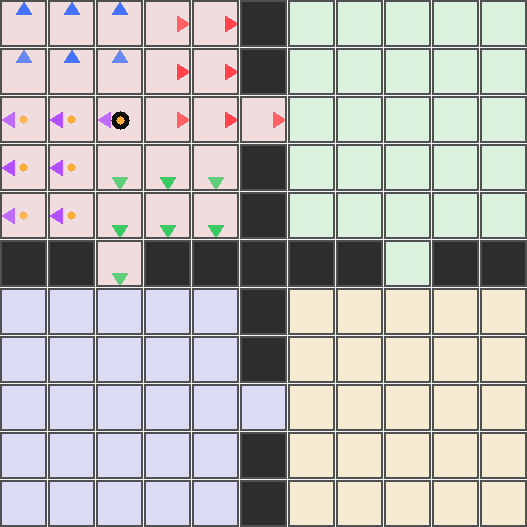}
        \par\smallskip{\footnotesize (b) Compact dynamics, \(r=16\)\par}
    \end{minipage}
    \caption{
    Action-layout visualization in the four-room gridworld. The black circle marks the current state.
    Each glyph is drawn at a possible next state under the corresponding action
    (\uptri{} up, \downtri{} down, \lefttri{} left, \righttri{} right, \staycirc{} stay),
    and its opacity indicates the transition probability. Cell colors indicate the four rooms, and dark cells are walls. Compact bilinear features share
    transition structure across groups of states while preserving wall-respecting dynamics.
    }
    \label{fig:exp-lowrank-environments}
\end{figure}

The preceding study aligns policy updates but not environment interactions. We next compare the algorithms under the same budget of $B=50{,}000$ real trajectories on entropy maximization and three-target optimization. To examine the role of batch size, we distribute this budget across five on-policy configurations: $N\in\{10,100,500,1000,5000\}$ fresh trajectories per update, corresponding to $K=B/N\in\{5000,500,100,50,10\}$ updates. The off-policy method collects one trajectory per update, reuses its accumulated experience through backward feature regression, and performs $50{,}000$ updates. All configurations use the same environment, tabular policy class, horizon, and previously selected objective-specific hyperparameters, which are not retuned for each batch size. The horizontal axis in \cref{fig:exp-budget} records real-environment trajectories, rather than policy updates.

For entropy maximization, $\OCUPOOFF$ achieves the lowest final average regret under the matched interaction budget. The on-policy methods exhibit substantial sensitivity to batch size, with larger batches generally reducing the number of optimization opportunities and slowing improvement in terms of environment interactions. Under these fixed hyperparameters, the results illustrate the interaction-efficiency benefits of off-policy data reuse for entropy maximization. For three-target optimization, $\OCUPOOFF$ also achieves substantially lower average regret than all on-policy configurations, approaching zero regret within the matched interaction budget. These comparisons evaluate the specified configurations rather than the best performance obtainable after separately tuning each batch size.

\subsection{Effect of Policy Representation}\label{subsec:exp-actor}

The off-policy algorithm accommodates a log-linear policy class defined by the choice of policy features $\boldsymbol{x}(s,a)$. We investigate how the dimension of this representation affects optimization while keeping the underlying full-rank four-room MDP, transition-model features, objective, and interaction budget fixed. We compare two policy-feature dimensions, $d_x\in\{104,605\}$: $d_x=104$ uses the full bilinear transition-feature representation, while $d_x=605$ is tabular. Each configuration runs $\OCUPOOFF$ for $K=1{,}000$ updates, collecting one real trajectory per update. All use the same off-policy hyperparameters selected for the tabular class, without retuning for $d_x=104$, with a full-rank actor coreset constructed separately in each feature space.

\Cref{fig:exp-actor-learning} shows that, under these hyperparameters, the tabular policy
representation achieves lower average regret on entropy
maximization, whereas the two representations exhibit similar
performance on the three-target objective.

\subsection{Compact Bilinear Representations}\label{subsec:exp-lowrank}

We next examine whether the optimization framework extends to environments whose transition kernels admit substantially lower-dimensional bilinear representations. We construct four room-structured MDPs with \(d_\phi=d_\psi=r\in\{16,24,36,64\}\). Free states are partitioned into \(r\) groups within rooms by deterministic median splits. For each state--action pair, the next state is drawn from a Gaussian-shaped distribution over the group that contains the wall-respecting successor under the full-rank dynamics, centered at a representative cell of that group. An attempted move into a wall selects the group of the current cell, and every action at a wall cell selects the group with the nearest representative. The construction is deterministic, and each resulting kernel is row-stochastic with rank \(r\). This preserves walls and coarse directional action semantics while sharing transition structure across states. Each rank corresponds to a different valid transition kernel, rather than an approximation to the same full-rank MDP. The learning algorithms use the resulting bilinear features and estimate the compact core \(\mathbf M^\star\in\mathbb R^{r\times r}\) from experience. Both algorithms use tabular policy features.
\Cref{fig:exp-lowrank-environments} illustrates the
resulting action-conditioned transition structure.

We evaluate both algorithms on maximum entropy and three-target optimization, using \(K=1{,}000\) iterations, \(N=2{,}000\) trajectories per update for \(\OCUPOON\), and the same five held-out seeds. Each objective--rank--algorithm combination uses hyperparameters selected independently with the tuning protocol described above (\cref{tab:exp-lowrank-hparams}). We compute a separate numerical reference for each MDP
when evaluating average regret.

\begin{figure}[p]
    \centering
    \includegraphics[width=0.8\linewidth]{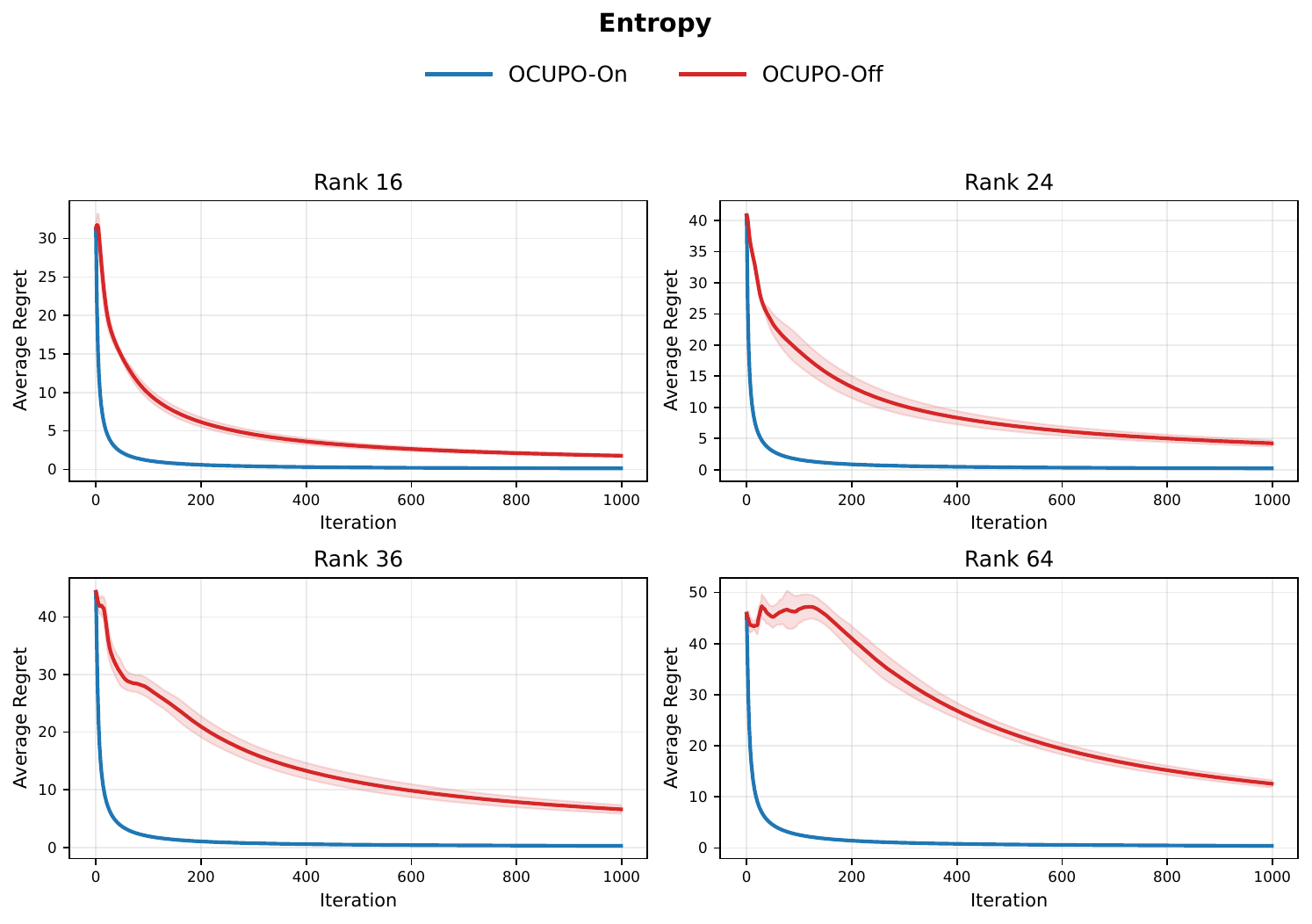}
    \par\bigskip
    \includegraphics[width=0.8\linewidth]{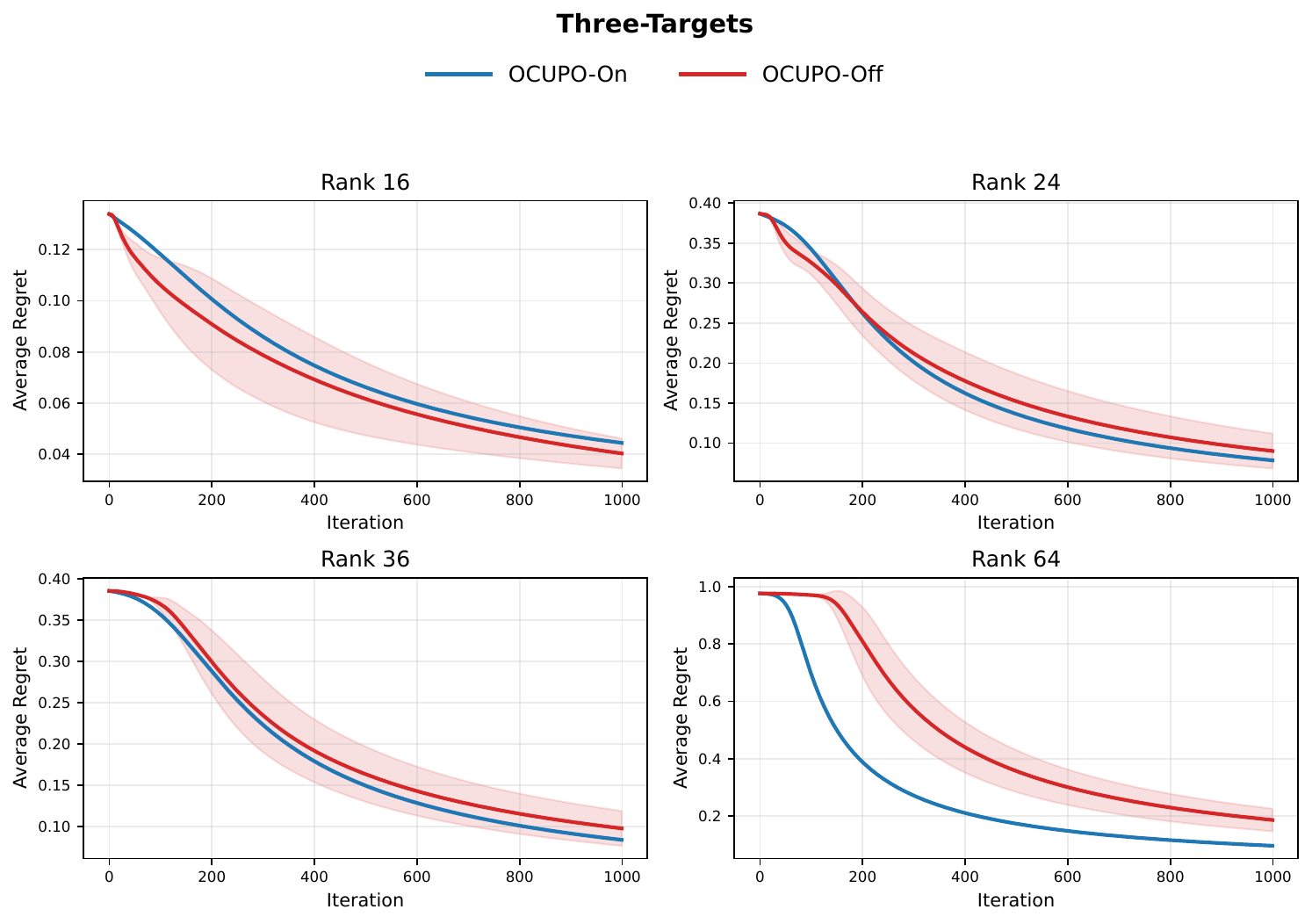}
    \caption{Average regret under compact bilinear dynamics with transition ranks \(r\in\{16,24,36,64\}\), using a separate numerical reference for each MDP. The top and bottom panels, titled Entropy and Three-Targets, show the results for maximum entropy and three-target optimization, and each subplot corresponds to one transition rank.}
    \label{fig:exp-lowrank-curves}
\end{figure}

\Cref{fig:exp-lowrank-curves} shows that both algorithms improve the occupancy objectives across the tested bilinear ranks. For entropy maximization, the on-policy method generally improves rapidly by iteration count, whereas the off-policy method has a longer transient before stabilizing. The three-target curves also improve across ranks, with differences in their early optimization behavior and final average-regret values. Together, the experiments illustrate optimization with compact bilinear cores without learning an explicit state--action transition table.

\subsection{Linear-MDP Extension}\label{subsec:exp-lin}
Finally, we examine whether the framework remains effective when only the state--action features are known. We evaluate $\OCUPOLIN$ on maximum entropy and three-target optimization across the full-rank four-room MDP ($r=104$) and the four compact MDPs of~\cref{subsec:exp-lowrank} ($r\in\{16,24,36,64\}$). In each MDP, $\OCUPOLIN$ receives only the features $\boldsymbol{\phi}$ and estimates the transition model by sparse transition estimation (\cref{alg:ste}) rather than through a core matrix. For each setting, we run $K=1{,}000$ updates with one trajectory per update, using hyperparameters tuned independently on the grid of $\OCUPOOFF$ (\cref{tab:exp-fullrank-hparams,tab:exp-lowrank-hparams}) and five held-out seeds. In these experiments, we use a practical variant of $\OCUPOLIN$ in which \OPE{} uses the fixed radius $\alpha_{k,h}^{\mathrm{Lin}}=\alpha(H-h)G$, as in $\OCUPOOFF$, instead of the data-dependent radius in~\cref{eq:lin_value_radius_outline}.

\begin{figure}[p]
    \centering
    \includegraphics[width=0.94\linewidth]{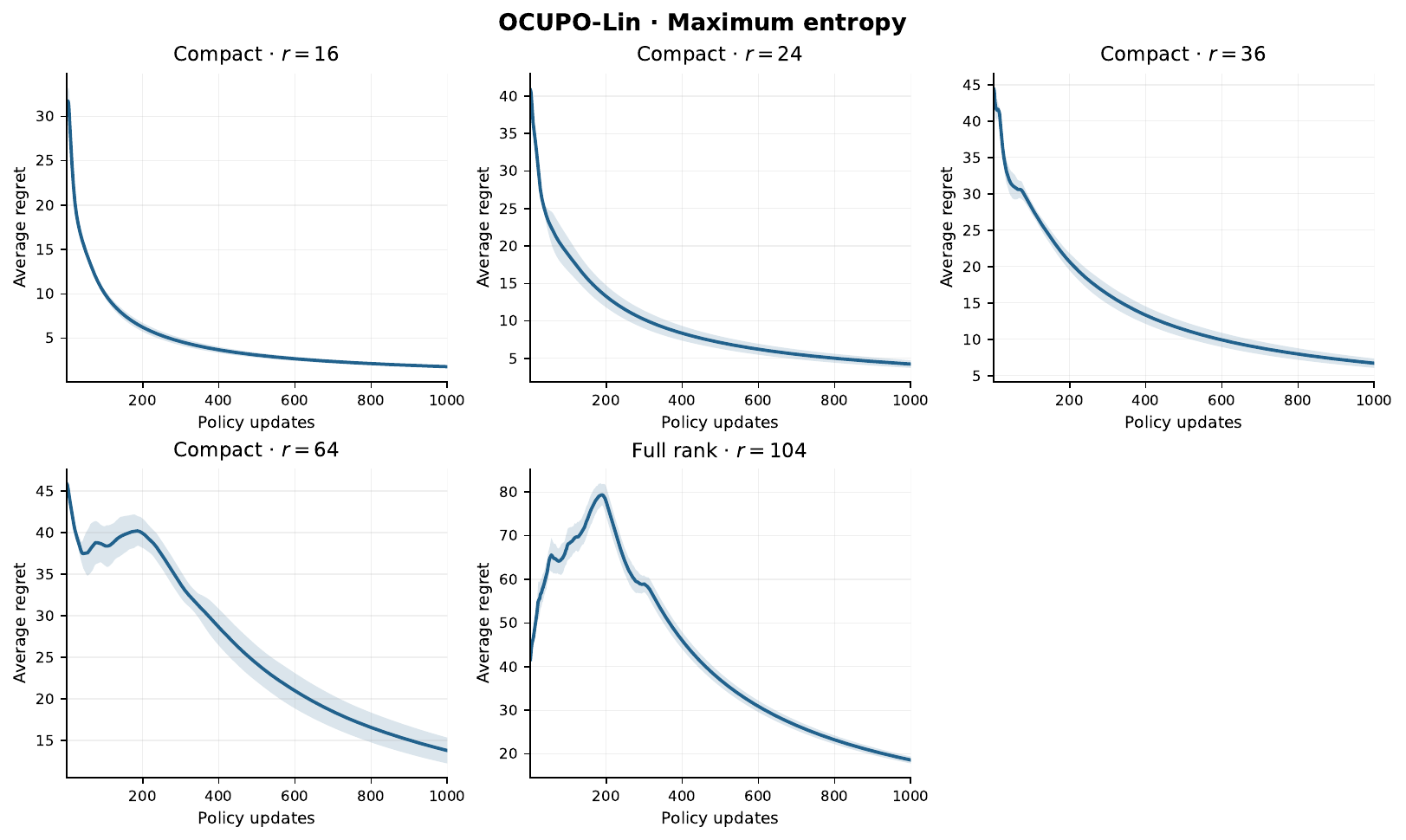}
    \par\bigskip
    \includegraphics[width=0.94\linewidth]{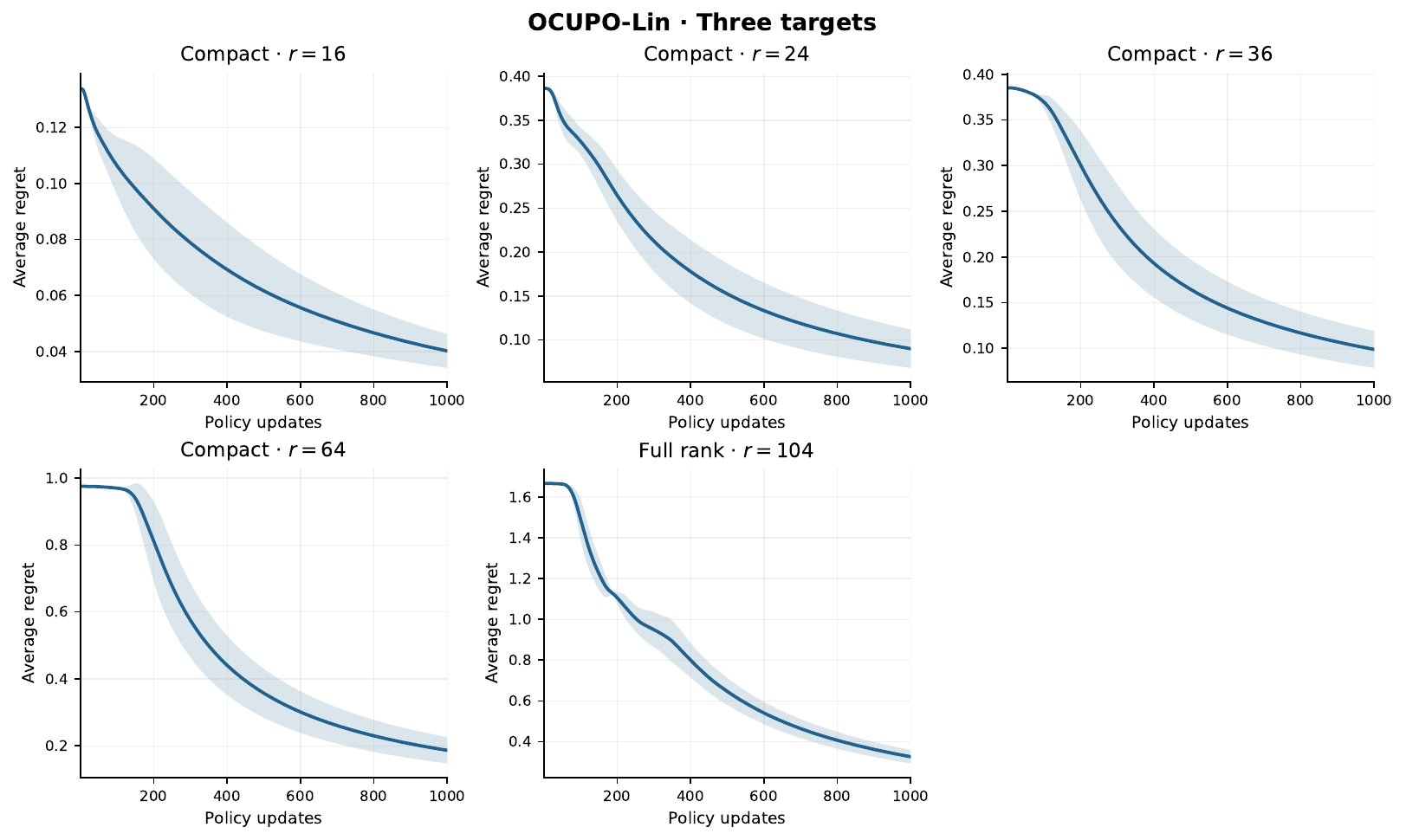}
    \caption{Average-regret learning curves for $\OCUPOLIN$. The top and bottom panels show maximum entropy and three-target optimization, and each subplot shows a single transition model: four compact ranks and the full-rank four-room MDP ($r=104$). Curves show the mean and pointwise $95\%$ confidence intervals across five seeds, and each MDP has its own numerical reference.}
    \label{fig:exp-lin}
\end{figure}

\Cref{fig:exp-lin} shows the resulting average-regret learning curves. Since each MDP has its own numerical reference, we compare the curves within each panel rather than across ranks. For maximum entropy, average regret decreases almost monotonically for $r\in\{16,24,36\}$. For $r=64$ and $r=104$, it first stagnates or increases during roughly the first $200$ updates and then decreases steadily. For three-target optimization, average regret decreases in all five MDPs, although for $r\geq36$ it remains close to its initial value during the early updates. Together, these results illustrate that the practical variant of $\OCUPOLIN$ reduces average regret on both objectives in every tested MDP while using only the state--action features.

\paragraph{Summary}
Across the considered objectives, both $\OCUPOON$ and
$\OCUPOOFF$ reduce average regret and produce the desired
occupancy patterns. Under the selected hyperparameters, the matched-budget experiments illustrate
the interaction-efficiency benefits of off-policy data reuse
for entropy maximization, while the policy-representation and
compact-dynamics studies illustrate the applicability of the
framework across different feature representations.

\begin{table}[p]
\centering
\caption{Hyperparameter search grids. $\OCUPOLIN$ uses the off-policy grids.}
\label{tab:exp-grids}
\small
\setlength{\tabcolsep}{4pt}
\renewcommand{\arraystretch}{1.12}
\begin{tabular}{@{}lp{0.78\linewidth}@{}}
\toprule
Parameter & Candidate values \\
\midrule
On-policy \(\eta\) & \(10^{-5}\), \(3\cdot10^{-5}\), \(10^{-4}\), \(3\cdot10^{-4}\), \(7\cdot10^{-4}\), \(10^{-3}\), \(1.5\cdot10^{-3}\), \(1.7\cdot10^{-3}\), \(2\cdot10^{-3}\), \(3\cdot10^{-3}\), \(5\cdot10^{-3}\), \(10^{-2}\), \(2\cdot10^{-2}\), \(5\cdot10^{-2}\), \(10^{-1}\) \\
Off-policy \(\eta\) & \(3\cdot10^{-6}\), \(10^{-5}\), \(3\cdot10^{-5}\), \(10^{-4}\), \(3\cdot10^{-4}\), \(7\cdot10^{-4}\), \(10^{-3}\), \(1.7\cdot10^{-3}\), \(3\cdot10^{-3}\), \(5\cdot10^{-3}\), \(10^{-2}\), \(2\cdot10^{-2}\), \(3\cdot10^{-2}\), \(5\cdot10^{-2}\), \(10^{-1}\) \\
On-policy \(\alpha\) & \(0\), \(10^{-4}\), \(10^{-3}\), \(10^{-2}\), \(0.03\), \(0.1\), \(0.3\), \(0.5\), \(0.7\), \(0.9\), \(1\), \(3\), \(10\), \(100\) \\
Off-policy \(\alpha\) & \(0\), \(10^{-4}\), \(10^{-3}\), \(0.003\), \(0.01\), \(0.03\), \(0.1\), \(0.3\), \(0.5\), \(1\), \(3\), \(10\) \\
Ridge \(\lambda\) & \(10^{-6}\), \(10^{-5}\), \(3\cdot10^{-5}\), \(10^{-4}\), \(10^{-3}\), \(10^{-2}\), \(10^{-1}\), \(1\) \\
\bottomrule
\end{tabular}
\end{table}

\begin{table}[p]
\centering
\caption{Selected full-rank hyperparameters from \(500\)-iteration tuning runs. \(\OCUPOLIN\) was not evaluated on imitation.}
\label{tab:exp-fullrank-hparams}
\small
\setlength{\tabcolsep}{5pt}
\renewcommand{\arraystretch}{1.12}
\begin{tabular}{@{}lrrr@{\hspace{14pt}}rrr@{\hspace{14pt}}rrr@{}}
\toprule
 & \multicolumn{3}{c}{Entropy} & \multicolumn{3}{c}{Three targets} & \multicolumn{3}{c}{Imitation} \\
\cmidrule(r){2-4}\cmidrule(lr){5-7}\cmidrule(l){8-10}
Algorithm & \(\eta\) & \(\alpha\) & \(\lambda\) & \(\eta\) & \(\alpha\) & \(\lambda\) & \(\eta\) & \(\alpha\) & \(\lambda\) \\
\midrule
\(\OCUPOON\) & \(0.01\) & \(0.01\) & \(3\cdot10^{-5}\) & \(0.1\) & \(10^{-4}\) & \(3\cdot10^{-5}\) & \(0.1\) & \(0\) & \(10^{-5}\) \\
\(\OCUPOOFF\) & \(0.05\) & \(0.003\) & \(10^{-6}\) & \(0.1\) & \(10^{-4}\) & \(10^{-5}\) & \(0.03\) & \(10^{-3}\) & \(3\cdot10^{-5}\) \\
\(\OCUPOLIN\) & \(0.05\) & \(0.003\) & \(10^{-6}\) & \(0.1\) & \(10^{-4}\) & \(10^{-5}\) & \multicolumn{1}{c}{--} & \multicolumn{1}{c}{--} & \multicolumn{1}{c}{--} \\
\bottomrule
\end{tabular}
\end{table}

\begin{table}[p]
\centering
\caption{Selected hyperparameters for compact dynamics, tuned independently for each objective, rank, and algorithm.}
\label{tab:exp-lowrank-hparams}
\small
\setlength{\tabcolsep}{4pt}
\renewcommand{\arraystretch}{1.0}
\begin{tabular}{@{}rlrrr@{\hspace{14pt}}rrr@{}}
\toprule
 & & \multicolumn{3}{c}{Entropy} & \multicolumn{3}{c}{Three targets} \\
\cmidrule(r){3-5}\cmidrule(l){6-8}
Rank & Algorithm & \(\eta\) & \(\alpha\) & \(\lambda\) & \(\eta\) & \(\alpha\) & \(\lambda\) \\
\midrule
16 & \(\OCUPOON\) & \(0.02\) & \(0.01\) & \(10^{-5}\) & \(0.1\) & \(0.001\) & \(10^{-5}\) \\
 & \(\OCUPOOFF\) & \(0.005\) & \(0.01\) & \(10^{-5}\) & \(0.1\) & \(10^{-4}\) & \(10^{-5}\) \\
 & \(\OCUPOLIN\) & \(0.005\) & \(0.01\) & \(10^{-5}\) & \(0.1\) & \(10^{-4}\) & \(10^{-5}\) \\
\addlinespace[2pt]
24 & \(\OCUPOON\) & \(0.02\) & \(0.01\) & \(10^{-5}\) & \(0.1\) & \(10^{-4}\) & \(10^{-6}\) \\
 & \(\OCUPOOFF\) & \(0.005\) & \(0.03\) & \(10^{-5}\) & \(0.1\) & \(10^{-4}\) & \(10^{-6}\) \\
 & \(\OCUPOLIN\) & \(0.005\) & \(0.03\) & \(10^{-6}\) & \(0.1\) & \(10^{-4}\) & \(10^{-6}\) \\
\addlinespace[2pt]
36 & \(\OCUPOON\) & \(0.02\) & \(0.01\) & \(3\cdot10^{-5}\) & \(0.1\) & \(10^{-4}\) & \(3\cdot10^{-5}\) \\
 & \(\OCUPOOFF\) & \(0.05\) & \(0.03\) & \(10^{-5}\) & \(0.1\) & \(10^{-4}\) & \(10^{-4}\) \\
 & \(\OCUPOLIN\) & \(0.1\) & \(0.03\) & \(10^{-5}\) & \(0.1\) & \(10^{-4}\) & \(10^{-4}\) \\
\addlinespace[2pt]
64 & \(\OCUPOON\) & \(0.02\) & \(0.01\) & \(3\cdot10^{-5}\) & \(0.1\) & \(10^{-4}\) & \(10^{-3}\) \\
 & \(\OCUPOOFF\) & \(0.05\) & \(0.01\) & \(10^{-6}\) & \(0.1\) & \(10^{-4}\) & \(10^{-3}\) \\
 & \(\OCUPOLIN\) & \(0.01\) & \(0.01\) & \(10^{-5}\) & \(0.1\) & \(10^{-4}\) & \(10^{-3}\) \\
\bottomrule
\end{tabular}
\end{table}

\begin{table}[p]
\centering
\caption{Bounds on the optimal utility. The upper bound is the reference for average regret.}
\label{tab:exp-reference}
\small
\setlength{\tabcolsep}{5pt}
\renewcommand{\arraystretch}{1.0}
\begin{tabular}{@{}lrrrrr@{}}
\toprule
Objective & Rank & Lower bound & Upper bound & Gap & Tolerance \\
\midrule
Entropy & 16 & \(159.285146\) & \(159.287127\) & \(0.001981\) & \(0.002\) \\
Entropy & 24 & \(158.564598\) & \(158.566526\) & \(0.001927\) & \(0.002\) \\
Entropy & 36 & \(153.629883\) & \(153.631856\) & \(0.001973\) & \(0.002\) \\
Entropy & 64 & \(151.203029\) & \(151.213093\) & \(0.010065\) & \(0.011\) \\
Entropy & 104 & \(160.060834\) & \(160.082546\) & \(0.021712\) & \(0.025\) \\
\midrule
Three targets & 16 & \(-2.862274\) & \(-2.860280\) & \(0.001995\) & \(0.002\) \\
Three targets & 24 & \(-2.610437\) & \(-2.608438\) & \(0.001999\) & \(0.002\) \\
Three targets & 36 & \(-2.613720\) & \(-2.611721\) & \(0.001999\) & \(0.002\) \\
Three targets & 64 & \(-2.025444\) & \(-2.023445\) & \(0.002000\) & \(0.002\) \\
Three targets & 104 & \(-1.333335\) & \(-1.331569\) & \(0.001766\) & \(0.002\) \\
\midrule
KL imitation & 81 & \(-3.838\cdot10^{-7}\) & \(0.001901\) & \(0.001901\) & \(0.002\) \\
\bottomrule
\end{tabular}
\end{table}

\end{document}